%% file: main.tex
\pdfoutput=1
\documentclass[dvipsnames,format=sigconf,authorversion,nonacm]{acmart}

\usepackage{listings}
\usepackage{subcaption}
\usepackage{booktabs}
\usepackage{tabularx}
\usepackage{todonotes}
\usepackage[export]{adjustbox}
\usepackage{multirow}

\usepackage{mathtools}

\usepackage{pifont}
\newcommand{\cmark}{\ding{51}}%
\newcommand{\xmark}{\ding{55}}%

\usepackage{float}
\usepackage{dblfloatfix}

\AtBeginDocument{%
  \providecommand\BibTeX{{%
    \normalfont B\kern-0.5em{\scshape i\kern-0.25em b}\kern-0.8em\TeX}}}

\setcopyright{acmcopyright}
\copyrightyear{2023}
\acmYear{2023}
\acmDOI{X}

\acmConference[GECCO '23]{GECCO '23:  Genetic and Evolutionary Computation Conference}{July 15--19, 2023}{Lisbon, Portugal}
\acmBooktitle{Portugal '23: Genetic and Evolutionary Computation Conference, July 15--19, 2023, Portugal, Lisbon}
\acmPrice{15.00}
\acmISBN{XXXXXXXXXX}

\begin{document}

\setlength{\emergencystretch}{2.5em}


\title[MOREA: a GPU-accelerated Evolutionary Algorithm for Multi-Objective Deformable Registration of 3D Medical Images]{MOREA: a GPU-accelerated Evolutionary Algorithm for Multi-Objective Deformable Registration of 3D Medical Images}





\author{Georgios Andreadis}
\affiliation{%
  \institution{Leiden University Medical Center}
  \city{Leiden}
  \country{The Netherlands}
}
\email{G.Andreadis@lumc.nl}
\author{Peter A.N. Bosman}
\affiliation{%
  \institution{Centrum Wiskunde \& Informatica}
  \city{Amsterdam}
  \country{The Netherlands}
}
\email{Peter.Bosman@cwi.nl}
\author{Tanja Alderliesten}
\affiliation{%
  \institution{Leiden University Medical Center}
  \city{Leiden}
  \country{The Netherlands}
}
\email{T.Alderliesten@lumc.nl}







\input{sections/0-abstract}

\keywords{deformable image registration, multi-objective optimization, smart mesh initialization, repair method, GOMEA}


\maketitle

\input{sections/1-introduction}
\input{sections/2-problem}
\input{sections/3-morvgomea}
\input{sections/4-approach}
\input{sections/5-experiments}
\input{sections/6-results-discussion}

\input{sections/7-conclusions}

\begin{acks}
The authors thank W. Visser-Groot and S.M. de Boer (Dept. of Radiation Oncology, LUMC, Leiden, NL) for their contributions to this study. 
This research is part of the research programme Open Technology Programme with project number 15586, which is financed by the Dutch Research Council (NWO), Elekta, and Xomnia.
Further, the work is co-funded by the public-private partnership allowance for top consortia for knowledge and innovation (TKIs) from the Dutch Ministry of Economic Affairs.
\end{acks}

\newpage

\appendix

\captionsetup[subfigure]{aboveskip=2pt}

\input{sup/sections/a-approach-technical-details}

\input{sup/sections/b-problem-specification}
\input{sup/sections/c-existing-approaches}
\input{sup/sections/d-full-results}

\bibliographystyle{ACM-Reference-Format}
\bibliography{references.bib}

\end{document}

%% file: sections/0-abstract.tex
\begin{abstract}

Finding a realistic deformation that transforms one image into another, in case large deformations are required, is considered a key challenge in medical image analysis.
Having a proper image registration approach to achieve this could unleash a number of applications requiring information to be transferred between images.
Clinical adoption is currently hampered by many existing methods requiring extensive configuration effort before each use, or not being able to (realistically) capture large deformations.
A recent multi-objective approach that uses the Multi-Objective Real-Valued Gene-pool Optimal Mixing Evolutionary Algorithm (MO-RV-GOMEA) and a dual-dynamic mesh transformation model has shown promise, exposing the trade-offs inherent to image registration problems and modeling large deformations in 2D.
This work builds on this promise and introduces MOREA: the first evolutionary algorithm-based multi-objective approach to deformable registration of 3D images capable of tackling large deformations.
MOREA includes a 3D biomechanical mesh model for physical plausibility and is fully GPU-accelerated.
We compare MOREA to two state-of-the-art approaches on abdominal CT scans of 4 cervical cancer patients, with the latter two approaches configured for the best results per patient.
Without requiring per-patient configuration, MOREA significantly outperforms these approaches on 3 of the 4 patients that represent the most difficult cases.

\end{abstract}

%% file: sections/1-introduction.tex
\section{Introduction}

In recent decades, the field of radiation oncology has experienced rapid developments.
Key to its modern practice are medical images acquired before, during, and after treatment.
Although these images are already guiding clinical decision-making in many ways, the transfer of information between multiple images that feature large deformations or content mismatches has proven to be a hard challenge and has eluded widespread clinical adoption.
In general, the challenge of Deformable Image Registration (DIR) is to find a realistic transformation that matches two or more image spaces to each other, as illustrated in Figure~\ref{fig:general}.
Given this transformation, other metadata could be transferred between images, such as annotated contours~\cite{Loi2018} or 3D radiation dose distributions~\cite{Mohammadi2019}, opening up opportunities to make radiation treatment more precise~\cite{Brock2017}.

The DIR problem consists of three main objectives: an image-based objective (for a visual comparison), a contour-based objective (for an assessment of object contour overlap), and a realism-based objective (to measure the energy required to perform the deformation).
These objectives are conflicting, especially when large deformations and content mismatches are at play~\cite{Alderliesten2015}. 
DIR is therefore an inherently multi-objective problem, making Evolutionary Algorithms (EAs) well-suited for its optimization~\cite{Deb2001}.

\input{figures/tex/illustrations/general-illustration.tex}

A diverse set of approaches to DIR has emerged~\cite{Brock2005,Avants2008,Weistrand2015}.
These all take a single-objective approach, requiring the user to choose the weights associated with the optimization objectives for each use, \textit{a priori}.
This can however hinder clinical adoption, since it has been shown that choosing good weights (and other parameters) for specific patients is difficult in general and can strongly influence registration quality~\cite{Pirpinia2017}.
Even when configured for the best results, many existing approaches struggle with large deformations and content mismatches between images because of limitations of their underlying transformation models and (often gradient-descent-based) optimization techniques.
This shortcoming forms a second obstacle to their translation into clinical workflows.
Therefore, there still is a need for a DIR approach that does not require \emph{a priori} objective weight configuration \emph{and} can tackle large deformations.

The need to configure objective weights \textit{a priori} has previously been addressed by taking a multi-objective approach~\cite{Alderliesten2012}.
This removes the need to select weights for the optimization objectives in a scalarized problem formulation \textit{a priori}, since a set of solutions can be produced that appropriately represents the trade-off between different conflicting objectives, allowing the user to select a solution from this set, \emph{a posteriori}.
To overcome the second obstacle, a flexible dual-dynamic triangular mesh transformation model that allows for inverse-consistent, biomechanical registration has been introduced~\cite{Alderliesten2013}.
This model can match structures on both images to capture large deformations.
The Multi-Objective Real-Valued Gene-pool Optimal Mixing Evolutionary Algorithm (MO-RV-GOMEA) has proven to be effective at performing DIR with this model for 2D images by decomposing the problem into local, partial evaluations~\cite{Bouter2017}.
The Graphics Processing Unit (GPU) is exceptionally well-suited to execute these partial evaluations in parallel, yielding significant speed-ups~\cite{Bouter2021}.
Recently, first steps have been taken to extend this GPU-accelerated approach to 3D images~\cite{Andreadis2022}, for which the benefits of partial evaluations may be even greater due to the increase in the amount of image information (from 65k pixels in 2D to more than 2 million voxels in 3D), leading to more, but also costlier partial evaluations.
While this extended approach has been shown to be capable of solving simple registration problems of single objects, it misses several crucial components required to tackle clinical problems that feature multiple interacting objects.

In this work, we therefore introduce MOREA, the first EA-based Multi-Objective Registration approach capable of registering 3D images with large deformations using a biomechanical model, without requiring \emph{a priori} configuration of objective weights.
In MOREA, a 3D tetrahedral mesh is initialized on interesting structures using a novel custom mesh generation approach, and a repair mechanism for folded meshes is embedded.
With MOREA we furthermore improve on prior modeling strategies~\cite{Andreadis2022} for all objectives to ensure desirable deformations will be achieved.




%% file: figures/tex/illustrations/general-illustration.tex
\begin{figure}[b]
    \vspace{-0.1cm}
    \centering
    \captionsetup[subfigure]{aboveskip=-2pt}
    \begin{subfigure}{.323\linewidth}
      \centering
      \includegraphics[width=\linewidth]{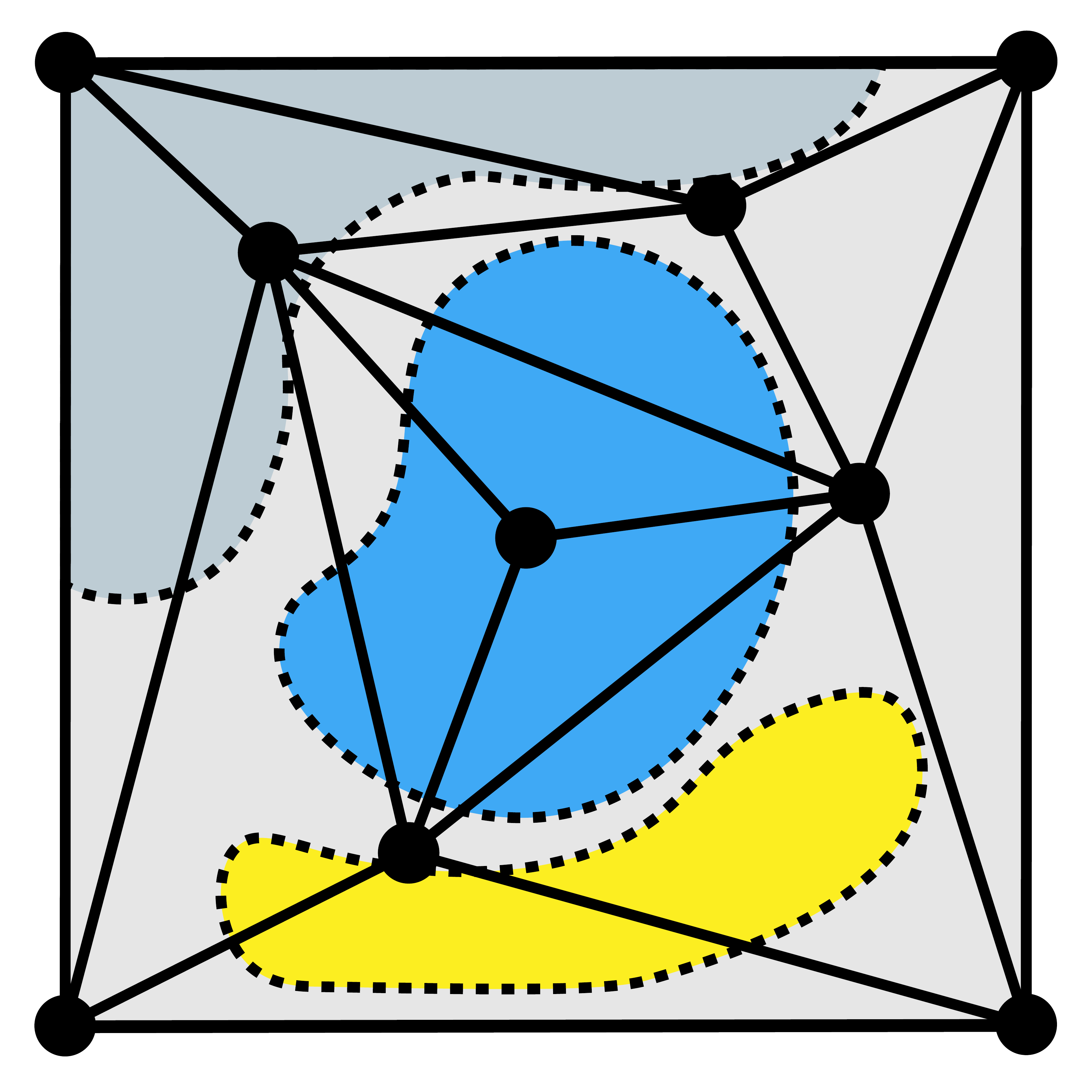}
      \caption{{\normalfont Source image}}
      \label{fig:general:source}
    \end{subfigure}%
    \hspace{0.003\linewidth}
    \begin{subfigure}{.323\linewidth}
      \centering
      \includegraphics[width=\linewidth]{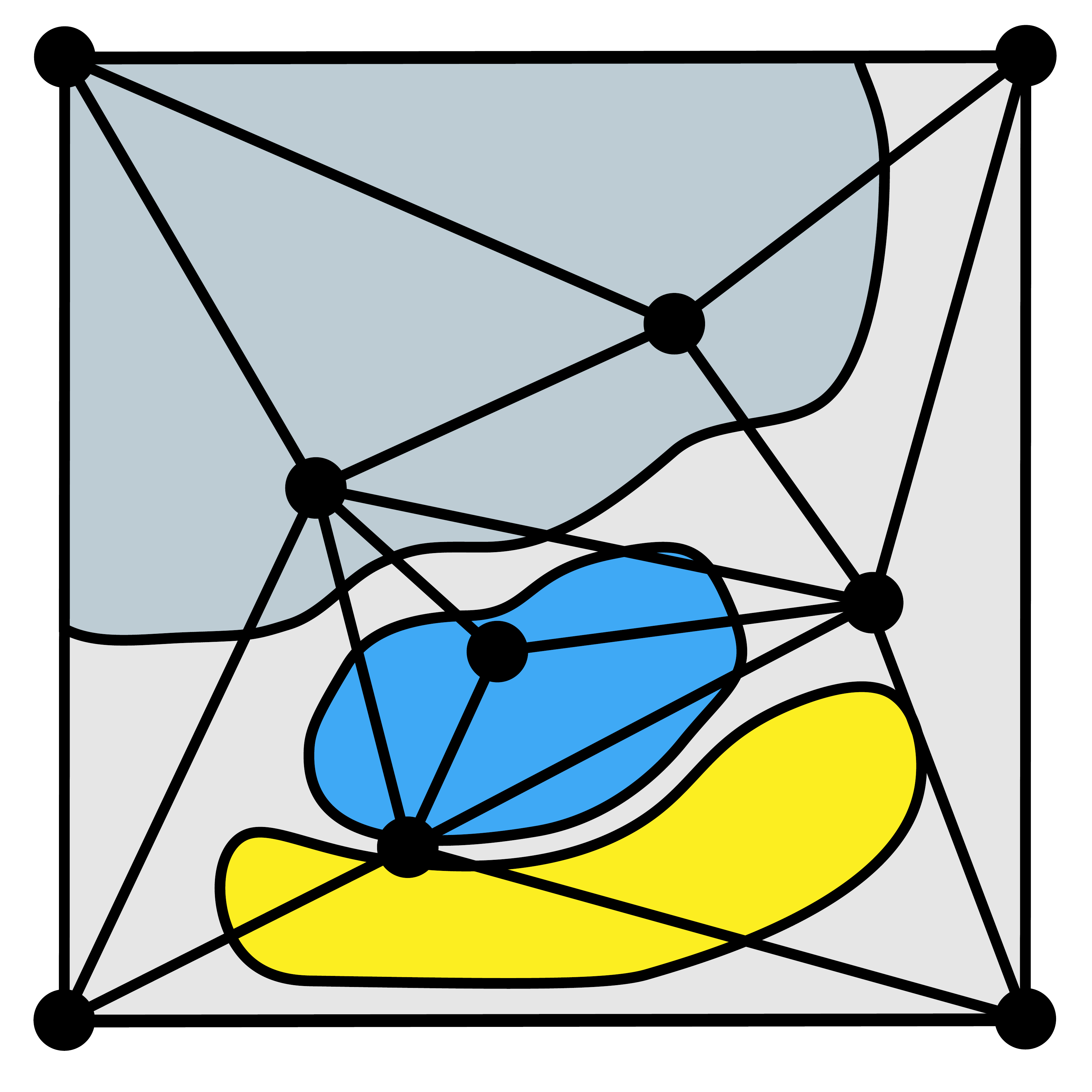}
      \caption{{\normalfont Target image}}
      \label{fig:general:target}
    \end{subfigure}%
    \hspace{0.003\linewidth}
    \begin{subfigure}{.323\linewidth}
      \centering
      \includegraphics[width=\linewidth]{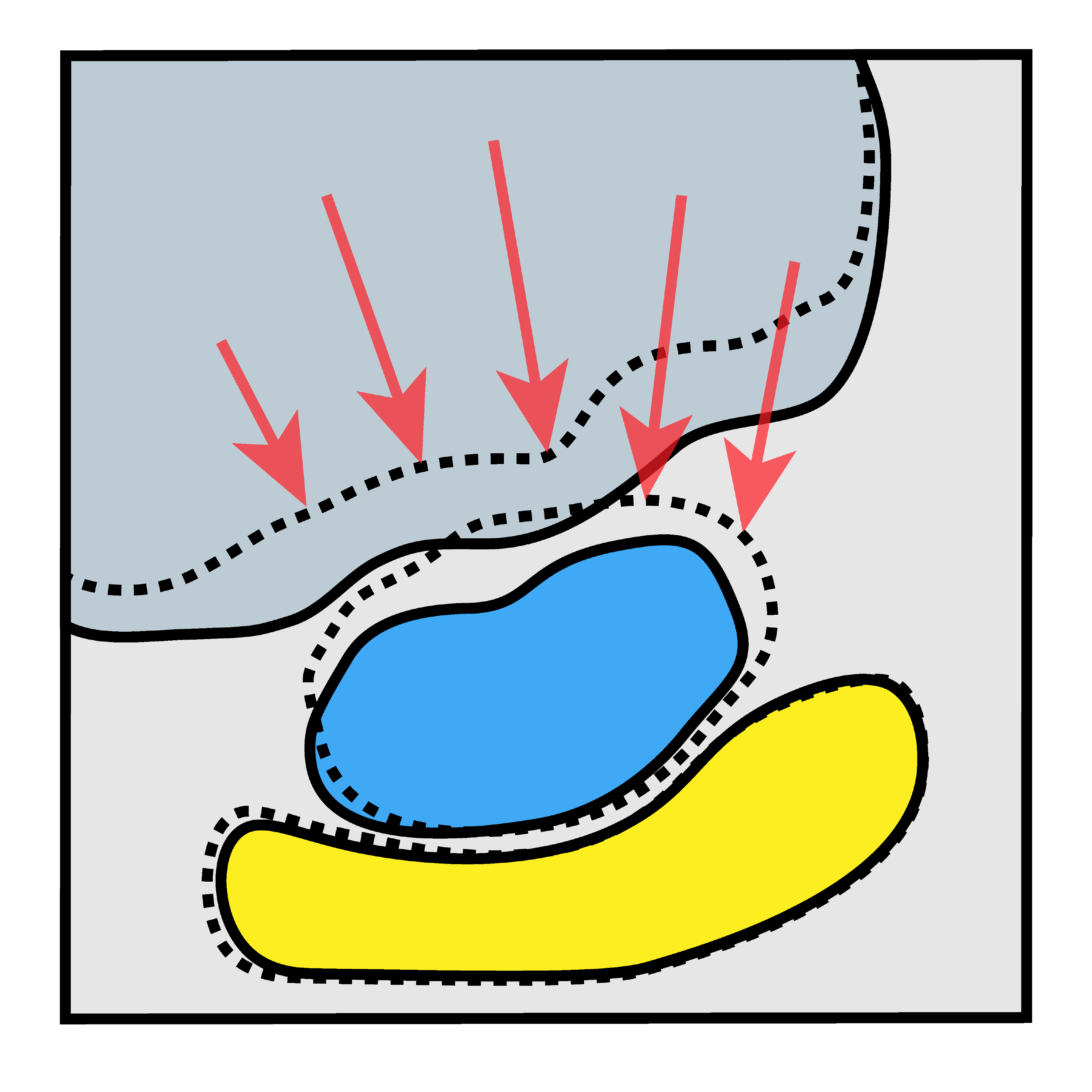}
      \caption{{\normalfont Example registration}}%
      \label{fig:general:deformed}
    \end{subfigure}
    \vspace{-0.3cm}
    \caption{Illustration of two images with large deformations and an example of a deformable image registration with MOREA's dual-dynamic mesh transformation model.}
    \label{fig:general}
\end{figure}

%% file: sections/2-problem.tex
\input{figures/tex/illustrations/competitors.tex}

\section{Deformable Image Registration for Large Deformations}
\label{sec:problem}

In this section, we define the DIR optimization problem (Section~\ref{sec:problem:definition}) and examine existing approaches (Section~\ref{sec:problem:related}).

\subsection{Problem Definition}
\label{sec:problem:definition}

The problem of DIR for a pair of 2 images is to find a non-rigid transformation $T$ that deforms a source image $I_s$ to match a target image $I_t$ as closely as possible~\cite{Sotiras2012}.
We distinguish between \emph{unidirectional} and \emph{symmetric} registration: in unidirectional registration, only $T(I_s) \approx I_t$ is optimized, while in symmetric registration, $T'(I_t) \approx I_s$ is also optimized~\cite{Sotiras2012}.
This can improve the physical viability of the registration.
Another desirable distinction for registrations is \emph{inverse-consistency}~\cite{Sotiras2012}, guaranteeing a one-to-one correspondence between any point in the source image and its corresponding point in the target image.

Registrations can generally be evaluated according to three classes of quality metrics.
\emph{Image intensity} metrics compare the predicted voxel intensity values of $T(I_s)$ to the voxel intensity values of $I_t$, using metrics such as cross-correlation or mutual information~\cite{Khalifa2011}.
\emph{Contour} metrics judge registration accuracy by applying $T$ to pairs of sets of points, representing contours ($C_s$ and $C_t$), and computing the distances between those point sets.
One example is the Chamfer distance~\cite{Beg2005}: for each pair $\langle C_s, C_t \rangle$, the longest minimum distance is calculated between points in $T(C_s)$ and any point in $C_t$.
DIR approaches can also use these contours at initialization time, to build transformation models for use during optimization. 
Finally, \emph{deformation magnitude} metrics express registration realism by measuring the force needed to apply the deformation, using a physical model of the image space~\cite{Fischer2008}.
This can serve as a regularization mechanism, discouraging the registration to overfit.

\subsection{Related Work}
\label{sec:problem:related}

These three quality metrics are conflicting objectives that form a trade-off~\cite{Alderliesten2015}.
A number of single-objective registration approaches have emerged in recent years, typically attempting to deal with  this trade-off by exploring different objective scalarizations.
This however has the downside of having to set objective weights, \emph{a priori}.
We categorize these existing approaches broadly according to the above defined classes of quality metrics, into classes of approaches mainly optimizing for (1) intensity match, (2) contour match, and (3) both matches simultaneously.
These and other features are compared for selected prominent approaches in Table~\ref{tab:competitors}.

An example of the first class, optimizing for intensity match, is the Elastix toolbox~\cite{Klein2010}.
It uses a B-spline based transformation model, which uses Bézier curves to model physical space. 
With this model, Elastix optimizes for intensity, regularized by deformation magnitude metrics.
While this is a good fit for many applications, we observe that registering more complex, large deformations with local discontinuities (such as studied in this work) can be difficult.
The ANTs SyN registration approach~\cite{Avants2008} was conceived to model such large deformations, featuring symmetric, inverse-consistent, and intensity-based registration using time-varying velocity fields.
A third intensity-based approach is the Demons algorithm~\cite{Thirion1998}, using principles from optical flow and Maxwell's Demons for inverse-consistent registration.
A more recent version of this approach also has a mechanism to handle content mismatch between images~\cite{Nithiananthan2012}.
Both the ANTs and Demons approach can in theory flexibly model large deformations, but lack biomechanical modeling capabilities and only operate on image intensity.
This can hamper reliably producing anatomically valid registrations~\cite{Loi2018}.

This is one reason to consider the second class of approaches.
One of these approaches is the Thin-Plate Splines Robust Point Matching approach (TPS-RPM), which deforms contours using a thin-plate spline model~\cite{Chui2000}.
Subsequent work validated this on an abdominal test case, registering a deforming bladder and two surrounding organs~\cite{VasquezOsorio2009}.
There is also a symmetric version of TPS-RPM, which improves robustness on large deformations~\cite{Bondar2010}.
Work conducted in parallel also applies a similar model to contours for abdominal registration problems~\cite{Schaly2004}.
While large deformations can be modeled, the biomechanical plausibility of the transformation is not guaranteed, and objective weights still require configuration.
Another contour-based approach is MORFEUS~\cite{Brock2005}, which registers a mesh representation of imaged objects using a Finite Element Method (FEM) solver.
It has shown promising results on brachytherapy applications in the abdomen~\cite{Rigaud2019}.
Although MORFEUS uses a biomechanical model, which improves realism, it does not take image intensities into account, thus losing detail between object surfaces and relying too heavily on (user-supplied) contours.

Recent work has targeted this shortcoming by proposing a combined contour-based and image-based approach: the ANAtomically CONstrained Deformation Algorithm (ANACONDA)~\cite{Weistrand2015} optimizes a fixed scalarization of image and contour terms by using the quasi-Newton algorithm.
This approach however lacks biomechanical modeling, and also introduces yet another parameter to configure.
Other hybrid attempts have also emerged, such as a combination of the Demons approach with local FEM meshes~\cite{Zhong2012}, or the use of an image-based registration step to derive tissue elasticities that are later used in an FEM-based registration approach~\cite{Li2013}.

In general, we see a gap: an approach that includes all registration aspects in one model.
As Table~\ref{tab:competitors} shows, we target this gap with MOREA by being both image-based and contour-based, featuring biomechanical modeling, and exploiting the multi-objective nature of the DIR problem.
These novelties are made possible by the flexibility and robustness of EAs, which are well-suited to optimize non-differentiable, multi-objective problems.
Additionally, the objective functions include millions of image voxel values and are therefore expensive to compute, calling for hardware acceleration.
Modern model-based EAs such as MO-RV-GOMEA feature excellent GPU compatibility, making them a good fit for optimizing the DIR problem.

%% file: figures/tex/illustrations/competitors.tex
\newcommand{\yes}{{\color{ForestGreen} \cmark}}
\newcommand{\no}{{\color{OrangeRed} \xmark}}

\begin{table*}
    \centering
    \setlength{\tabcolsep}{3pt}
    \begin{tabularx}{\linewidth}{Xccccccc}
        \toprule
        Feature & Elastix~\cite{Klein2010} & ANTs SyN~\cite{Avants2008} & Demons~\cite{Thirion1998} & TPS-RPM~\cite{Chui2000} & ANACONDA~\cite{Weistrand2015} & MORFEUS~\cite{Brock2005} & MOREA \textit{(this work)} \\
        \midrule
        Image-based     & \yes & \yes & \yes & \no & \yes & \no & \yes \\
        Contour-based   & \no & \no & \no & \yes & \yes & \yes & \yes \\
        Biomechanical   & \no & \no & \no & \no & \no & \yes & \yes \\
        Multi-objective & \no & \no & \no & \no & \no & \no & \yes \\
        \bottomrule
    \end{tabularx}
    \vspace{0.05cm}
    \caption{Comparison of selected prominent existing DIR approaches by supported registration features.}
    \vspace{-0.4cm}
    \label{tab:competitors}
\end{table*}

%% file: sections/3-morvgomea.tex
\section{MO-RV-GOMEA}
\label{sec:morvgomea}

The structure of Black-Box Optimization (BBO) problems only gets revealed through repeated function evaluations.
Gray-Box Optimization (GBO) problems, on the other hand, have a (partly) known problem structure, which can be exploited during optimization.
The GOMEA suite of EAs has proven to be exceptionally well suited for efficiently solving both benchmark and real-world GBO problems~\cite{Thierens2011}.
Its extension to multi-objective, real-valued problems, MO-RV-GOMEA~\cite{Bouter2021a}, has even found real-world adoption in clinical practice for prostate brachytherapy treatment planning~\cite{Bouter2019,Barten2023}.
We give an overview of the key working principles of MO-RV-GOMEA here.
A detailed description may be found in literature~\cite{Bouter2017b}.

Non-dominated solutions are preserved across generations in an elitist archive with a pre-specified capacity~\cite{Luong2012}.
Each generation starts with the selection of a subset of non-dominated solutions from the current population.
This selection is clustered into $k$ equally sized clusters.
For each cluster, MO-RV-GOMEA employs a linkage model that describes dependence relations between variables using a set of dependent variable sets, called Family of Subset (FOS) elements. 
This linkage model can be learned during optimization in a BBO setting, but in MOREA, we employ a static linkage model based on topological proximity of variables (see Section~\ref{sec:approach:init:mesh}).
Variation then proceeds by considering variables in FOS elements jointly in a procedure called \emph{optimal mixing}.
In this step, distributions are estimated for each FOS element in each cluster, and new, partial solutions are sampled from these distributions.
Newly sampled partial solutions are evaluated and accepted if their insertion into the parent solution results in a solution that dominates the parent solution or that is non-dominated in the current elitist archive.

%% file: sections/4-approach.tex
\section{Approach}
\label{sec:approach}

The approach outlined in this work builds on the recently proposed multi-objective approach for 3D images~\cite{Andreadis2022}.
In this section, we present the new techniques we have added, in modeling the problem~(Section~\ref{sec:approach:modeling}), initializing the population of solutions~(Section~\ref{sec:approach:init}), and optimizing the deformations~(Section~\ref{sec:approach:optimization}).

\subsection{Modeling}
\label{sec:approach:modeling}

\subsubsection{Enhancing realism with tissue-specific elasticities}
Adjacent work has indicated that using tissue-specific elasticities, instead of assuming one homogeneous elasticity for the entire image region, can enhance the realism of resulting deformations~\cite{Wognum2013,Rigaud2019}.
Following this insight, we extend the deformation magnitude objective used in existing work~\cite{Andreadis2022} by computing an elasticity factor for each tetrahedron, based on its underlying image region.
Implementation details for this computation are provided in Appendix~A.
We observe in exploratory experiments that this leads to better registration outcomes (see Appendix Section~C.3.1).

To compute the deformation magnitude objective, we consider all corresponding edges $e_s$ and $e_t$ of each tetrahedron $\delta \in \Delta$, belonging to the mesh on the source image and the mesh on the target image, respectively.
This includes 4 spoke edges that better capture flattening motion, giving a total of 10 edges per tetrahedron~\cite{Andreadis2022}.
Given the tetrahedron-specific elasticity constant $c_\delta$, the objective is computed as follows:

\vspace{-0.4cm}

\begin{align*}
f_{\text{\emph{magnitude}}} = \frac{1}{10 |\Delta|} \sum_{\delta \in \Delta} \left[ \sum_{(e_s,e_t) \in E_\delta}c_\delta (\lVert e_s \rVert - \lVert e_t \rVert)^2 \right] 
\end{align*}

\vspace{-0.1cm}

\setcounter{figure}{2} 
\input{figures/tex/illustrations/parallelization.tex}

\subsubsection{Robustly estimating image similarity}
The intensity objective we use is defined as a voxel-to-voxel comparison by taking the sum of squared intensity differences, with special handling for comparisons of foreground (i.e., non-zero intensity) and background (i.e., zero intensity) voxels.
We use a random sampling technique that is well-suited for GPU acceleration (defined in detail in Appendix~A).
Using the set of all sampled image points on both images, $P_s$ and $P_t$, and image intensities of source image $I_s$ and target image $I_t$, the objective is defined as follows:


\begin{align*}
f_{\text{\emph{intensity}}} & = \frac{1}{|P_s| + |P_t|} \left[ \sum_{p_s \in P_s} h(p_s, T(p_s)) + \sum_{p_t \in P_t} h(p_t, T'(p_t)) \right] \\
h(p_s,p_t) & = \begin{cases}
  (p_s - p_t)^2  & p_s > 0 \wedge p_t > 0 \\
  0 & p_s = 0 \wedge p_t = 0 \\
  1 & \text{otherwise}
\end{cases}
\end{align*}



\subsubsection{Approximating the guidance error}
In contrast to previous work where an exact guidance measure was used as one of the objectives~\cite{Andreadis2022}, in this work we have opted to introduce a measure that is an approximation thereof that can be much more efficiently computed using the GPU-accelerated sampling method that we already use for the calculation of the values for the image similarity objective. 
Preliminary experiments showed very similar results (when looking at the voxel displacement fields), also because a perfect guidance error is not necessarily the best solution.
In Appendix~A, we provide details regarding the implementation.

MOREA's guidance objective is computed at positions $P_s$ and $P_t$, using the set $G$ of all point set pairs $\langle C_s, C_t \rangle_i$ and the minimal point-to-point-set distance $d(p, C)$.
The total number of guidance points is indicated as $|G_s|$ and $|G_t|$, and a truncation radius as $r$.
The guidance objective is now defined as follows:

\vspace{-0.2cm}

\begin{align*}
f_{\text{\emph{guidance}}} = & \frac{1}{|P_s|+|P_t|}  \sum_{\langle C_s, C_t \rangle \in G} \Biggr[ \\
& \frac{|C_s|}{|G_s|} g(P_s, T, C_s, C_t) + \frac{|C_t|}{|G_t|} g(P_t, T', C_t, C_s) \Biggr] \\
g(P, \Phi, C, C') = & \sum_{\substack{p \in P \\ d(p, C) < r}} \Biggr[ \frac{r - d(p, C)}{r} (d(p, C) - d(\Phi(p), C'))^2 \Biggr] \\
\end{align*}

\vspace{-0.4cm}

\setcounter{figure}{1} 
\input{figures/tex/illustrations/constraint-violations.tex}
\setcounter{figure}{3} 

\subsubsection{Rapidly computing constraints}
MOREA's solutions represent meshes with hundreds of points, which can easily get entangled into folded configurations.
Such constraint violations should be prevented, to uphold the guarantee of inverse-consistency.
Prior work~\cite{Andreadis2022} used a strategy that proved error-prone in more complex meshes.
MOREA includes a novel fold detection method that is based on an observed phenomenon: a mesh fold will cause the sign of at least one tetrahedron's volume to change, as illustrated in Figure~\ref{fig:constraints} (the figure is in 2D, but this also holds in 3D).
Our method uses this phenomenon to detect folds and to measure their severity, opening up repair opportunities (see Section~\ref{sec:approach:optimization:repair}).
Implementation details for our method are provided in Appendix~A.

\subsection{Initialization of Registration Solutions}
\label{sec:approach:init}

Significant performance gains can be obtained if the initial guesses given to the optimizer are closer to desirable objective space regions than a random guess or grid-like initializations~\cite{Bosman2016}.
We introduce two techniques that provide such initial guesses.

\subsubsection{Exploiting problem structures with mesh initialization}
\label{sec:approach:init:mesh}
We initialize the meshes to align with objects in the image, adapting an existing method for 2D images~\cite{Bosman2016} and expanding it to facilitate parallelization on the GPU.
First, we place points on the contours of objects in the source image to capture their shape (see Fig.~\ref{fig:parallelization:points}).
We choose these points by greedily taking a spread-out subset from the contour annotations also used for the guidance objective, as well as a small fraction of randomly chosen points across the image.
Then, we perform a Delaunay tetrahedralization on these points, using the TetGen suite~\cite{Hang2015} (see Fig.~\ref{fig:parallelization:edges}).
This yields a mesh that we duplicate to the target image space to complete the dual-dynamic transformation model.

As laid out in Section~\ref{sec:morvgomea}, MO-RV-GOMEA evaluates groups of variables (i.e., FOS elements) jointly during variation.
Exploratory experiments have shown that using edges as FOS elements (i.e., groups of two connected points, with the variables encoding their coordinates), is beneficial for this problem.
If two FOS elements are completely independent because their variables are not needed for the partial evaluation of each set, variation and evaluation for these FOS elements can be done in parallel.
We conduct two further steps to facilitate parallel evaluation and optimization on the GPU.
First, we execute a greedy set cover algorithm\footnote{Source: \url{https://github.com/martin-steinegger/setcover}} to find a subset of edges that covers all points (see Fig.~\ref{fig:parallelization:set-cover}), so that each variable (point coordinate) undergoes variation.
We could alternatively use all edges, but this would lead to points being included in several FOS sets and thus undergoing variation multiple times per generation.
For parallelization purposes, it is more efficient to select an (approximately) minimal set of edges.

Given the edge subset found by the set cover, we now determine which FOS elements can be safely optimized in parallel.
For this, we build an interaction graph based on topological proximity~\cite{Bouter2021}, where two elements are connected if their sets of \emph{dependent tetrahedra} overlap, i.e., the tetrahedra that are reevaluated when an element is changed (see Fig.~\ref{fig:parallelization:dependencies}).
Given this graph, parallel groups are created with the DSATUR graph coloring algorithm~\cite{Brelaz1979} (see Fig.~\ref{fig:parallelization:graph-coloring}).
The dependent tetrahedra of each parallel group can be evaluated in parallel on the GPU, which has been proven to lead to speed-ups of more than 100x on 2D images~\cite{Bouter2021}.

Tetrahedral mesh quality can further be improved by specifying surfaces that should be included in the generated mesh.
We apply this principle to the bladder by generating a surface mesh using the Marching Cubes algorithm.
We then specify its triangular surfaces as constraints to the mesh generation algorithm, ensuring that bladder surface triangles are included in the mesh.
Exploratory experiments show superior performance when using this step (see Appendix~B.3.1).

\subsubsection{Ensuring diversity in initial population}
To promote diversity in the initial population, prior work generates random deviations for each point in the mesh, starting at a grid-initialized solution~\cite{Andreadis2022}.
We observe that this method can produce many folded mesh configurations in generated solutions, which get discarded and thus hamper convergence speed.
In this work, we use a radial-basis-function approach to introduce large deformations free of mesh folds.
Implementation details on how these fields are generated and applied to solution meshes are provided in Appendix~A.

\subsection{Repairing and Steering}
\label{sec:approach:optimization}

During optimization, we apply two techniques to improve the quality of solutions obtained, and the time needed to reach them.

\subsubsection{Repairing infeasible solutions}
\label{sec:approach:optimization:repair}
By default, infeasible solutions (i.e., solutions with either of the two meshes having one or more folds) are discarded.
This, however, can hamper the creation of high-quality offspring, as infeasible solutions may still provide useful information for higher-quality search space regions. 
We therefore devise a repair method that attempts to reverse folds on a point-by-point basis.
For each point in a folded tetrahedron, the method mutates the point using a Gaussian distribution scaled by its estimated distance to the surrounding 3D polygon.
After 64 samples, the change with the best constraint improvement is selected, if present.
If all samples result in a deterioration, repair is aborted.
The repair process for one point is illustrated in Figure~\ref{fig:constraints:repair}.


\subsubsection{Applying pressure with adaptive steering}
In general, an approximation set should be as diverse as possible while resembling the Pareto set as closely as possible.
In practice, however, not all regions of the Pareto front are of equal interest to users.
A user conducting medical DIR for images with large deformations is typically not interested in solutions with a small deformation magnitude.
The user is actually most interested in solutions with good guidance objective values, and we would like the algorithm to \emph{steer} its search towards that region in the objective space.
Following earlier work~\cite{Alderliesten2015}, we implement an adaptive steering strategy, which steers the front towards high-quality guidance solutions after an exploration period of 100 generations.
Given the best guidance objective value $s_G$ of any solution in the elitist archive, we only preserve solutions with guidance objective values between $[s_G; 1.5 s_G]$, i.e., this becomes a hard constraint.

%% file: figures/tex/illustrations/parallelization.tex
\newcommand{\aaa}{1000}
\begin{figure*}[b]
    \vspace{0.2cm}
    \centering
    \begin{subfigure}{.18\linewidth}
      \centering
      \includegraphics[width=\linewidth]{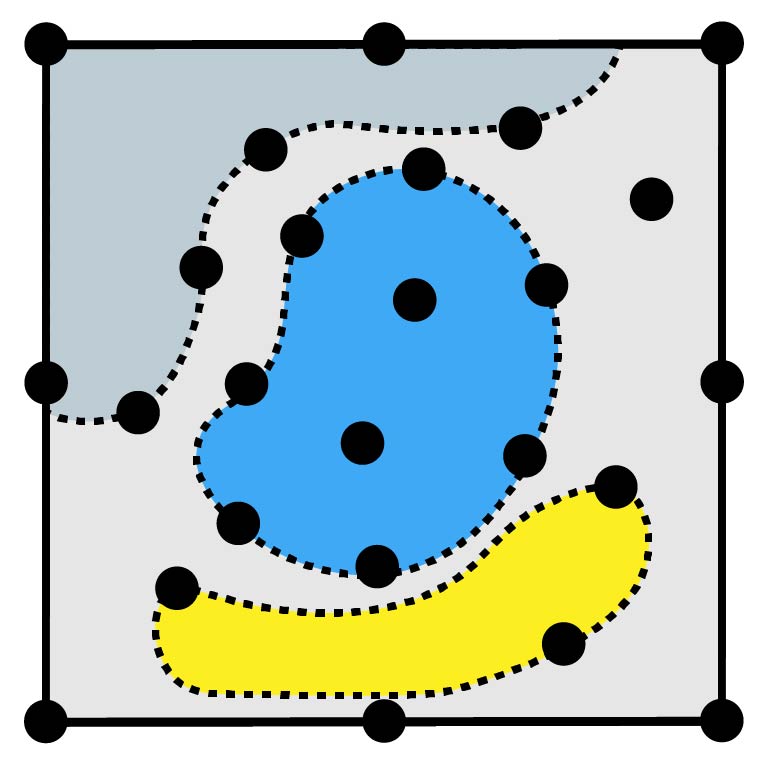}
      \caption{{\normalfont Points placed on potentially interesting positions.}}
      \label{fig:parallelization:points}
    \end{subfigure}%
    \hspace{0.02\linewidth}
    \begin{subfigure}{.18\linewidth}
      \centering
      \includegraphics[width=\linewidth]{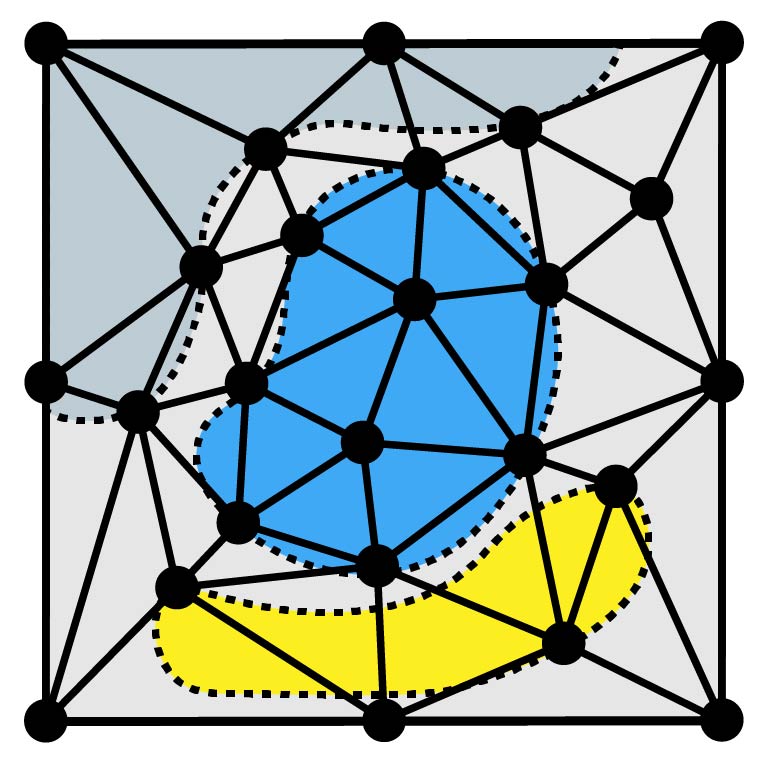}
      \caption{{\normalfont Custom mesh derived from these points.}}
      \label{fig:parallelization:edges}
    \end{subfigure}%
    \hspace{0.02\linewidth}
    \begin{subfigure}{.18\linewidth}
      \centering
      \includegraphics[width=\linewidth]{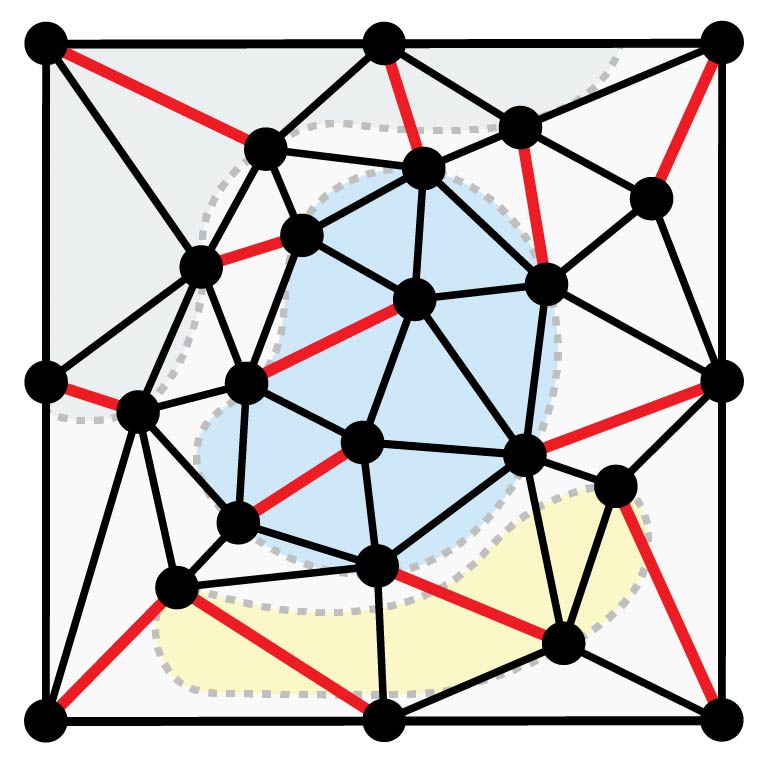}
      \caption{{\normalfont Edges selected for variation through set cover.}}%
      \label{fig:parallelization:set-cover}
    \end{subfigure}%
    \hspace{0.02\linewidth}
    \begin{subfigure}{.18\linewidth}
      \centering
      \includegraphics[width=\linewidth]{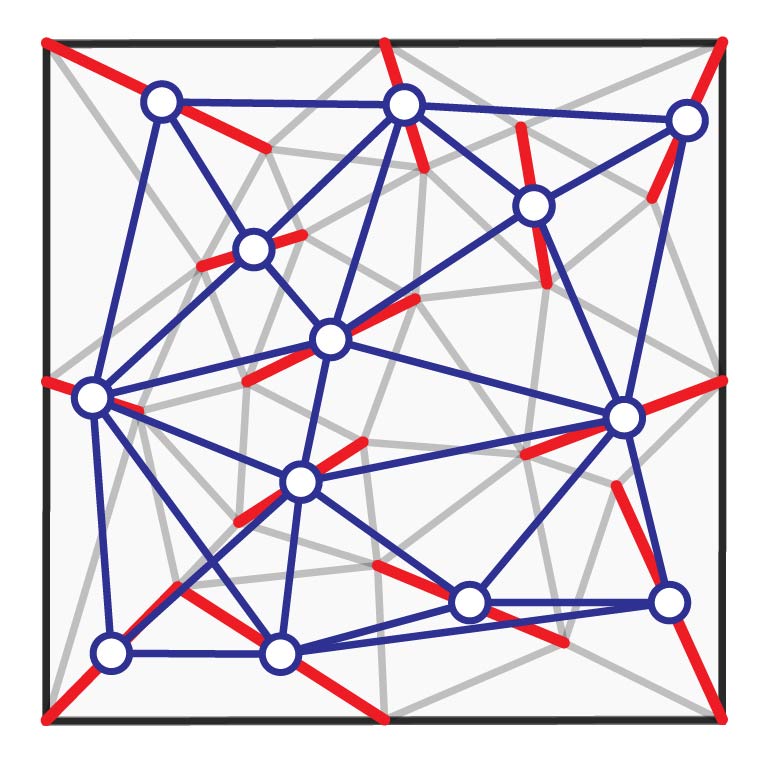}
      \caption{{\normalfont Interaction graph (blue) between selected edges.}}%
      \label{fig:parallelization:dependencies}
    \end{subfigure}%
    \hspace{0.02\linewidth}
    \begin{subfigure}{.18\linewidth}
      \centering
      \includegraphics[width=\linewidth]{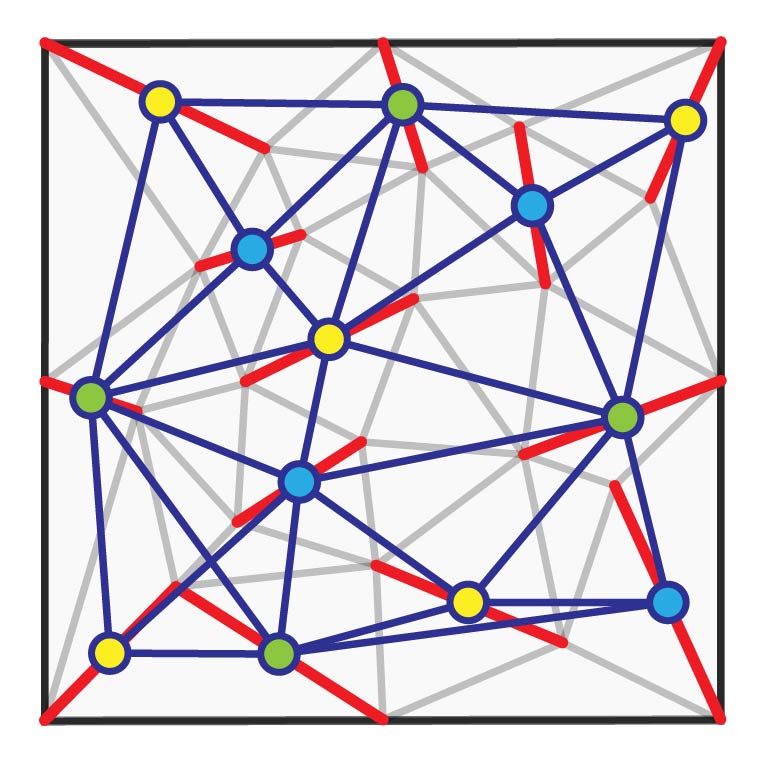}
      \caption{{\normalfont Graph coloring computed on interaction graph.}}%
      \label{fig:parallelization:graph-coloring}
    \end{subfigure}
    \vspace{-0.3cm}
    \caption{2D illustration of the mesh initialization process, which produces a custom mesh and determines which groups of edges (i.e., FOS elements) can be optimized in parallel. Selected edges are highlighted in red, interaction edges in blue.}
    \label{fig:parallelization}
    \vspace{-0.1cm}
\end{figure*}

%% file: figures/tex/illustrations/constraint-violations.tex
\begin{figure}
    \centering
    \begin{subfigure}{.32\linewidth}
      \centering
      \includegraphics[width=\linewidth]{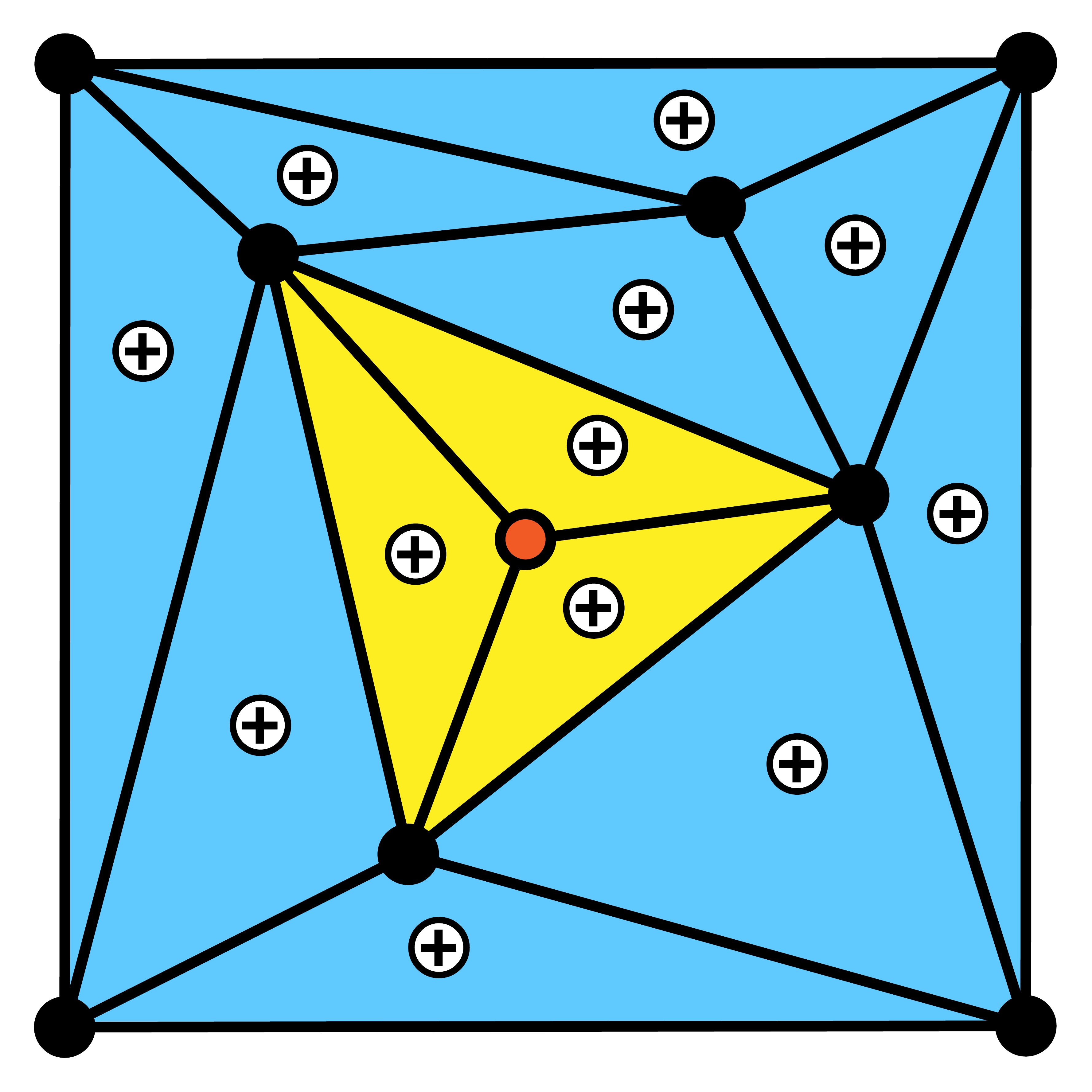}
      \caption{{\normalfont The initial configuration, with positive area signs for each triangle.}}
      \label{fig:constraints:initial}
    \end{subfigure}%
    \hspace{0.009\linewidth}
    \begin{subfigure}{.32\linewidth}
      \centering
      \includegraphics[width=\linewidth]{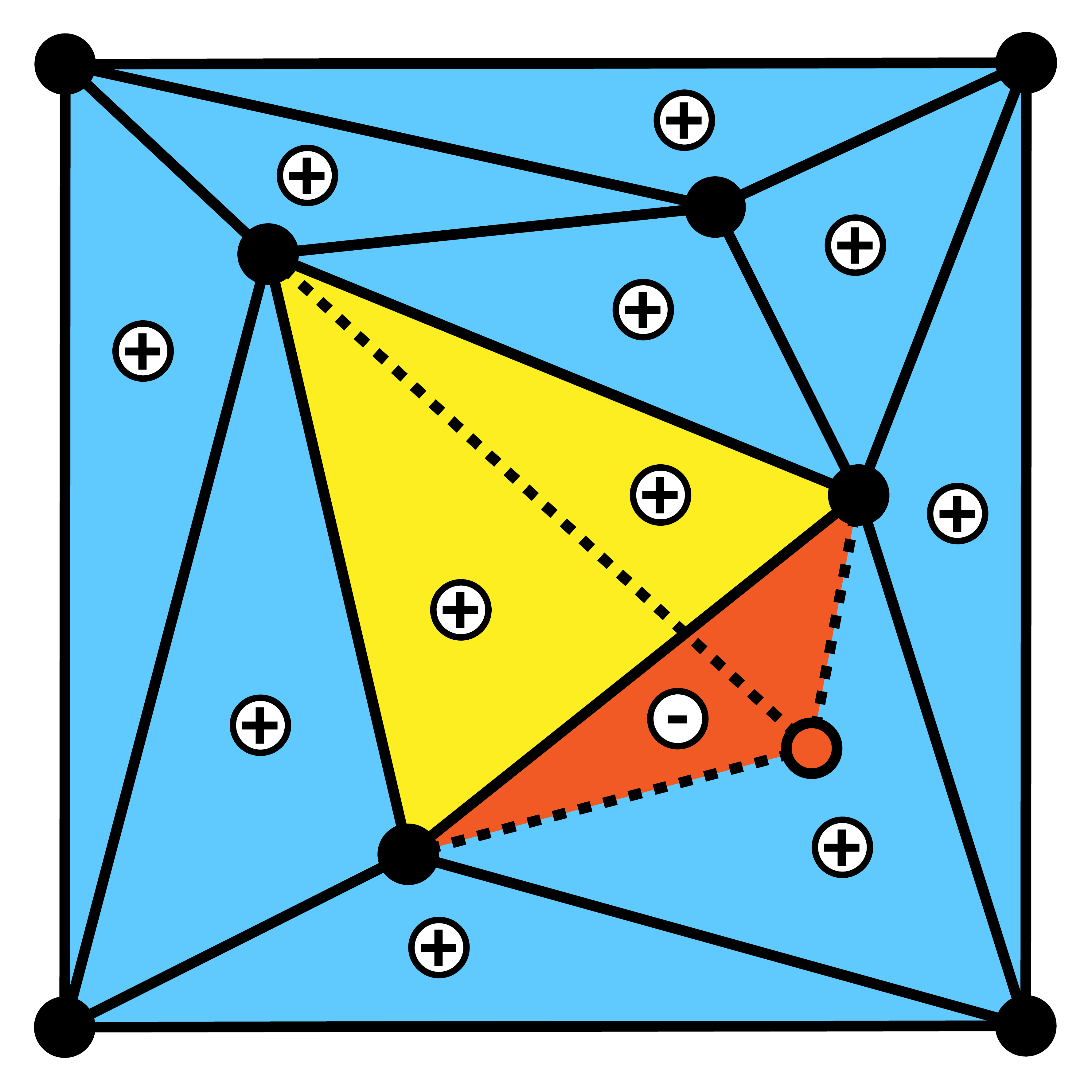}
      \caption{{\normalfont The fold, detected by a sign change in the folded (red) triangle.}}
      \label{fig:constraints:with-constraint}
    \end{subfigure}%
    \hspace{0.009\linewidth}
    \begin{subfigure}{.32\linewidth}
      \centering
      \includegraphics[width=\linewidth]{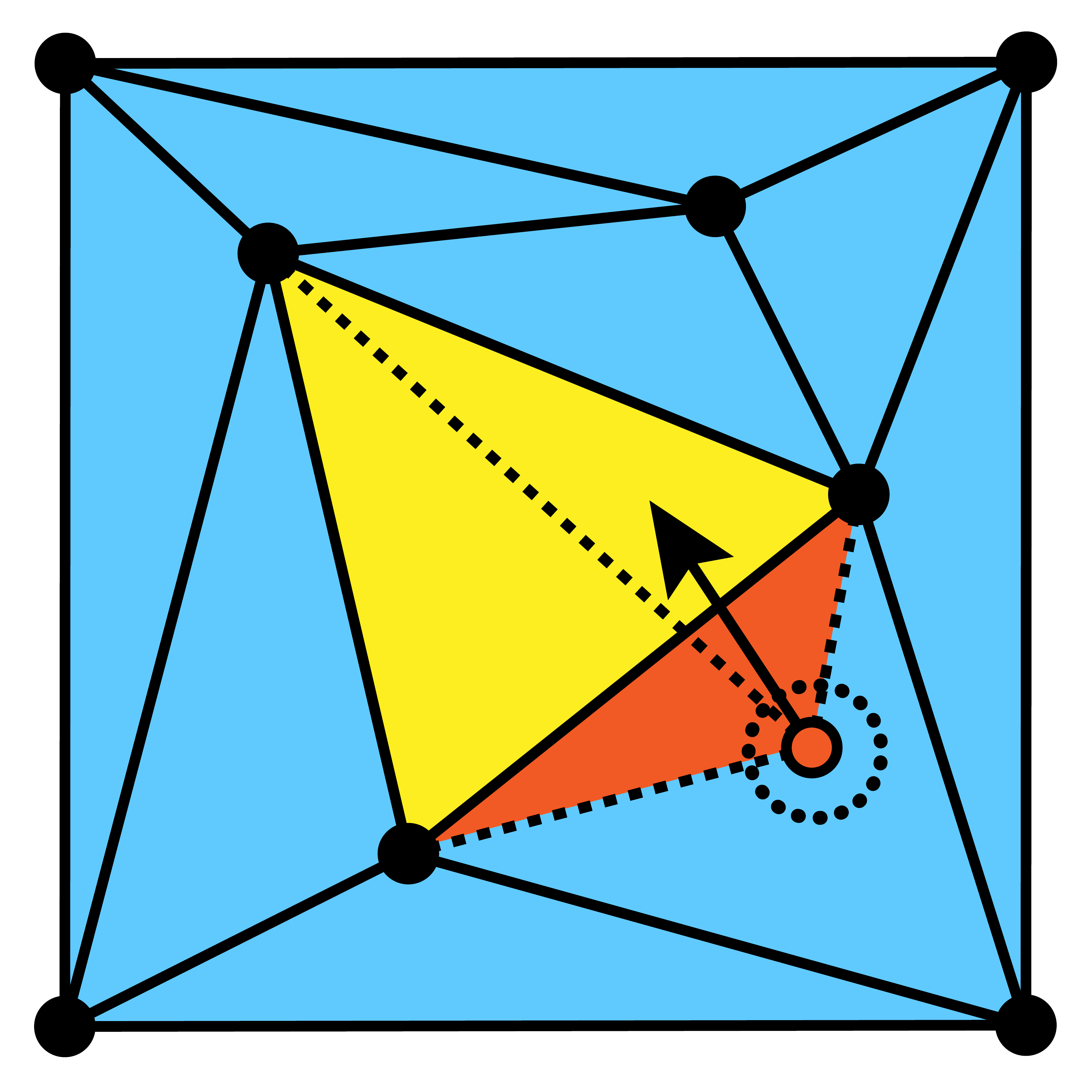}
      \caption{{\normalfont The repair method, resolving the fold by moving the red point.}}%
      \label{fig:constraints:repair}
    \end{subfigure}
    \vspace{-0.4cm}
    \caption{2D illustration of a mesh configuration with and without a constraint violation (fold). One of the triangles is folded, due to the red point having moved outside the central triangle, colored yellow. The folded area is colored red.}
    \label{fig:constraints}
    \vspace{-0.4cm}
\end{figure}

%% file: sections/5-experiments.tex
\section{Experiments}
\label{sec:experiments}

We compare MOREA to existing state-of-the-art registration approaches.
Due to the complexity of the problem, we do not impose one time limit on all approaches, but rather ensure that they have (reasonably) converged.
We repeat all approaches with all configurations 5 times, seeded reproducibly.
All MOREA registration runs are run on Dell Precision 7920R machines with NVIDIA RTX A5000 GPUs.
Additional information on experimental setup and results is provided in the appendix.

\subsection{Registration Problems}
\label{sec:experiments:problems}

We test all approaches on 4 clinical registration problems with large deformations (see Table~\ref{tab:reference-renders}).
We retrospectively select two successive Computerized Tomography (CT) scans of the abdominal area of cervical cancer patients, acquired for radiation treatment planning purposes, with a Philips Brilliance Big Bore scanner.
On the first CT scan, the bladder of the patient is filled, and on the second scan, the bladder is empty and thus has shrunken significantly.
This large deformation is challenging to register correctly while respecting the surrounding organs (e.g., rectum and bowel) and bony anatomy.
Patients 1--3 represent common cases in clinical practice, exhibiting large deformations and little to no margin between bladder and bowel in the full-bladder scan.
The bladder of Patient 4 largely preserves its shape and exhibits a wide margin between bladder and bowel, making registration easier.
This case, however, is also rarer in practice, and therefore less representative.

The axial slices of the CT scans have a thickness of 3\textit{mm}, with in-slice resolutions ranging between $(0.86,0.86)$\textit{mm} and $(1.07,1.07)$\textit{mm}.
Each scan is resampled to $(1.5,1.5,1.5)$\textit{mm} for consistency.
Afterward, each scan pair is rigidly registered (i.e., translated, rotated, or scaled linearly) to align the bony anatomies of both scans, using bone contours delineated by a radiation therapy technologist (RTT).
Each pair is cropped to an axis-aligned bounding box surrounding the bladder with a 30\textit{mm} margin, taking the maximal bounds from both images.
This restricts the registration to the region where treatment was delivered, including the surrounding organs at risk.

\input{figures/tex/results/reference-renders.tex}

Contours of key organs in each scan have been annotated by an RTT and verified by a radiation oncologist.
The sets of points defining these contours serve as input to the guidance objective of MOREA.
We also use these clinical contours to generate binary masks for each organ and the bones by filling 2D polygonal estimates formed by contours on each slice.
As is common in practice, these contours can overlap, since organs are delineated independently and are often surrounded by a small safety margin.
Registration approaches therefore need to be robust enough to handle this overlap.
Several anatomically relevant corresponding landmarks have been annotated by an RTT and verified by a radiation oncologist on both scans, for evaluation purposes (see Appendix~D).

\subsection{Registration Approaches}
\label{sec:experiments:approaches}

We consider a number of existing, popular registration approaches for which executable code is available.
For these approaches, we follow a two-phase configuration process.
First, we explore relevant coarse-grained settings for a single patient scan pair (of Patient~1), to find a suitable configuration for the imaging modality and problem difficulty.
Then, we conduct fine-grained configuration on the remaining settings (e.g., objective scalarization weights) for each patient scan pair.
We describe the resulting configuration for each approach below, including the general coarse-grained configuration of MOREA.
A detailed overview of how we reached these configurations, with additional configuration experiments, can be found in Appendix~C.

\subsubsection{Elastix}
We configure Elastix to conduct a regularized, multi-resolution~\cite{Unser1993} image registration.
Recommended settings\footnote{Based on an official parameter settings database: \url{https://elastix.lumc.nl/modelzoo/}} did not yield satisfactory results on our scans, therefore we first register composite mask images onto each other for each patient.
This is used as starting point for optimization on the original image intensities.
As a fine-grained configuration step for each patient, we configure the weight assigned to the deformation magnitude objective in a fixed sweep of exponentially increasing weights of $[0, 0.001, 0.01, \ldots, 10.0]$, as is done in related work~\cite{Bondar2010}.

\subsubsection{ANTs SyN}
For the ANTs SyN algorithm, the recommended settings\footnote{Based on technical documentation: \url{https://github.com/ANTsX/ANTs/wiki/Anatomy-of-an-antsRegistration-call}} for multi-resolution registration also were not satisfactory, which led us to conduct initial configuration experiments with several key parameters, listed in Appendix~C.
We also add a composite mask in an additional image channel that is registered alongside the image.
For each patient, we test the same regularization weight of the overall deformation by testing the same weights as for Elastix.

\subsubsection{This work: MOREA}
MOREA uses a single-resolution approach and is configured to generate a mesh of 600 points (i.e., the problem is 3600-dimensional), using the strategies for mesh generation described in Section~\ref{sec:approach:init}.
We set the elitist archive capacity to 2000 and use 10 clusters during optimization, with a runtime budget of 500 generations, during which the EA converges (see Appendix~D).
As MOREA is a multi-objective approach returning an approximation set of registrations, we do not need to configure it further for each patient.

\subsection{Evaluation of Registrations}
\label{sec:experiments:evaluation}

Solutions to complex registration problems, such as the problems in this study, require a multi-faceted evaluation.
Below, we outline two main methods for evaluating registrations: surface-based accuracy and visual inspection.
Additional methods are described in Appendix Section~B.2 and applied in Appendices~C and~D.

\input{figures/tex/results/best-renders.tex}

\input{figures/tex/results/dvfs.tex}

\subsubsection{Surface-based registration accuracy}
A key part of evaluating registration accuracy is to assess how well the surfaces (contours) of objects align~\cite{Brock2017}.
We use the Hausdorff distance, which represents the largest minimal distance between any two points on two object surfaces.
This can be interpreted as the severity of the worst surface match.
To account for potential deformation inaccuracies at the border regions of the image, we discard a margin of 15\textit{mm} on each side for the computation of this metric.
Since this is smaller than the earlier cropping margin of 30\textit{mm}, the bladder and regions around it are left untouched by this second crop.

\subsubsection{Visual inspection}
Surface-based accuracy analysis is complemented by a visual inspection, since a registration with a good contour match can still have undesirable deformations in regions between contours.
This inspection includes viewing slices of the target image overlaid with the source contours transformed using the computed forward DVF of the registration.
To also inspect the deformation between contours, we also visualize the full deformation:
First, we render the DVF itself with a quiver plot.
Second, we overlay a regular grid onto a slice and deform it with the DVF, which gives a different perspective.

\subsection{Comparison of Registrations}
All registration solutions from all approaches are compared using the same evaluation pipeline, to ensure a fair comparison.
Each approach is configured to output its registrations in the form of a forward and an inverse DVF, which define the deformation on the source and the target image, respectively.
Existing approaches either directly or indirectly can be configured to output such DVFs.
For MOREA, we rasterize the deformation encoded by the two deformed meshes of a solution, using an existing rasterization method~\cite{Gascon2013}.
Since we are comparing single-objective approaches to a multi-objective approach (MOREA), we need to select solutions from MOREA's approximation set.
We conduct this \emph{a posteriori} selection by starting at the solution with the best guidance objective value and manually navigating through the approximation front to find a solution with a good trade-off between contour quality and realism.

We also conduct statistical testing using the two-sided Mann-Whitney U test (a standard non-parametric test) to compare MOREA to ANTs and Elastix.
The Hausdorff distance of the bladder contour is used as the test metric, as it describes the largest deforming organ.
To correct for multiple tests in the pair-wise comparisons, we apply Bonferroni correction to the $\alpha$-level and reduce it from 0.05 to 0.025.

%% file: figures/tex/results/reference-renders.tex
\newcommand{\referencerenderwidth}{3.55cm}

\begin{table}
    \centering
    \setlength\tabcolsep{0pt}
    \begin{tabularx}{\linewidth}{Xcc}
        \toprule
        Instance & Source & Target \\
        \midrule
        Patient 1 & \includegraphics[width=\referencerenderwidth,valign=m]{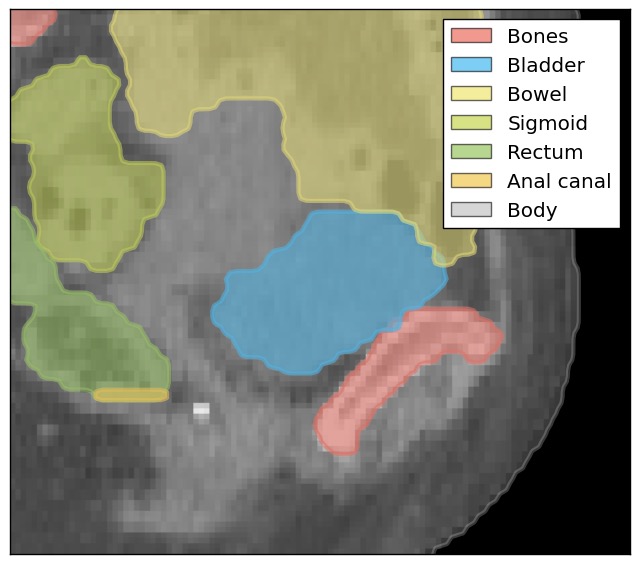} & \includegraphics[width=\referencerenderwidth,valign=m]{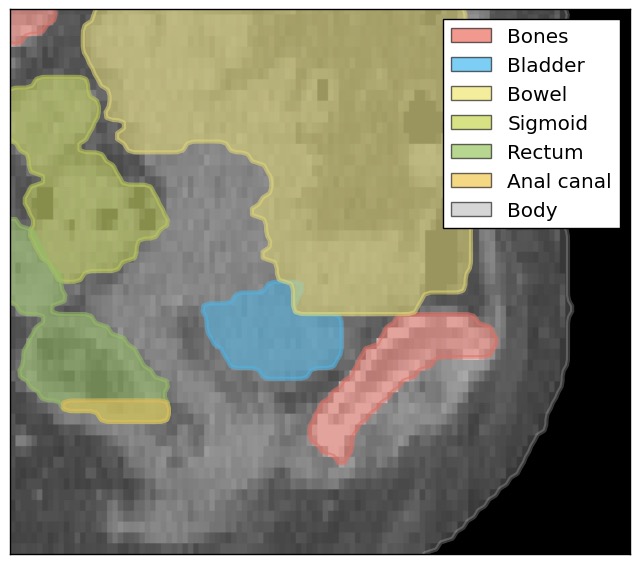}\vspace{0.05cm}\\
        
        Patient 2 & \includegraphics[width=\referencerenderwidth,valign=m]{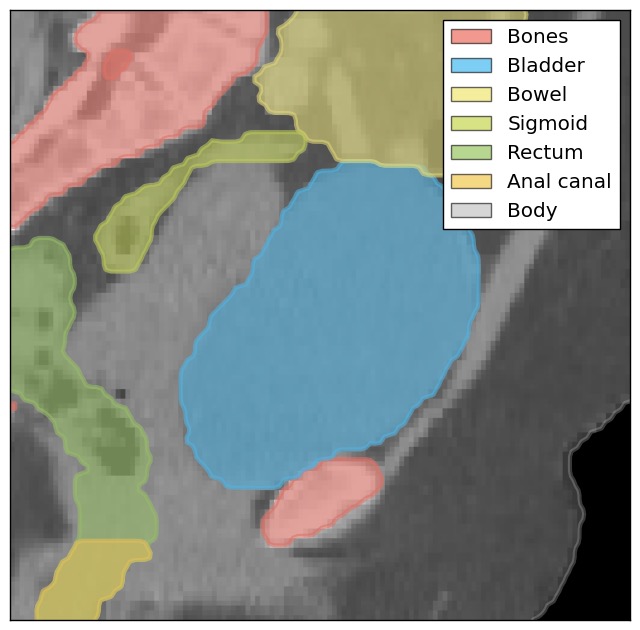} & \includegraphics[width=\referencerenderwidth,valign=m]{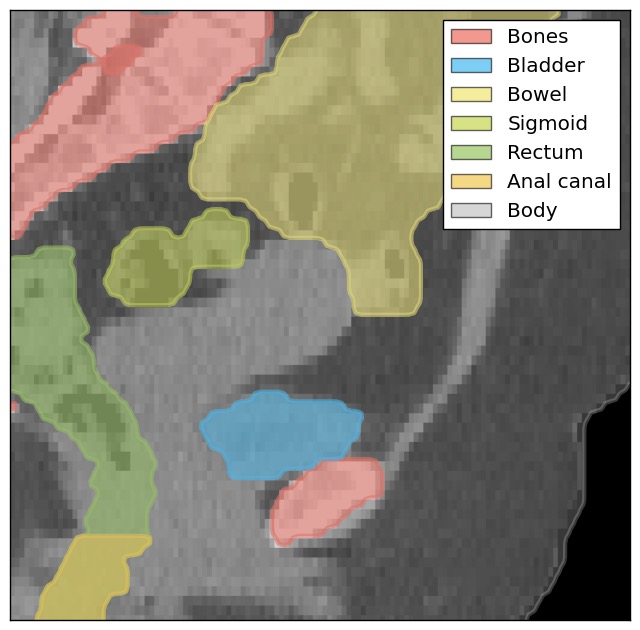}\vspace{0.05cm}\\
        
        Patient 3 & \includegraphics[width=\referencerenderwidth,valign=m]{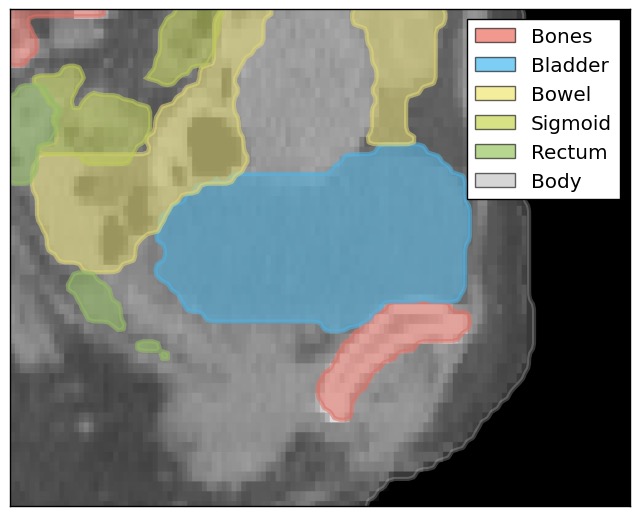} & \includegraphics[width=\referencerenderwidth,valign=m]{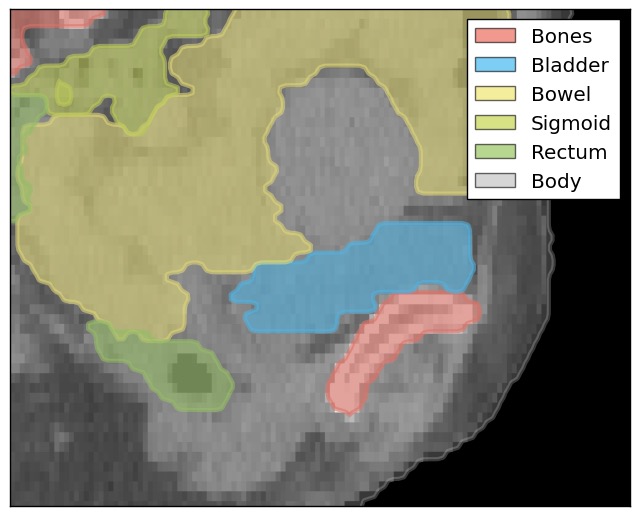}\vspace{0.05cm}\\
        
        Patient 4 & \includegraphics[width=\referencerenderwidth,valign=m]{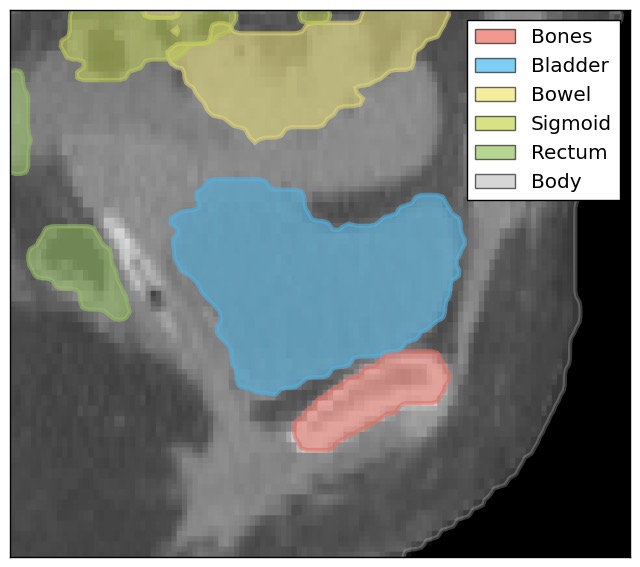} & \includegraphics[width=\referencerenderwidth,valign=m]{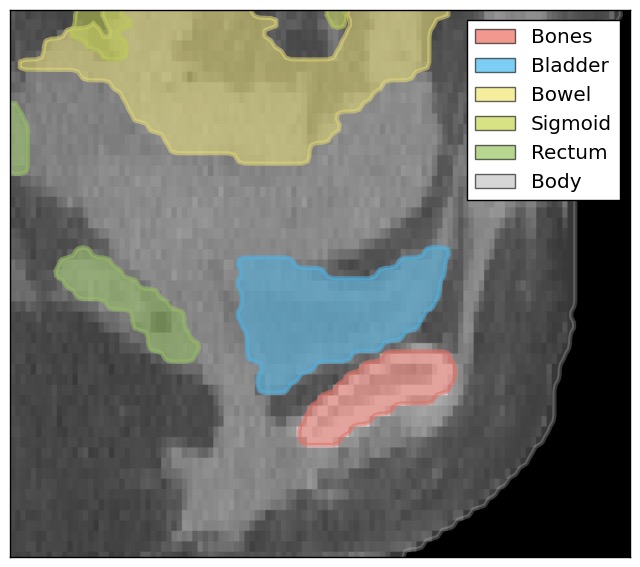}\vspace{0.05cm}\\
        \bottomrule
    \end{tabularx}
    \vspace{0.05cm}
    \caption{Sagittal slices of all registration problems, with organs contoured in different colors.}
    \label{tab:reference-renders}
    \vspace{-0.5cm}
\end{table}

%% file: figures/tex/results/best-renders.tex
\begin{figure*}
\centering
\captionsetup[subfigure]{aboveskip=0pt}
\begin{subfigure}[b]{.245\linewidth}
  \centering
  \includegraphics[width=\linewidth]{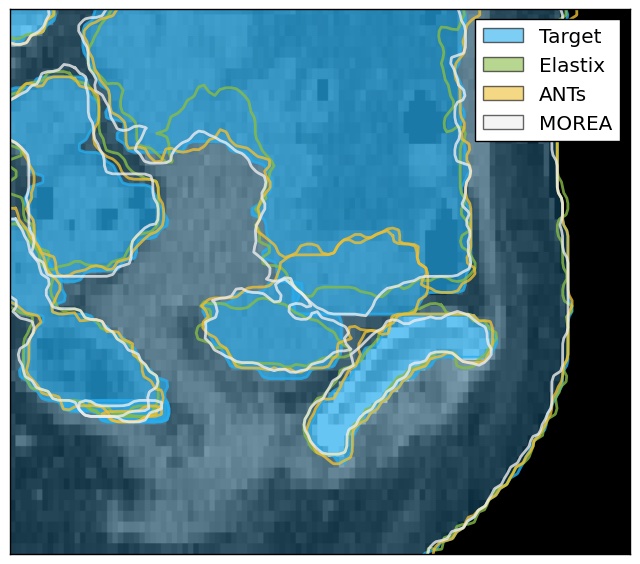}
  \caption{{\normalfont Patient 1}}
\end{subfigure}%
\begin{subfigure}[b]{.245\linewidth}
  \centering
  \includegraphics[width=\linewidth]{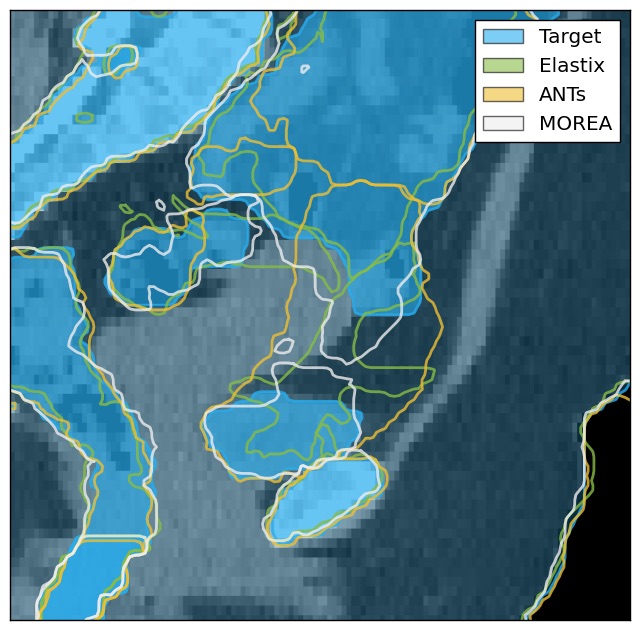}
  \caption{{\normalfont Patient 2}}
\end{subfigure}%
\begin{subfigure}[b]{.245\linewidth}
  \centering
  \includegraphics[width=\linewidth]{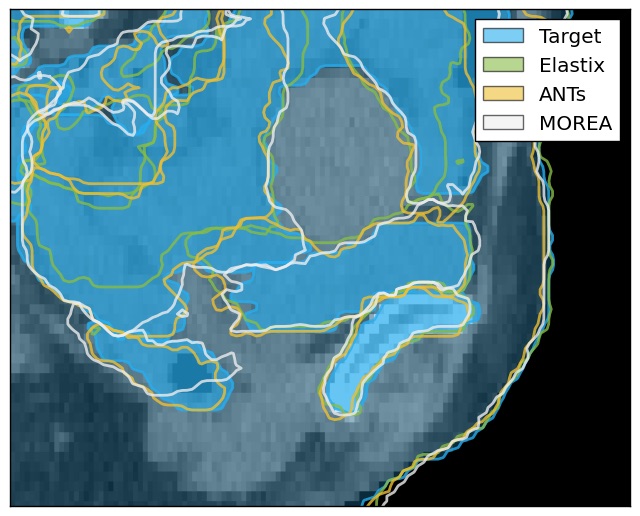}
  \caption{{\normalfont Patient 3}}
\end{subfigure}%
\begin{subfigure}[b]{.245\linewidth}
  \centering
  \includegraphics[width=\linewidth]{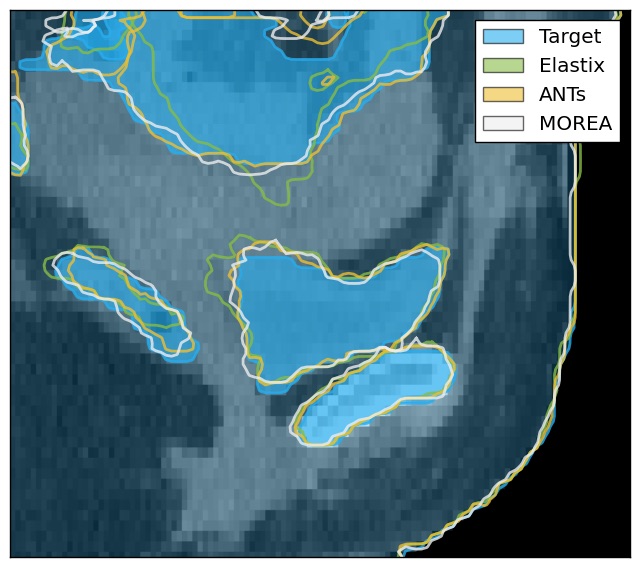}
  \caption{{\normalfont Patient 4}}
\end{subfigure}%
\vspace{-0.3cm}
\caption{A selection of the best predicted deformations for each patient, represented by deformed contours rendered onto the target image with its reference contours (i.e., target in blue). Annotated slices showing all organs are provided in Table~\ref{tab:reference-renders}.}
\label{fig:best-renders}
\end{figure*}

%% file: figures/tex/results/dvfs.tex

\begin{figure*}
\centering
\begin{subfigure}[b]{.3\linewidth}
  \centering
  \includegraphics[width=\linewidth]{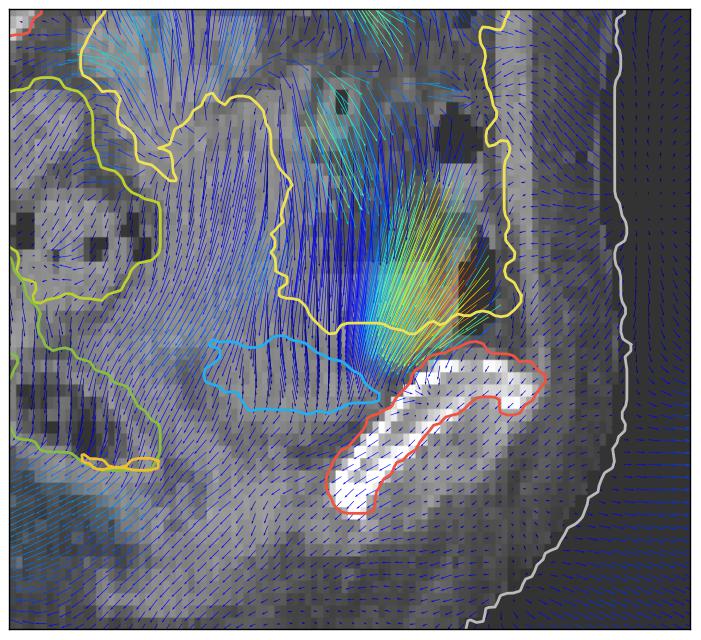}
  \caption{Elastix}
\end{subfigure}%
\begin{subfigure}[b]{.3\linewidth}
  \centering
  \includegraphics[width=\linewidth]{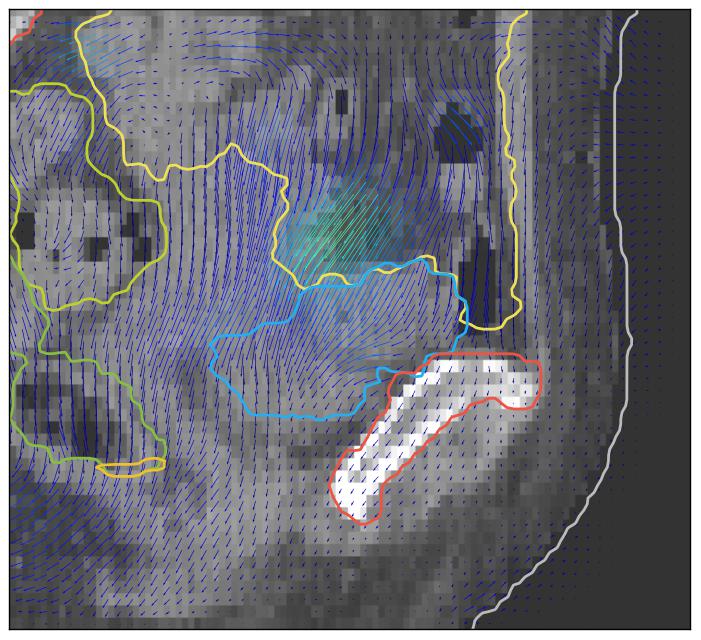}
  \caption{ANTs}
\end{subfigure}%
\captionsetup[subfigure]{aboveskip=0pt}
\begin{subfigure}[b]{.3\linewidth}
  \centering
  \includegraphics[width=\linewidth]{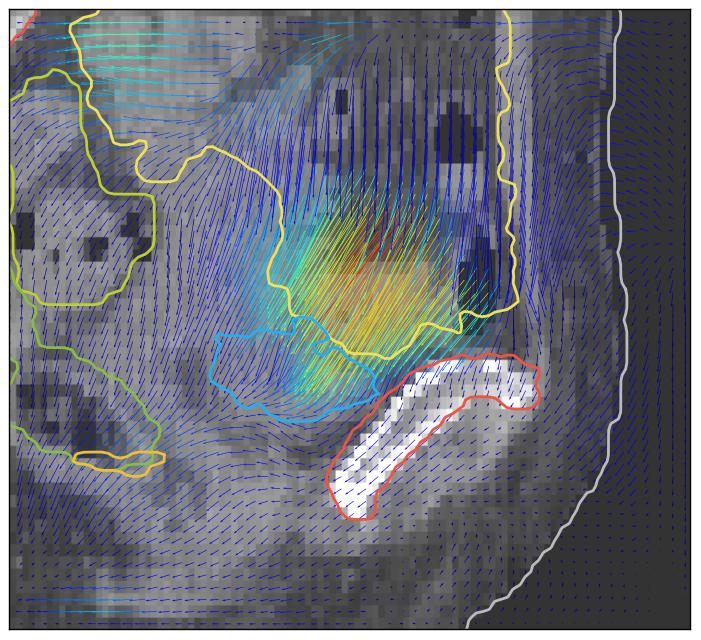}
  \caption{MOREA}
\end{subfigure}
\begin{subfigure}[b]{.05\linewidth}
  \includegraphics[height=4.96cm]{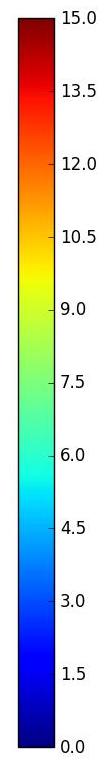}
  \vspace{0.3cm}
\end{subfigure}
\vspace{-0.3cm}
\caption{Forward deformation vector fields and deformed contours of selected predicted deformations on Patient 1, for all 3 approaches (down-sampled for visibility). Arrow colors represent deformation magnitudes, in voxels (1 voxel $=$ 1.5\emph{mm}).}
\label{fig:dvfs}
\end{figure*}

%% file: sections/6-results-discussion.tex
\input{figures/tex/results/significance.tex}

\section{Results and Discussion}
\label{sec:results-discussion}

Figure~\ref{fig:best-renders} shows selected outcomes from each per-patient fine-grained configuration experiment, along with a solution from MOREA's approximation front for each patient.
For Elastix, we select the runs with regularization weights 1.0, 1.0, 10.0, and 10.0 on Patients 1--4, respectively, and for ANTs, we select all runs with weight 0.
The full results of our configuration experiments for both existing approaches can be inspected in Appendix Sections~B.1.2 and~B.2.2.
Convergence plots for Patient~1, which show how all approaches have converged to the results presented here, can be found in Appendix~D.
As described in Section~\ref{sec:experiments:problems}, there is an intrinsic difference in difficulty between the scans.
In general, we observe MOREA generally outperforming other approaches on the more difficult patients (1--3), as can be seen visually in the deformed contours shown in Figure~\ref{fig:best-renders} and in the additional renders and analyses provided in Appendix~D.

For \textbf{Patient~1}, we also render DVF slices in Figure~\ref{fig:dvfs}, showing the transformation computed for each region of one slice.
We observe that the deformations returned by Elastix and ANTs only deform the top region of the bladder.
MOREA is the only approach which distributes this deformation across the entire bladder, which is a more realistic deformation in this flexible volume.
Figure~\ref{fig:front} plots the approximation set that is produced by MOREA on Patient 1, highlighting 3 solutions with slightly different deformations.
This illustrates the range of solutions presented to the user, all of which spread the deformation across the bladder.

\input{figures/tex/results/front.tex}

\textbf{Patient~2}, which features the largest volume change in the bladder, seems to prove the most difficult:
MOREA comes closest to modeling its deformation (see Fig.~\ref{fig:best-renders}), although this comes at the cost of the bowel also being moved downwards.
A probable cause is the little space (i.e., margin) left between the two organs in the source image.
Here, MOREA's result exposes a more fundamental problem that affects all approaches: structures separated by little to no margin in one image cannot be separated in the other image with a transformation model consisting of a single mesh.
The change of bladder shape in \textbf{Patient~3} is less severe than for Patient~2, but still proves challenging for Elastix and ANTs (see Fig.~\ref{fig:best-renders}).
Especially the back region (located left of the image center) does not match the target.
\textbf{Patient~4} represents a relatively easy registration problem, with little change in the shape of the bladder and a clear margin between bladder and bowel (see Fig.~\ref{tab:reference-renders}).
On this problem, visual inspection shows that ANTs and MOREA both find a good bladder contour fit, while Elastix struggles with both bladder and bowel.

Examining these results quantitatively, we conduct significance tests on the Hausdorff distance of the bladder, listed in Table~\ref{tab:significance}.
In all patients, the contour match of the bladder as deformed by MOREA is significantly superior to Elastix's contour match.
ANTs models the contour of the bladder significantly less accurately than MOREA in 3 out of 4 cases, with the fourth case (Patient~4) not having a significantly different result.
Appendix~D lists significance test results for all organs, which confirm these trends, but also show that MOREA's Hausdorff distance can sometimes be significantly higher than that of ANTs or Elastix.
This does not however need to imply worse registration performance, as a qualitative analysis shows.
For example, the deformed shape of the sigmoid of Patient~2 found by ANTs is strongly off (see Figure~\ref{fig:best-renders}).
However, its metric value is deemed significantly better than MOREA's, even though MOREA is closer to the target in terms of general shape.

%% file: figures/tex/results/significance.tex
\begin{table}
    \centering
    \begin{tabularx}{\linewidth}{Xrr}
        \toprule
        Problem & MOREA vs. Elastix & MOREA vs. ANTs \\
        \midrule
        Patient 1 & \textbf{0.011} (\texttt{+}) & \textbf{0.007} (\texttt{+}) \\
        Patient 2 & \textbf{0.007} (\texttt{+}) & \textbf{0.007} (\texttt{+}) \\
        Patient 3 & \textbf{0.012} (\texttt{+}) & \textbf{0.007} (\texttt{+}) \\
        Patient 4 & \textbf{0.007} (\texttt{+}) & 0.195 (\texttt{-}) \\
        \bottomrule
    \end{tabularx}
    \vspace{0.05cm}
    \caption{\emph{p}-values of pair-wise comparisons of Hausdorff distances for the bladder between approaches. A plus (\texttt{+}) indicates a better mean with MOREA, a minus (\texttt{-}) the opposite. Significant results are highlighted.}
    \label{tab:significance}
    \vspace{-0.5cm}
\end{table}

%% file: figures/tex/results/front.tex
\begin{figure}
    \centering
    \includegraphics[width=0.95\linewidth]{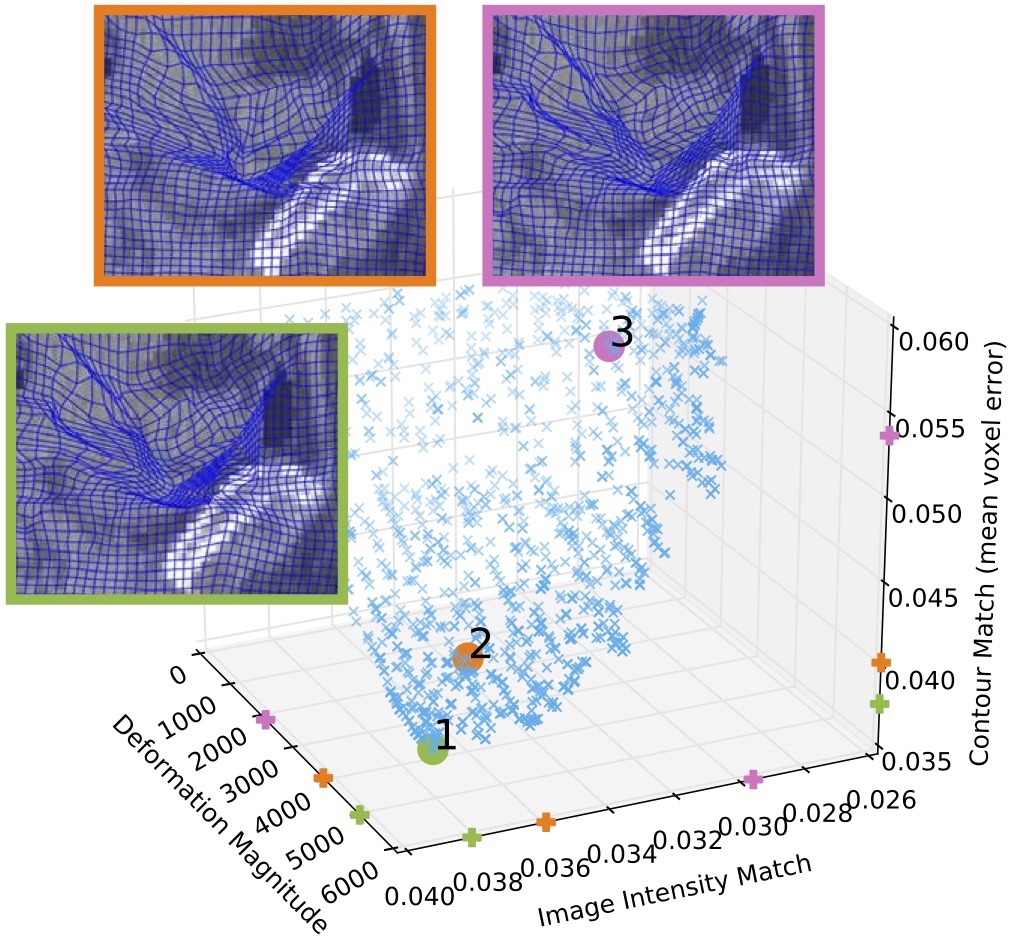}
    \vspace{-0.2cm}
    \caption{Approximation front produced by MOREA on Patient~1. We render 3 zoomed-in registration solutions.}
    \label{fig:front}
\end{figure}

%% file: sections/7-conclusions.tex
\section{Conclusions}
\label{sec:conclusions}

This work uniquely brings multiple lines of research in the field of deformable image registration together.
We have introduced a registration approach, MOREA, that is both contour-based and image-based, uses a biomechanical model, and performs multi-objective optimization.
This combination uniquely positions MOREA to tackle challenging 3D image registration problems with large deformations and content mismatches.
MOREA was built on the MO-RV-GOMEA model-based evolutionary algorithm with several problem-specific extensions, such as GPU acceleration, solution repair, and object-aligned mesh initialization.
Our experiments have shown promising results on 4 cervical cancer patient scans, reaching higher contour registration accuracy than two state-of-the-art approaches on 3 of the 4 patients, representing the most difficult cases.
Importantly, the deformation produced by MOREA seems to be more uniformly spread across objects than the deformations produced by existing approaches, which is deemed to be more realistic.

Solutions obtained by MOREA still contain local inaccuracies which does leave room for improvement, in particular in regions where organs interface.
In fact, the results of this study expose a more fundamental problem in DIR, which is the inability of typical DIR models to capture local discontinuities and content mismatches.
This motivates future research into the modeling of independent organ motion, following recent work on this topic~\cite{Risser2013,Pace2012}.
MOREA's extensible, biomechanical model could be well-positioned for expansions to capture these phenomena.
Given such an expanded approach, a larger validation study, with more patients and involving domain experts, could help close the gap to clinical practice.


%% file: sup/sections/a-approach-technical-details.tex
\section{Technical Implementation Details for the MOREA Approach}

In this appendix, we provide additional technical implementation details for the MOREA approach proposed in Section~4.

\subsection{Modeling the deformation magnitude}

MOREA's deformation magnitude objective models heterogeneous elasticities for different image regions.
For each tetrahedron $\delta$, we establish the elasticity of its underlying image region by sampling from object-specific binary masks (see Figure~\ref{fig:heterogeneous-elasticities}).
These masks are computed for each object by filling the interior of its contour (available as guidance), yielding a discrete object segmentation.
We compute the overlap that each object mask has with the tetrahedron $\delta$, which produces one fraction per object.
In the example given in Figure~\ref{fig:heterogeneous-elasticities}, this would be a fraction of 0.4 for the object corresponding to this mask.
These object fractions are multiplied by pre-determined elasticity factors for different tissue types, yielding an overall element-specific factor for $\delta$.
At present, only bones and bladder are assigned custom factors.
The magnitude objective value for $\delta$ is multiplied by this factor to better model the actual energy required to deform this image region.

\subsection{Modeling the image similarity}

The image intensity objective of MOREA is defined as a sum of squared intensity differences at certain sample points.
Modeling the partial objective value of one tetrahedron requires determining which image voxels to sample.
The existing prototype~\cite{Andreadis2022} tries to find all voxels with center points lying inside the tetrahedron, using a line-search-inspired method.
We observe, however, that this discrete association of voxels with tetrahedra leads to undesirable behavior around tetrahedral surfaces, with voxels sometimes being associated with multiple or no neighboring tetrahedra.
This phenomenon can be used to improve the sampled value while not improving or even deteriorating the true value.

In our approach, we therefore introduce a random-sampling based method which samples the image space continuously, interpolating intensity values between voxel centers.
This is also better-suited for GPU acceleration, since there are less decision points at which execution needs to pause.
We uniformly sample $N$ points in each tetrahedron using its barycentric coordinate system, with $N$ being determined by the volume of the tetrahedron.
For each point, we sample 4 random real numbers $r_i \in [0;1]$ and take $-\log(r_i)$ for a uniform spread.
We then normalize the coordinates by their sum, to ensure that they lie in the tetrahedron.
Instead of a conventional random number generator, we use the Sobol sequence, for a more even spread of sample points.
We ensure reproducibility by seeding the Sobol sequence for each tetrahedron with a seed derived from its coordinates.
Therefore, the same positions are always sampled per tetrahedron configuration.

\input{sup/figures/tex/illustrations/heterogeneous-elasticities.tex}

\input{sup/figures/tex/illustrations/guidance.tex}

\subsection{Modeling the guidance error}

The guidance error objective of MOREA approximates the contour match of a solution.
Previous work~\cite{Andreadis2022} computes the extent of a contour match by considering each point in $C_s$ and computing the distance of its corresponding version in target space to the closest point in the set $C_t$.
This requires iterating over all points in $C_s$, establishing which tetrahedron they are located in, and computing the transformation at that point.
We introduce a new, continuous guidance formulation that approximates point-wise distances and proved to be faster and more robust, in preliminary experiments.

During the random sampling process used for the intensity objective on the source image $I_s$, we also consider the same locations on a distance map of $C_s$, which gives the closest point to the source contour (see Figure~\ref{fig:guidance}).
The distance at that point in the map of $C_s$ is subtracted from the distance at the corresponding point in the map of $C_t$, and weighted inversely by the distance to the source contour.
The distances are truncated to a radius around each guidance point, measuring 2.5\% of the width of the image, so that far away movements do not influence the guidance error of a point set.
We normalize the guidance error of each point set by the number of points in that set compared to the total number of guidance points, to counteract biases towards more well-defined or larger contours.

\subsection{Accurately detecting mesh folds}

A function detecting constraint violations needs to have high precision (i.e., accurately identify all violations) and low latency (i.e., quickly return its answer).
It should furthermore be defined continuously, so that the method can also assess the severity of violations.
This is important for methods that repair violations.

Prior work on mesh-based 3D image registration~\cite{Andreadis2022} uses a ray-intersection method, testing if a point is inside a so-called bounding polygon.
This method has proven error-prone in 3D in preliminary experiments, producing false positives and negatives.
We therefore develop a new method for detecting folds in a tetrahedral mesh, based on the signed volumes of its tetrahedra~\cite{Ericson2004}.
Our method calculates the signed volume of each tetrahedron in the initial mesh configuration, to establish a set of reference signs.
When a point is moved, we recalculate the signed volumes of all tetrahedra that this affects and compare them to the respective reference signs.
The signs of at least one tetrahedron will flip if a fold has occurred.
We use this phenomenon to detect mesh constraint violations and to compute the severity of each violation, using the absolute value of the violating signed volume.

\subsection{Ensuring diversity in the initial population}

Even with a smartly initialized mesh, the diversity of the population at generation 0 plays an important role~\cite{Maaranen2007}.
Prior work uses one reference solution and generates random deviations by sampling around each mesh point with increasingly large variance~\cite{Andreadis2022}.
For low-resolution meshes, this method can be effective, but for higher-resolution meshes, this method can lead to many constraint violations in the generated solutions (i.e., folded mesh configurations).
We introduce a method for initialization noise that generates large deformations free of constraint violations, inspired by approaches using radial basis functions in other domains~\cite{Zhang2020}.
Our method places a number of Gaussian kernels on both source and target images and models a sense of gravity from mesh points towards these kernels.
These forces are applied in incremental rounds, as long as they do not cause constraint violations.
A deformation vector field generated by this strategy is depicted in Figure~\ref{fig:random-init-dvf}.

\input{sup/figures/tex/illustrations/init-dvf.tex}

%% file: sup/figures/tex/illustrations/heterogeneous-elasticities.tex
\begin{figure}
    \centering
    \hspace{2.54cm}%
    \includegraphics[width=0.7\linewidth]{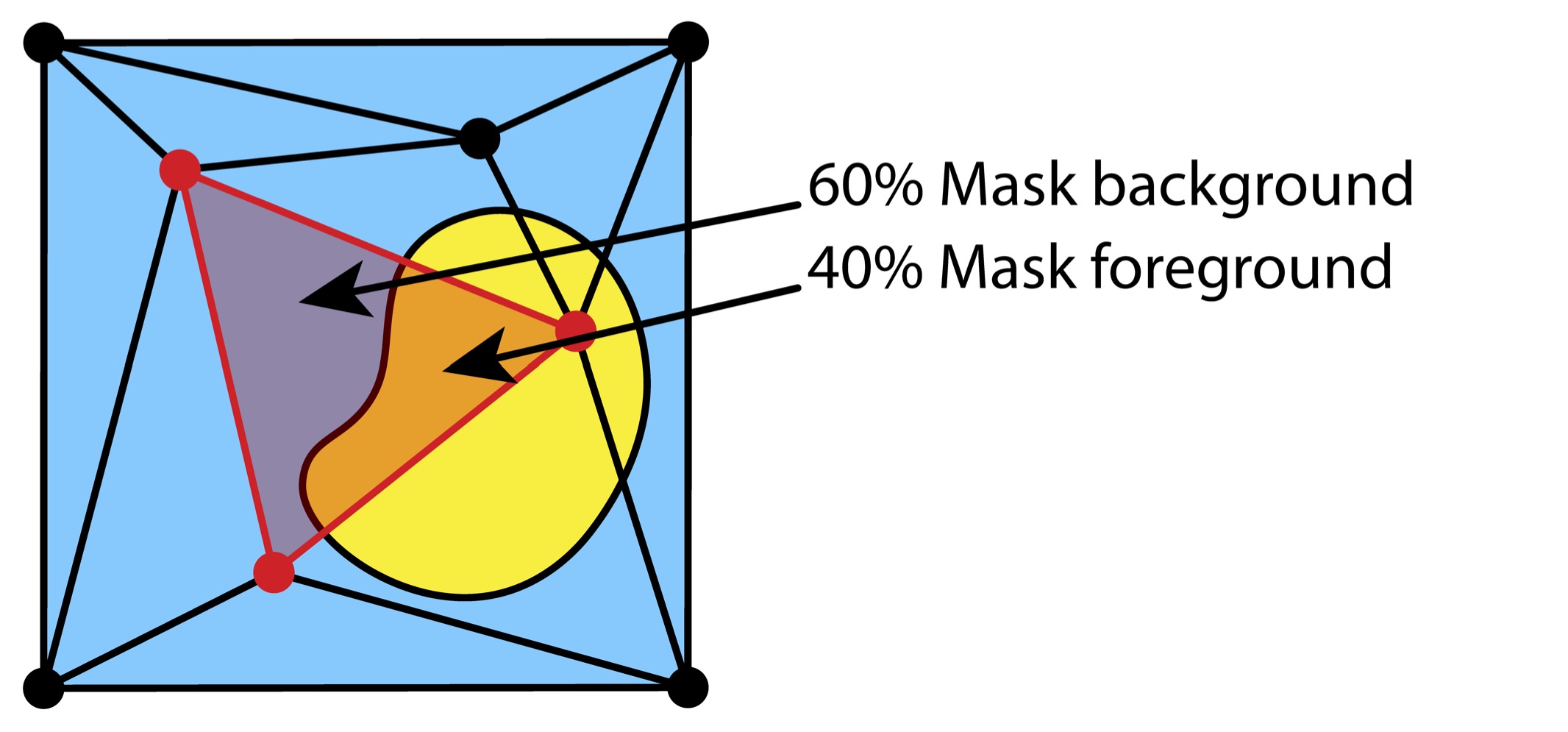}
    \vspace{-0.7cm}
    \caption{2D illustration of how one tetrahedral element (here: the red triangle) overlaps with the mask of an organ. The computed overlap fractions are used to establish the elasticity factor for this tetrahedron's deformation magnitude.}
    \label{fig:heterogeneous-elasticities}
    \vspace{-0.2cm}
\end{figure}

%% file: sup/figures/tex/illustrations/guidance.tex
\begin{figure}
    \centering
    \begin{subfigure}{.48\linewidth}
      \centering
      \includegraphics[width=\linewidth]{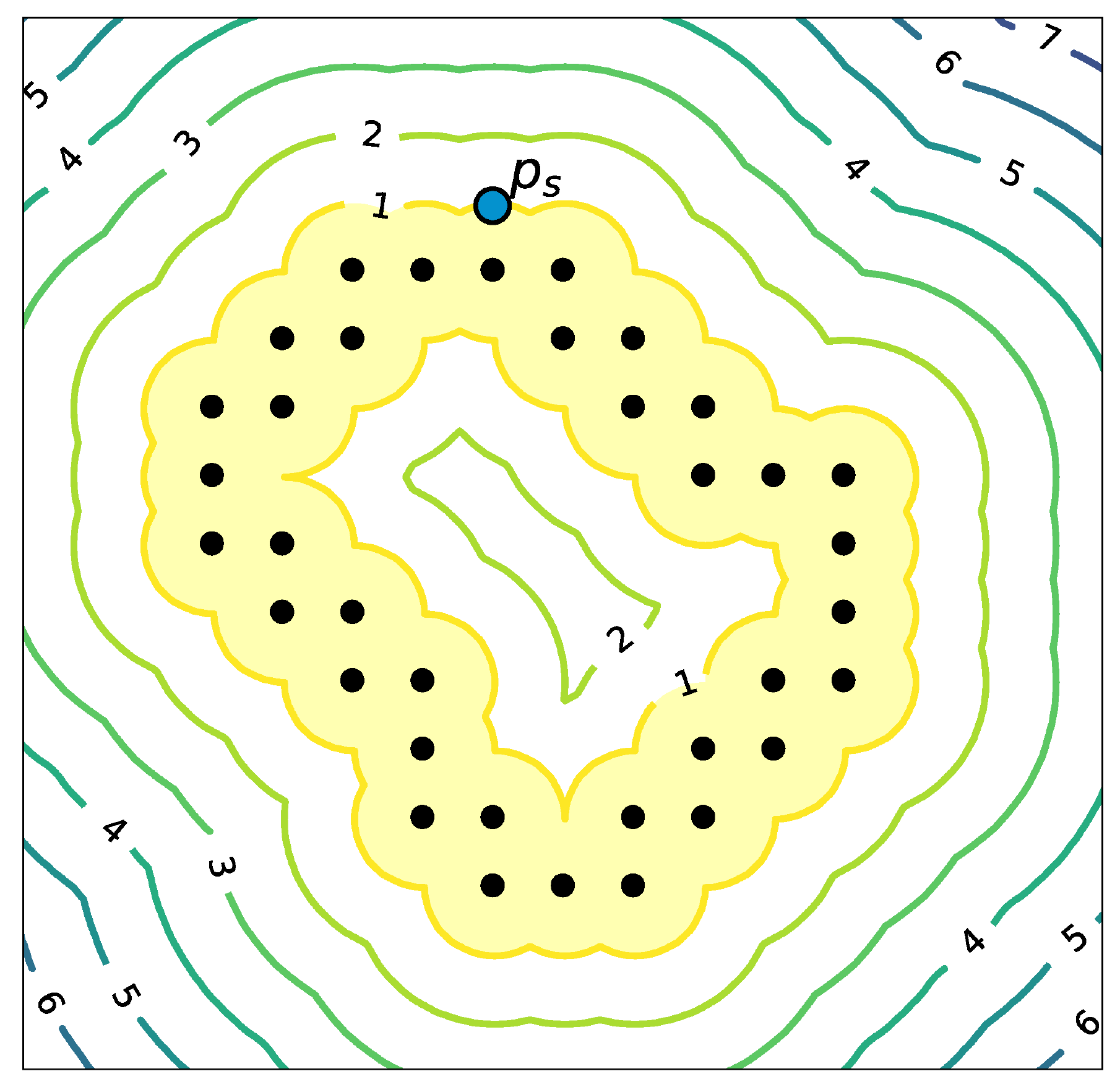}
      \caption{{\normalfont Source contour point set.}}
      \label{fig:guidance:source}
    \end{subfigure}%
    \hspace{0.009\linewidth}
    \begin{subfigure}{.48\linewidth}
      \centering
      \includegraphics[width=\linewidth]{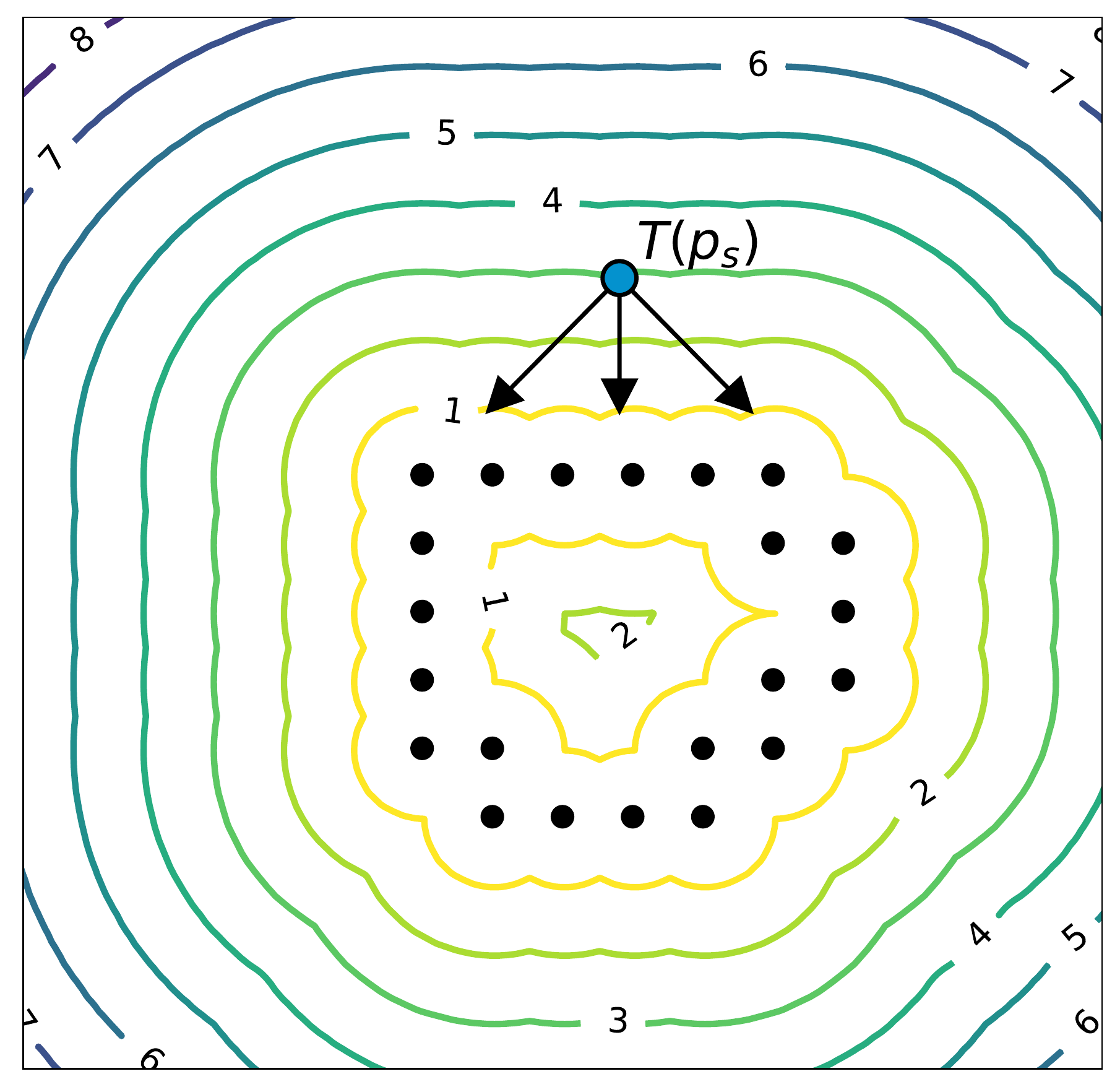}
      \caption{{\normalfont Target contour point set.}}
      \label{fig:guidance:target}
    \end{subfigure}
    \vspace{-0.25cm}
    \caption{Two point sets of object contours in a source and target image, with minimal distance maps visualized using isolines. A randomly sampled point $p_s$ is close to the source contour, but the transformed $T(p_s)$ is farther away from the target contour. The yellow shaded area represents the truncation area beyond which sampled points are discarded.}
    \label{fig:guidance}
    \vspace{-0.4cm}
\end{figure}

%% file: sup/figures/tex/illustrations/init-dvf.tex
\begin{figure}
    \centering
    \includegraphics[width=0.95\linewidth]{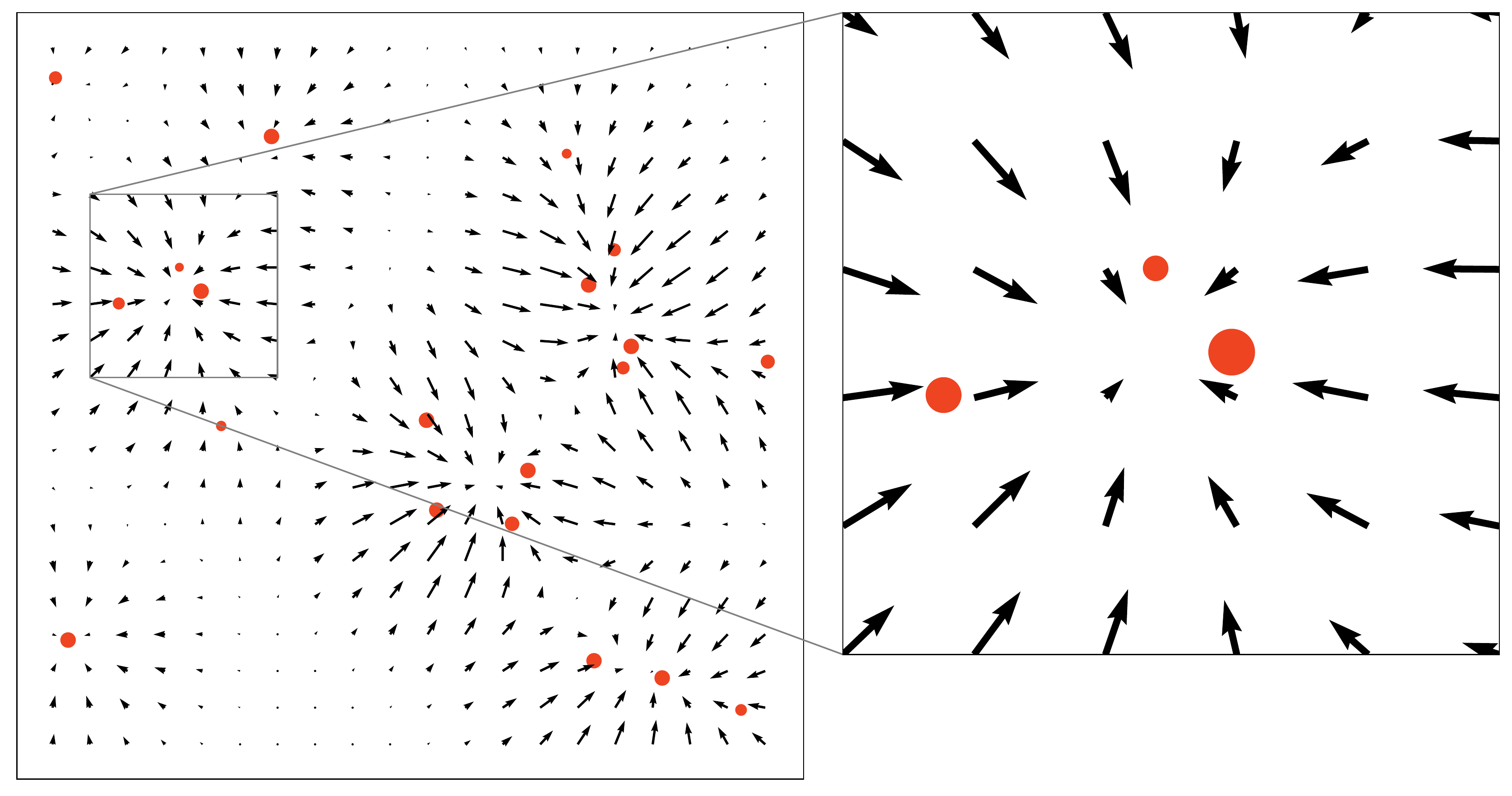}
    \vspace{-0.2cm}
    \caption{A 2D vector field produced by our radial-basis-function approach used to generate solutions. Red dots mark attractors, with their size indicating their weight.}
    \label{fig:random-init-dvf}
    \vspace{-0.2cm}
\end{figure}

%% file: sup/sections/b-problem-specification.tex
\newpage~
\newpage
\section{Extended Problem Specification}
\label{sec:full-problem-spec}

In this appendix, we provide additional information on the registration problems used in this study and specify additional methods for evaluation and comparison of registration quality.

\subsection{Additional Problem Information} 
Table~\ref{tab:ct-resolutions} lists the in-slice resolutions of the CT scans used.
This is the physical resolution of each slice prior to our resampling step to $(1.5,1.5)$\emph{mm}.
We also provide additional views on each medical image: For each patient, Table~\ref{tab:reference-renders-full} lists two slices per source and target image.
This provides a useful additional perspective, since some movements are better visible from a different angle.



\subsection{Additional Evaluation Methods}
\label{sec:full-problem-spec:evaluation}

We evaluate each solution with four types of methods, based on (1) surface-based registration accuracy, (2) visual inspection using 2D and 3D visualizations, (3) volume-based registration accuracy, (4) landmark registration accuracy.
Method types (1) and (2) have been described in Section~5.3.
Here, we give an additional strategy for (1), and outline additional methods (3) and (4).

\subsubsection{Surface-based registration accuracy}
Alternatively to the Hausdorff distance, the 95th percentile of the Hausdorff distance is another indicator we use in our study.
This represents the distance for which it holds that 95\% of all surface point distances are smaller than this distance.
Both Hausdorff and Hausdorff 95th percentile metrics are computed using the \texttt{pymia} PyPI package.

\subsubsection{Volume-based registration accuracy}
Adjacent to surface accuracy, we are interested in the accuracy of individual volumes (e.g., organs, bones) represented in the images.
A common metric for this is the Dice coefficient, which represents the fraction of volume overlap compared to total volumes.
Using binary masks of each annotated object in the images, we compute this metric on a voxel-by-voxel basis.
We compare the binary masks corresponding to the target image against binary masks of the source image transformed using the computed deformation.
With the same reasoning as for surface-based evaluation (see Section~5.3), we discard the same border margin when evaluating volume-based metrics.

\subsubsection{Landmark registration accuracy}
A set of corresponding landmarks not provided to the algorithm during optimization can be used to locally assess the accuracy of a registration.
For each pair of landmarks, we transform the source landmark using the forward transformation to target space, and compute landmark accuracy as the Euclidean distance between the transformed source landmark and its corresponding target landmark.
This is a common accuracy measure in image registration studies~\cite{Eiben2016,Brock2017}, but can be less accurate as an indicator of overall registration quality, since landmarks are placed on visible anatomical structures that often have limited movement, as is the case in our scans.

\newpage

\subsection{Comparing Multi-Object Metrics}

The metrics of individual organs cannot be adequately interpreted in isolation, as organ motions are related and therefore form trade-offs.
We visualize these trade-offs by plotting scores for different organs in one parallel coordinates plot, similar to the color-coded heatmap comparison presented in~\cite{Klein2009}.
These line plots help inform decisions that need to take registration quality across registration targets into account.

\input{sup/figures/tex/results/ct-resolutions.tex}

\input{sup/figures/tex/results/reference-renders-full.tex}

\newpage
~\newpage
~\newpage

%% file: sup/figures/tex/results/ct-resolutions.tex
\begin{table}[b]
    \centering
    \begin{tabularx}{\linewidth}{lXr}
        \toprule
        Patient & Scan & In-slice Resolution \\
        \midrule
        \multirow{2}{*}{Patient 1} & Full bladder & $(0.86,0.86)$\emph{mm}\\
        & Empty bladder & $(0.98,0.98)$\emph{mm} \\
        
        \midrule
        \multirow{2}{*}{Patient 2} & Full bladder & $(1.04,1.04)$\emph{mm}\\
        & Empty bladder & $(1.07,1.07)$\emph{mm} \\
        
        \midrule
        \multirow{2}{*}{Patient 3} & Full bladder & $(0.98,0.98)$\emph{mm}\\
        & Empty bladder & $(0.98,0.98)$\emph{mm} \\
        
        \midrule
        \multirow{2}{*}{Patient 4} & Full bladder & $(1.04,1.04)$\emph{mm}\\
        & Empty bladder & $(1.00,1.00)$\emph{mm} \\
        \bottomrule
    \end{tabularx}
    \vspace{0.05cm}
    \caption{In-slice resolutions for the slices of each CT scan, prior to resampling them to $(1.5,1.5)$\emph{mm}.}
    \label{tab:ct-resolutions}
    \vspace{-0.4cm}
\end{table}

%% file: sup/figures/tex/results/reference-renders-full.tex
\newcommand{\referencerenderwidthfull}{4cm}

\begin{table*}[t]
    \centering
    \setlength\tabcolsep{0pt}
    \begin{tabular}{lcccc}
        \toprule
        Instance & Source image: sagittal & Target image: sagittal & Source image: coronal & Target image: coronal \\
        \midrule
        Patient 1 \hspace{0.4cm} & \includegraphics[width=\referencerenderwidthfull,valign=m]{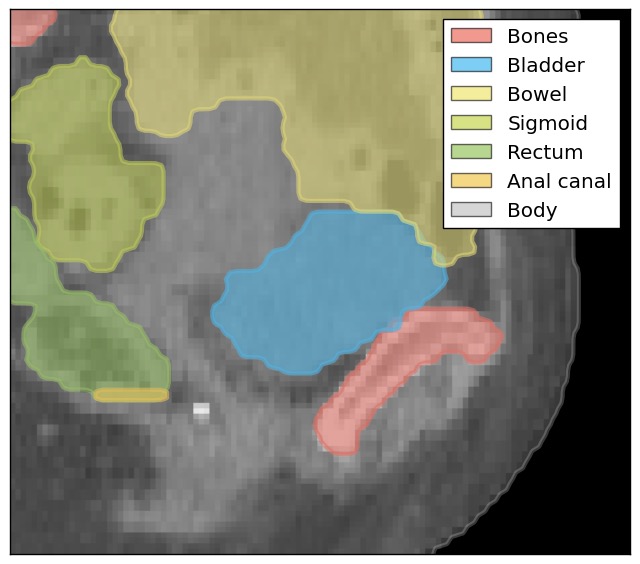} & \includegraphics[width=\referencerenderwidthfull,valign=m]{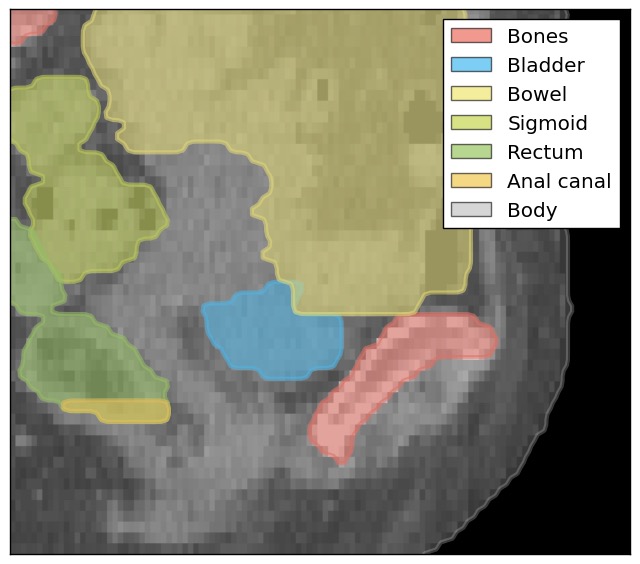} & \includegraphics[width=\referencerenderwidthfull,valign=m]{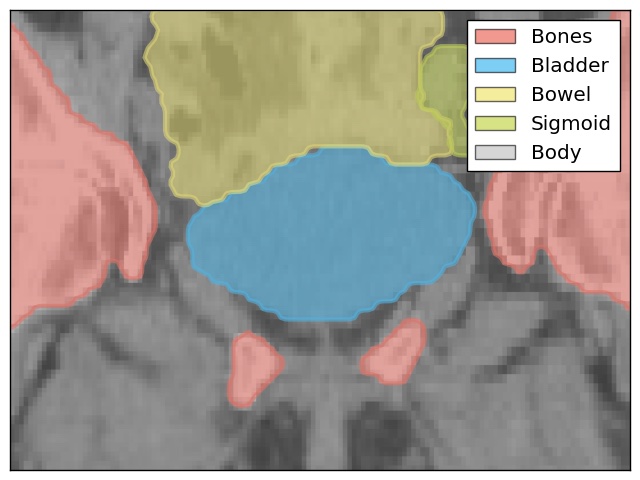} & \includegraphics[width=\referencerenderwidthfull,valign=m]{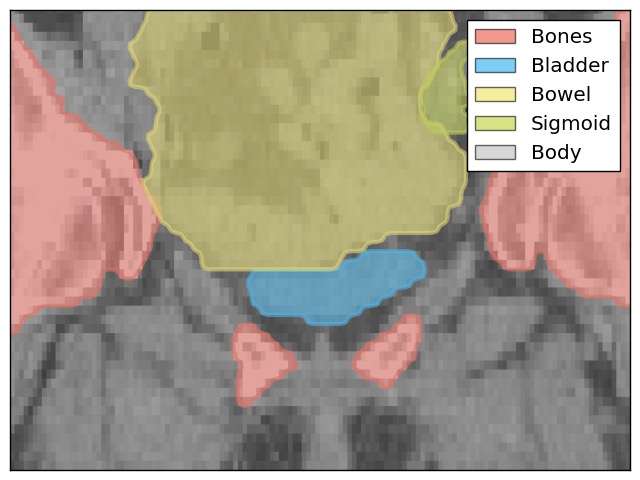} \vspace{0.05cm}\\
        
        Patient 2 \hspace{0.4cm} & \includegraphics[width=\referencerenderwidthfull,valign=m]{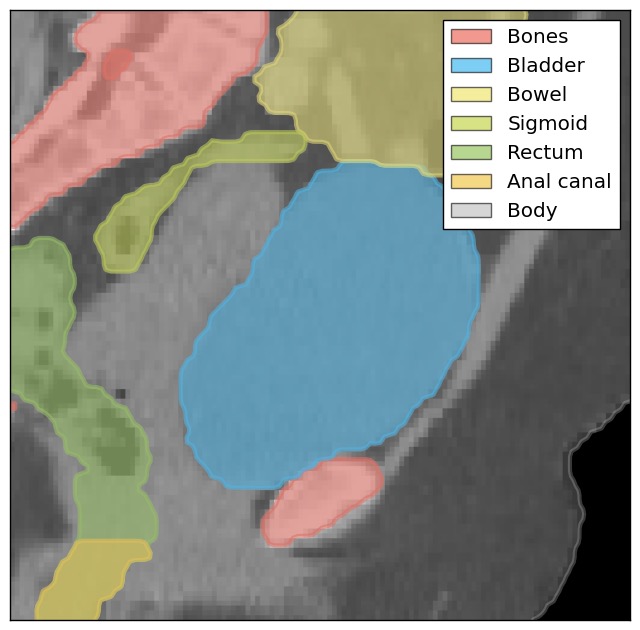} & \includegraphics[width=\referencerenderwidthfull,valign=m]{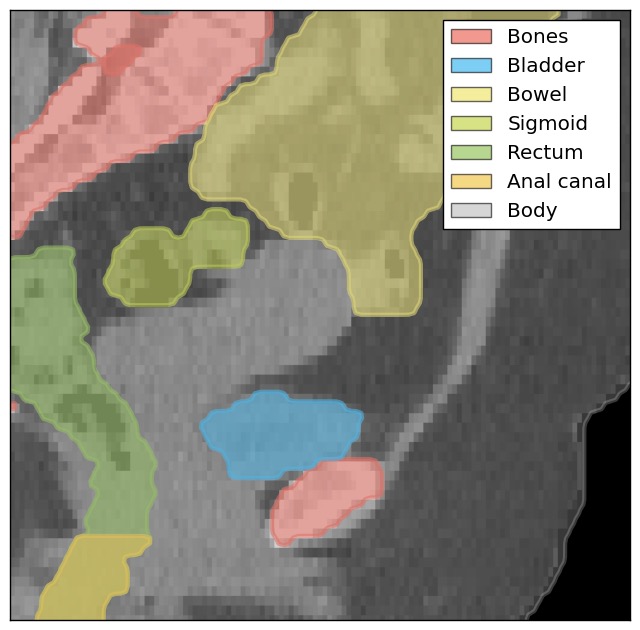} & \includegraphics[width=\referencerenderwidthfull,valign=m]{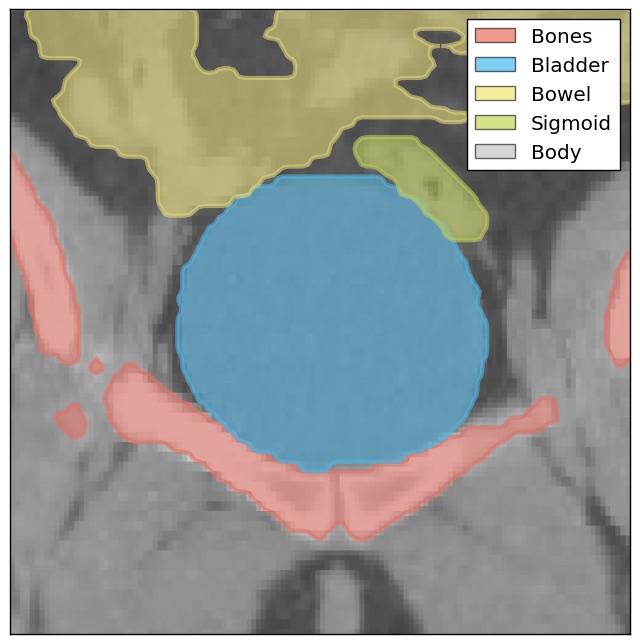} & \includegraphics[width=\referencerenderwidthfull,valign=m]{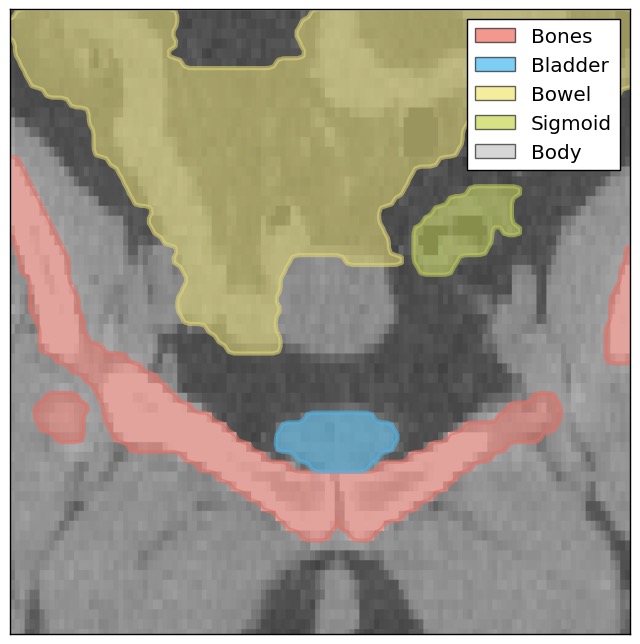} \vspace{0.05cm}\\
        
        Patient 3 \hspace{0.4cm} & \includegraphics[width=\referencerenderwidthfull,valign=m]{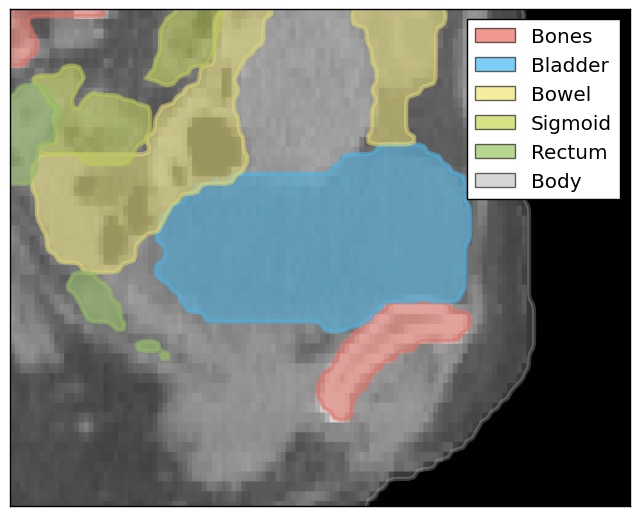} & \includegraphics[width=\referencerenderwidthfull,valign=m]{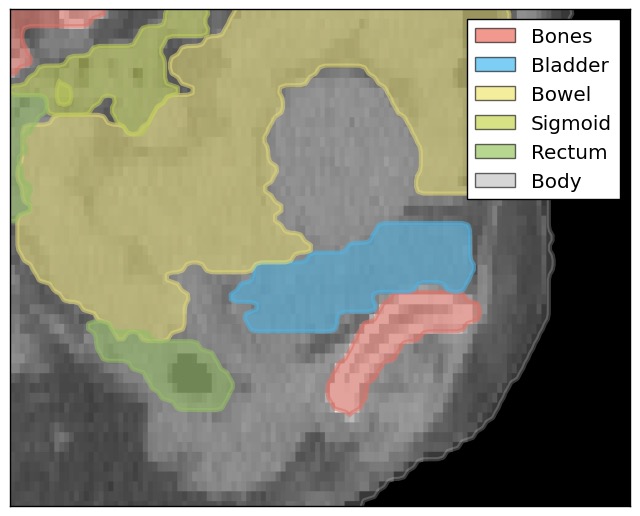} & \includegraphics[width=\referencerenderwidthfull,valign=m]{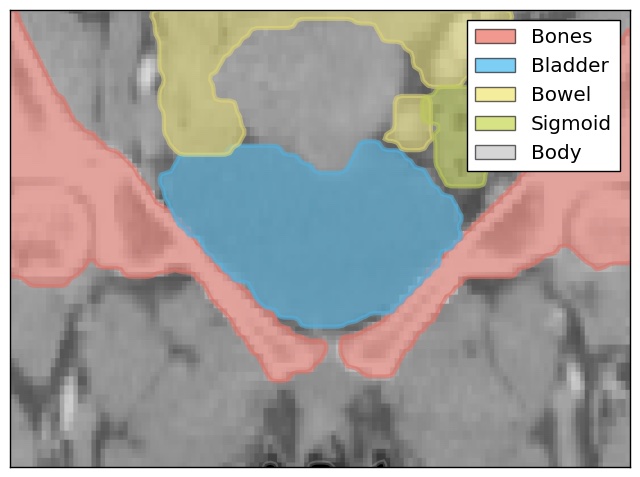} & \includegraphics[width=\referencerenderwidthfull,valign=m]{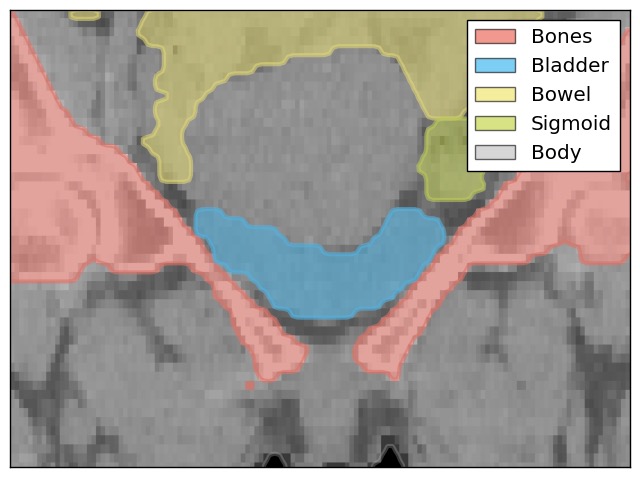} \vspace{0.05cm}\\
        
        Patient 4 \hspace{0.4cm} & \includegraphics[width=\referencerenderwidthfull,valign=m]{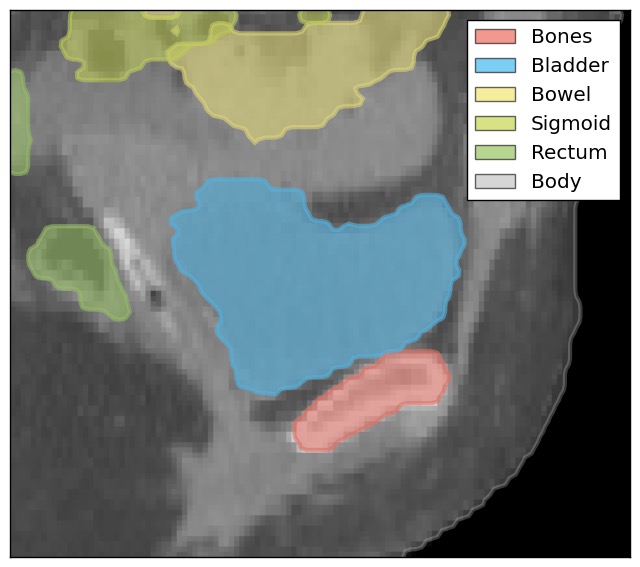} & \includegraphics[width=\referencerenderwidthfull,valign=m]{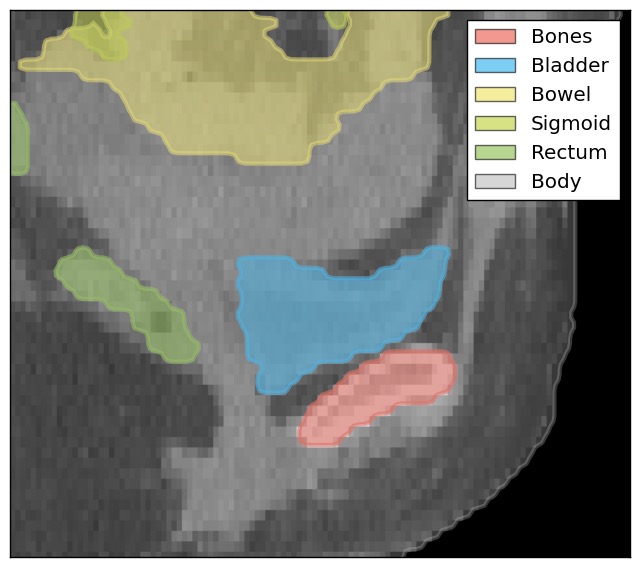} & \includegraphics[width=\referencerenderwidthfull,valign=m]{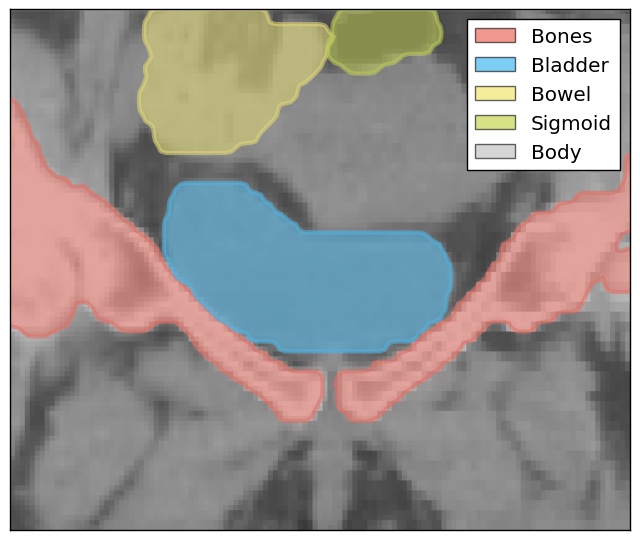} & \includegraphics[width=\referencerenderwidthfull,valign=m]{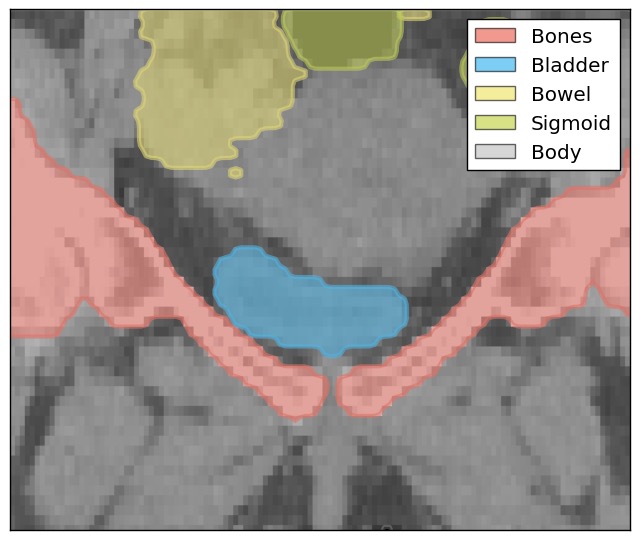} \vspace{0.05cm}\\
        \bottomrule
    \end{tabular}
    \vspace{0.05cm}
    \caption{Slices of all registration problems, with organs contoured. Sagittal: side view; coronal: front-to-back view.}
    \label{tab:reference-renders-full}
    \vspace{-0.4cm}
\end{table*}

%% file: sup/sections/c-existing-approaches.tex
\section{Configuration of Compared Approaches}
\label{sec:approach-configurations}

\subsection{Elastix}
\label{sec:approach-configurations:elastix}

We use Elastix version 5.0.0.
Based on parameter settings from the Elastix Model Zoo\footnote{\url{https://elastix.lumc.nl/modelzoo/}}, we apply multi-resolution Elastix registration to our registration problems with a range of configurations, trying to find the optimal configuration for each problem  (see Section~\ref{sec:approach-configurations:parameter-files:elastix} for our parameter files).
Inspired by an approach implementing symmetric registration in Elastix using a group-wise methodology~\cite{Bartel2019}, we also experiment with a symmetric variant which registers both images to a common image mid-space.
For all setups, we relax convergence requirements by increasing the number of iterations per resolution to 10,000, which is significantly larger (5 times) than the computational budget given in most reference files.
This is done to give Elastix sufficient opportunity to model the large deformations present.
We also stabilize optimization by increasing the number of image sampling points from the frequently used 10,000 to 20,000.
Although increasing the computational complexity, this should make image intensity approximations used internally during optimization more accurate and computed gradients more reliable.

Elastix computes the \textit{inverse} transform by default, meaning a vector field defined in fixed (target) space leading to moving (source) space.
To compute the \textit{forward} transform, which is needed to transform annotations from moving (source) to fixed (target) space, we rerun the registrations with the given parameter files and the computed transform as initial transform, but replace the metric(s) with the \textit{DisplacementMagnitudePenalty} metric.
This effectively finds the \textit{forward} transform of the computed \textit{inverse} transform.
Exporting this \textit{forward} transform in isolation, by removing the initial transform pointer from the parameter file, yields the desired DVF.

Elastix does not support the optimization of object contour matches, which are optimized by the MOREA approach through the guidance objective.
To ensure a fair comparison, we attempt to input this information as a pair of composite mask images to implicitly pass on contour information.
Each mask image is made by combining the different binary object masks available for each scan, giving each object segmentation a different homogeneous intensity value.
In runs where this feature is enabled, we precede the CT image registration run with a registration of these prepared composite masks.

\input{sup/figures/tex/a-elastix/pre-tuning-symmetry.tex}

\input{sup/figures/tex/a-elastix/pre-tuning-renders-symmetry.tex}

\subsubsection{Coarse-grained configuration experiments}

First, we conduct an initial set of runs on Patient~1 to establish a suitable base configuration for this problem modality and difficulty.
We explore the influence of registration direction (unidirectional vs. symmetric) and the use of a composite mask registration step (with vs. without), assuming a regularization weight of 0.001, to give Elastix flexibility for large deformations (a large weight on the deformation magnitude weight can hinder large deformations).

In Figure~\ref{fig:elastix:pre-tuning:symmetry}, we plot the performance of Elastix using symmetric and unidirectional registration, reporting two different metrics (Dice score and 95th percentile of the Hausdorff distance).
We observe that unidirectional registration generally performs similarly or better compared to symmetric registration, except for the rectum and anal canal, in terms of Dice score. 
Due to the relatively large performance gain in the bladder (the most strongly deforming organ), we choose unidirectional registration at this point.
This choice is supported by visual inspection of Figure~\ref{fig:elastix:pre-tuning:renders:symmetry}, which shows slightly better performance on the bladder in the coronal slice.

We now turn to the use of a composite mask registration step, in an attempt to get larger deformations by simplifying the information input to Elastix.
Figure~\ref{fig:elastix:pre-tuning:masks} shows the same metrics, but with and without the use of such a step (while using unidirectional registration).
The results do not identify one clear superior approach, since the Dice score of the \textit{with-mask} configuration is generally superior but the Hausdorff 95th percentile is lower for the \textit{without-mask} configuration.
Figure~\ref{fig:elastix:pre-tuning:renders:masks} indicates that adding a mask step improves the modeling of the base region of the bladder, but the middle region is merely contracted sideways without moving the top region downwards, thereby not resulting in anatomically realistic deformations.
Nevertheless, we choose this version over the version without mask registration step, since the large deformation needed is modeled more closely with the step added.

\subsubsection{Fine-grained configuration experiments per patient}
\label{sec:approach-configurations:elastix:main-tuning}

For each patient, we try exponentially increasing regularization weights; an exponential regularization weight sweep that is also used in similar work~\cite{Bondar2010}.
The Dice scores on each patient are reported in Figure~\ref{fig:elastix:main-tuning:dice} and the 95th percentiles of the Hausdorff distance in
Figure~\ref{fig:elastix:main-tuning:hausdorff-95}.
Renders for each problem are provided in Figures~\ref{fig:elastix:main-tuning:renders:603}--\ref{fig:elastix:main-tuning:renders:618}.

We observe that the optimal regularization weight varies strongly between different registration problems. 
While the scans of Patient 1 (Fig.~\ref{fig:elastix:main-tuning:renders:603}) are best served with a weight of 1.0 out of the tried settings, the scans of Patient 3 (Fig.~\ref{fig:elastix:main-tuning:renders:617}) seem better off with a weight of 10.0.

\input{sup/figures/tex/a-elastix/pre-tuning-masks.tex}

\input{sup/figures/tex/a-elastix/pre-tuning-renders-masks.tex}

\input{sup/figures/tex/a-elastix/main-tuning-dice.tex}

\input{sup/figures/tex/a-elastix/main-tuning-hausdorff-95.tex}

\input{sup/figures/tex/a-elastix/main-tuning-renders.tex}

\clearpage

\subsubsection{Parameter files}
\label{sec:approach-configurations:parameter-files:elastix}

Below, we list the parameter files that we used for the different variants of Elastix registration.
Tokens starting with the \texttt{\$} character denote variables that are resolved before we pass the file to Elastix (e.g., a random seed that we increment at every repeat).

\input{sup/figures/tex/a-elastix/parameter-files.tex}

\clearpage

\subsection{ANTs SyN}
\label{sec:approach-configurations:ants}

We use ANTs SyN algorithm version 2.4.2.
We bootstrap a registration command using the \texttt{antsRegistrationSyN.sh} script and customize it to fit our problem (see Section~\ref{sec:approach-configurations:parameter-files:ants} for our run commands).
Following official recommendations\footnote{\url{https://github.com/ANTsX/ANTs/wiki/Anatomy-of-an-antsRegistration-call}}, we consider the following settings to be left tunable for this problem:
(1)~what region radius to use for the cross correlation metric, 
(2)~whether to use composite masks as an additional image modality channel during registration,
(3)~what gradient step size to use,
(4)~what regularization weight to assign to local deformations between time steps, and
(5)~what regularization weight to assign to the total deformation.
We configure the first four parameters for Patient~1, and then configure the fifth parameter for each patient, separately.

In our setup, we relaxed convergence limits compared to guidelines to allow for longer, and hopefully more accurate registration.
In terms of metrics, we do not use the point set registration metric that is mentioned in the manual, as the manual states that this metric is not currently supported in ANTs SyN.

We encountered that ANTs SyN random seed does not have any effect on the outcome of registration with the Cross Correlation (CC) measure, even with a random sampling strategy.
The current version seems fully deterministic, but without taking the random seed into account, therefore always producing the same output, regardless of the seed.
This is problematic, since we would like to get multiple outputs that expose how the registration approach reacts to slightly varying inputs.
To mitigate the lack of control on the determinism of the registration, we slightly perturb the sigma smoothing factors (see Listing~\ref{listing:ants-command-with-masks}) with very small (deterministically random) deltas.
$\Delta_3$ is normally distributed and capped between $[-0.1,0.1]$, $\Delta_2$ between $[-0.05,0.05]$, and $\Delta_1$ between $[-0.01,0.01]$.

\subsubsection{Coarse-grained configuration experiments}

We conduct an initial set of coarse-grained configuration experiments on Patient~1 with the ANTs SyN algorithm.
The officially recommended settings serve as our baseline: a cross-correlation radius of 4 voxels, a gradient step size of 0.1, registration of only the image itself (no additional channels), and an update regularization weight of 3.0.
For each of these settings, we experiment with different deviations from the baseline.

\noindentparagraph{Cross correlation radius}
First, we investigate the impact of a different cross correlation radius.
Larger values should improve registration accuracy, since more context information is taken into account when computing the cross correlation of a sample.
Figure~\ref{fig:ants:pre-tuning:cc} confirms this expectation, although it shows little impact overall.
Most organs show little deviation in score, but the anal canal is registered more accurately in terms of Dice score when the radius is increased.
We observe that there are diminishing returns here, e.g., a change of radius from 7 to 8 provides only marginal improvement.
Still, we decide to use the largest setting tested (8 voxels, meaning 12\emph{mm} in the case of the clinical problems), since this setting provides the best outcome and there is no time limit on registration in our study.
The visual render in Figure~\ref{fig:ants:pre-tuning:renders:cc} shows the visual impact of this setting, which can be described as limited.

\input{sup/figures/tex/a-ants/pre-tuning-cc.tex}

\input{sup/figures/tex/a-ants/pre-tuning-renders-cc.tex}

\input{sup/figures/tex/a-ants/pre-tuning-masks.tex}

\input{sup/figures/tex/a-ants/pre-tuning-renders-masks.tex}

\noindentparagraph{Composite mask channel}
Second, we explore the effect of including a composite mask image channel during registration.
Figure~\ref{fig:ants:pre-tuning:masks} provides evidence that including a mask channel has added value in terms of Dice score for registration of all organs.
The difference in performance is only slightly visible in Figure~\ref{fig:ants:pre-tuning:renders:masks}, but the difference in metric values motivates our decision to use a mask channel in the upcoming patient-specific configuration steps.

\input{sup/figures/tex/a-ants/pre-tuning-step-size.tex}

\input{sup/figures/tex/a-ants/pre-tuning-renders-step-size.tex}

\noindentparagraph{Gradient step size}
Third, we examine the impact of using a different gradient step size on the registration performance of ANTs.
A larger step size between time points in ANTs' registration could lead to larger deformations becoming feasible, since optimization is less likely to get stuck in local minima.
Figure~\ref{fig:ants:pre-tuning:step-size} indicates that choosing a larger step size than the recommended value of 0.1 can be beneficial, with 1.0 providing a good trade-off for different organs.
Larger step sizes such as 5.0 cause the algorithm to overshoot the target and strongly deform a number of organs, as can be seen in the contour renders (Figure~\ref{fig:ants:pre-tuning:renders:step-size}).
We choose a gradient step size of 1.0 for its good trade-off between performance targets.

\input{sup/figures/tex/a-ants/pre-tuning-update-weight.tex}

\input{sup/figures/tex/a-ants/pre-tuning-renders-update-weight.tex}

\noindentparagraph{Update regularization weight}
Finally, we use the deduced settings from the previous three sweeps to test which update regularization weight performs best.
Figure~\ref{fig:ants:pre-tuning:update-weight} shows best overall performance for 4.0, in both metrics.
Visually, Figure~\ref{fig:ants:pre-tuning:renders:update-weight} indicates that weights 4.0 and 5.0 lead to the best registration outcomes, with little visible difference between the two.
Based on visual and quantitative results, we choose an update regularization weight of 4.0 for the patient-specific configuration experiments.

\subsubsection{Fine-grained configuration experiments per patient}
\label{sec:approach-configurations:ants:main-tuning}

We try exponentially increasing total regularization weights for all problem instances.
Figures~\ref{fig:ants:main-tuning:dice} and~\ref{fig:ants:main-tuning:hausdorff-95} plot the Dice scores and Hausdorff 95th percentiles for each problem instance, and Figures~\ref{fig:ants:main-tuning:renders:603}--\ref{fig:ants:main-tuning:renders:618} show renders of the deformed contours that ANTs predicts for these instances.
We observe that regularization has a strong impact on performance in all examined cases, but that often the (relatively) better outcomes are still acquired without regularization.
Figures~\ref{fig:ants:main-tuning:renders:603}--\ref{fig:ants:main-tuning:renders:617} show ANTs failing to model the large deformation taking place in the bladder and its surrounding organs, regardless of the regularization.
The Dice and Hausdorff metric results underscore these observations.
In Figure~\ref{fig:ants:main-tuning:renders:618}, ANTs shows that it can model the bladder deformation quite closely, but it should be noted that this is morphologically also the easiest problem.

\input{sup/figures/tex/a-ants/main-tuning-dice.tex}
\input{sup/figures/tex/a-ants/main-tuning-hausdorff-95.tex}
\input{sup/figures/tex/a-ants/main-tuning-renders.tex}

\subsubsection{Run commands}
\label{sec:approach-configurations:parameter-files:ants}

We list the two commands that we used for registration with ANTs.
Tokens starting with the \texttt{\$} character denote variables that are resolved before we execute these commands.
Note that the random seed, even though given to the command, is not functional and does not change the output.

\input{sup/figures/tex/a-ants/commands.tex}

\clearpage

\subsection{This Work: MOREA}
\label{sec:approach-configurations:morea}

We describe several coarse-grained configuration experiments that we conducted with MOREA on Patient~1.
The base parameter file we derived from these experiments can be found in Section~\ref{sec:approach-configurations:morea:parameters}.
We do not conduct fine-grained configuration steps, since MOREA is a multi-objective approach.

For MOREA's guidance objective, we perform an additional pre-processing step on each scan, to address the discrepancy between resolutions in different dimensions.
The initial resampling step bringing each scan to a uniform voxel resolution of 1.5\emph{mm} leads to the between-slice dimension being over-sampled (originally, slices are 3\emph{mm} apart).
Contour annotations are placed only on slices, which means that the new slices added by resampling to 1.5\emph{mm}, between original slices, do not have contour information.
These slice ``gaps'' in the contours of objects can be exploited during optimization.
We address this with an intermediate step, building a 3D model of each object across slices and generating border points from this model.

\subsubsection{Coarse-grained configuration experiments}

\noindentparagraph{Heterogeneous elasticity}

In Section~4.1, we describe a model that enables capturing biomechanical properties of different tissue types in the deformation magnitude objective.
The core principle of this biomechanical model is to ascribe heterogeneous elasticities to different regions of image space, corresponding with objects (e.g., organs and bones) present.
In this first configuration experiment, we compare the performance of this model with the performance of the model which is used by prior work~\cite{Andreadis2022}, assuming homogeneous elasticity of image space.
This experiment was conducted without a contour on the body, later experiments do have this contour.

The metric results in Figure~\ref{fig:morea:pre-tuning:magnitude} indicate that the heterogeneous model generally receives higher Dice scores and similar Hausdorff 95th percentiles.
Figure~\ref{fig:morea:pre-tuning:renders:magnitude} shows renderings of selected solutions with the heterogeneous and homogeneous models, which confirm this trend.
We observe in both slices that heterogeneous elasticity especially shows improved performance on the bladder deformation, potentially due to the increased elasticity that this models assigns to the bladder.

\input{sup/figures/tex/a-morea/pre-tuning-magnitude.tex}

\input{sup/figures/tex/a-morea/pre-tuning-renders-magnitude.tex}

\noindentparagraph{Mesh generation}

Using the biomechanical model that experiments in the previous subsection covered, we now investigate the impact of different mesh point placement strategies.
The strategy used to create meshes from these points is described in Section~4.2.1.

In this experiment, compare how well a random (Sobol-sequence based) placement compares to a contour-based strategy where points are sampled per contour and a contour-based strategy which has special handling for the bladder's surface.
Figure~\ref{fig:morea:pre-tuning:placement:dice} shows the bladder being modeled best by the last strategy, with contour-based strategies in general performing better than random, across organs.
The renders in Figure~\ref{fig:morea:pre-tuning:renders:placement} indicate that a random placement method can model the general deformation, but is too coarse to accurately treat details of specific organs and parts of the bones.
Both contour-based strategies perform well, but around the bladder's surface, the strategy with special surface constraints excels.

\noindentparagraph{Supplying guidance information}

The multi-objective line of registration approaches, which MOREA continues, can have a third objective that captures guidance (contour) match.
In this experiment, we assess what the impact of this objective is on the quality of registrations.

The quantitative results in Figure~\ref{fig:morea:pre-tuning:disable-guidance} leave little doubt that the adoption of a guidance objective is crucial to modeling large deformations.
Without it, the bladder remains largely in place, as can be seen in Figure~\ref{fig:morea:pre-tuning:renders:disable-guidance}. 
It seems that in this problem, image information is not sufficient to guide the optimization.

\input{sup/figures/tex/a-morea/pre-tuning-placement.tex}

\input{sup/figures/tex/a-morea/pre-tuning-renders-placement.tex}

\input{sup/figures/tex/a-morea/pre-tuning-disable-guidance.tex}

\input{sup/figures/tex/a-morea/pre-tuning-renders-disable-guidance.tex}

\clearpage

\subsubsection{Parameter file}
\label{sec:approach-configurations:morea:parameters}
We pass parameters to MOREA in a self-written parameter file format.
Below we list the parameter file used as basis for the experiments listed in this work.

\input{sup/figures/tex/a-morea/parameter-files.tex}

\clearpage

%% file: sup/figures/tex/a-elastix/pre-tuning-symmetry.tex
\begin{figure}
\centering
\begin{subfigure}{\linewidth}
  \centering
  \includegraphics[width=\linewidth]{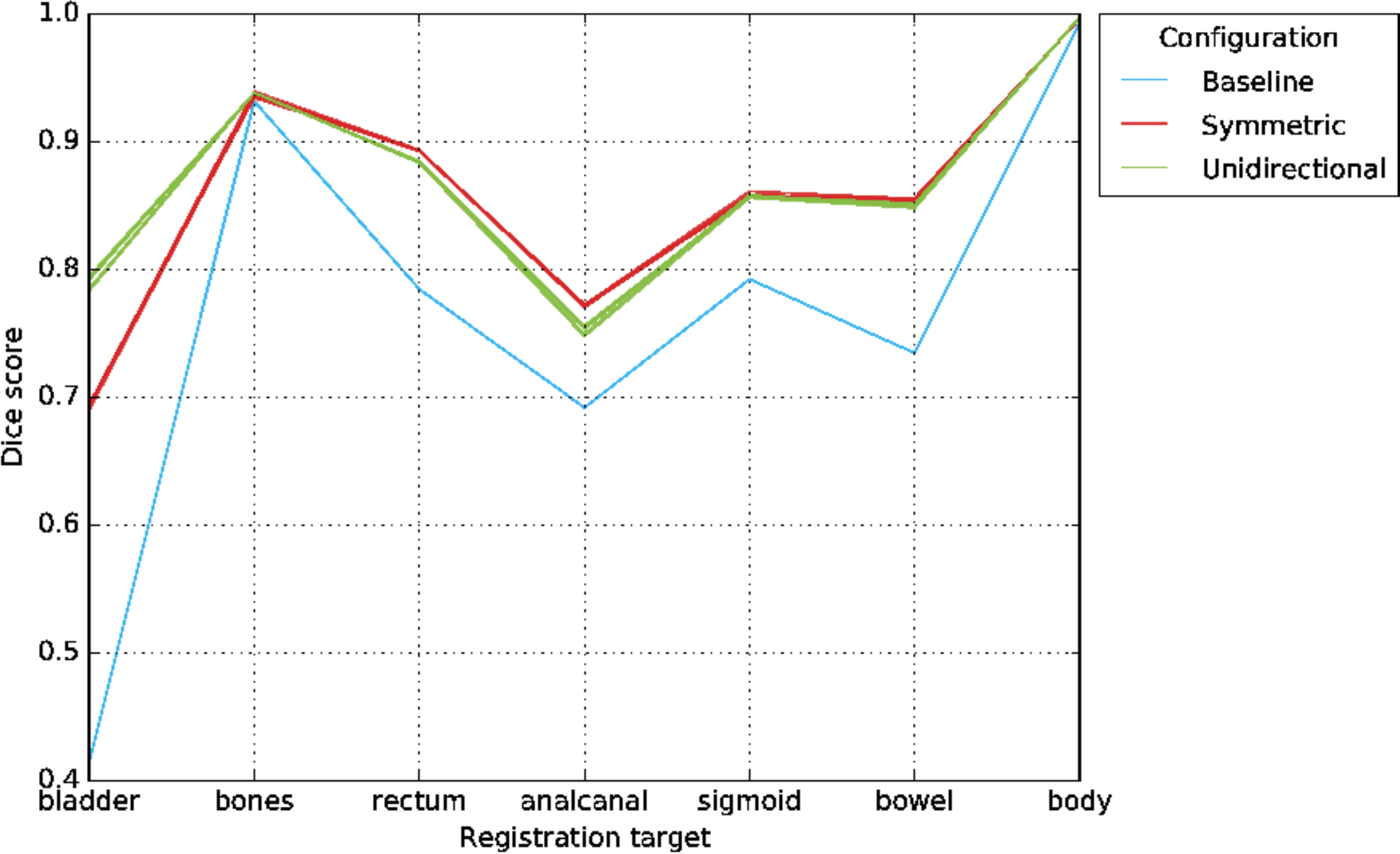}
  \caption{Dice scores.}
  \label{fig:elastix:pre-tuning:symmetry:dice}
\end{subfigure}
\begin{subfigure}{\linewidth}
  \centering
  \includegraphics[width=\linewidth]{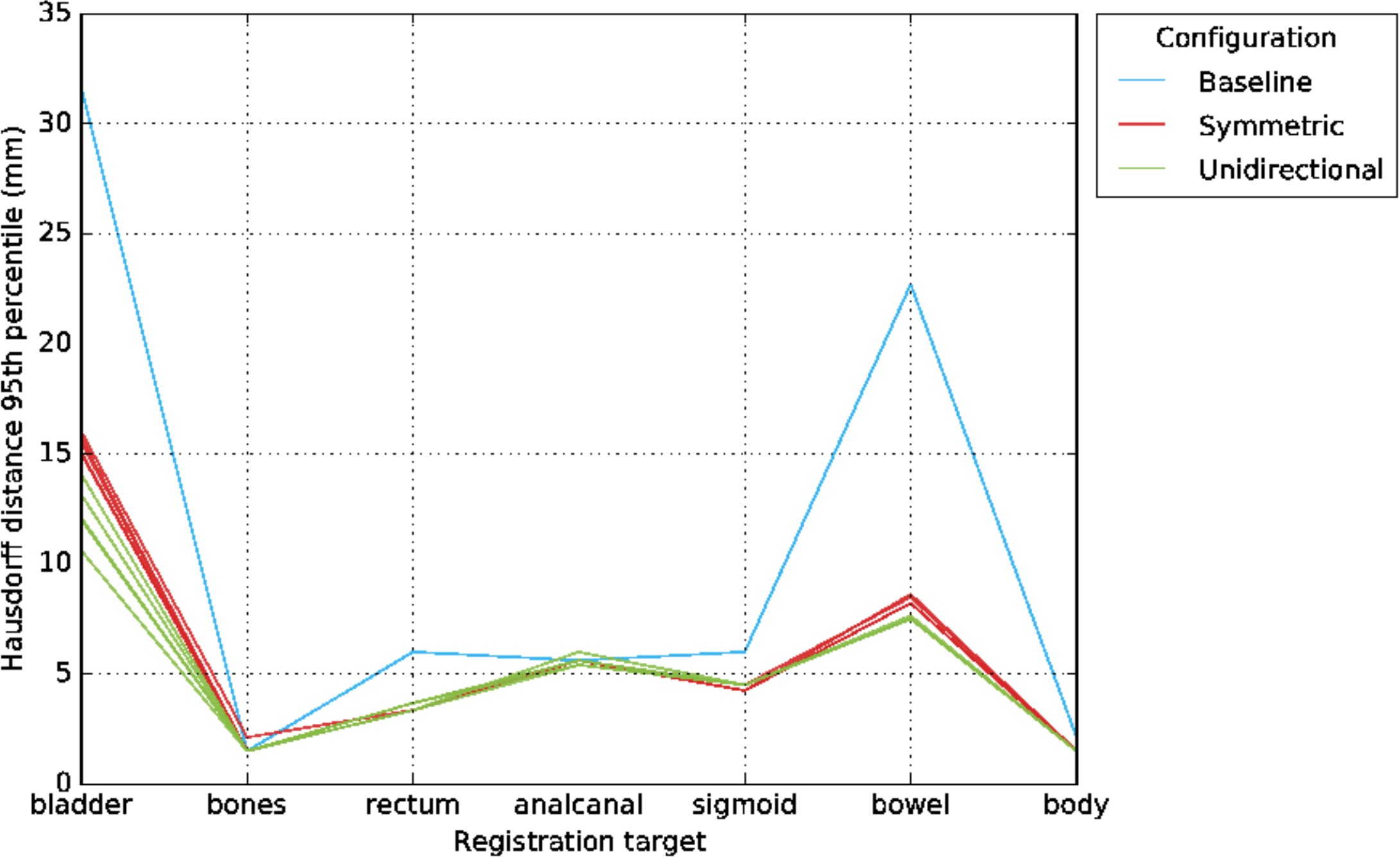}
  \caption{95th percentiles of the Hausdorff distance.}
  \label{fig:elastix:pre-tuning:symmetry:hausdorff-95}
\end{subfigure}
\vspace{-0.25cm}
\caption{Comparison of symmetric and unidirectional registration in Elastix, for multiple runs. The baseline score after rigid registration is plotted in blue.}
\label{fig:elastix:pre-tuning:symmetry}
\end{figure}

%% file: sup/figures/tex/a-elastix/pre-tuning-renders-symmetry.tex
\begin{figure}
\centering
\begin{subfigure}[b]{.49\linewidth}
  \centering
  \includegraphics[width=\linewidth]{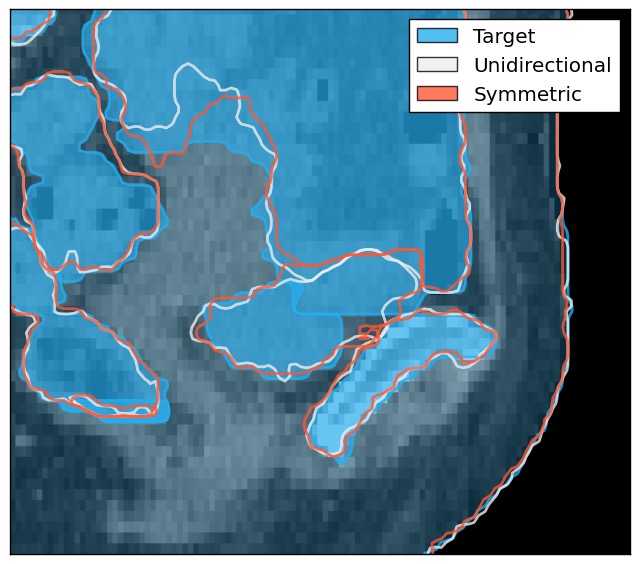}
  \caption{Sagittal slice.}
  \label{fig:elastix:pre-tuning:renders:symmetry:sagittal}
\end{subfigure}%
\begin{subfigure}[b]{.49\linewidth}
  \centering
  \includegraphics[width=\linewidth]{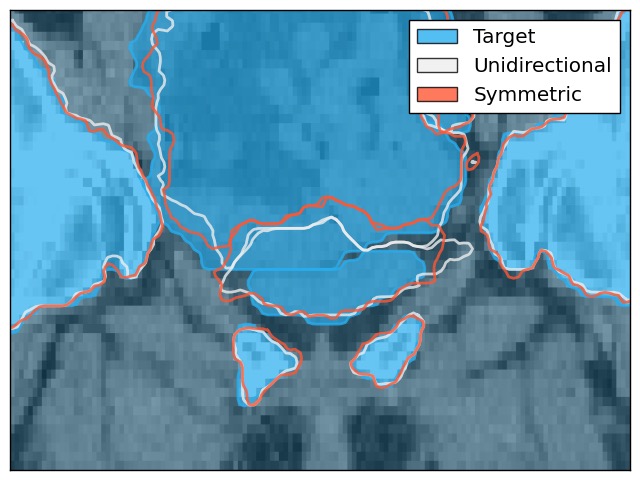}
  \caption{Coronal slice.}
  \label{fig:elastix:pre-tuning:renders:symmetry:coronal}
\end{subfigure}
\vspace{-0.25cm}
\caption{Visual renders of deformations predicted by Elastix configurations using unidirectional and symmetric registration, without mask registration step.}
\label{fig:elastix:pre-tuning:renders:symmetry}
\end{figure}

%% file: sup/figures/tex/a-elastix/pre-tuning-masks.tex
\begin{figure}
\centering
\begin{subfigure}{\linewidth}
  \centering
  \includegraphics[width=\linewidth]{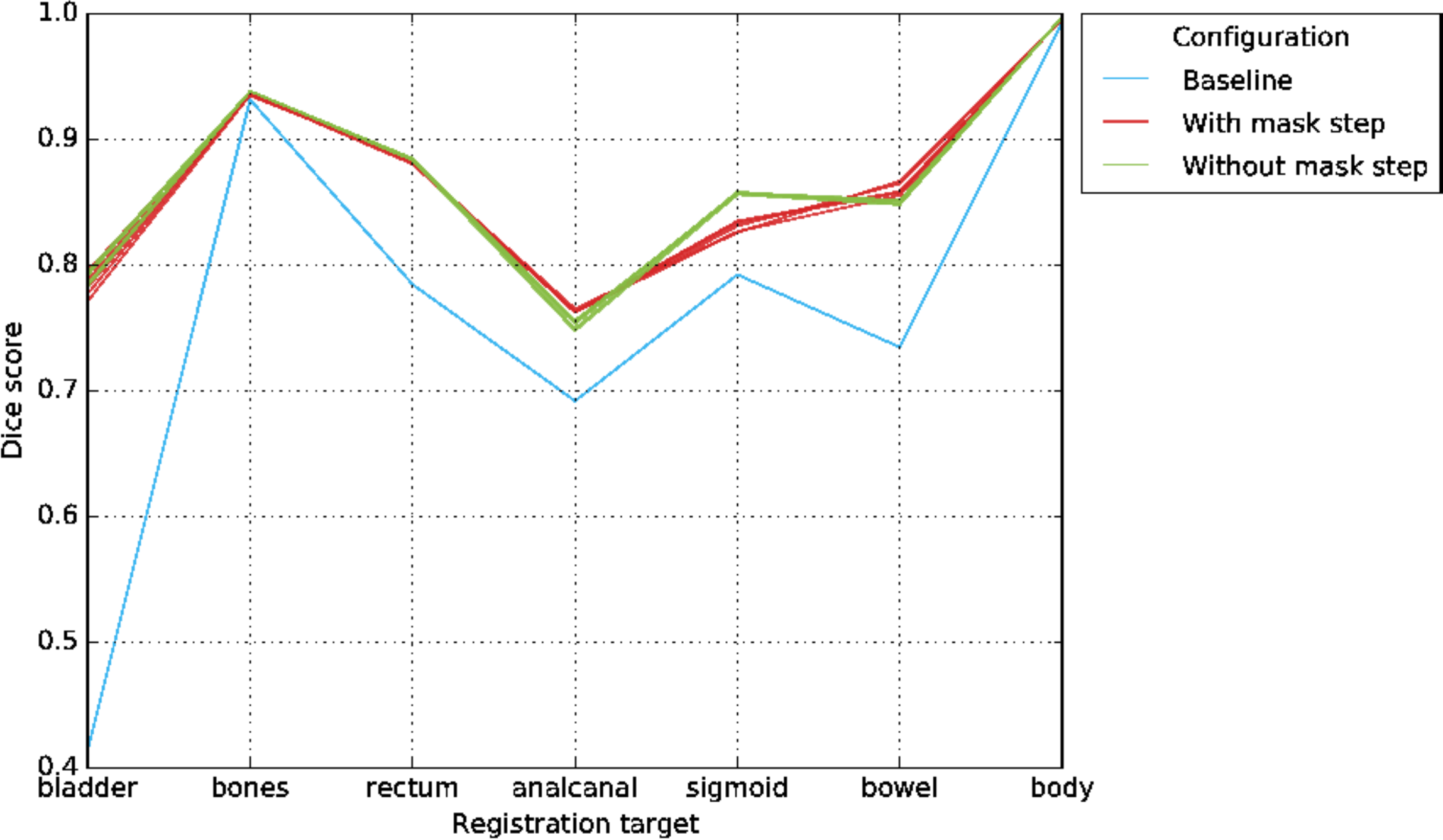}
  \caption{Dice scores.}
  \label{fig:elastix:pre-tuning:masks:dice}
\end{subfigure}
\begin{subfigure}{\linewidth}
  \centering
  \includegraphics[width=\linewidth]{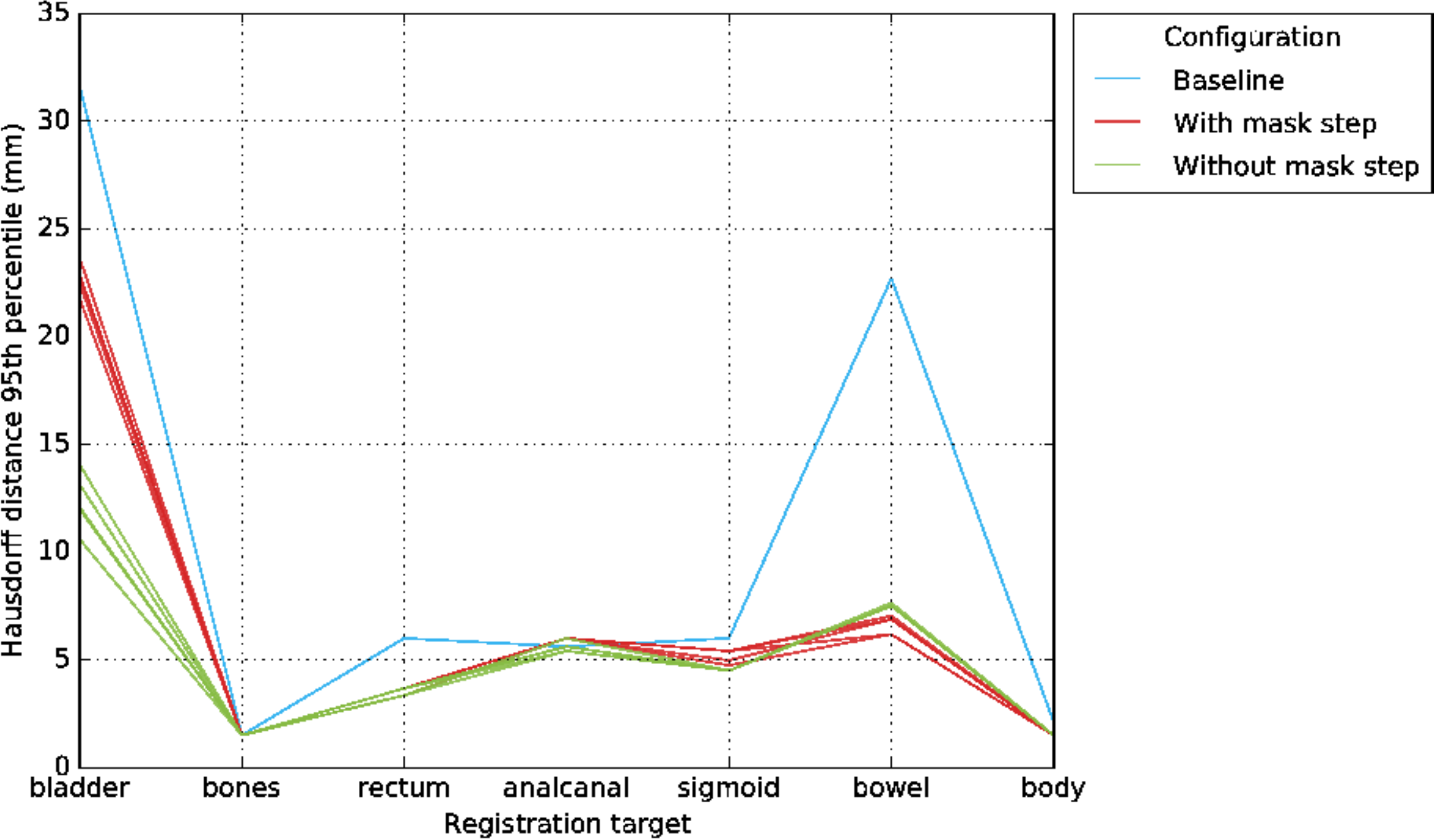}
  \caption{95th percentiles of the Hausdorff distance.}
  \label{fig:elastix:pre-tuning:masks:hausdorff-95}
\end{subfigure}
\vspace{-0.25cm}
\caption{Comparison of unidirectional registration with and without a composite mask registration step in Elastix, for multiple runs. The baseline score after rigid registration is plotted in blue.}
\label{fig:elastix:pre-tuning:masks}
\end{figure}

%% file: sup/figures/tex/a-elastix/pre-tuning-renders-masks.tex
\begin{figure}
\centering
\begin{subfigure}[b]{.49\linewidth}
  \centering
  \includegraphics[width=\linewidth]{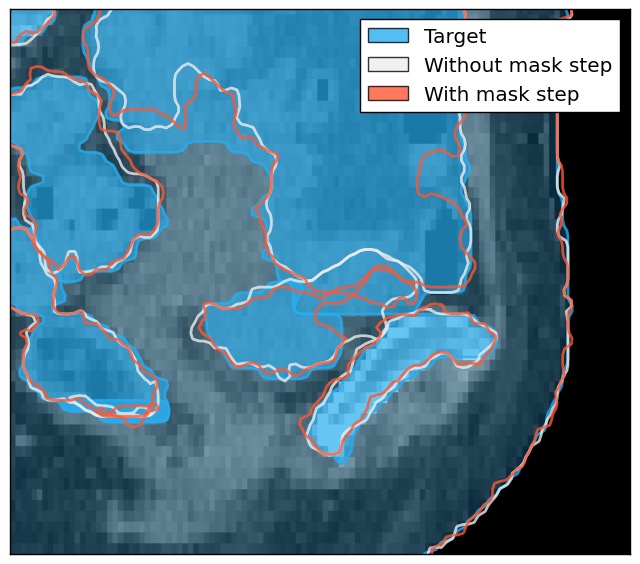}
  \caption{Sagittal slice.}
  \label{fig:elastix:pre-tuning:renders:masks:sagittal}
\end{subfigure}%
\begin{subfigure}[b]{.49\linewidth}
  \centering
  \includegraphics[width=\linewidth]{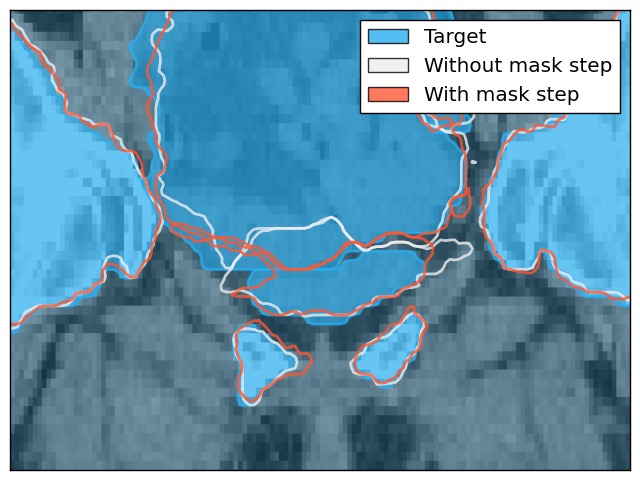}
  \caption{Coronal slice.}
  \label{fig:elastix:pre-tuning:renders:masks:coronal}
\end{subfigure}
\vspace{-0.25cm}
\caption{Visual renders of deformations predicted by Elastix configurations with and without a composite mask registration step, using unidirectional registration.}
\label{fig:elastix:pre-tuning:renders:masks}
\end{figure}

%% file: sup/figures/tex/a-elastix/main-tuning-dice.tex
\begin{figure*}
\centering
\begin{subfigure}{.4\linewidth}
  \centering
  \includegraphics[width=\linewidth]{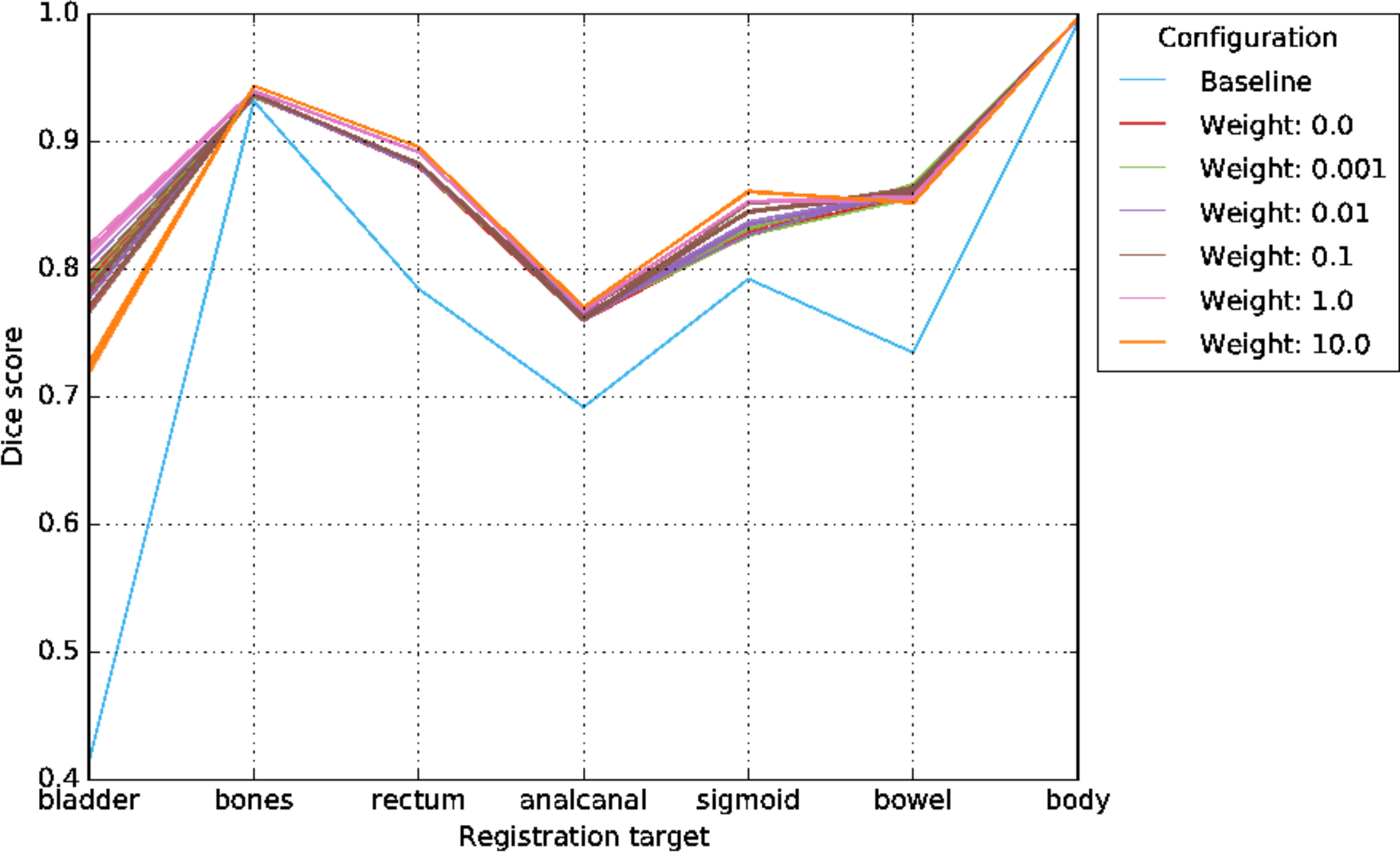}
  \caption{Patient 1.}
  \label{fig:elastix:main-tuning:dice:603}
\end{subfigure}%
\hspace{0.019\linewidth}
\begin{subfigure}{.4\linewidth}
  \centering
  \includegraphics[width=\linewidth]{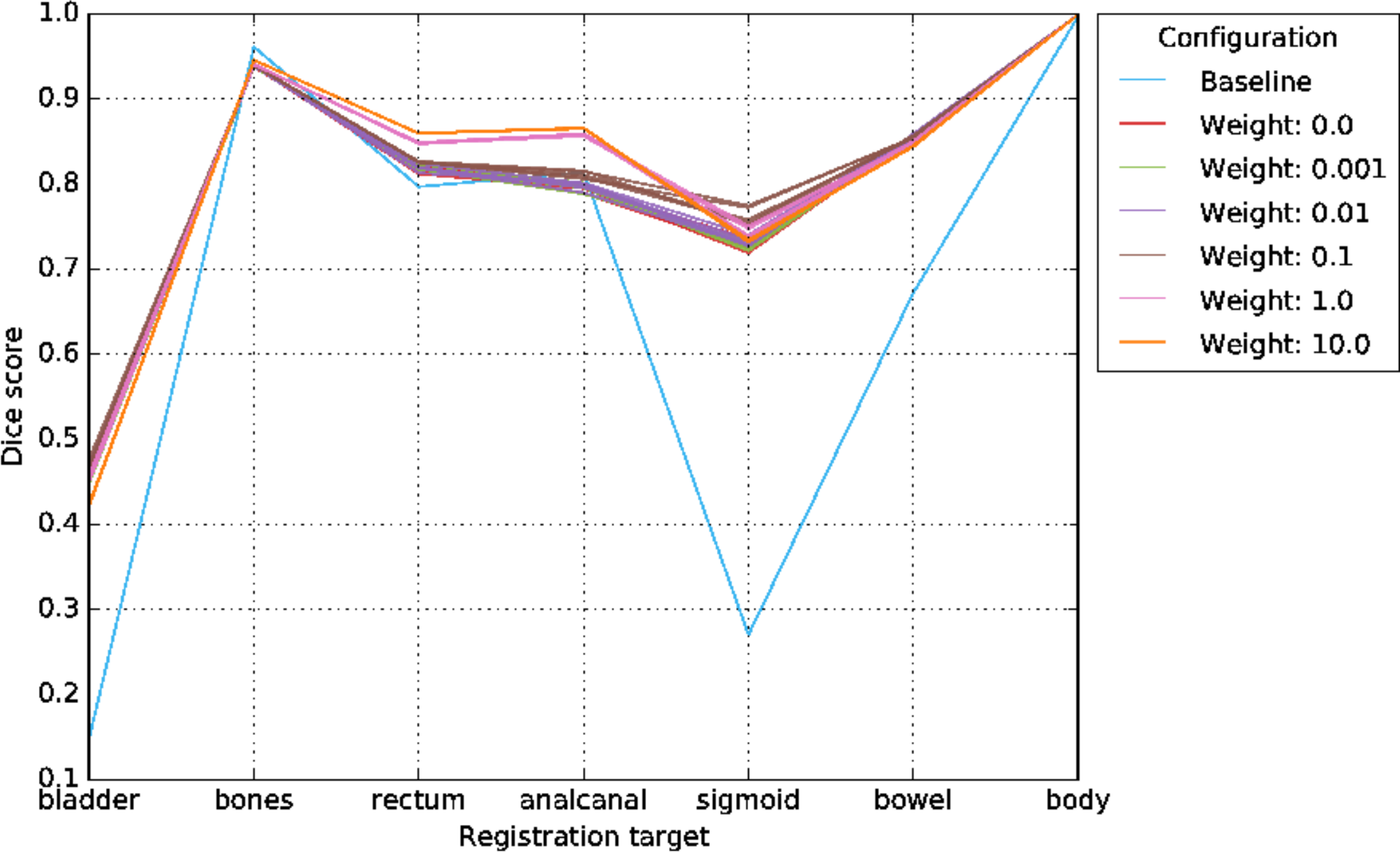}
  \caption{Patient 2.}
  \label{fig:elastix:main-tuning:dice:611}
\end{subfigure}%
\hspace{0.019\linewidth}
\begin{subfigure}{.4\linewidth}
  \centering
  \includegraphics[width=\linewidth]{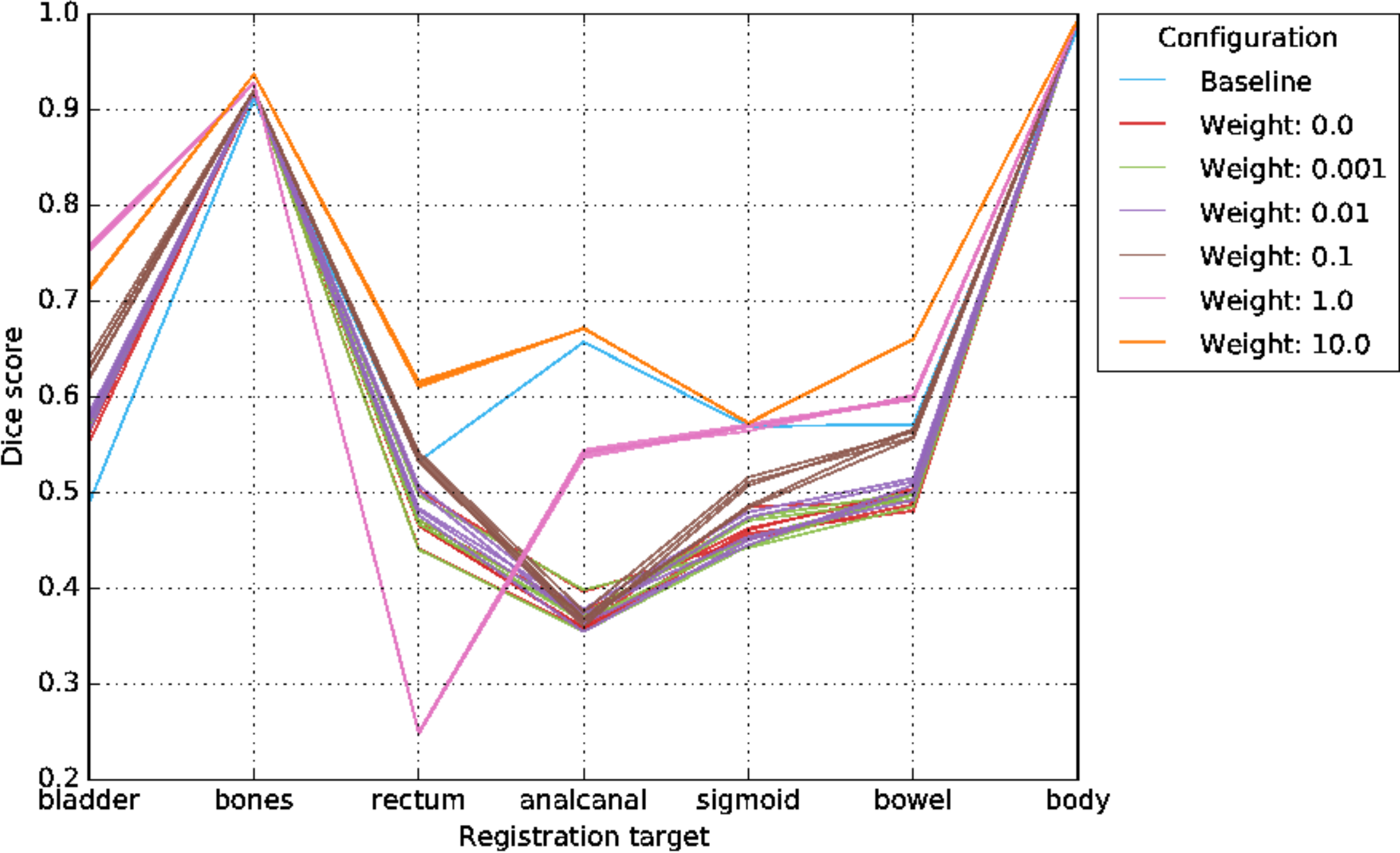}
  \caption{Patient 3.}
  \label{fig:elastix:main-tuning:dice:617}
\end{subfigure}%
\hspace{0.019\linewidth}
\begin{subfigure}{.4\linewidth}
  \centering
  \includegraphics[width=\linewidth]{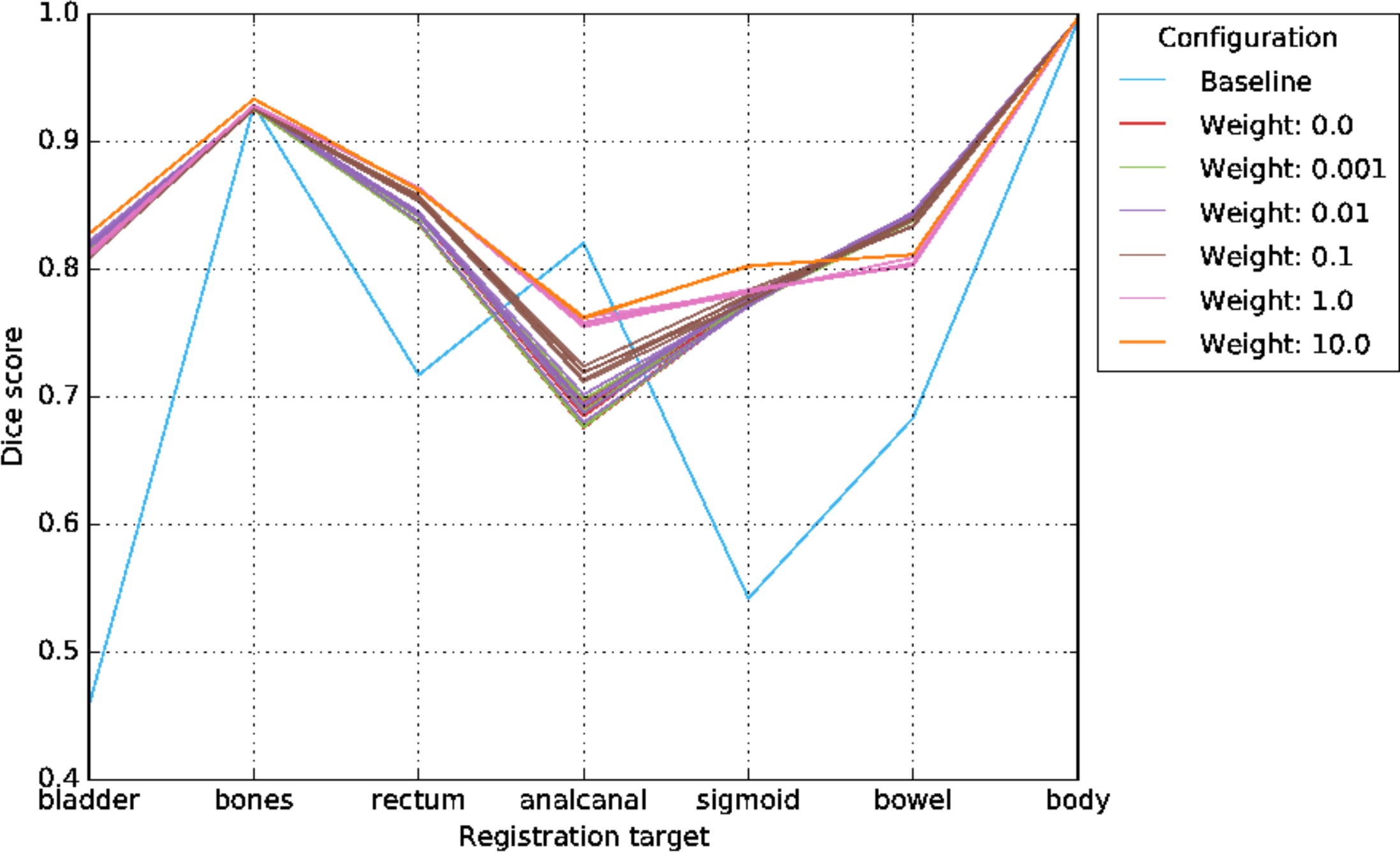}
  \caption{Patient 4.}
  \label{fig:elastix:main-tuning:dice:618}
\end{subfigure}%
\vspace{-0.25cm}
\caption{Dice scores for per-patient fine-grained configuration runs in Elastix. The baseline score after rigid registration is plotted in blue.}
\label{fig:elastix:main-tuning:dice}
\end{figure*}

%% file: sup/figures/tex/a-elastix/main-tuning-hausdorff-95.tex
\begin{figure*}
\centering
\begin{subfigure}{.4\linewidth}
  \centering
  \includegraphics[width=\linewidth]{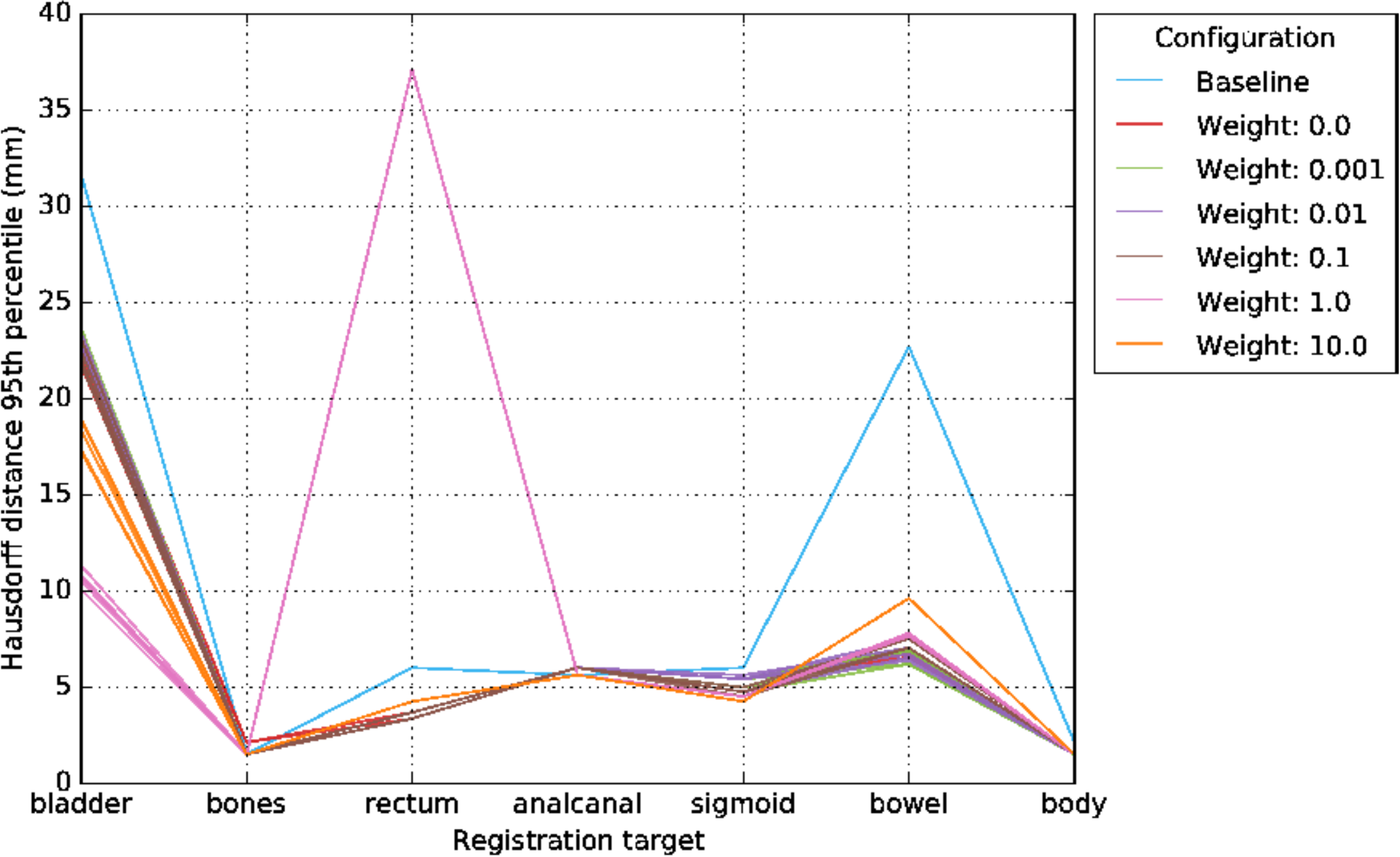}
  \caption{Patient 1.}
  \label{fig:elastix:main-tuning:hausdorff-95:603}
\end{subfigure}%
\hspace{0.019\linewidth}
\begin{subfigure}{.4\linewidth}
  \centering
  \includegraphics[width=\linewidth]{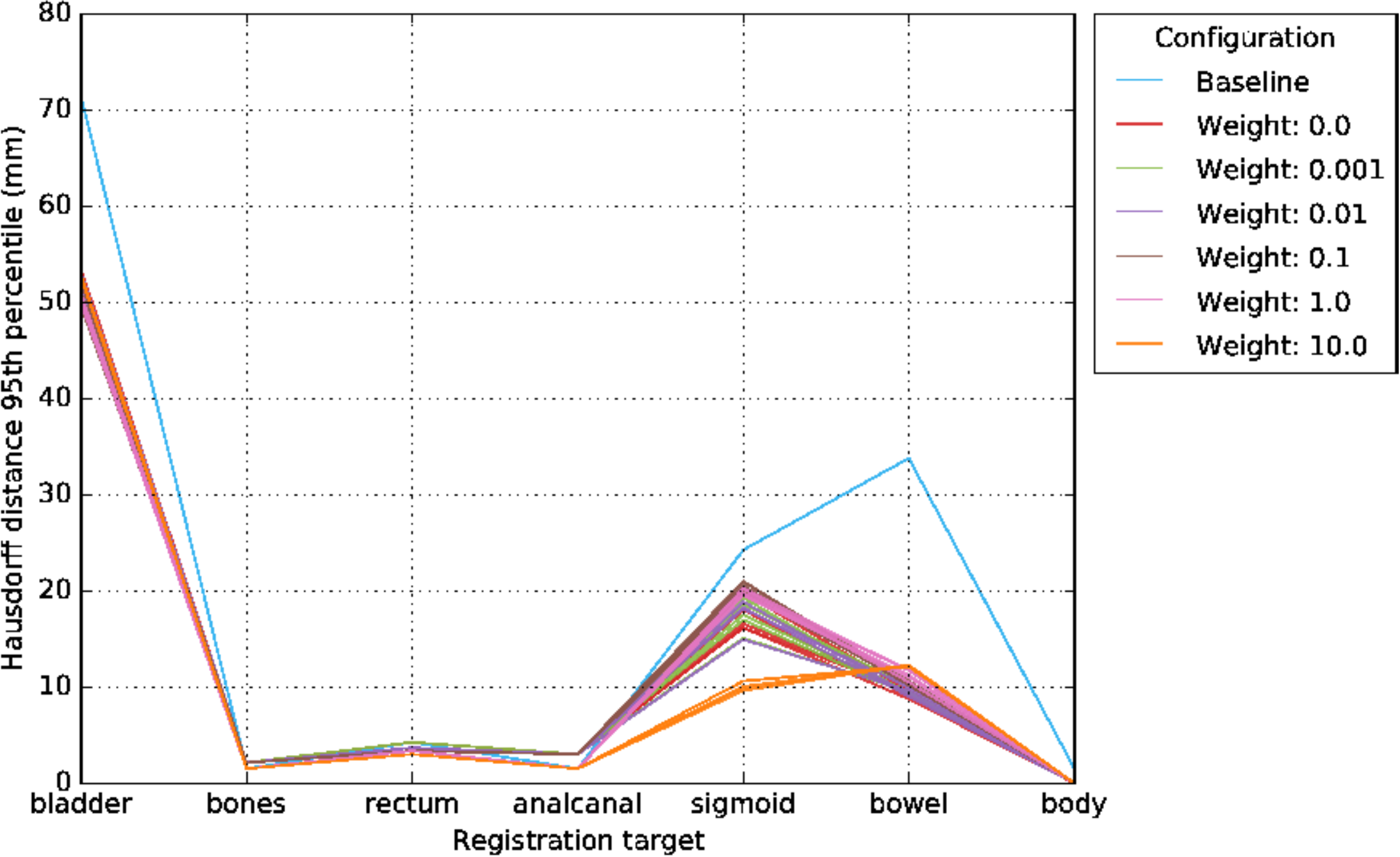}
  \caption{Patient 2.}
  \label{fig:elastix:main-tuning:hausdorff-95:611}
\end{subfigure}%
\hspace{0.019\linewidth}
\begin{subfigure}{.4\linewidth}
  \centering
  \includegraphics[width=\linewidth]{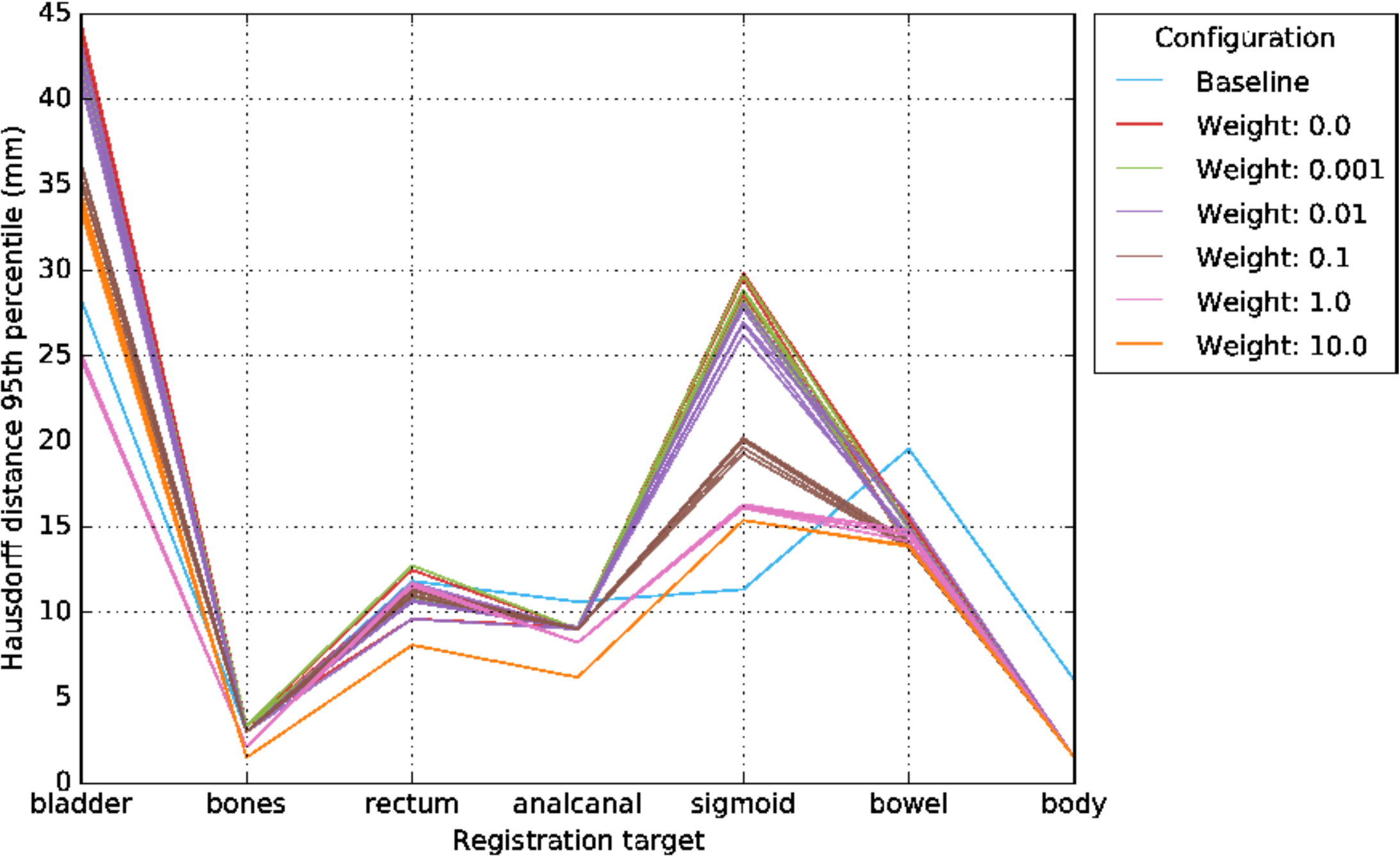}
  \caption{Patient 3.}
  \label{fig:elastix:main-tuning:hausdorff-95:617}
\end{subfigure}%
\hspace{0.019\linewidth}
\begin{subfigure}{.4\linewidth}
  \centering
  \includegraphics[width=\linewidth]{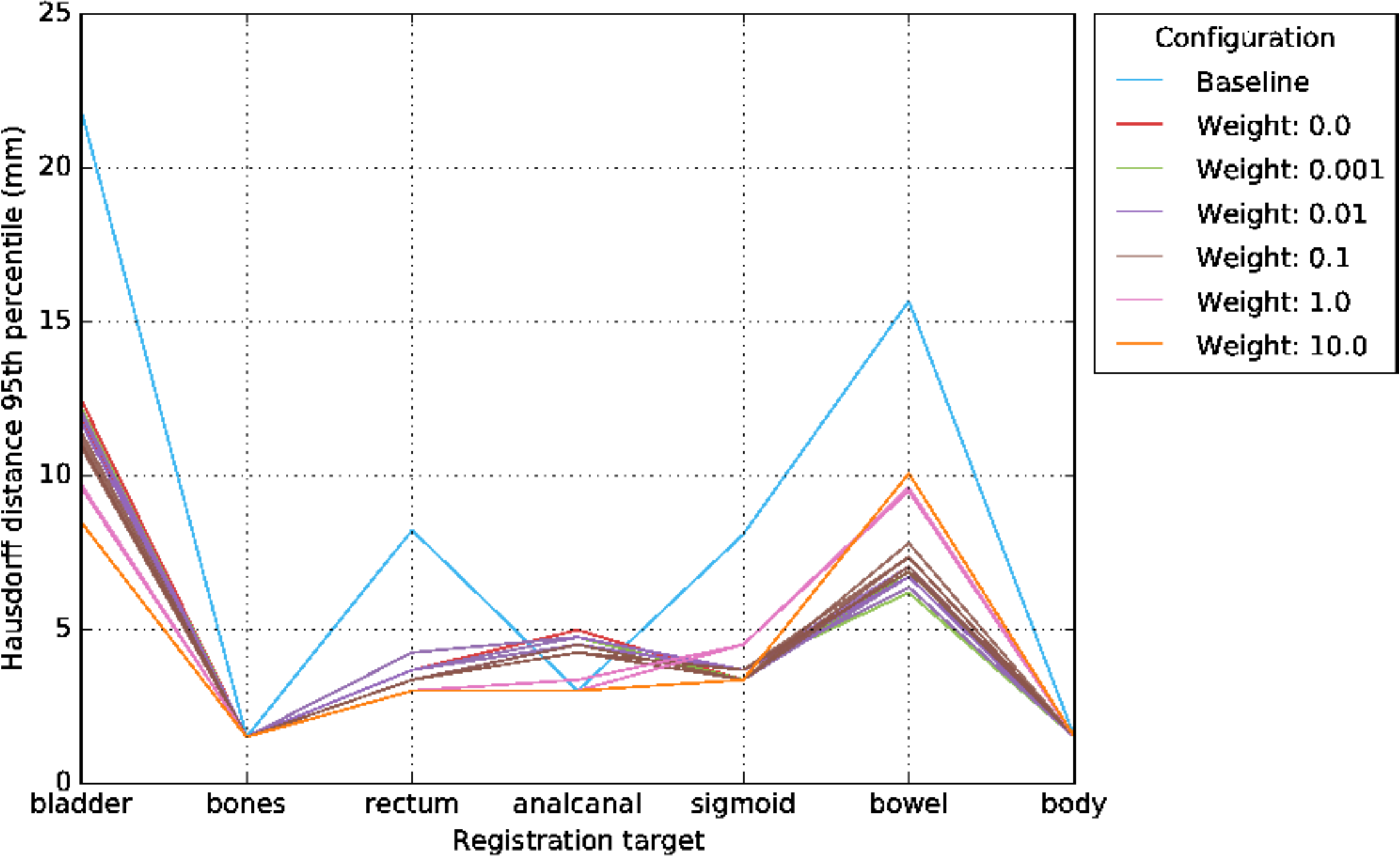}
  \caption{Patient 4.}
  \label{fig:elastix:main-tuning:hausdorff-95:618}
\end{subfigure}%
\vspace{-0.25cm}
\caption{Hausdorff 95th percentiles for per-patient fine-grained configuration runs in Elastix. The baseline score after rigid registration is plotted in blue.}
\label{fig:elastix:main-tuning:hausdorff-95}
\end{figure*}

%% file: sup/figures/tex/a-elastix/main-tuning-renders.tex
\begin{figure}
\centering
\begin{subfigure}[b]{.49\linewidth}
  \centering
  \includegraphics[width=\linewidth]{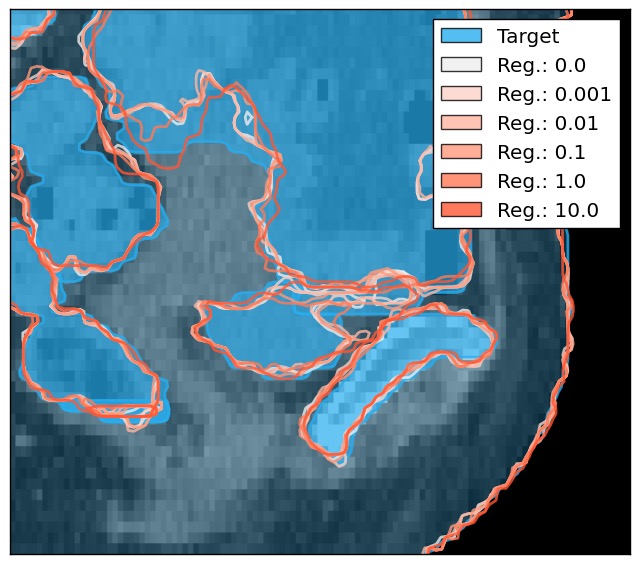}
  \caption{Sagittal slice.}
\end{subfigure}%
\begin{subfigure}[b]{.49\linewidth}
  \centering
  \includegraphics[width=\linewidth]{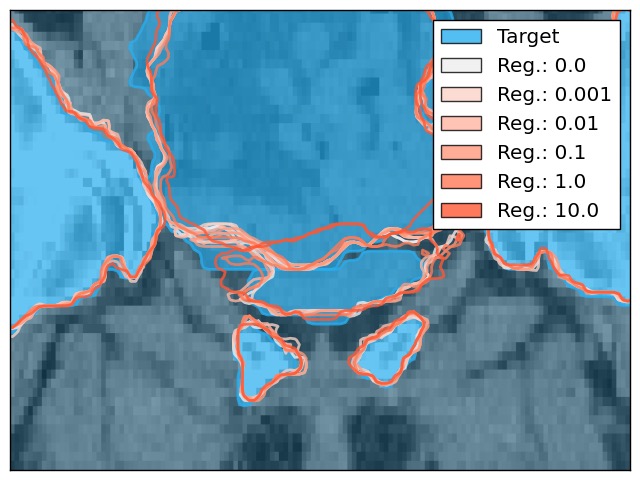}
  \caption{Coronal slice.}
\end{subfigure}
\vspace{-0.25cm}
\caption{Visual renders of deformations predicted by Elastix with different regularization weights, on Patient 1.}
\label{fig:elastix:main-tuning:renders:603}
\end{figure}

\begin{figure}
\centering
\begin{subfigure}[b]{.49\linewidth}
  \centering
  \includegraphics[width=\linewidth]{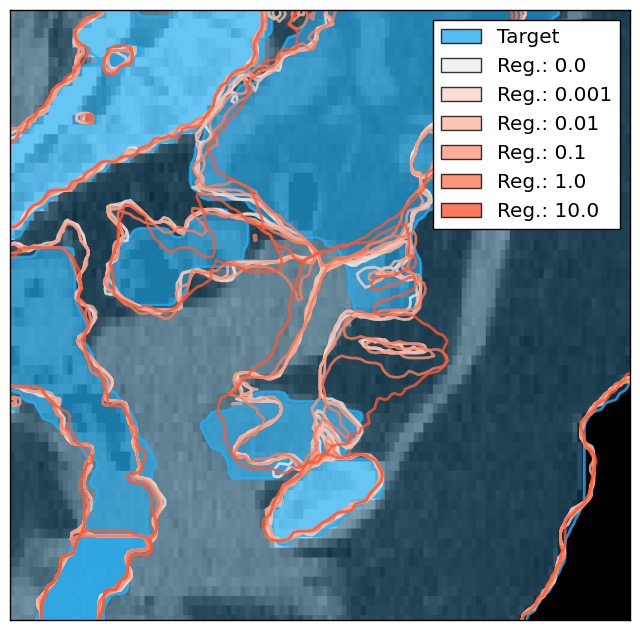}
  \caption{Sagittal slice.}
\end{subfigure}%
\begin{subfigure}[b]{.49\linewidth}
  \centering
  \includegraphics[width=\linewidth]{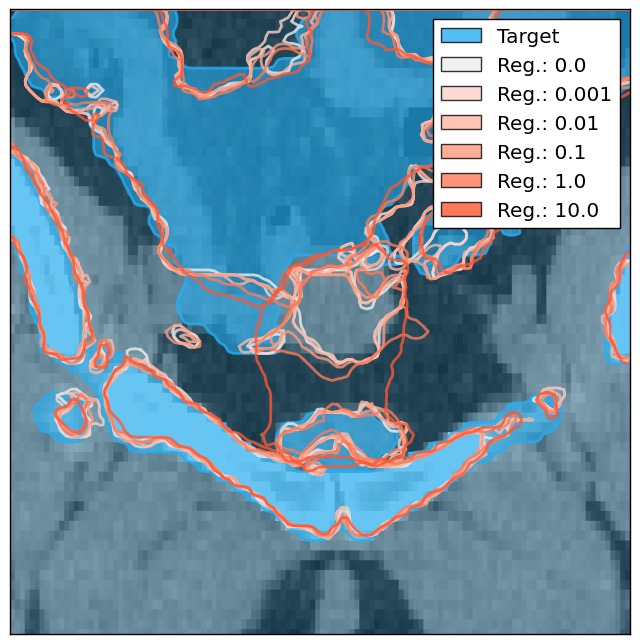}
  \caption{Coronal slice.}
\end{subfigure}
\vspace{-0.25cm}
\caption{Visual renders of deformations predicted by Elastix with different regularization weights, on Patient 2.}
\label{fig:elastix:main-tuning:renders:611}
\end{figure}

\begin{figure}
\centering
\begin{subfigure}[b]{.49\linewidth}
  \centering
  \includegraphics[width=\linewidth]{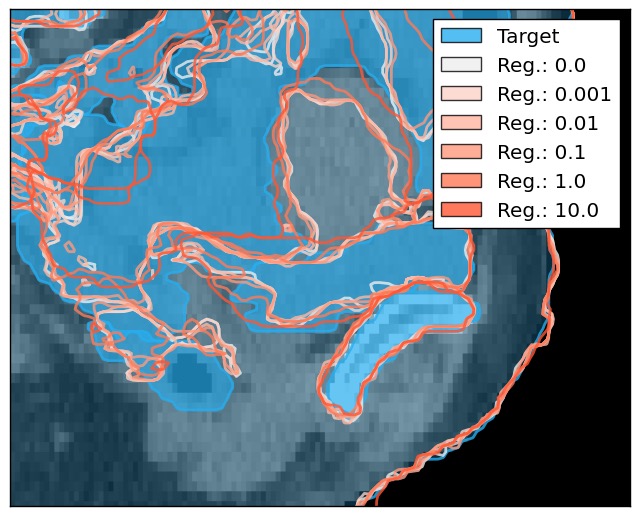}
  \caption{Sagittal slice.}
\end{subfigure}%
\begin{subfigure}[b]{.49\linewidth}
  \centering
  \includegraphics[width=\linewidth]{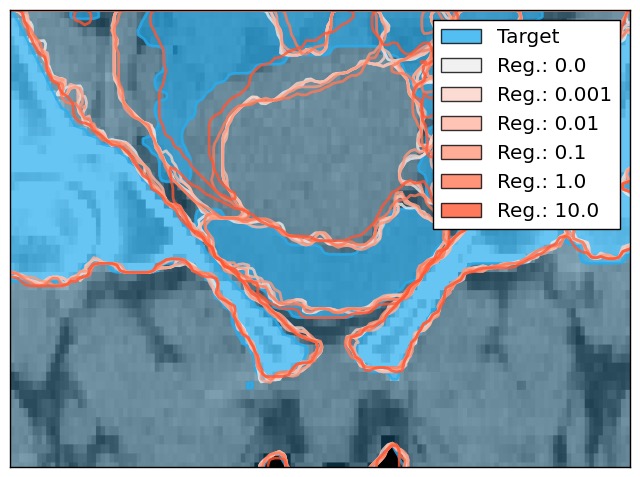}
  \caption{Coronal slice.}
\end{subfigure}
\vspace{-0.25cm}
\caption{Visual renders of deformations predicted by Elastix with different regularization weights, on Patient 3.}
\label{fig:elastix:main-tuning:renders:617}
\end{figure}

\begin{figure}
\centering
\begin{subfigure}[b]{.49\linewidth}
  \centering
  \includegraphics[width=\linewidth]{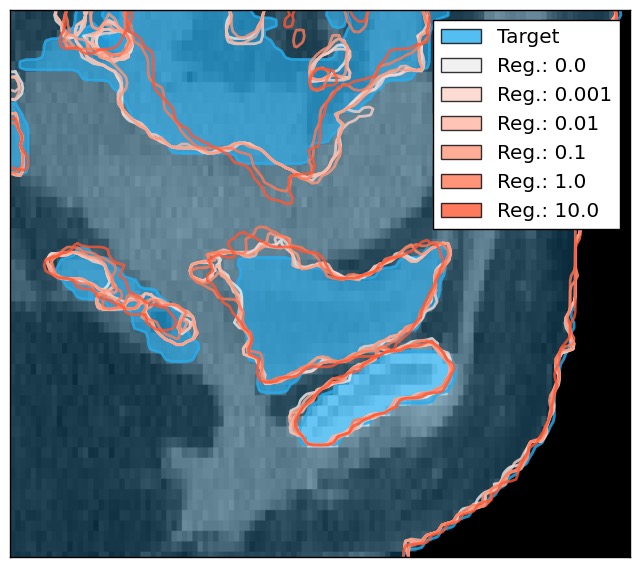}
  \caption{Sagittal slice.}
\end{subfigure}%
\begin{subfigure}[b]{.49\linewidth}
  \centering
  \includegraphics[width=\linewidth]{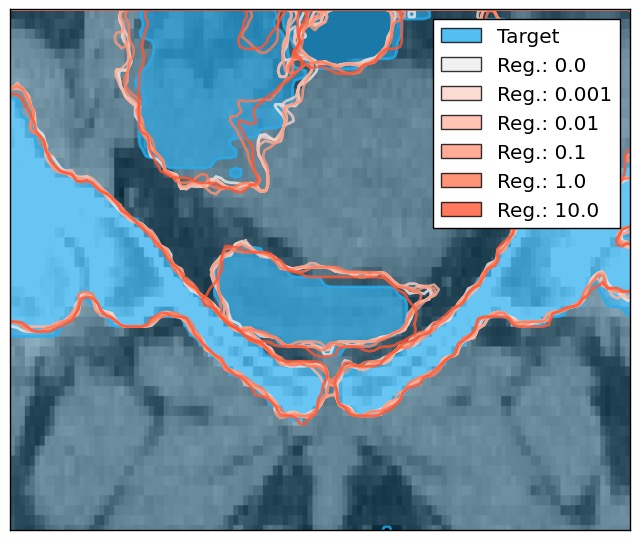}
  \caption{Coronal slice.}
\end{subfigure}
\vspace{-0.25cm}
\caption{Visual renders of deformations predicted by Elastix with different regularization weights, on Patient 4.}
\label{fig:elastix:main-tuning:renders:618}
\end{figure}


%% file: sup/figures/tex/a-elastix/parameter-files.tex
\begin{lstlisting}[caption={Forward transform parameters for conventional, unidirectional deformation.},label={listing:forward-unidirectional}]
// ImageTypes
(FixedImagePixelType "short")
(FixedImageDimension 3)
(MovingImagePixelType "short")
(MovingImageDimension 3)

// Multi resolution
(Registration "MultiMetricMultiResolutionRegistration")
(HowToCombineTransforms "Compose")
(NumberOfHistogramBins 32)
(NumberOfResolutions 4)
(MaximumNumberOfIterations 10000)

// Optimizer
(Optimizer "AdaptiveStochasticGradientDescent")
(AutomaticParameterEstimation "true")
(UseAdaptiveStepSizes "true")
(CheckNumberOfSamples "true")
(UseDirectionCosines "true")
(RandomSeed $random_seed)

// Metric
(Metric "AdvancedMattesMutualInformation" "TransformBendingEnergyPenalty")
(Metric0Weight 1.0)
(Metric1Weight $regularization_weight)

// Components
(FixedImagePyramid "FixedSmoothingImagePyramid")
(MovingImagePyramid "MovingSmoothingImagePyramid")
(Interpolator "BSplineInterpolator")
(ResampleInterpolator "FinalBSplineInterpolator")
(Resampler "DefaultResampler")
(Transform "BSplineTransform")

// Transform
(FinalGridSpacingInPhysicalUnits 2.0)

// Sampling
(ImageSampler "RandomCoordinate")
(NewSamplesEveryIteration "true")
(NumberOfSpatialSamples 20000)

// Interpolation and resampling
(BSplineInterpolationOrder 1)
(FinalBSplineInterpolationOrder 3)
(DefaultPixelValue 0)

// Output and other
(WriteTransformParametersEachIteration "false" "false" "false" "false" "false")
(WriteTransformParametersEachResolution "true" "true" "true" "true" "true")
(ShowExactMetricValue "false" "false" "false" "false" "false")
(WriteResultImageAfterEachResolution "false")
(WriteResultImage "true")
(ResultImagePixelType "short")
(ResultImageFormat "nii.gz")
\end{lstlisting}

\newpage

\begin{lstlisting}[caption={Forward transform parameters for symmetric deformation.},label={listing:forward-symmetric}]
// ImageTypes
(FixedImagePixelType "short")
(FixedInternalImagePixelType "short")
(FixedImageDimension 4)
(MovingImagePixelType "short")
(MovingInternalImagePixelType "short")
(MovingImageDimension 4)

// Multi resolution
(Registration "MultiResolutionRegistration")
(HowToCombineTransforms "Compose")
(NumberOfHistogramBins 32)
(NumberOfResolutions 4)
(MaximumNumberOfIterations 10000)
(MaximumNumberOfSamplingAttempts 10)

// Optimizer
(Optimizer "AdaptiveStochasticGradientDescent")
(AutomaticParameterEstimation "true")
(UseAdaptiveStepSizes "true")
(CheckNumberOfSamples "true")
(UseDirectionCosines "true")
(RandomSeed \$random_seed)

// Metric
(Metric "$metric")

(NumEigenValues 1)
(TemplateImage "ArithmeticAverage" "ArithmeticAverage")
(Combination "Sum" "Sum")
(SubtractMean "true")
(MovingImageDerivativeScales 1.0 1.0 1.0 0.0)

// Components
(FixedImagePyramid "FixedSmoothingImagePyramid")
(MovingImagePyramid "MovingSmoothingImagePyramid")
(ImagePyramidSchedule 8 8 8 0 4 4 4 0 2 2 2 0 1 1 1 0)
(Interpolator "ReducedDimensionBSplineInterpolator")
(ResampleInterpolator "FinalReducedDimensionBSplineInterpolator")
(Resampler "DefaultResampler")
(Transform "BSplineStackTransform")

// Transform
(FinalGridSpacingInPhysicalUnits 2.0)

// Sampling
(ImageSampler "RandomCoordinate")
(NewSamplesEveryIteration "true")
(NumberOfSpatialSamples 20000)

// Interpolation and resampling
(BSplineTransformSplineOrder 1)
(FinalBSplineInterpolationOrder 3)
(DefaultPixelValue 0)

// Output and other
(WriteTransformParametersEachIteration "false" "false" "false" "false")
(WriteTransformParametersEachResolution "true" "true" "true" "true")
(ShowExactMetricValue "false" "false" "false" "false")
(WriteResultImageAfterEachResolution "false")
(WriteResultImage "true")
(ResultImagePixelType "short")
(ResultImageFormat "nii.gz")
\end{lstlisting}

%% file: sup/figures/tex/a-ants/pre-tuning-cc.tex
\begin{figure}
\centering
\begin{subfigure}{\linewidth}
  \centering
  \includegraphics[width=\linewidth]{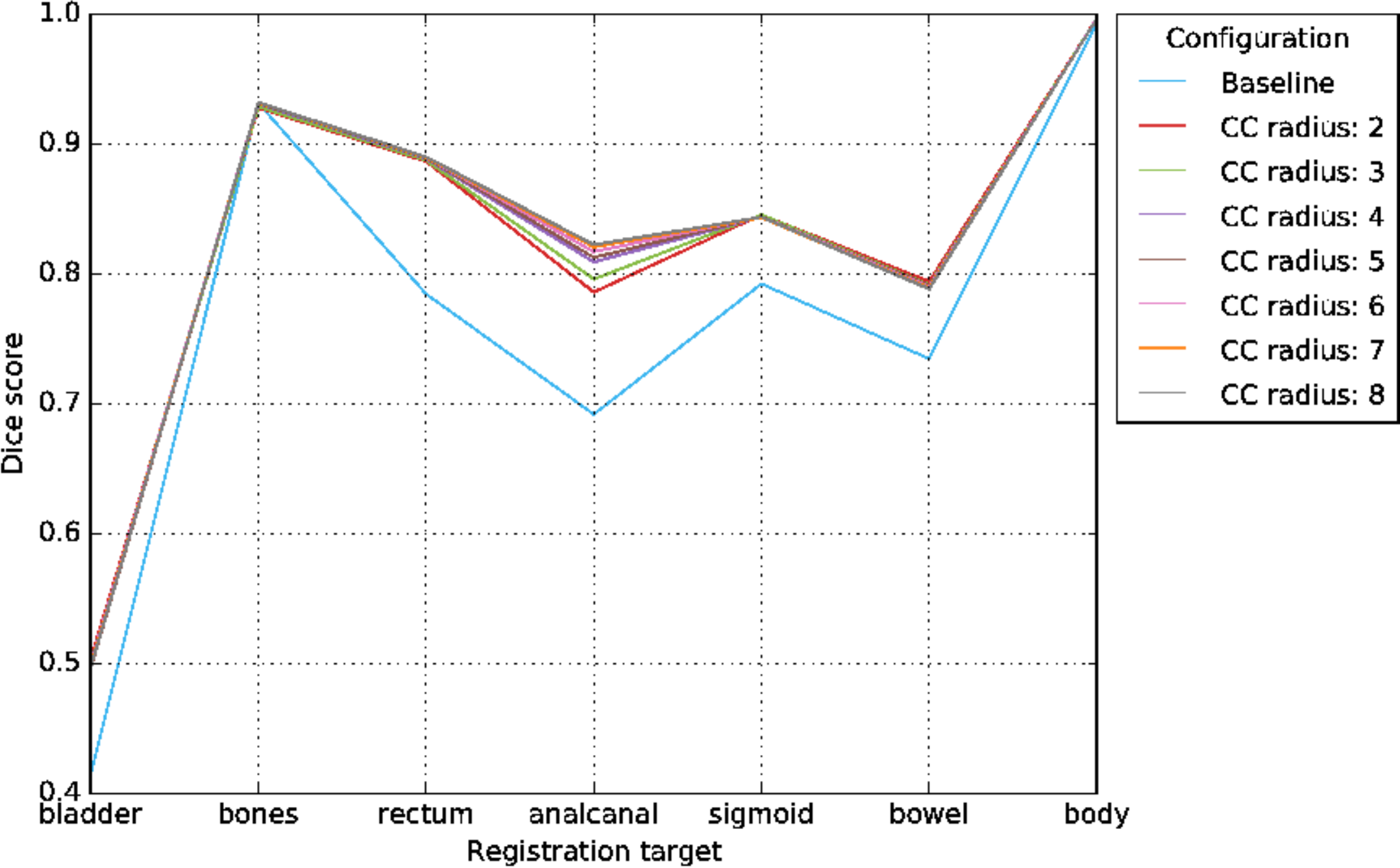}
  \caption{Dice scores.}
  \label{fig:ants:pre-tuning:cc:dice}
\end{subfigure}
\begin{subfigure}{\linewidth}
  \centering
  \includegraphics[width=\linewidth]{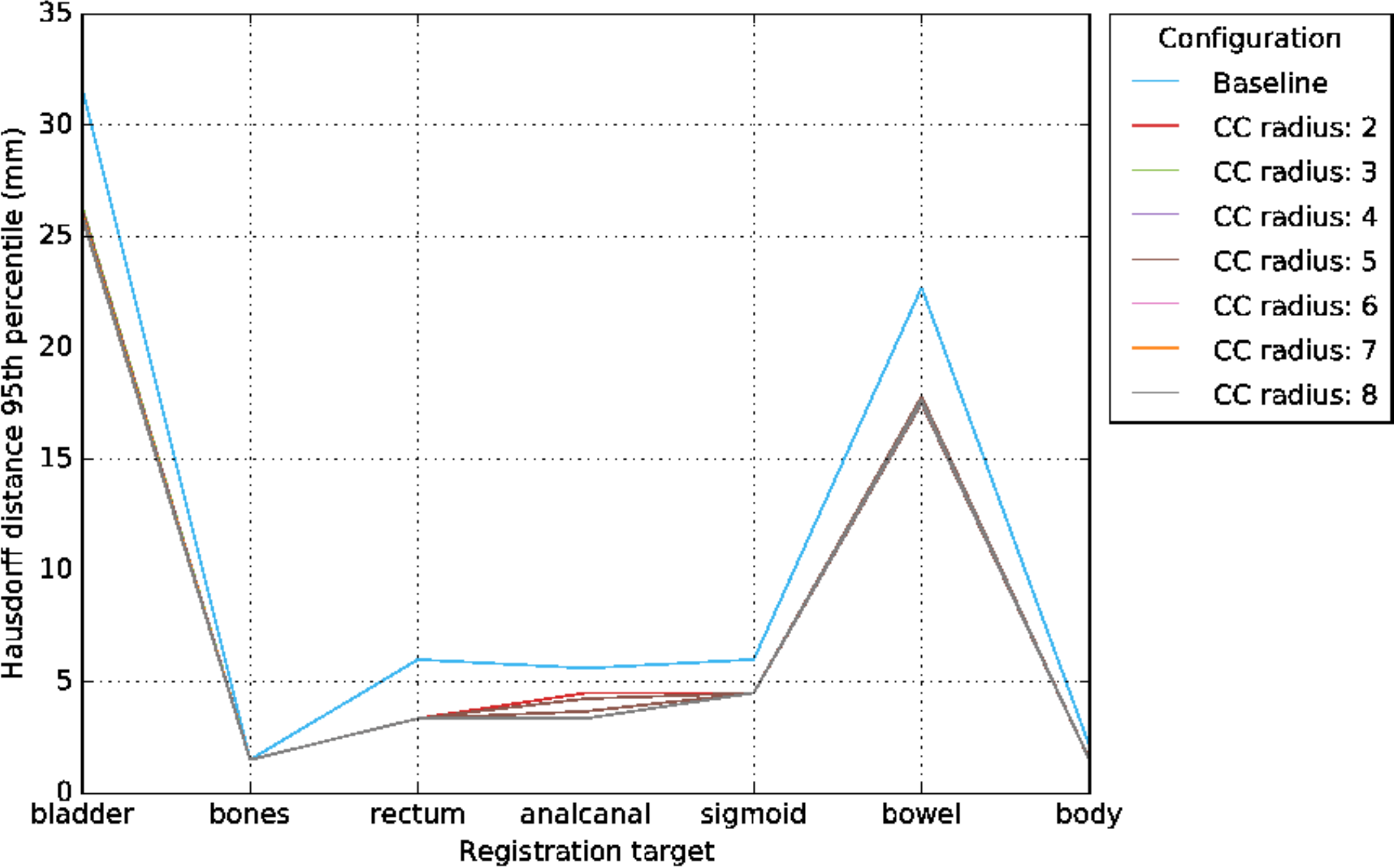}
  \caption{95th percentiles of the Hausdorff distance.}
  \label{fig:ants:pre-tuning:cc:hausdorff-95}
\end{subfigure}
\vspace{-0.4cm}
\caption{Comparison of registrations with different region radii for the ANTs cross correlation metric. The baseline score after rigid registration is plotted in blue.}
\label{fig:ants:pre-tuning:cc}
\end{figure}

%% file: sup/figures/tex/a-ants/pre-tuning-renders-cc.tex

\begin{figure}
\centering
\begin{subfigure}[b]{0.49\linewidth}
  \centering
  \includegraphics[width=\linewidth]{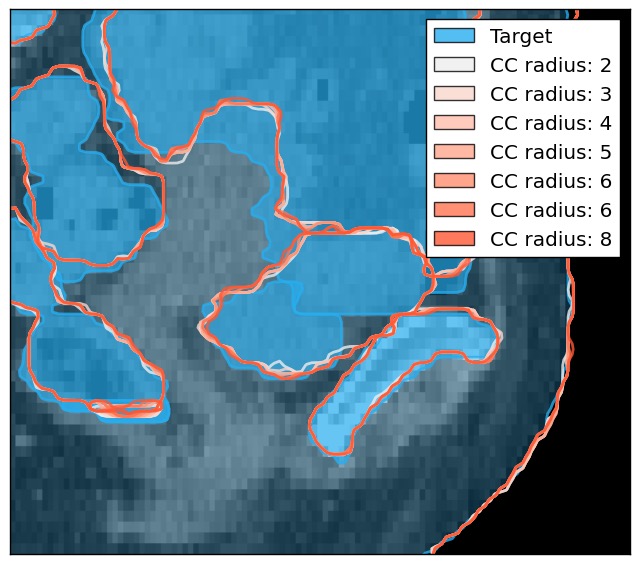}
  \caption{Sagittal slice.}
  \label{fig:ants:pre-tuning:renders:cc:sagittal}
\end{subfigure}%
\begin{subfigure}[b]{0.49\linewidth}
  \centering
  \includegraphics[width=\linewidth]{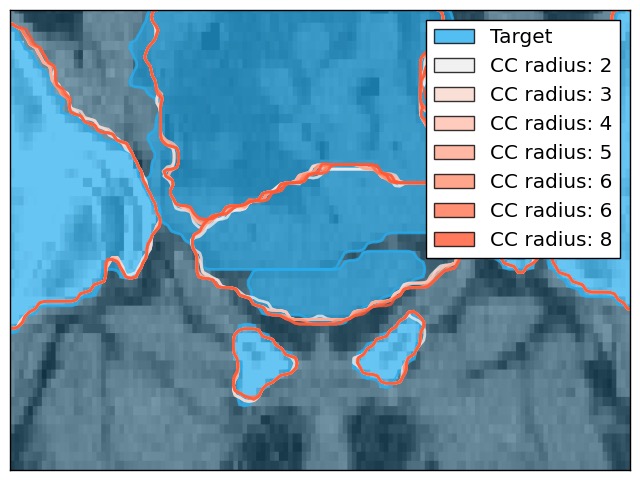}
  \caption{Coronal slice.}
  \label{fig:ants:pre-tuning:renders:cc:coronal}
\end{subfigure}
\vspace{-0.25cm}
\caption{Visual renders of deformations predicted by ANT configurations with different CC radii.}
\label{fig:ants:pre-tuning:renders:cc}
\vspace{-0.4cm}
\end{figure}

%% file: sup/figures/tex/a-ants/pre-tuning-masks.tex
\begin{figure}
\centering
\begin{subfigure}{\linewidth}
  \centering
  \includegraphics[width=\linewidth]{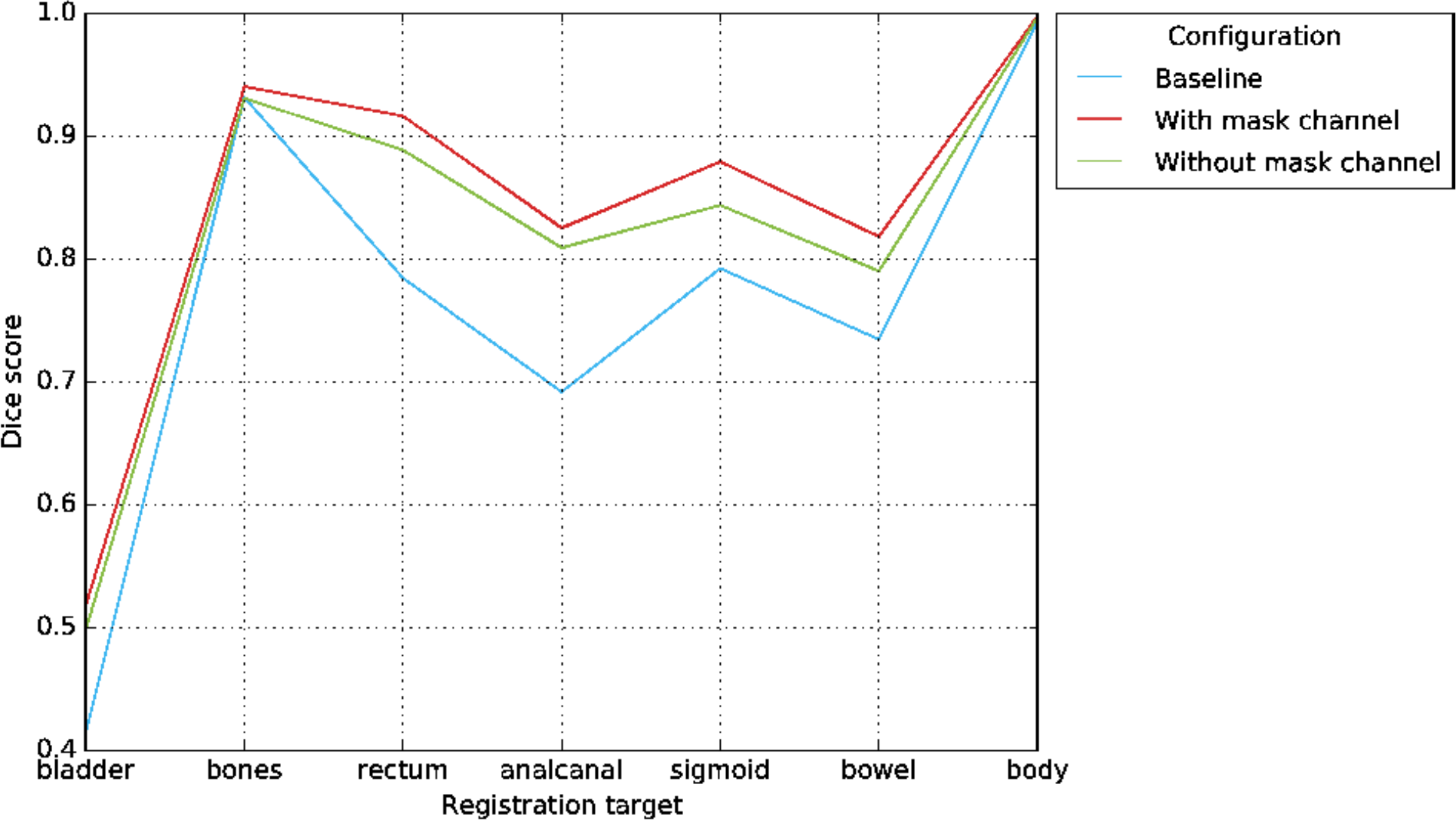}
  \caption{Dice scores.}
  \label{fig:ants:pre-tuning:masks:dice}
\end{subfigure}
\begin{subfigure}{\linewidth}
  \centering
  \includegraphics[width=\linewidth]{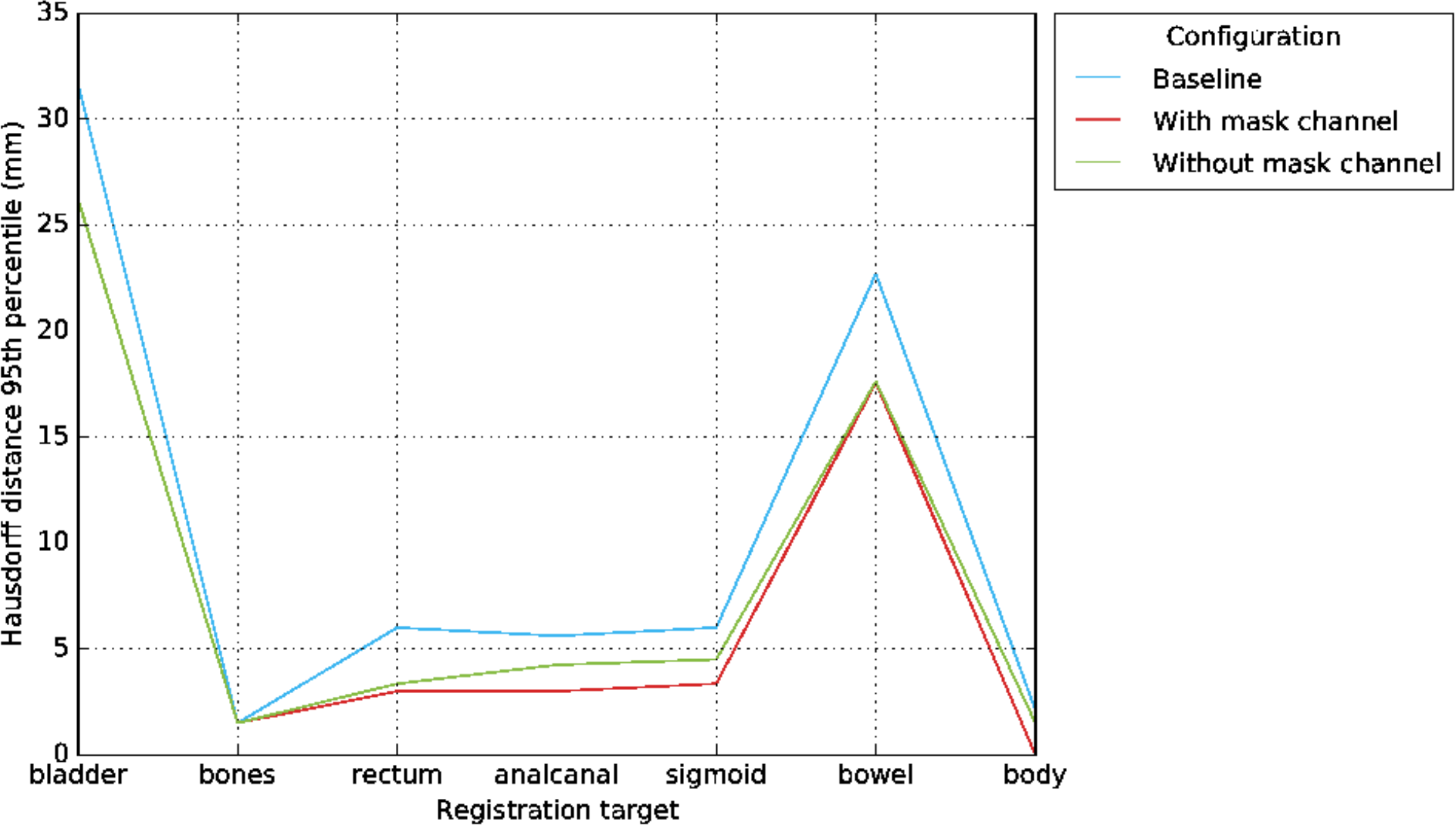}
  \caption{95th percentiles of the Hausdorff distance.}
  \label{fig:ants:pre-tuning:masks:hausdorff-95}
\end{subfigure}
\vspace{-0.4cm}
\caption{Comparison of registrations with and without a composite mask channel in ANTs. The baseline score after rigid registration is plotted in blue.}
\label{fig:ants:pre-tuning:masks}
\end{figure}

%% file: sup/figures/tex/a-ants/pre-tuning-renders-masks.tex
\begin{figure}
\centering
\begin{subfigure}[b]{0.49\linewidth}
  \centering
  \includegraphics[width=\linewidth]{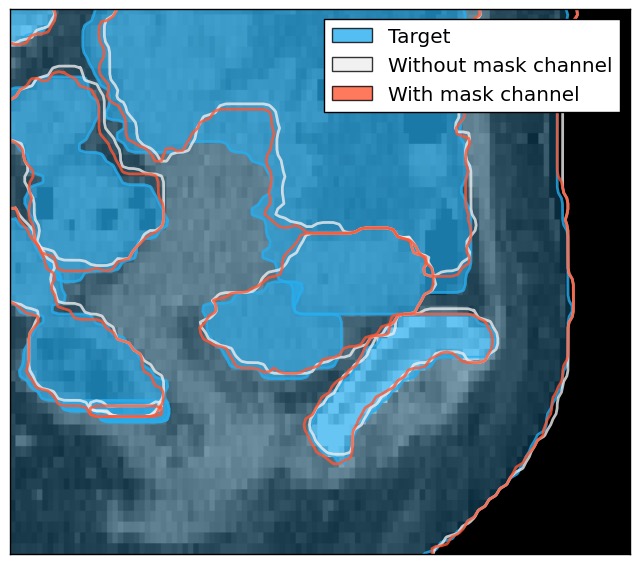}
  \caption{Sagittal slice.}
  \label{fig:ants:pre-tuning:renders:masks:sagittal}
\end{subfigure}%
\begin{subfigure}[b]{0.49\linewidth}
  \centering
  \includegraphics[width=\linewidth]{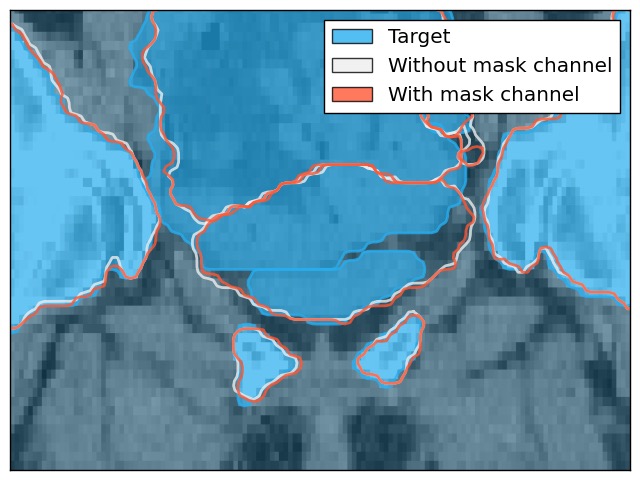}
  \caption{Coronal slice.}
  \label{fig:ants:pre-tuning:renders:masks:coronal}
\end{subfigure}
\vspace{-0.25cm}
\caption{Visual renders of deformations predicted by ANT configurations with and without a composite mask channel.}
\label{fig:ants:pre-tuning:renders:masks}
\vspace{-0.4cm}
\end{figure}

%% file: sup/figures/tex/a-ants/pre-tuning-step-size.tex
\begin{figure}
\centering
\begin{subfigure}{\linewidth}
  \centering
  \includegraphics[width=\linewidth]{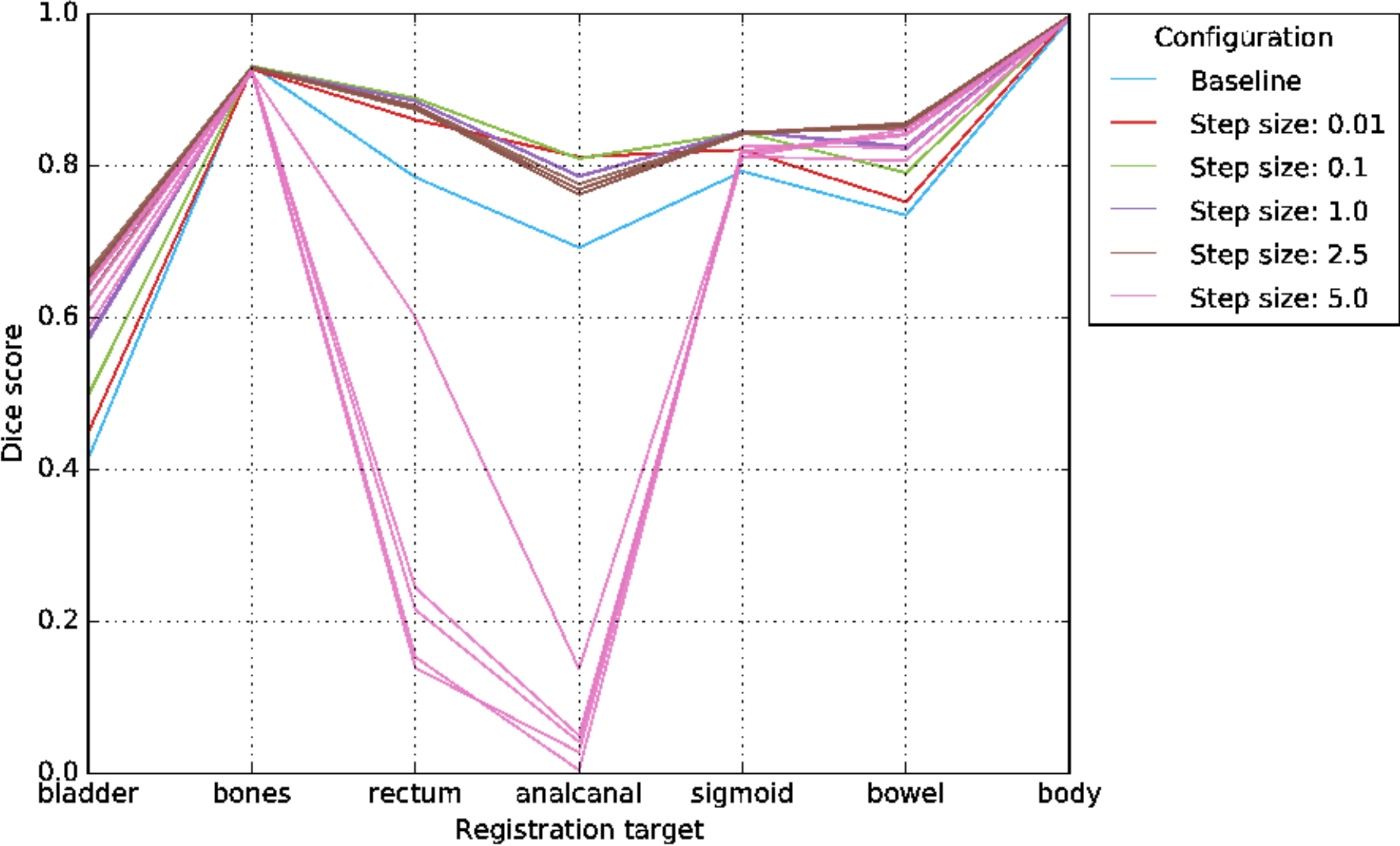}
  \caption{Dice scores.}
  \label{fig:ants:pre-tuning:step-size:dice}
\end{subfigure}
\begin{subfigure}{\linewidth}
  \centering
  \includegraphics[width=\linewidth]{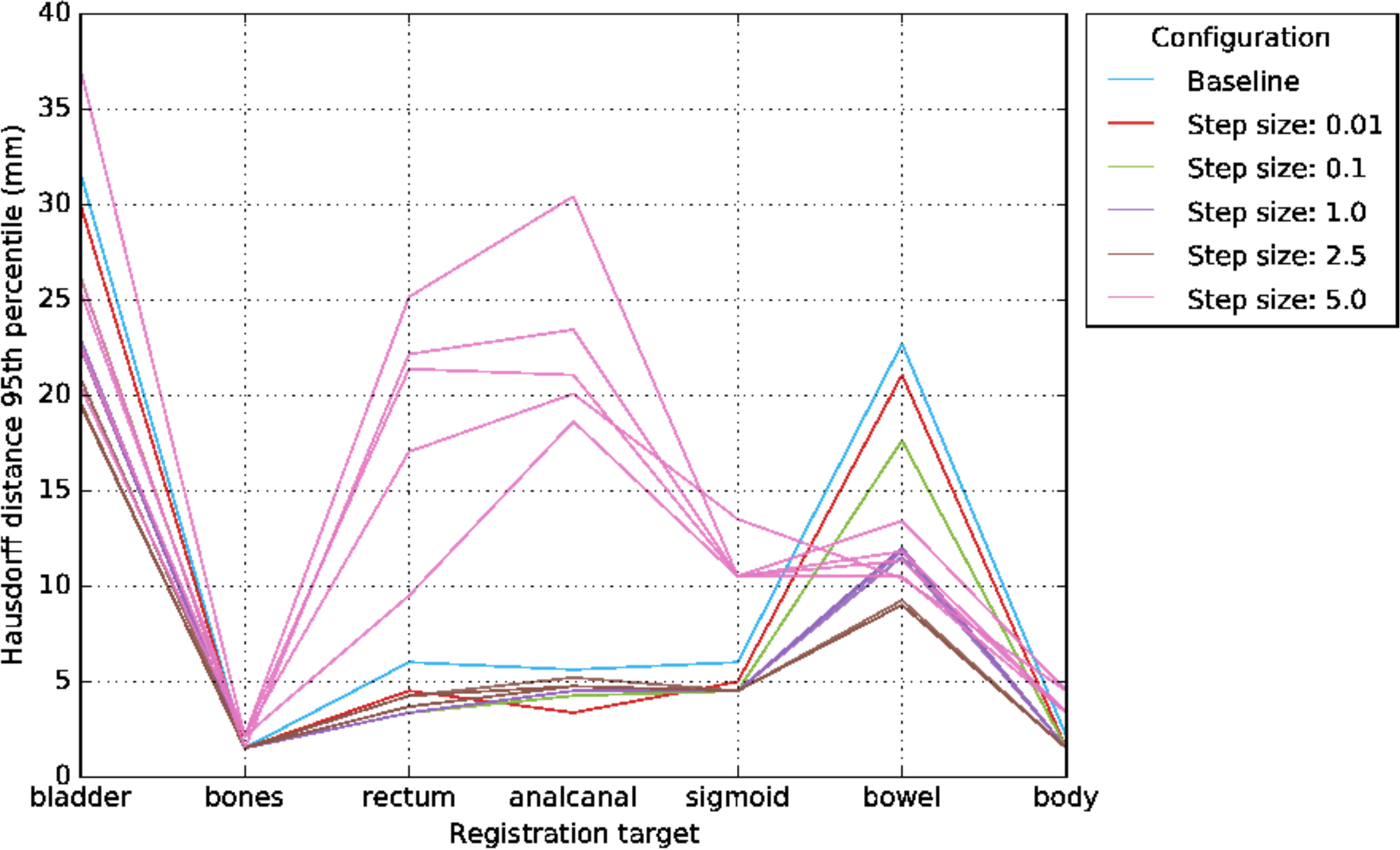}
  \caption{95th percentiles of the Hausdorff distance.}
  \label{fig:ants:pre-tuning:step-size:hausdorff-95}
\end{subfigure}
\vspace{-0.4cm}
\caption{Comparison of ANTs registrations with different gradient step sizes between time points. The baseline score after rigid registration is plotted in blue.}
\label{fig:ants:pre-tuning:step-size}
\end{figure}

%% file: sup/figures/tex/a-ants/pre-tuning-renders-step-size.tex
\begin{figure}
\centering
\begin{subfigure}[b]{0.49\linewidth}
  \centering
  \includegraphics[width=\linewidth]{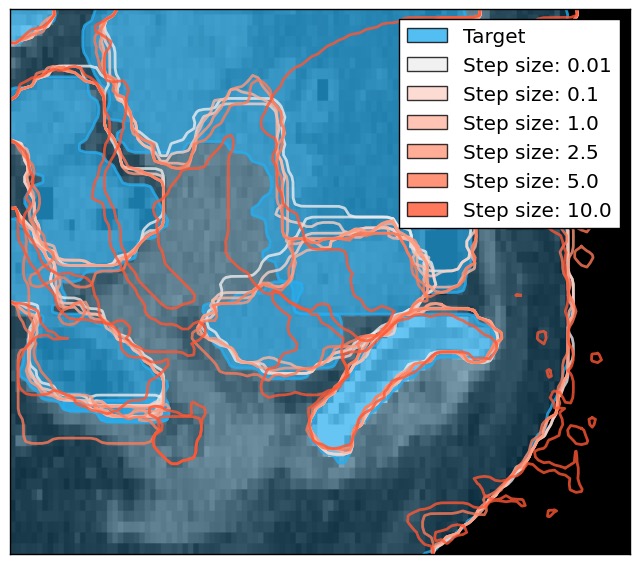}
  \caption{Sagittal slice.}
  \label{fig:ants:pre-tuning:renders:step-size:sagittal}
\end{subfigure}%
\begin{subfigure}[b]{0.49\linewidth}
  \centering
  \includegraphics[width=\linewidth]{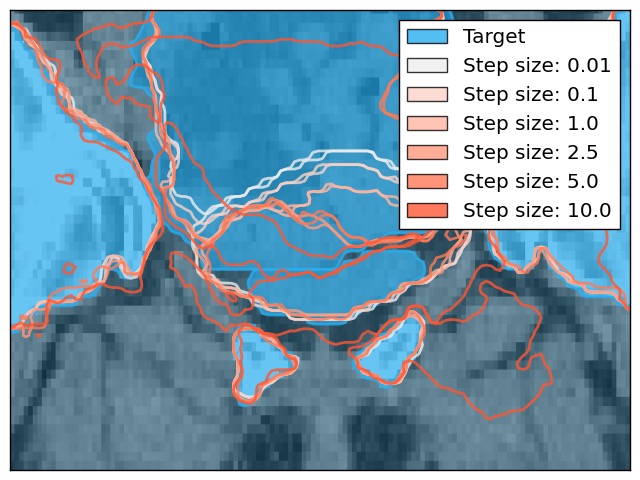}
  \caption{Coronal slice.}
  \label{fig:ants:pre-tuning:renders:step-size:coronal}
\end{subfigure}
\vspace{-0.25cm}
\caption{Visual renders of deformations predicted by ANT configurations with different gradient step sizes.}
\label{fig:ants:pre-tuning:renders:step-size}
\vspace{-0.4cm}
\end{figure}

%% file: sup/figures/tex/a-ants/pre-tuning-update-weight.tex
\begin{figure}
\centering
\begin{subfigure}{\linewidth}
  \centering
  \includegraphics[width=\linewidth]{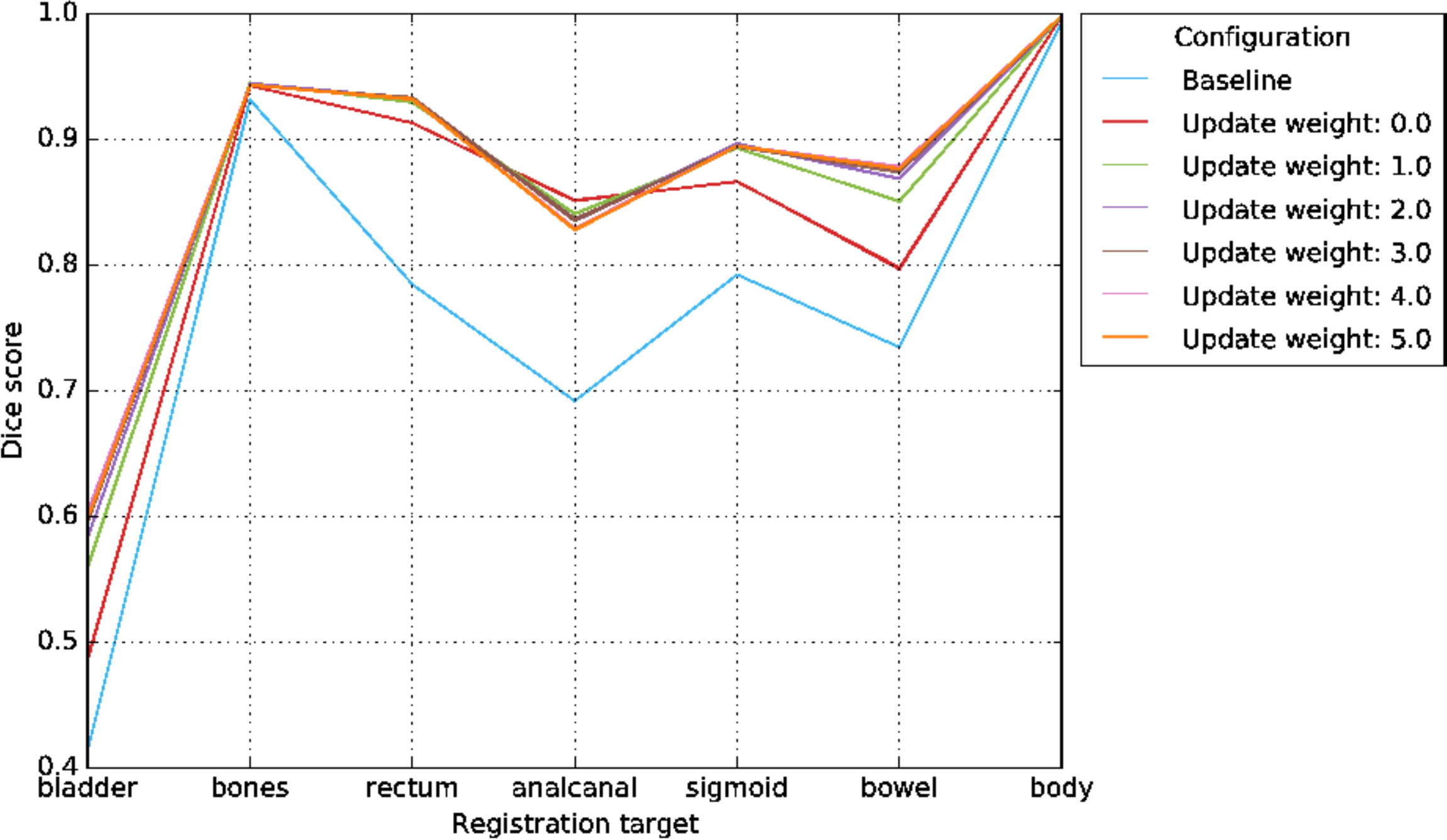}
  \caption{Dice scores.}
  \label{fig:ants:pre-tuning:update-weight:dice}
\end{subfigure}
\begin{subfigure}{\linewidth}
  \centering
  \includegraphics[width=\linewidth]{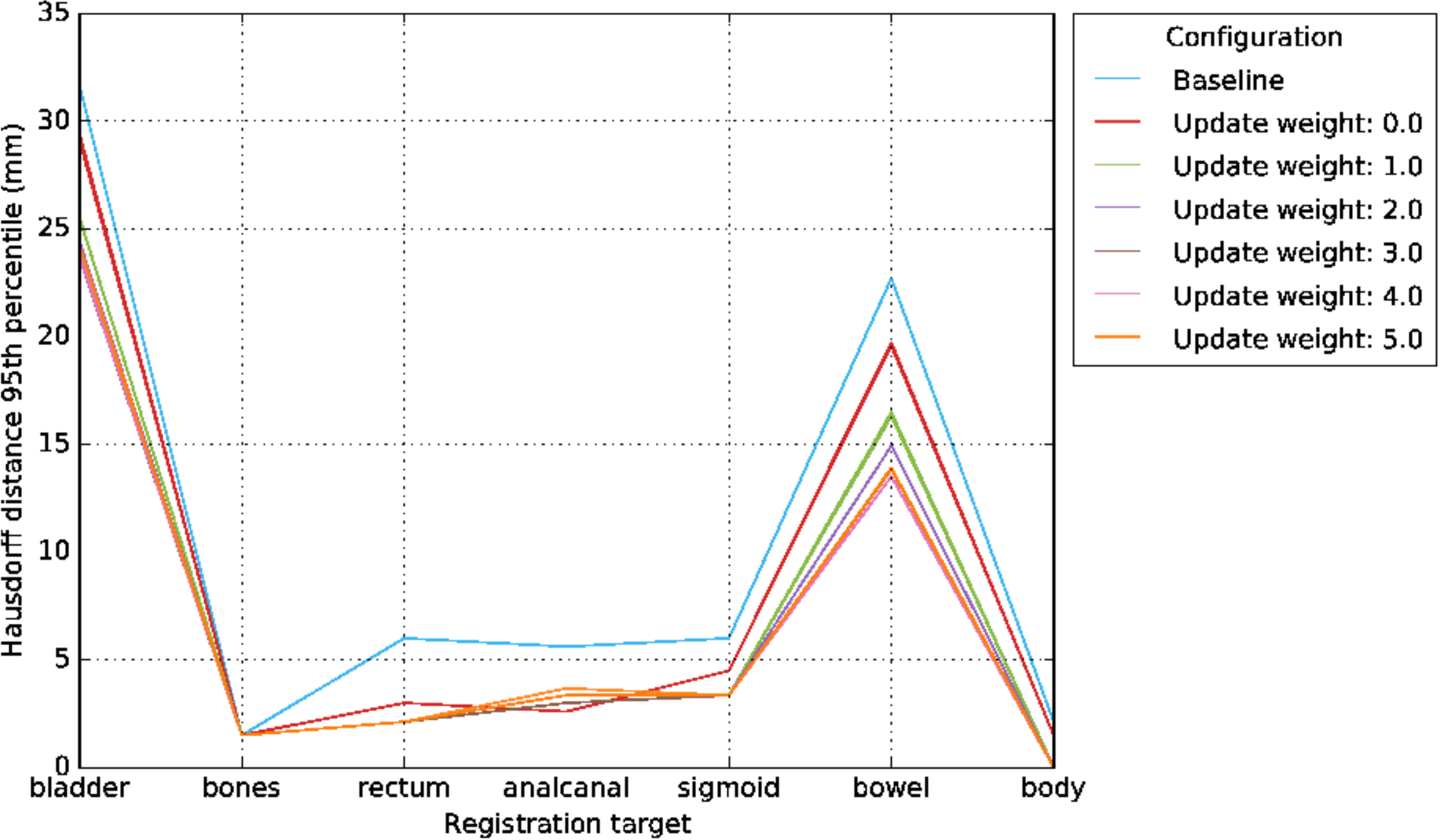}
  \caption{95th percentiles of the Hausdorff distance.}
  \label{fig:ants:pre-tuning:update-weight:hausdorff-95}
\end{subfigure}
\vspace{-0.5cm}
\caption{Comparison of ANTs registrations with different update regularization weights between time points. The baseline score after rigid registration is plotted in blue.}
\label{fig:ants:pre-tuning:update-weight}
\vspace{-0.2cm}
\end{figure}

%% file: sup/figures/tex/a-ants/pre-tuning-renders-update-weight.tex
\begin{figure}
\centering
\begin{subfigure}[b]{0.49\linewidth}
  \centering
  \includegraphics[width=\linewidth]{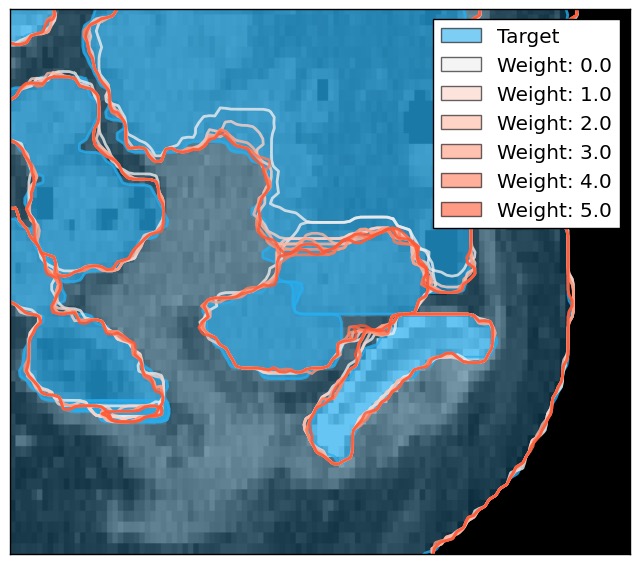}
  \caption{Sagittal slice.}
  \label{fig:ants:pre-tuning:renders:update-weight:sagittal}
\end{subfigure}%
\begin{subfigure}[b]{0.49\linewidth}
  \centering
  \includegraphics[width=\linewidth]{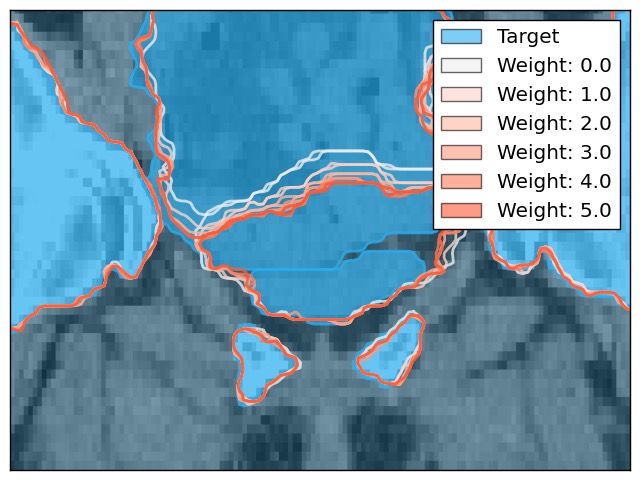}
  \caption{Coronal slice.}
  \label{fig:ants:pre-tuning:renders:update-weight:coronal}
\end{subfigure}
\vspace{-0.25cm}
\caption{Visual renders of deformations predicted by ANT configurations with different update regularization weights.}
\label{fig:ants:pre-tuning:renders:update-weight}
\vspace{-0.4cm}
\end{figure}

%% file: sup/figures/tex/a-ants/main-tuning-dice.tex
\begin{figure*}
\centering
\begin{subfigure}{.44\linewidth}
  \centering
  \includegraphics[width=\linewidth]{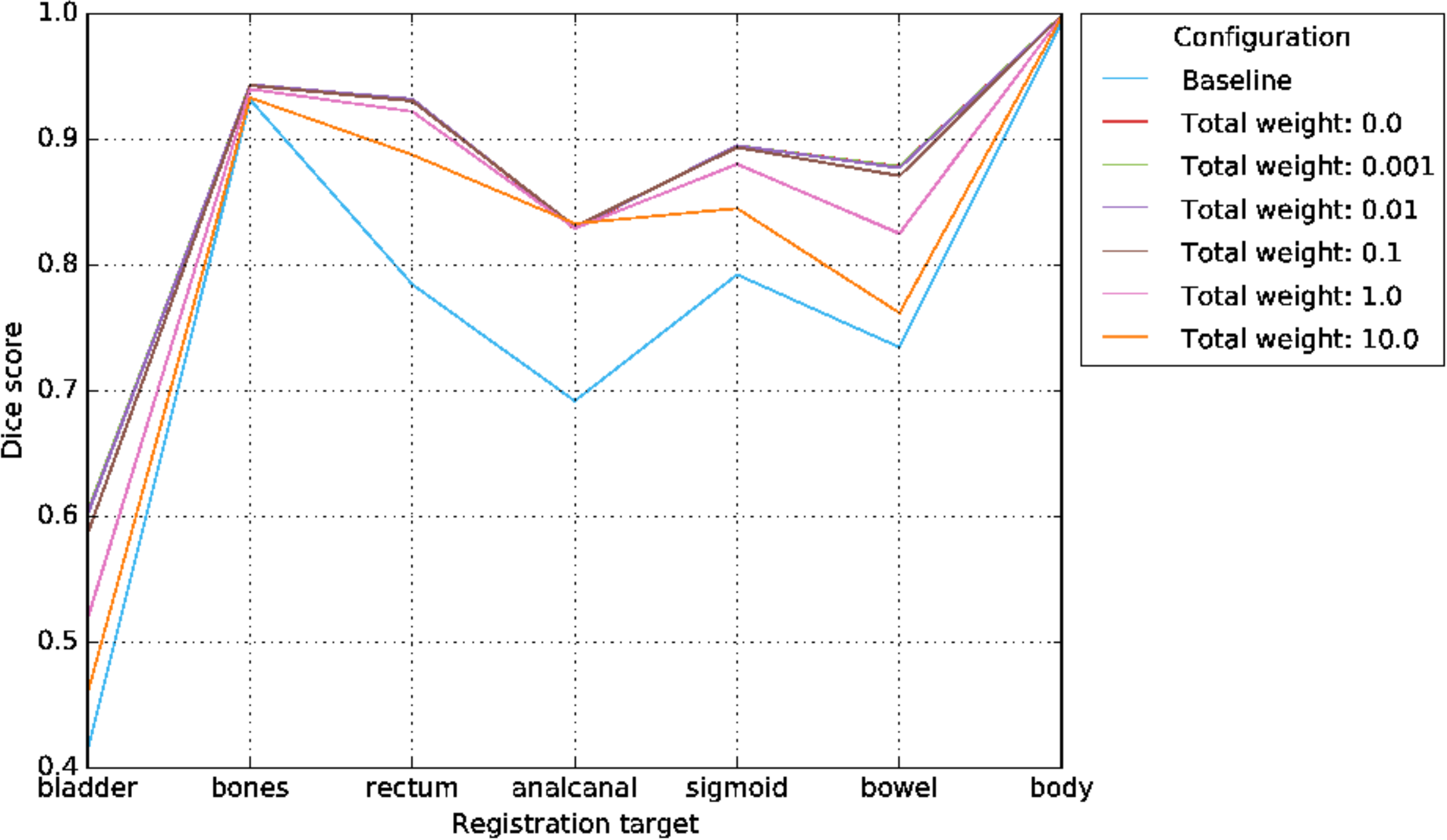}
  \caption{Patient 1.}
  \label{fig:ants:main-tuning:dice:603}
\end{subfigure}%
\hspace{0.019\linewidth}
\begin{subfigure}{.44\linewidth}
  \centering
  \includegraphics[width=\linewidth]{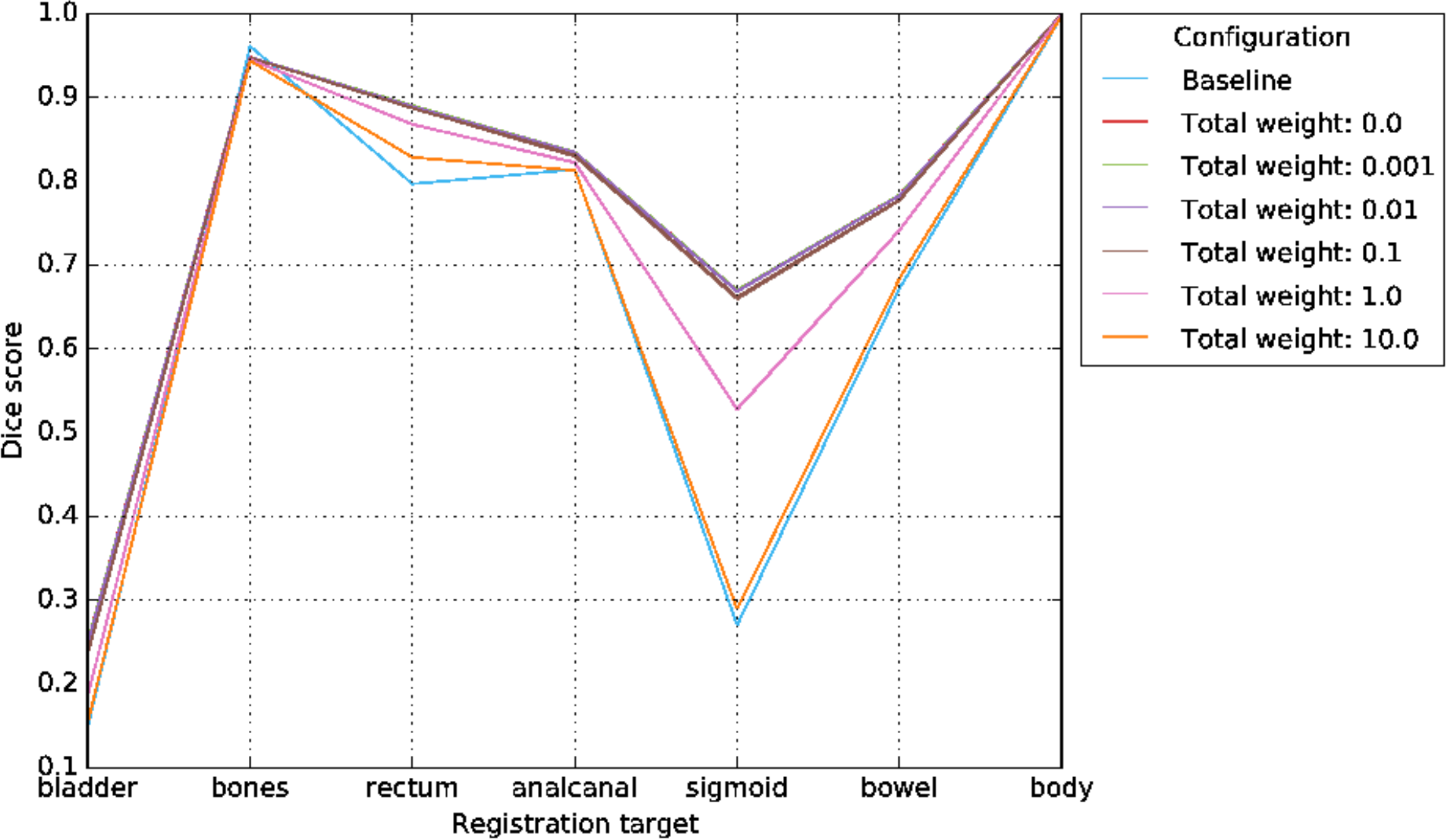}
  \caption{Patient 2.}
  \label{fig:ants:main-tuning:dice:611}
\end{subfigure}%
\hspace{0.019\linewidth}
\begin{subfigure}{.44\linewidth}
  \centering
  \includegraphics[width=\linewidth]{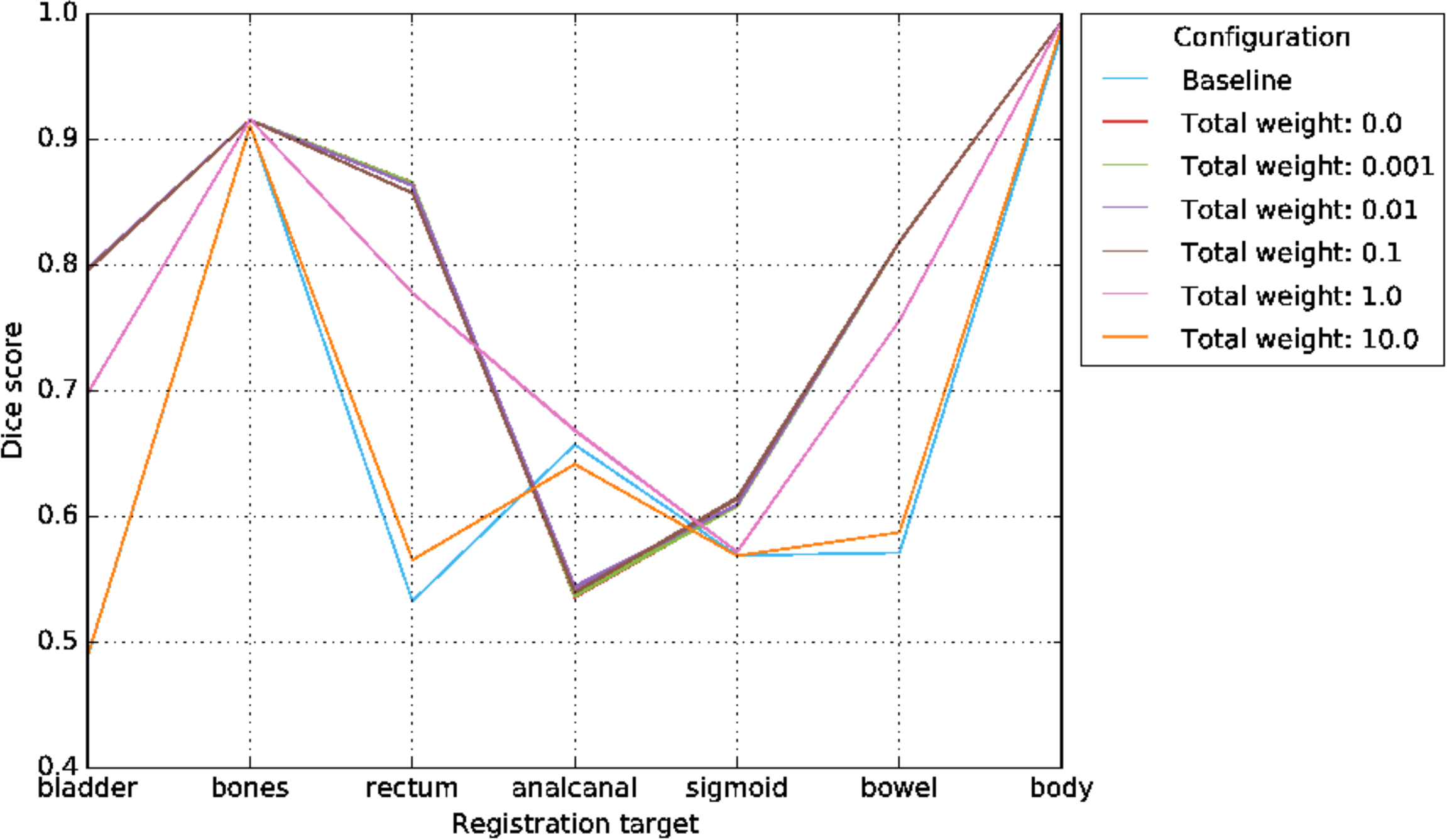}
  \caption{Patient 3.}
  \label{fig:ants:main-tuning:dice:617}
\end{subfigure}%
\hspace{0.019\linewidth}
\begin{subfigure}{.44\linewidth}
  \centering
  \includegraphics[width=\linewidth]{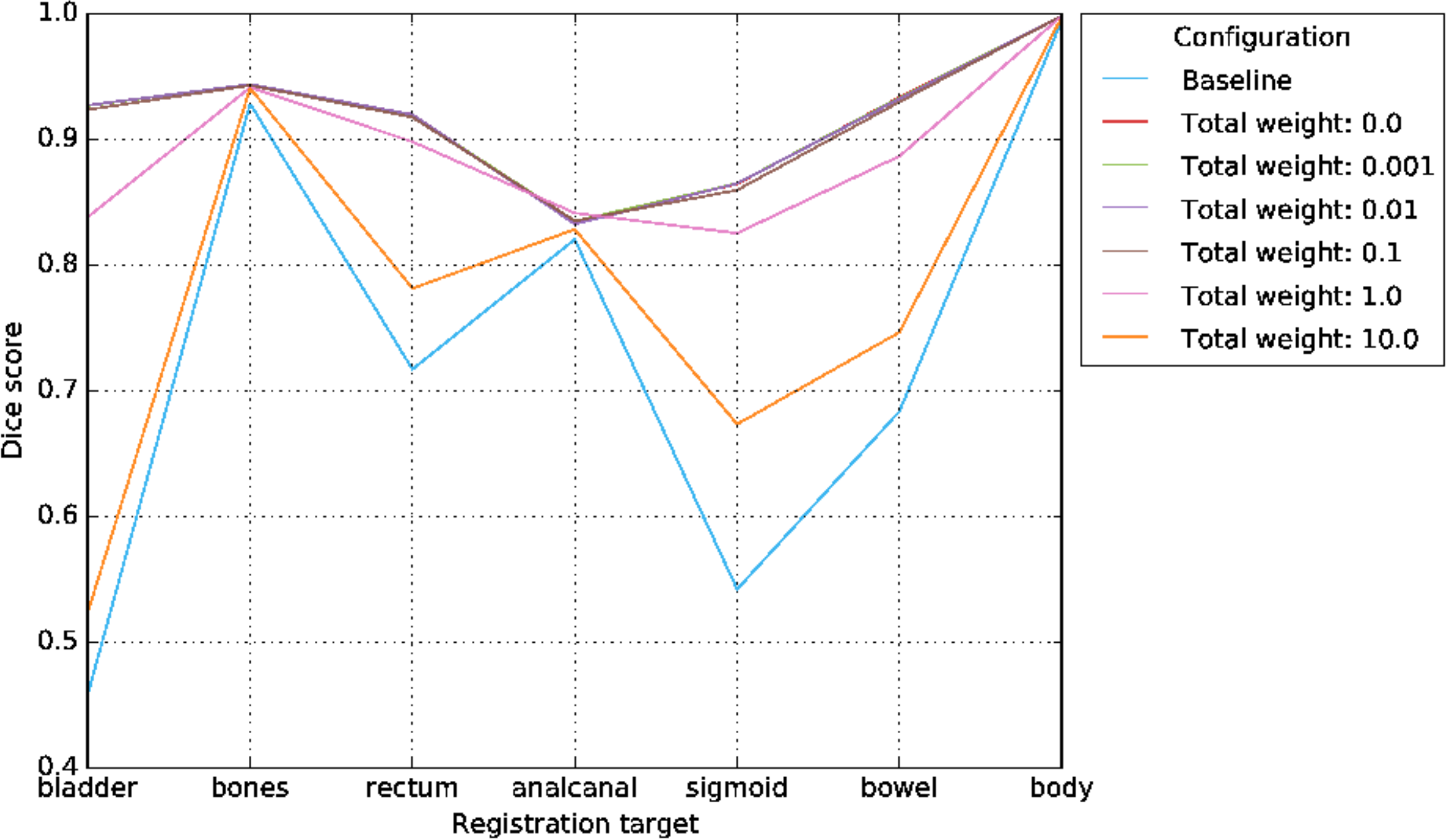}
  \caption{Patient 4.}
  \label{fig:ants:main-tuning:dice:618}
\end{subfigure}%
\vspace{-0.4cm}
\caption{Dice scores for per-patient fine-grained configuration runs in ANTs, with the baseline after rigid registration in blue.}
\label{fig:ants:main-tuning:dice}
\vspace{-0.1cm}
\end{figure*}

%% file: sup/figures/tex/a-ants/main-tuning-hausdorff-95.tex
\begin{figure*}
\centering
\begin{subfigure}{.44\linewidth}
  \centering
  \includegraphics[width=\linewidth]{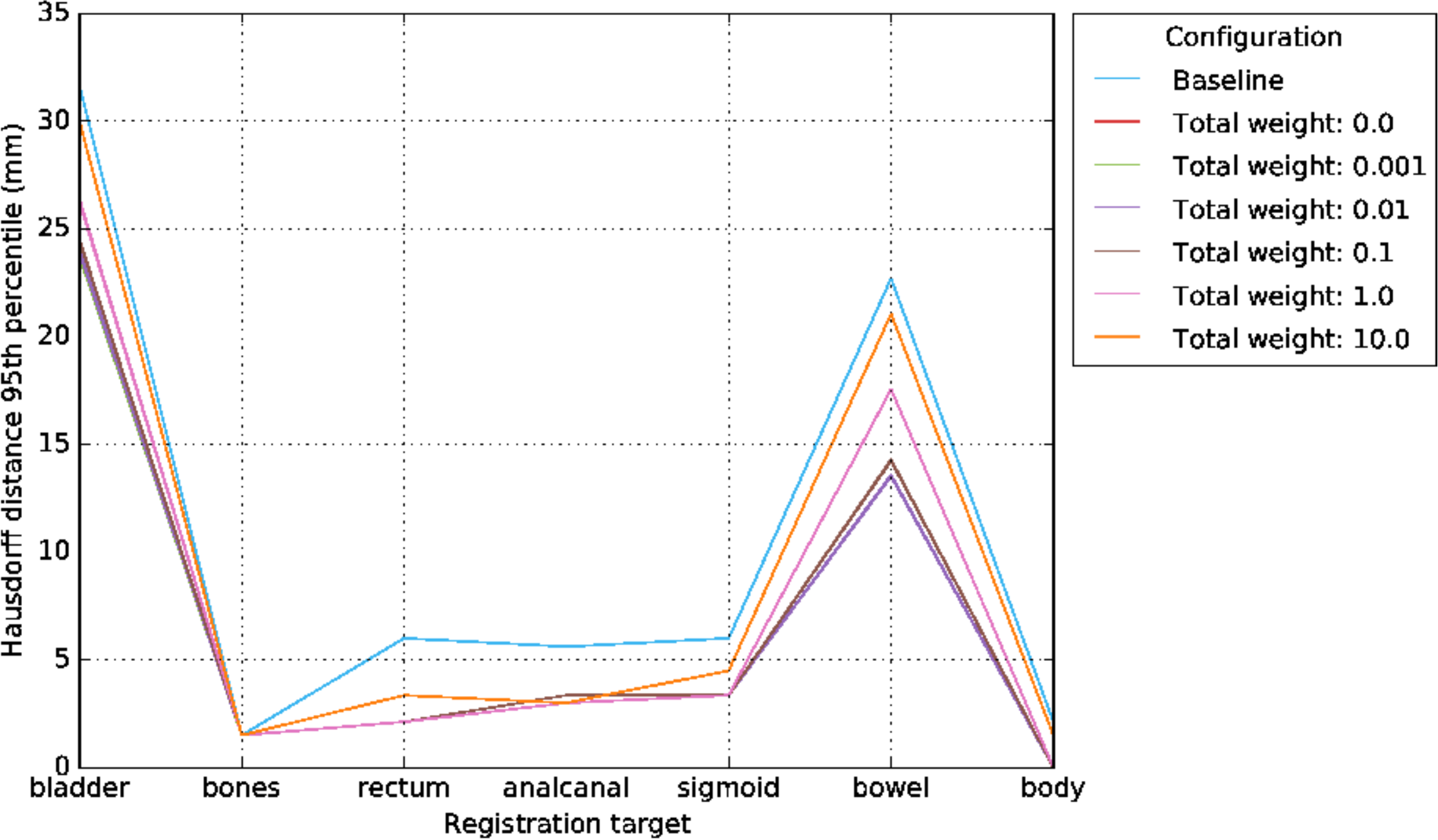}
  \caption{Patient 1.}
  \label{fig:ants:main-tuning:hausdorff-95:603}
\end{subfigure}%
\hspace{0.019\linewidth}
\begin{subfigure}{.44\linewidth}
  \centering
  \includegraphics[width=\linewidth]{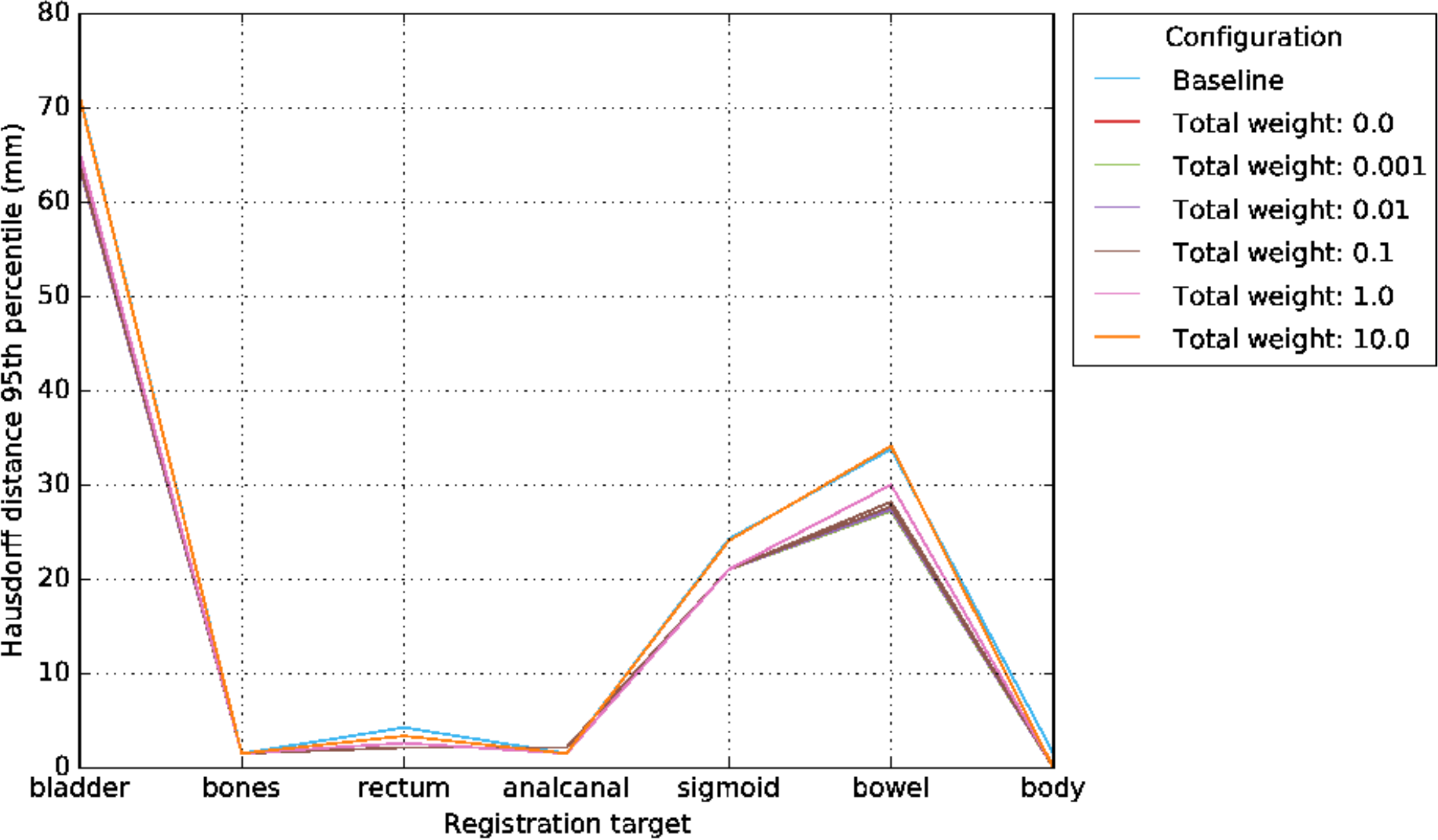}
  \caption{Patient 2.}
  \label{fig:ants:main-tuning:hausdorff-95:611}
\end{subfigure}%
\hspace{0.019\linewidth}
\begin{subfigure}{.44\linewidth}
  \centering
  \includegraphics[width=\linewidth]{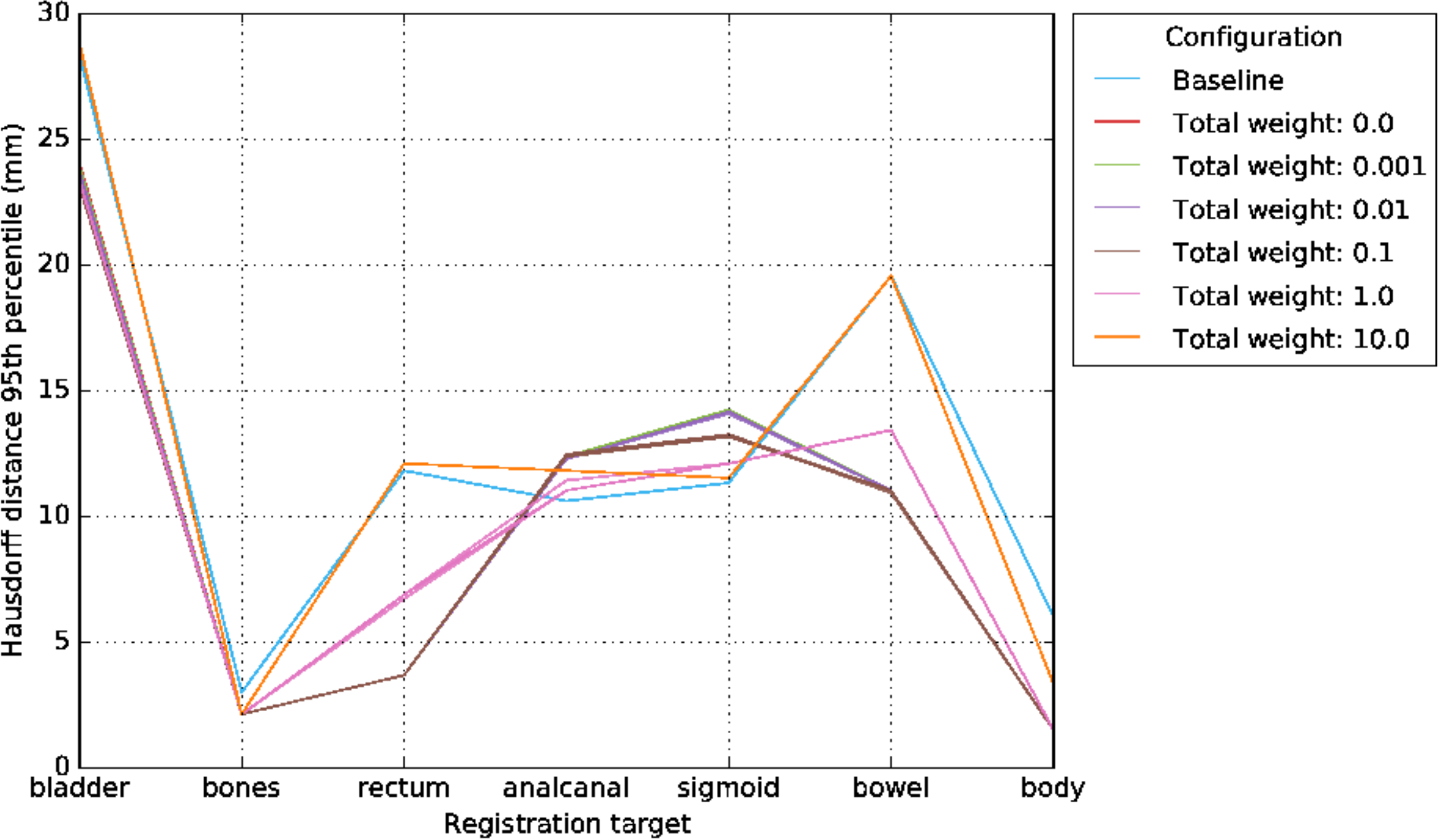}
  \caption{Patient 3.}
  \label{fig:ants:main-tuning:hausdorff-95:617}
\end{subfigure}%
\hspace{0.019\linewidth}
\begin{subfigure}{.44\linewidth}
  \centering
  \includegraphics[width=\linewidth]{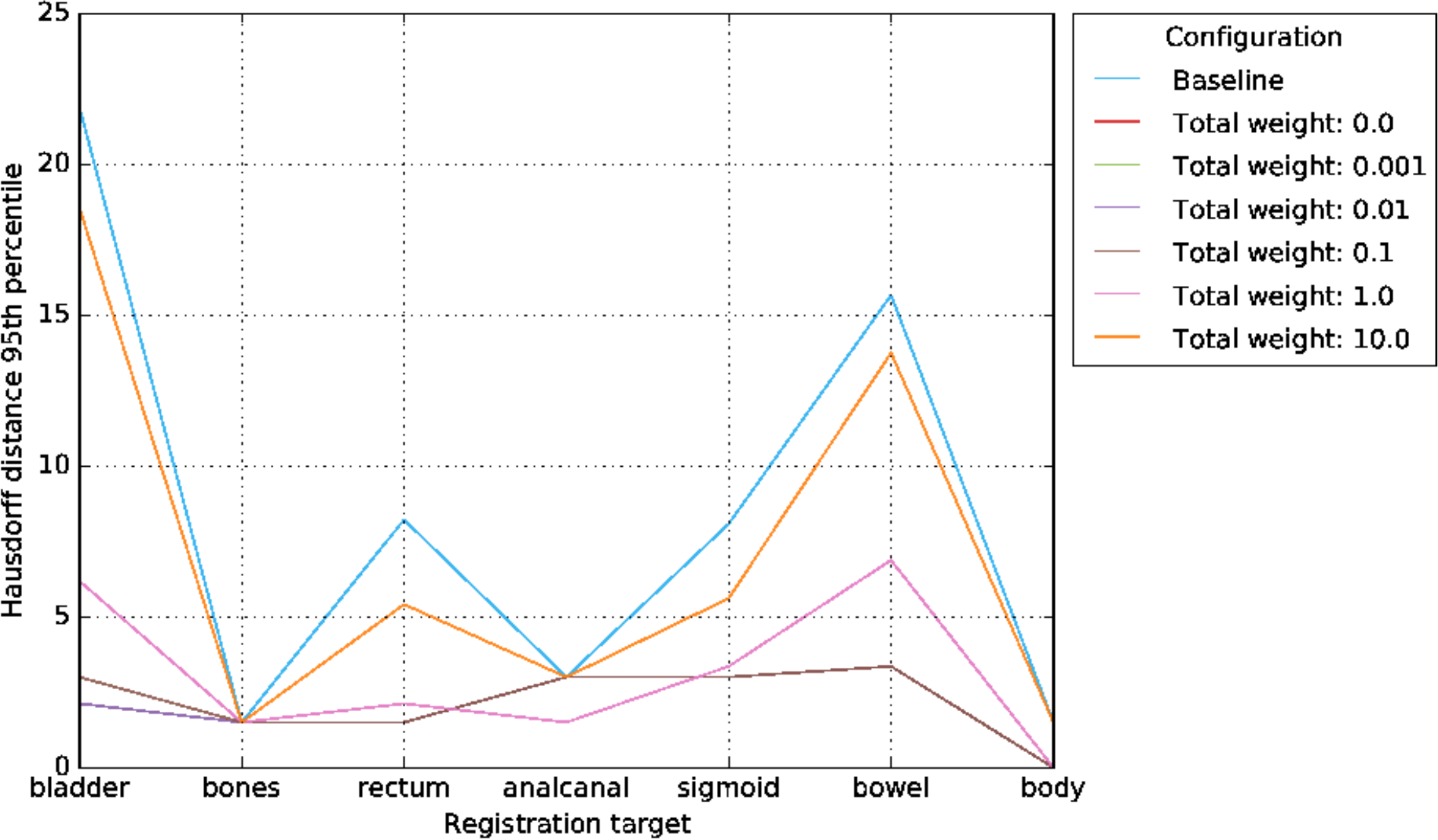}
  \caption{Patient 4.}
  \label{fig:ants:main-tuning:hausdorff-95:618}
\end{subfigure}%
\vspace{-0.4cm}
\caption{Hausdorff 95th percentiles for per-patient fine-grained configuration runs in ANTs, with the baseline after rigid registration in blue.}
\label{fig:ants:main-tuning:hausdorff-95}
\vspace{-0.4cm}
\end{figure*}

%% file: sup/figures/tex/a-ants/main-tuning-renders.tex
\begin{figure}
\centering
\begin{subfigure}[b]{.49\linewidth}
  \centering
  \includegraphics[width=\linewidth]{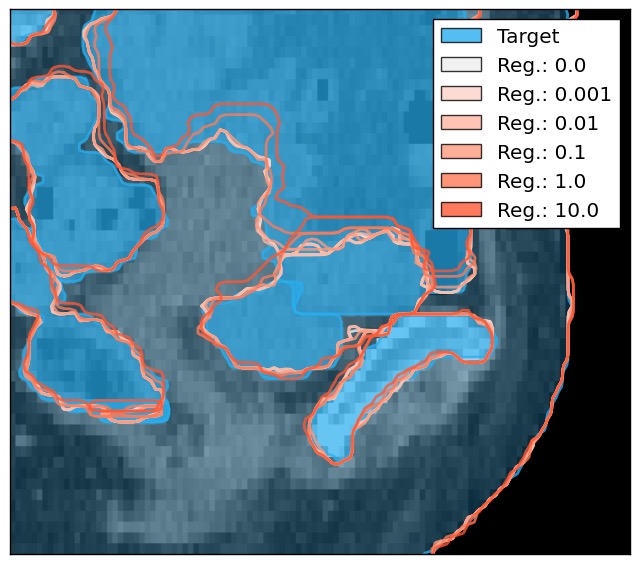}
  \caption{Sagittal slice.}
\end{subfigure}%
\begin{subfigure}[b]{.49\linewidth}
  \centering
  \includegraphics[width=\linewidth]{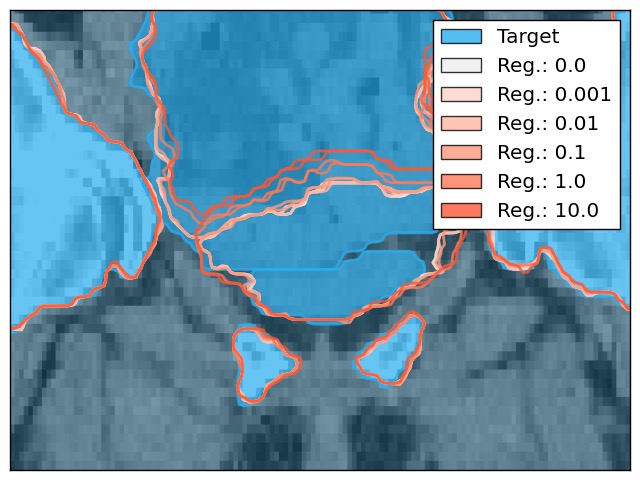}
  \caption{Coronal slice.}
\end{subfigure}
\vspace{-0.25cm}
\caption{Visual renders of deformations predicted by ANTs with different total regularization weights, on Patient 1.}
\label{fig:ants:main-tuning:renders:603}
\end{figure}

\begin{figure}
\centering
\begin{subfigure}[b]{.49\linewidth}
  \centering
  \includegraphics[width=\linewidth]{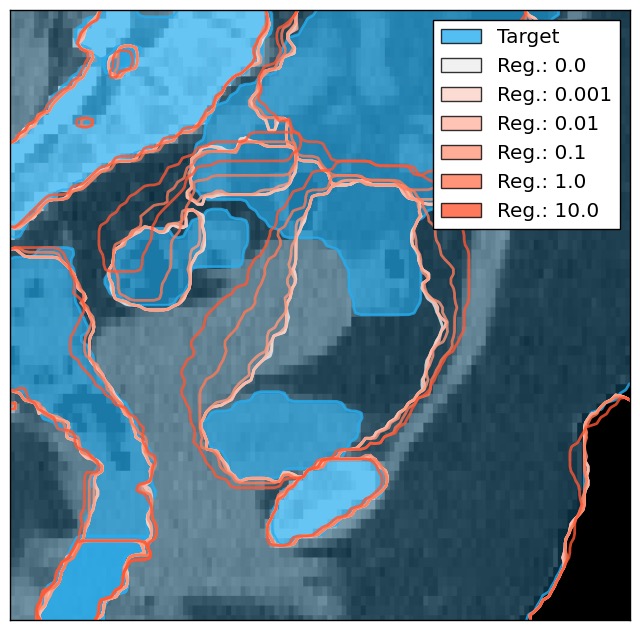}
  \caption{Sagittal slice.}
\end{subfigure}%
\begin{subfigure}[b]{.49\linewidth}
  \centering
  \includegraphics[width=\linewidth]{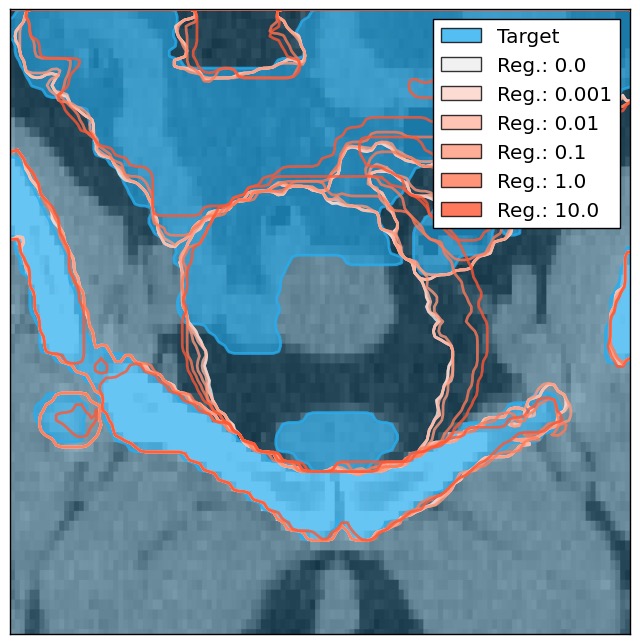}
  \caption{Coronal slice.}
\end{subfigure}
\vspace{-0.25cm}
\caption{Visual renders of deformations predicted by ANTs with different total regularization weights, on Patient 2.}
\label{fig:ants:main-tuning:renders:611}
\end{figure}

\begin{figure}
\centering
\begin{subfigure}[b]{.49\linewidth}
  \centering
  \includegraphics[width=\linewidth]{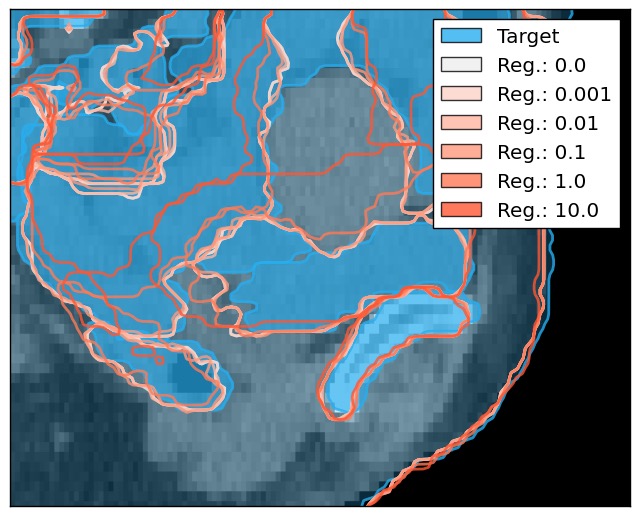}
  \caption{Sagittal slice.}
\end{subfigure}%
\begin{subfigure}[b]{.49\linewidth}
  \centering
  \includegraphics[width=\linewidth]{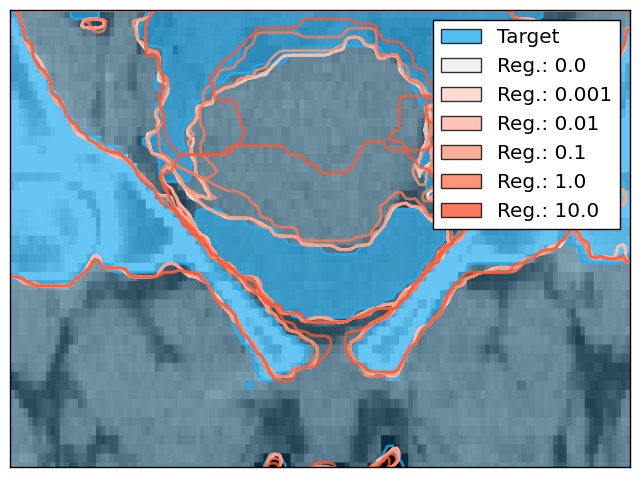}
  \caption{Coronal slice.}
\end{subfigure}
\vspace{-0.25cm}
\caption{Visual renders of deformations predicted by ANTs with different total regularization weights, on Patient 3.}
\label{fig:ants:main-tuning:renders:617}
\end{figure}

\begin{figure}
\centering
\begin{subfigure}[b]{.49\linewidth}
  \centering
  \includegraphics[width=\linewidth]{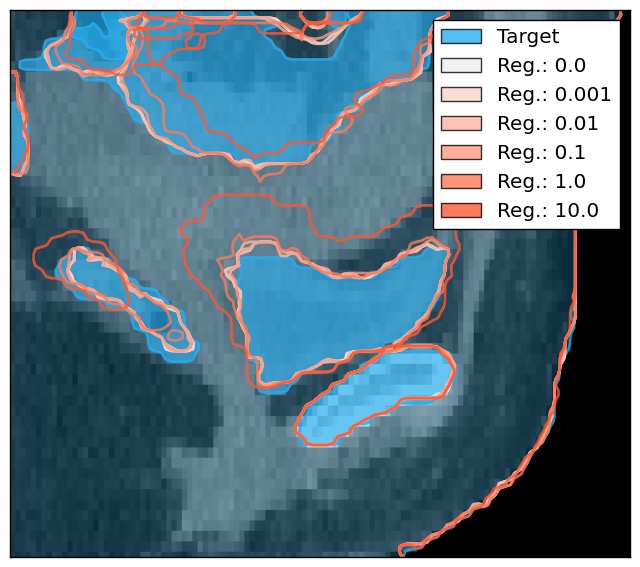}
  \caption{Sagittal slice.}
\end{subfigure}%
\begin{subfigure}[b]{.49\linewidth}
  \centering
  \includegraphics[width=\linewidth]{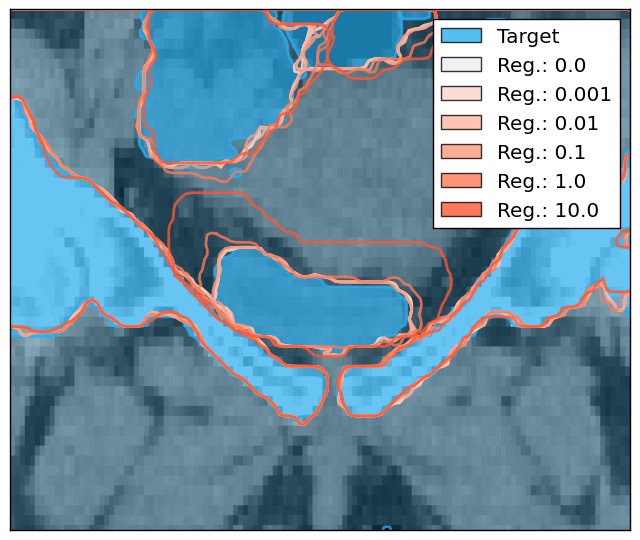}
  \caption{Coronal slice.}
\end{subfigure}
\vspace{-0.25cm}
\caption{Visual renders of deformations predicted by ANTs with different total regularization weights, on Patient 4.}
\label{fig:ants:main-tuning:renders:618}
\end{figure}


%% file: sup/figures/tex/a-ants/commands.tex
\begin{lstlisting}[language=bash,caption={ANTs registration command for multivariate registration \textit{with} composite masks.},label={listing:ants-command-with-masks}]
$ANTSPATH/antsRegistration
    --verbose 1
    --random-seed $random_seed
    --dimensionality 3
    --float 0
    --collapse-output-transforms 1
    --output [ , Warped.nii.gz, InverseWarped.nii.gz ]
    --interpolation Linear
    --use-histogram-matching 0
    --winsorize-image-intensities [ 0.005, 0.995 ]
    --initial-moving-transform [ $fixed_composite_mask, $moving_composite_mask, 1 ]
    --transform SyN[ $gradient_step_size, $update_regularization_weight, $total_regularization_weight ]
    --metric CC[ $fixed_composite_mask, $moving_composite_mask, 1, $cross_correlation_radius ]
    --metric CC[ $fixed_image, $moving_image, 1, $cross_correlation_radius ]
    --convergence [ 2000x1000x500x250, 1e-6, 10 ]
    --shrink-factors 8x4x2x1
    --smoothing-sigmas {3+delta_3}x{2+delta_2}x{1+delta_1}x0vox
\end{lstlisting}

\begin{lstlisting}[language=bash,caption={ANTs registration command for multivariate registration \textit{without} composite masks.},label={listing:ants-command-without-masks}]
$ANTSPATH/antsRegistration
    --verbose 1
    --random-seed $random_seed
    --dimensionality 3
    --float 0
    --collapse-output-transforms 1
    --output [ , Warped.nii.gz, InverseWarped.nii.gz ]
    --interpolation Linear
    --use-histogram-matching 0
    --winsorize-image-intensities [ 0.005, 0.995 ]
    --initial-moving-transform [ $fixed_image, $moving_image, 1 ]
    --transform SyN[ $gradient_step_size, $update_regularization_weight, $total_regularization_weight ]
    --metric CC[ $fixed_image, $moving_image, 1, $cross_correlation_radius ]
    --convergence [ 2000x1000x500x250, 1e-6, 10 ]
    --shrink-factors 8x4x2x1
    --smoothing-sigmas {3+delta_3}x{2+delta_2}x{1+delta_1}x0vox
\end{lstlisting}

%% file: sup/figures/tex/a-morea/pre-tuning-magnitude.tex
\begin{figure}
\centering
\begin{subfigure}{\linewidth}
  \centering
  \includegraphics[width=\linewidth]{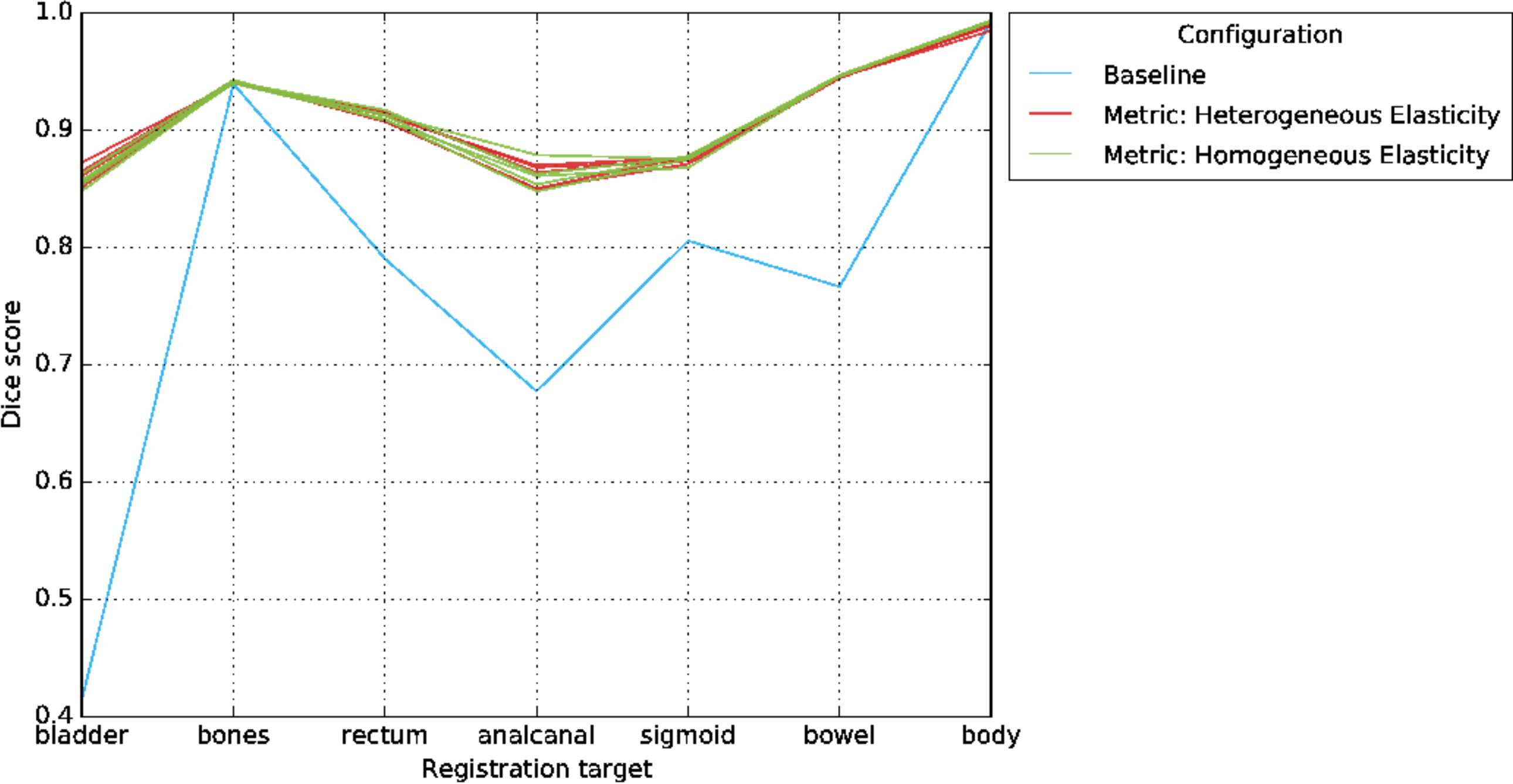}
  \caption{Dice scores.}
  \label{fig:morea:pre-tuning:magnitude:dice}
\end{subfigure}
\begin{subfigure}{\linewidth}
  \centering
  \includegraphics[width=\linewidth]{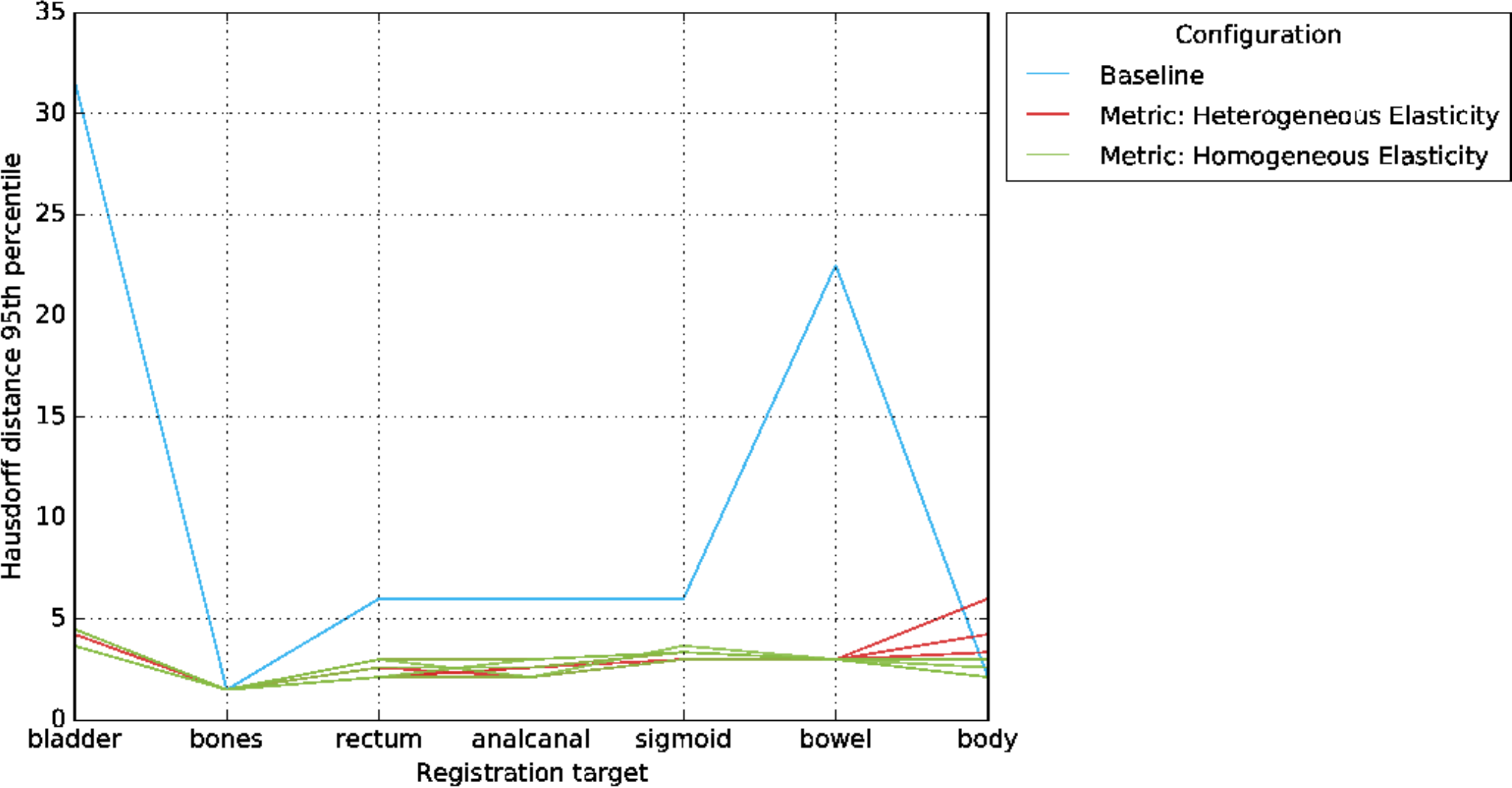}
  \caption{95th percentiles of the Hausdorff distance.}
  \label{fig:morea:pre-tuning:magnitude:hausdorff-95}
\end{subfigure}
\vspace{-0.5cm}
\caption{Comparison of the use of heterogeneous elasticities in the deformation magnitude objective of MOREA against the prior use of a homogeneous elasticity model, for multiple runs. The baseline score after rigid registration is plotted in blue.}
\label{fig:morea:pre-tuning:magnitude}
\end{figure}

%% file: sup/figures/tex/a-morea/pre-tuning-renders-magnitude.tex
\begin{figure}
\centering
\begin{subfigure}[b]{.49\linewidth}
  \centering
  \includegraphics[width=\linewidth]{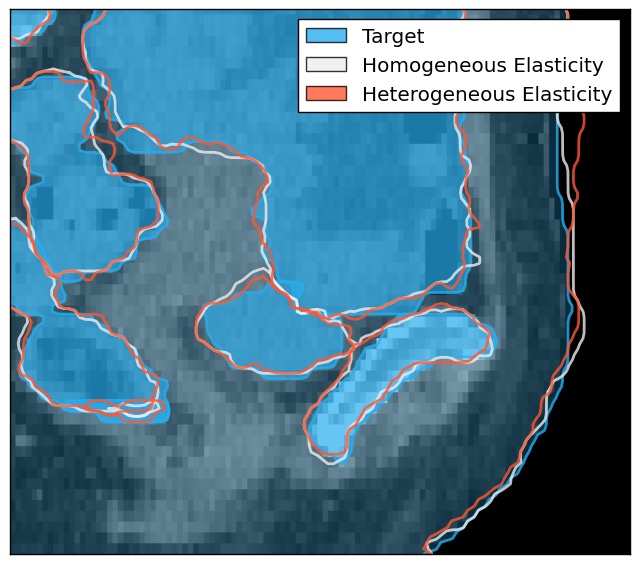}
  \caption{Sagittal slice.}
  \label{fig:morea:pre-tuning:renders:magnitude:sagittal}
\end{subfigure}%
\begin{subfigure}[b]{.49\linewidth}
  \centering
  \includegraphics[width=\linewidth]{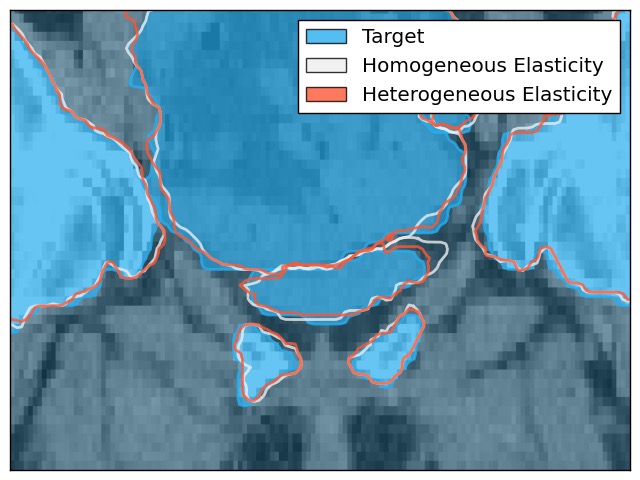}
  \caption{Coronal slice.}
  \label{fig:morea:pre-tuning:renders:magnitude:coronal}
\end{subfigure}
\vspace{-0.25cm}
\caption{Visual renders of deformations predicted by MOREA with a heterogeneous elastic deformation model and a homogeneous model.}
\label{fig:morea:pre-tuning:renders:magnitude}
\end{figure}

%% file: sup/figures/tex/a-morea/pre-tuning-placement.tex
\begin{figure}
\centering
\begin{subfigure}{\linewidth}
  \centering
  \includegraphics[width=\linewidth]{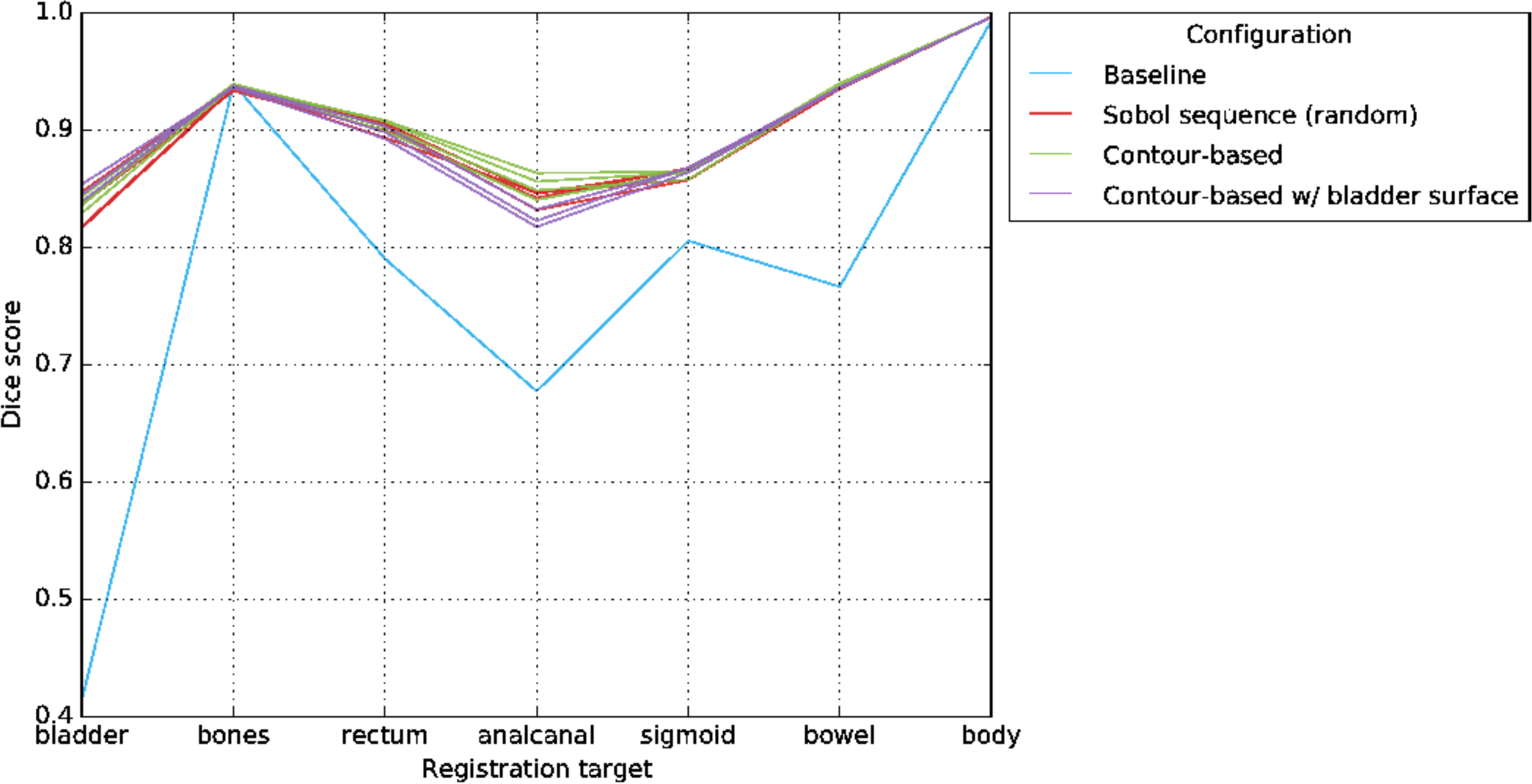}
  \caption{Dice scores.}
  \label{fig:morea:pre-tuning:placement:dice}
\end{subfigure}
\begin{subfigure}{\linewidth}
  \centering
  \includegraphics[width=\linewidth]{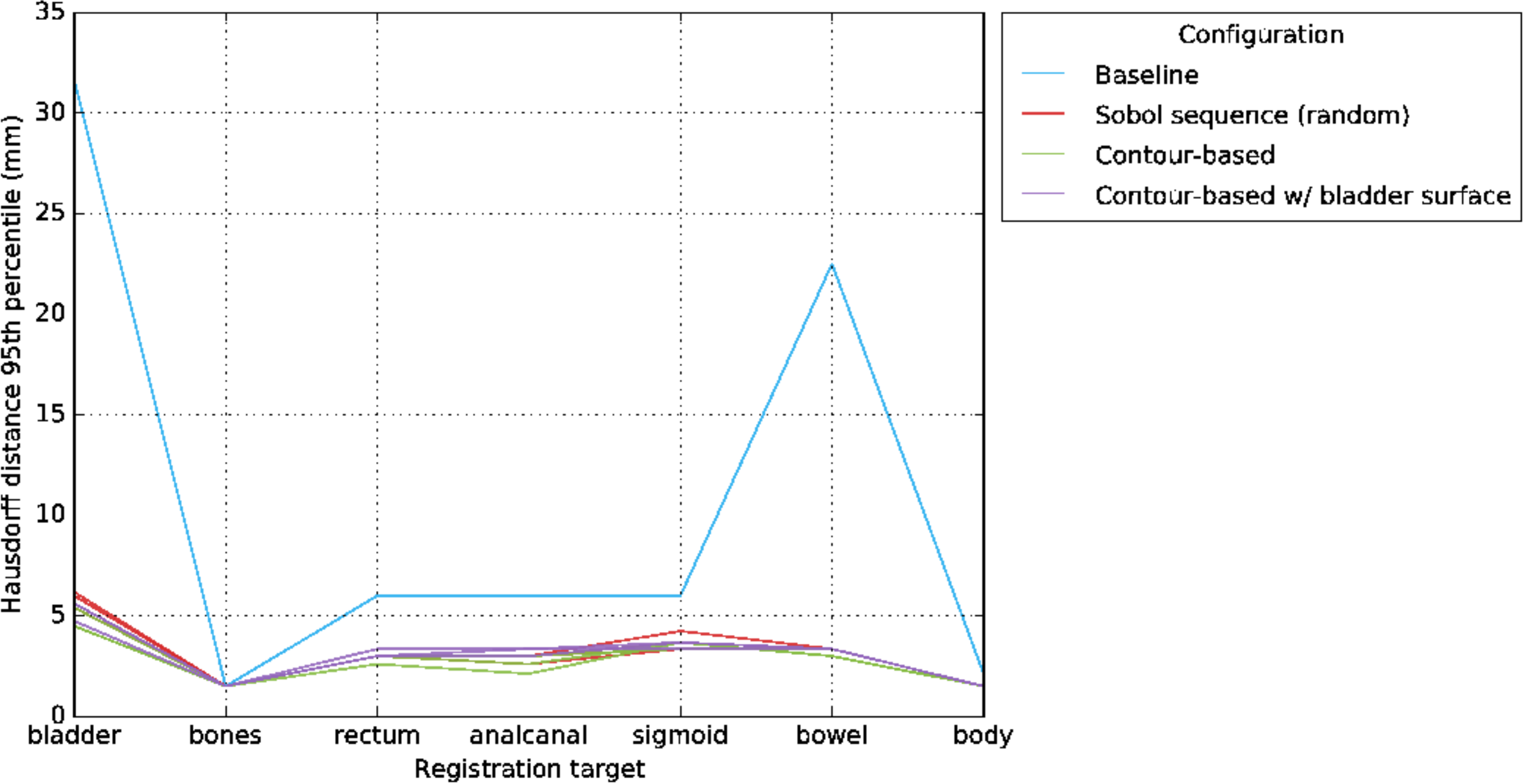}
  \caption{95th percentiles of the Hausdorff distance.}
  \label{fig:morea:pre-tuning:placement:hausdorff-95}
\end{subfigure}
\vspace{-0.25cm}
\caption{Comparison of different mesh point placement strategies, for multiple runs. The baseline score after rigid registration is plotted in blue.}
\label{fig:morea:pre-tuning:placement}
\end{figure}

%% file: sup/figures/tex/a-morea/pre-tuning-renders-placement.tex
\begin{figure}
\centering
\begin{subfigure}[b]{.49\linewidth}
  \centering
  \includegraphics[width=\linewidth]{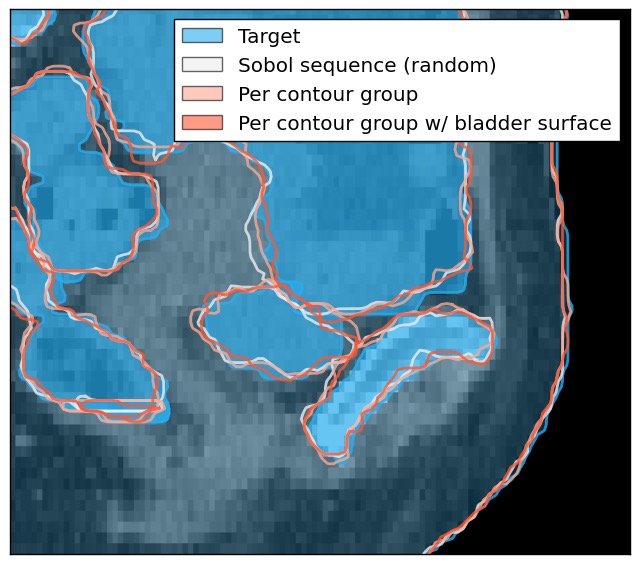}
  \caption{Sagittal slice.}
  \label{fig:morea:pre-tuning:renders:placement:sagittal}
\end{subfigure}%
\begin{subfigure}[b]{.49\linewidth}
  \centering
  \includegraphics[width=\linewidth]{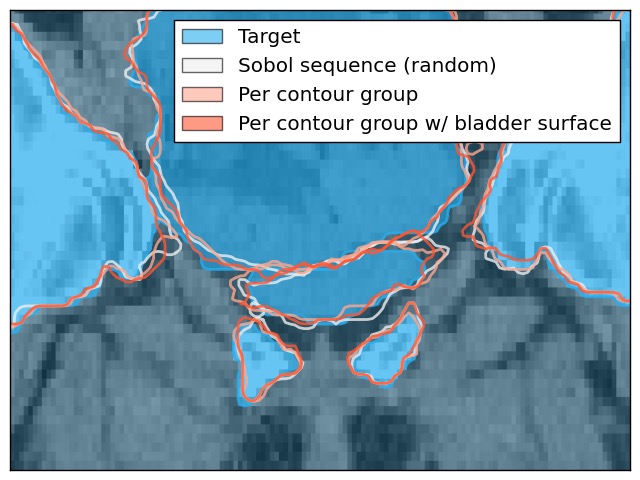}
  \caption{Coronal slice.}
  \label{fig:morea:pre-tuning:renders:placement:coronal}
\end{subfigure}
\vspace{-0.25cm}
\caption{Visual renders of deformations predicted by MOREA with different mesh point placement strategies.}
\label{fig:morea:pre-tuning:renders:placement}
\end{figure}

%% file: sup/figures/tex/a-morea/pre-tuning-disable-guidance.tex
\begin{figure}
\centering
\begin{subfigure}{\linewidth}
  \centering
  \includegraphics[width=\linewidth]{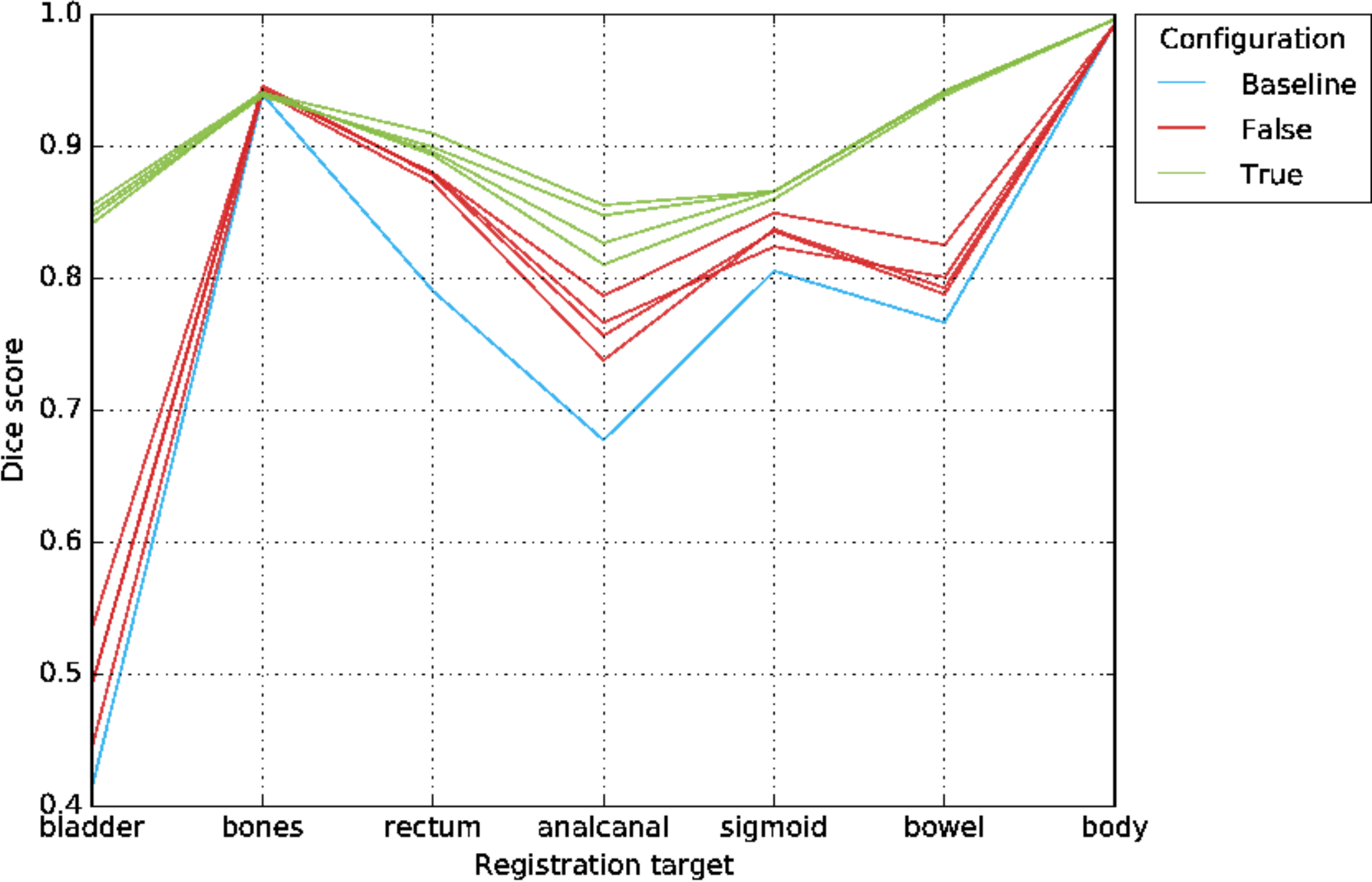}
  \caption{Dice scores.}
  \label{fig:morea:pre-tuning:disable-guidance:dice}
\end{subfigure}
\begin{subfigure}{\linewidth}
  \centering
  \includegraphics[width=\linewidth]{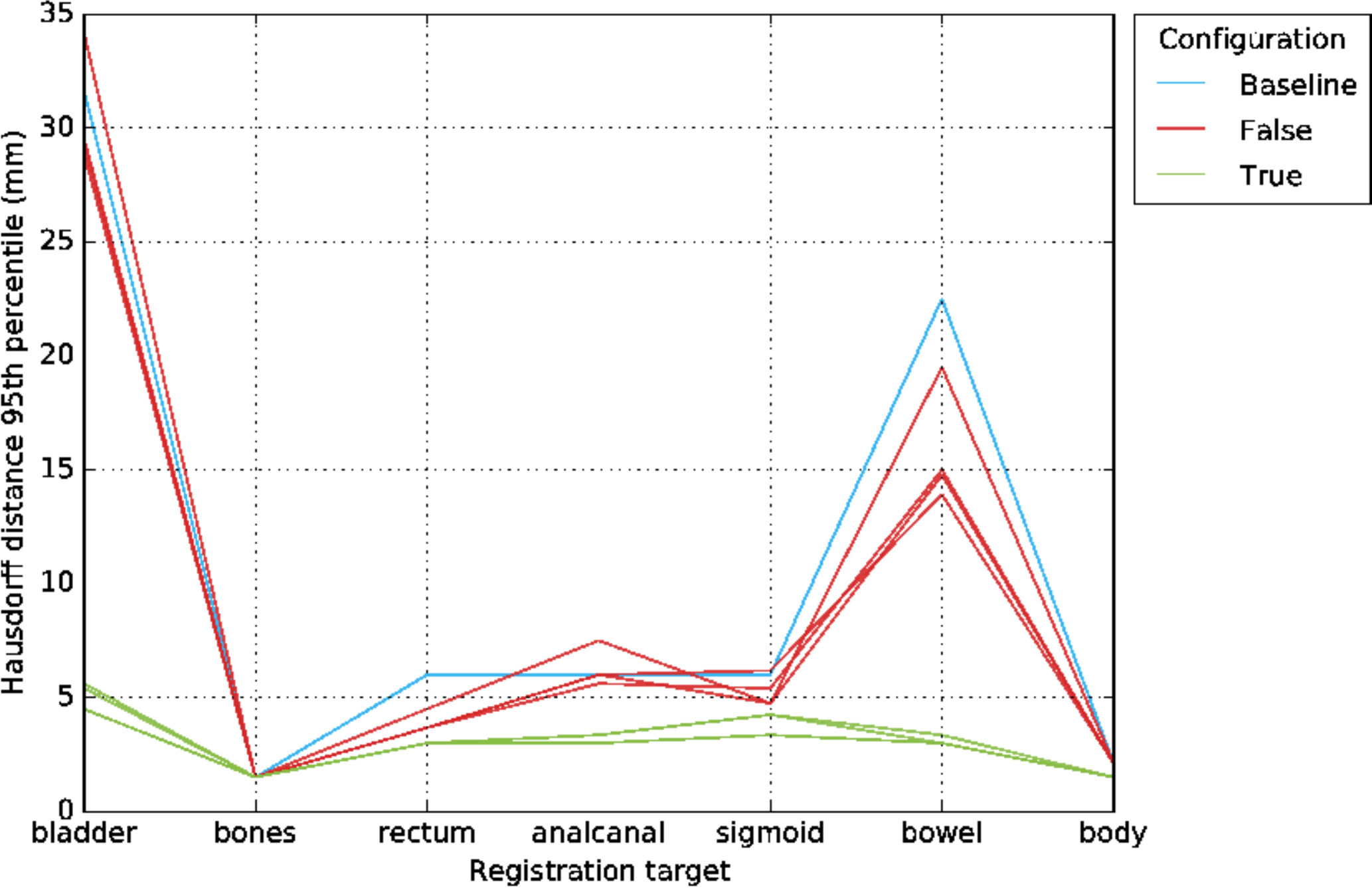}
  \caption{95th percentiles of the Hausdorff distance.}
  \label{fig:morea:pre-tuning:disable-guidance:hausdorff-95}
\end{subfigure}
\vspace{-0.25cm}
\caption{Comparison of MOREA registrations with and without guidance information, for multiple runs. The baseline score after rigid registration is plotted in blue.}
\label{fig:morea:pre-tuning:disable-guidance}
\end{figure}

%% file: sup/figures/tex/a-morea/pre-tuning-renders-disable-guidance.tex
\begin{figure}
\centering
\begin{subfigure}[b]{.49\linewidth}
  \centering
  \includegraphics[width=\linewidth]{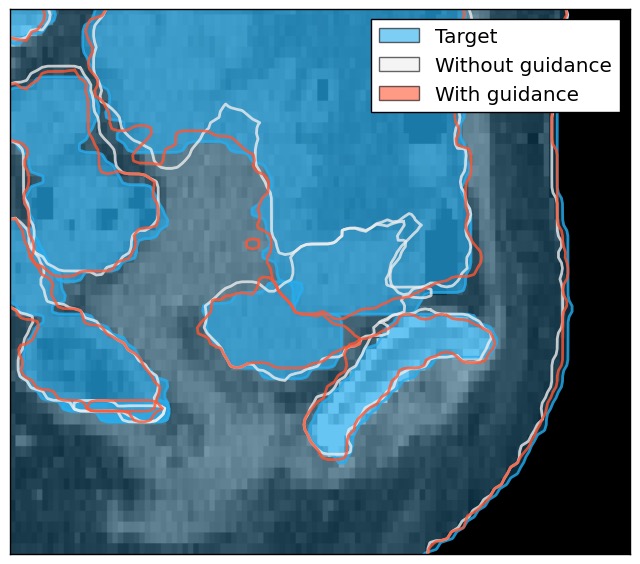}
  \caption{Sagittal slice.}
  \label{fig:morea:pre-tuning:renders:disable-guidance:sagittal}
\end{subfigure}%
\begin{subfigure}[b]{.49\linewidth}
  \centering
  \includegraphics[width=\linewidth]{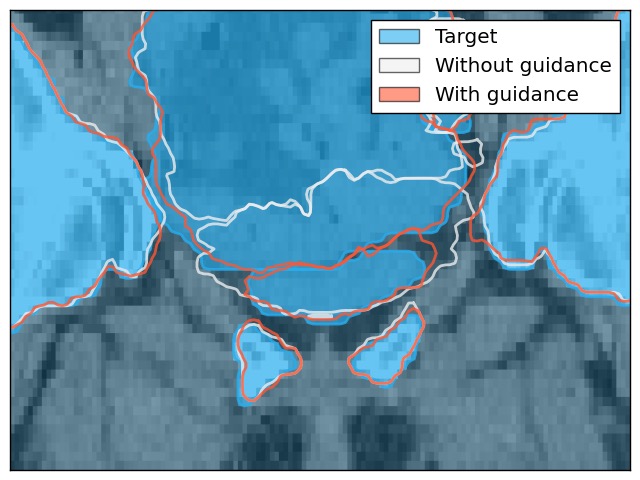}
  \caption{Coronal slice.}
  \label{fig:morea:pre-tuning:renders:disable-guidance:coronal}
\end{subfigure}
\vspace{-0.25cm}
\caption{Visual renders of deformations predicted by MOREA with and without guidance enabled.}
\label{fig:morea:pre-tuning:renders:disable-guidance}
\end{figure}

%% file: sup/figures/tex/a-morea/parameter-files.tex
\begin{lstlisting}[caption={Parameter file used as basis for the main MOREA experiments.},label={listing:morea-parameters}]
sweep_descriptor = "$experiment_descriptor"
num_runs = 5
problem_id = "$problem_id"
zip = true

problem_guidance_enabled = true
problem_guidance_selection = "-1"

cuda_compute_level = 80
cuda_gpu_id = 0

ea_num_generations = 500
ea_population_size = 700
ea_num_clusters = 10
ea_archive_size = 2000

ea_adaptive_steering_enabled = true
ea_adaptive_steering_activated_at_num_generations = 100
ea_adaptive_steering_guidance_threshold = 1.5

morea_init_noise_method = "global-gaussian"
morea_init_noise_factor = 1.0

morea_mesh_generation_method = "annotation-group-random-bladder-10"
morea_mesh_num_points = 600
morea_max_num_mesh_levels = 1
morea_num_generations_per_level = 0

morea_magnitude_metric = "biomechanical"
morea_image_metric = "squared-differences"
morea_guidance_metric = "continuous-per-group"
morea_sampling_rate = 1.0

morea_fos_type = "edges"
morea_symmetry_mode = "transform-both"
morea_dual_dynamic_mode = "dual"
morea_repair_method = "gaussian"
morea_ams_strategy = "none"
morea_num_disruption_kernels = 0
morea_disruption_frequency = 0
\end{lstlisting}

%% file: sup/sections/d-full-results.tex
\section{Full Experimental Results}
\label{sec:full-experimental-results}

In this appendix, we list more extensive results of the experiments presented in Section~6.
Figure~\ref{fig:comparison:dice:full} and~\ref{fig:comparison:hausdorff:full} give full metric results for all patients, comparing the three approaches with parallel coordinate plots.
Table~\ref{tab:significance-full} lists significance test results for all organ Hausdorff distances.
A visual perspective is provided by Table~\ref{tab:best-renders-full}, which shows an additional slice per patient, overlaid with the predicted deformations.
Below, we analyze convergence behavior (Section~\ref{sec:full-experimental-results:convergence}) and landmark performance (Section~\ref{sec:full-experimental-results:landmarks}).

\subsection{Convergence Behavior}
\label{sec:full-experimental-results:convergence}
We plot the convergence behavior of each approach on Patient~1 in Figure~\ref{fig:convergence} to show how each approach has converged before yielding the results we show here.
Elastix and ANTs both have a multi-resolution approach.
To deal with the discontinuities in multi-stage resolution, we mark resolution switches in those plots with red vertical lines.
Our configuration of Elastix also has a mask registration step, meaning that there are in total 8 segments (4 resolutions of mask registration and 4 resolutions of image registration).
The scaling of the value to be optimized is not always normalized across resolutions, which explains the jumps in value ranges between resolutions.
Note that ANTs uses a separate ``convergence value'' to determine when it has converged, plotted in Figure~\ref{fig:convergence:ants:conv}.
For MOREA, we plot the achieved hypervolume and the best guidance objective value achieved.
The sudden decrease in hypervolume at generation 100 is related to the adaptive steering strategy used, which purges any solutions with unfavorable guidance objective values from the elitist archive.

\subsection{Landmark Accuracy}
\label{sec:full-experimental-results:landmarks}

We list landmark registration accuracy on all 4 patients in Table~\ref{tab:tres}.
We aggregate all errors of all landmarks across repeats for one patient and approach, and compute the mean and standard deviation on this sample set.
Since these landmarks are generally placed on visible, anatomically stable locations, and typically not in strongly deforming regions, this accuracy should be interpreted as a measure of how well the method preserves certain anatomical structures.
This measure is therefore less suitable as a measure of how well the registration problem is ``solved'', for which visual (DVF and rendered) inspection is still key.
For some landmarks, the precise location can be ambiguously defined or less visible on certain patients.
These landmarks are, however, still accurately placeable between scans by using the visual context they are situated in and taking consistent placement decisions for each pair of scans.

Generally, we observe that Elastix performs worse than ANTs and MOREA, and MOREA always improves or roughly maintains the baseline landmark registration error.
We do not see a consistent correlation between actual registration performance on large deforming objects and target registration error values, due to the aforementioned reasons.

\input{sup/figures/tex/results/significance-full.tex}

\input{sup/figures/tex/results/tres.tex}

\clearpage

\input{sup/figures/tex/results/best-renders-full.tex}

\input{sup/figures/tex/results/comparison-dice-full.tex}

\input{sup/figures/tex/results/comparison-hausdorff-full.tex}

\input{sup/figures/tex/results/convergence-plots.tex}

\clearpage

%% file: sup/figures/tex/results/significance-full.tex
\begin{table}
    \centering
    \begin{tabularx}{\linewidth}{Xlrr}
        \toprule
        Problem & Contour & MOREA / Elastix & MOREA / ANTs \\
        \midrule
\multirow{7}{*}{Patient 1} 
& bladder & \textbf{0.011} (+) & \textbf{0.007} (+)\\
& bones & \textbf{0.009} (+) & \textbf{0.006} (+)\\
& rectum & \textbf{0.007} (+) & \textbf{0.007} (+)\\
& anal canal & \textbf{0.007} (+) & \textbf{0.007} (+)\\
& sigmoid & \textbf{0.007} (+) & \textbf{0.007} (+)\\
& bowel & \textbf{0.010} (+) & \textbf{0.011} (+)\\
& body & \textbf{0.006} (+) & \textbf{0.006} (-)\\
\midrule

\multirow{7}{*}{Patient 2} 
& bladder & \textbf{0.007} (+) & \textbf{0.007} (+)\\
& bones & \textbf{0.007} (+) & \textbf{0.007} (+)\\
& rectum & 0.118 (+) & \textbf{0.007} (-)\\
& anal canal & 0.123 (-) & 0.180 (-)\\
& sigmoid & \textbf{0.007} (+) & \textbf{0.007} (-)\\
& bowel & 0.401 (+) & \textbf{0.007} (+)\\
& body & 0.655 (+) & 1.000 (-)\\
\midrule

\multirow{7}{*}{Patient 3} 
& bladder & \textbf{0.012} (+) & \textbf{0.007} (+)\\
& bones & \textbf{0.007} (+) & \textbf{0.007} (+)\\
& rectum & 0.290 (+) & \textbf{0.007} (-)\\
& anal canal & 0.118 (-) & \textbf{0.007} (+)\\
& sigmoid & \textbf{0.007} (+) & \textbf{0.007} (+)\\
& bowel & \textbf{0.007} (+) & 0.056 (+)\\
& body & \textbf{0.007} (+) & 0.118 (+)\\
\midrule

\multirow{7}{*}{Patient 4} 
& bladder & \textbf{0.007} (+) & 0.195 (-)\\
& bones & \textbf{0.007} (-) & \textbf{0.007} (-)\\
& rectum & \textbf{0.010} (-) & \textbf{0.007} (-)\\
& anal canal & 0.606 (+) & \textbf{0.007} (-)\\
& sigmoid & \textbf{0.009} (+) & 0.118 (+)\\
& bowel & 0.119 (+) & 0.119 (-)\\
& body & \textbf{0.020} (-) & \textbf{0.020} (-)\\
        \bottomrule
    \end{tabularx}
    \vspace{0.05cm}
    \caption{\textbf{p}-values of pair-wise comparisons of Hausdorff distances for all contours between approaches, computed by the two-sided Mann-Whitney U test. A plus (\texttt{+}) indicates a better mean with MOREA, a minus (\texttt{-}) the opposite. Significant results are highlighted according to an $\alpha$ of 0.025.}
    \label{tab:significance-full}
    \vspace{-0.4cm}
\end{table}

%% file: sup/figures/tex/results/tres.tex
\begin{table}[t]
    \centering
    \begin{tabularx}{\linewidth}{Xrrrr}
        \toprule
        Problem & Baseline & Elastix & ANTs & MOREA \\
        \midrule
Patient 1 & 4.8 $\pm$ 3.1 & 5.6 $\pm$ 2.8 & 4.2 $\pm$ 2.0 & 4.8 $\pm$ 2.0\\
Patient 2 & 7.5 $\pm$ 4.0 & 11.8 $\pm$ 7.3 & 7.7 $\pm$ 4.3 & 7.8 $\pm$ 3.8\\
Patient 3 & 9.5 $\pm$ 6.7 & 6.4 $\pm$ 2.0 & 7.7 $\pm$ 2.6 & 6.5 $\pm$ 1.9\\
Patient 4 & 14.1 $\pm$ 9.5 & 8.1 $\pm$ 4.3 & 6.3 $\pm$ 3.4 & 6.8 $\pm$ 4.0\\
        \bottomrule
    \end{tabularx}
    \vspace{0.05cm}
    \caption{Target registration errors (mean and standard deviation) for the shown registrations of each approach on each patient, across repeats. All errors are specified in \emph{mm}.}
    \label{tab:tres}
    \vspace{-0.4cm}
\end{table}

%% file: sup/figures/tex/results/best-renders-full.tex
\newcommand{\bestrenderwidthfull}{5.5cm}
\newcommand{\patientspacefull}{0.3cm}

\begin{table*}
    \centering
    \setlength\tabcolsep{0pt}
    \begin{tabular}{lcc}
        \toprule
        Instance & Transformed: sagittal & Transformed: coronal \\
        \midrule
        Patient 1 \hspace{\patientspacefull} & \includegraphics[width=\bestrenderwidthfull,valign=m]{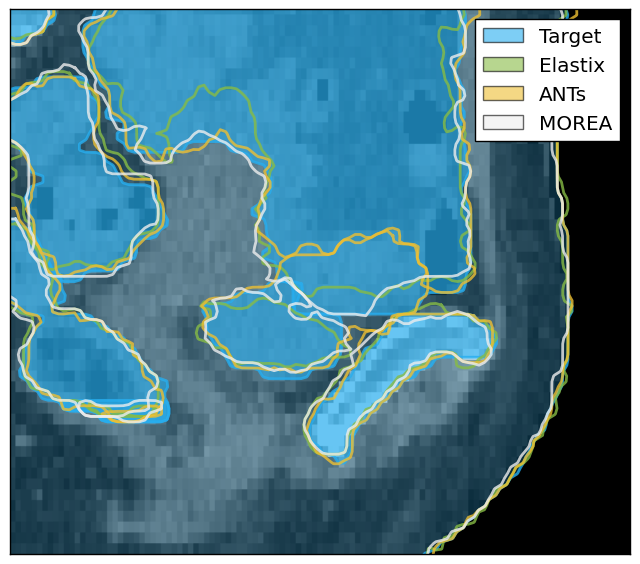} & \includegraphics[width=\bestrenderwidthfull,valign=m]{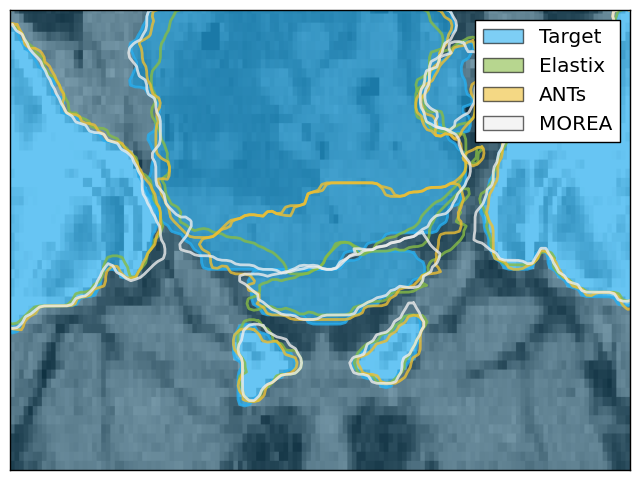}\vspace{0.05cm}\\
        
        Patient 2 \hspace{\patientspacefull} & \includegraphics[width=\bestrenderwidthfull,valign=m]{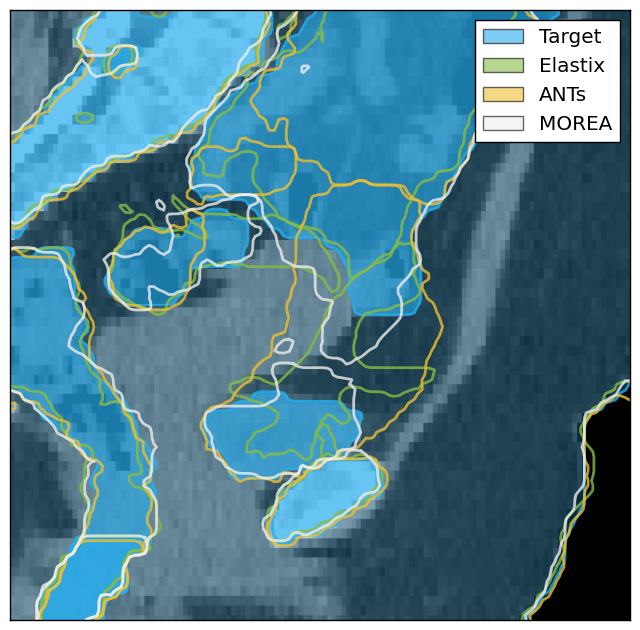} & \includegraphics[width=\bestrenderwidthfull,valign=m]{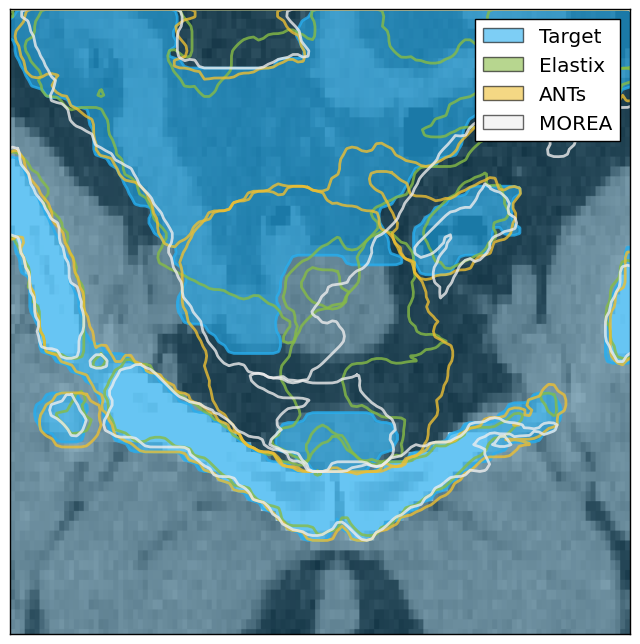}\vspace{0.05cm}\\
        
        Patient 3 \hspace{\patientspacefull} & \includegraphics[width=\bestrenderwidthfull,valign=m]{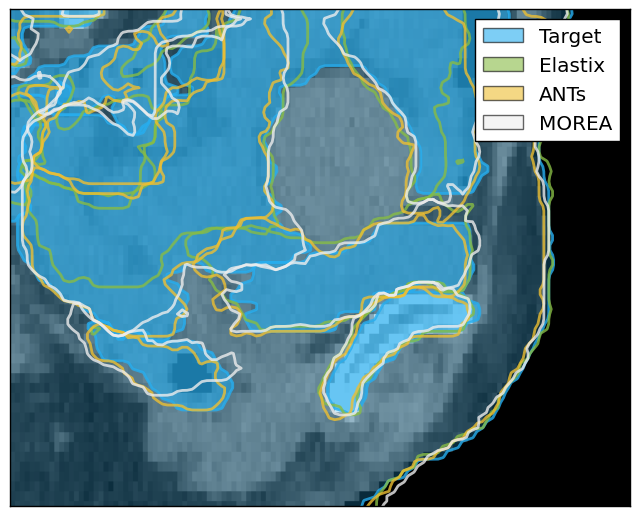} & \includegraphics[width=\bestrenderwidthfull,valign=m]{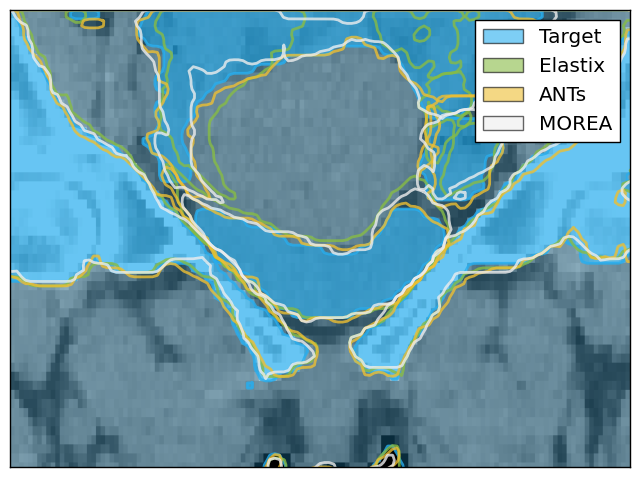}\vspace{0.05cm}\\
        
        Patient 4 \hspace{\patientspacefull} & \includegraphics[width=\bestrenderwidthfull,valign=m]{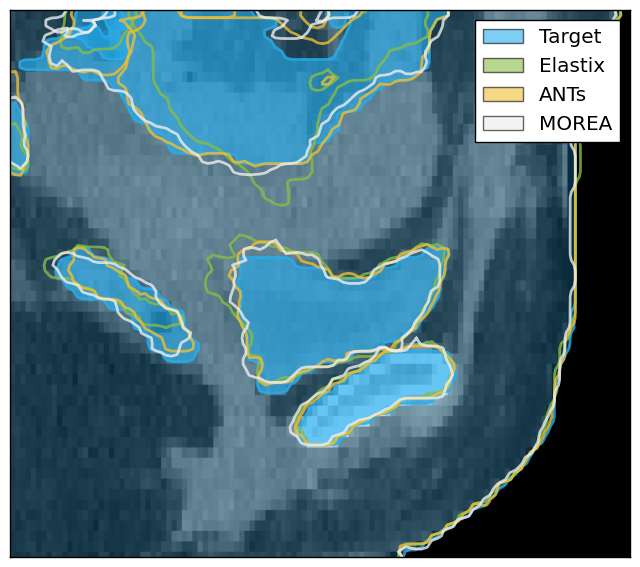} & \includegraphics[width=\bestrenderwidthfull,valign=m]{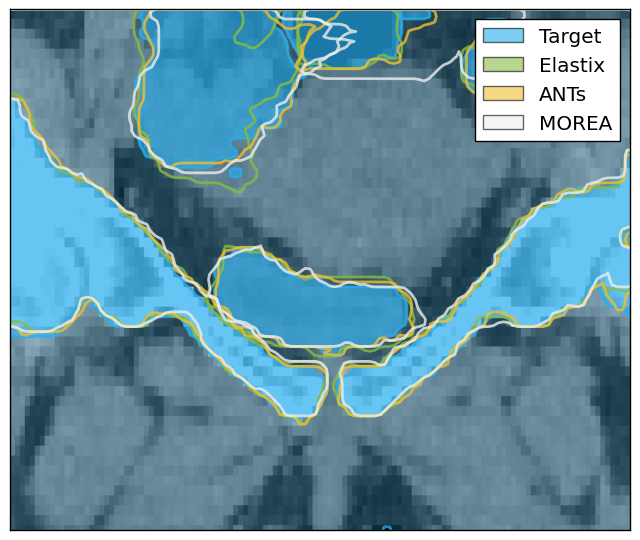}\vspace{0.05cm}\\
        \bottomrule
    \end{tabular}
    \vspace{0.05cm}
    \caption{A selection of the best predicted deformations of the compared registration approaches, represented by deformed contours compared to the target contours and image.}
    \label{tab:best-renders-full}
    \vspace{-0.4cm}
\end{table*}

%% file: sup/figures/tex/results/comparison-dice-full.tex
\begin{figure*}
\centering
\begin{subfigure}{.4\linewidth}
  \centering
  \includegraphics[width=\linewidth]{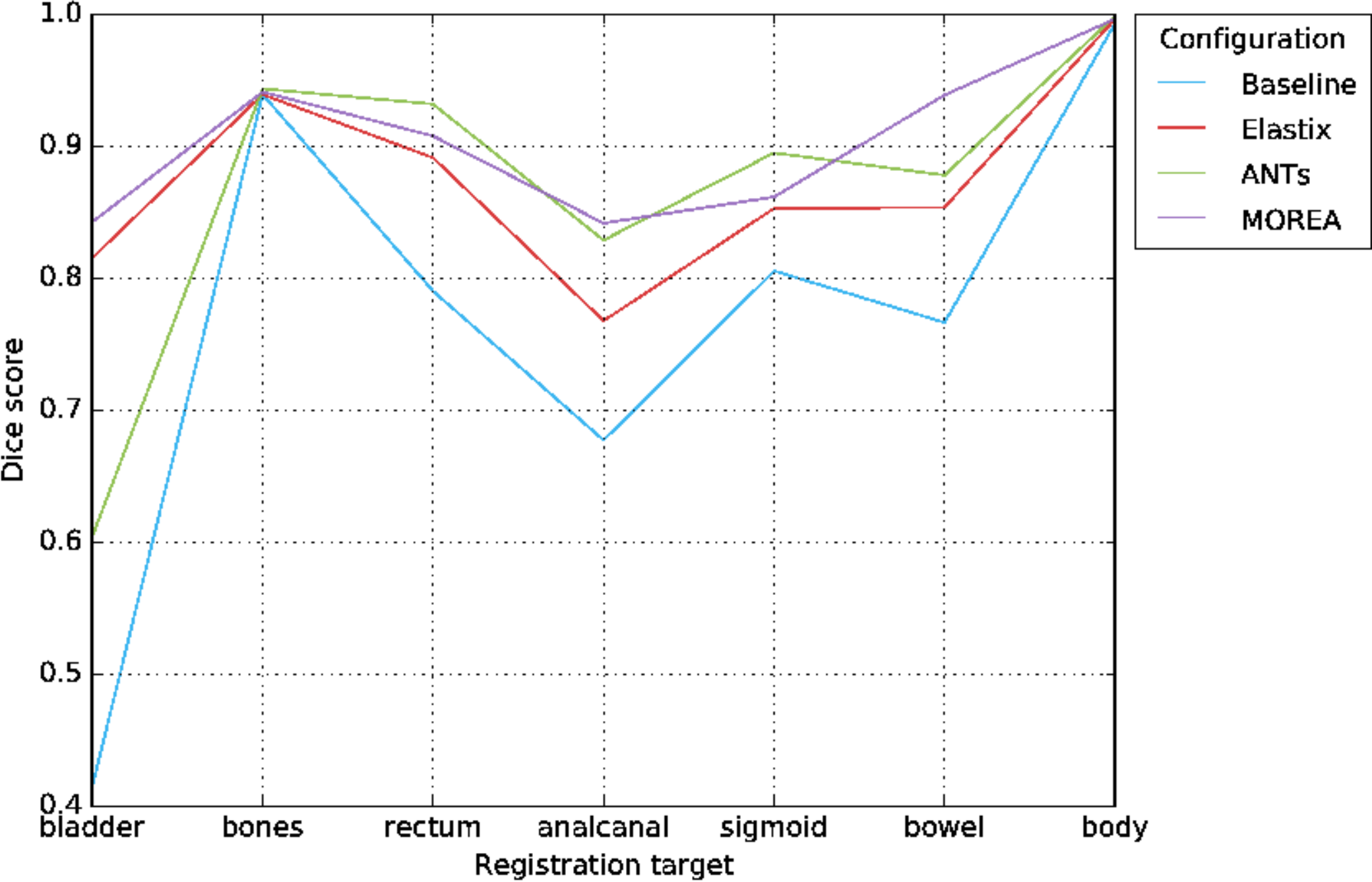}
  \caption{Patient 1.}
  \label{fig:comparison:dice:603}
\end{subfigure}%
\hspace{0.019\linewidth}
\begin{subfigure}{.4\linewidth}
  \centering
  \includegraphics[width=\linewidth]{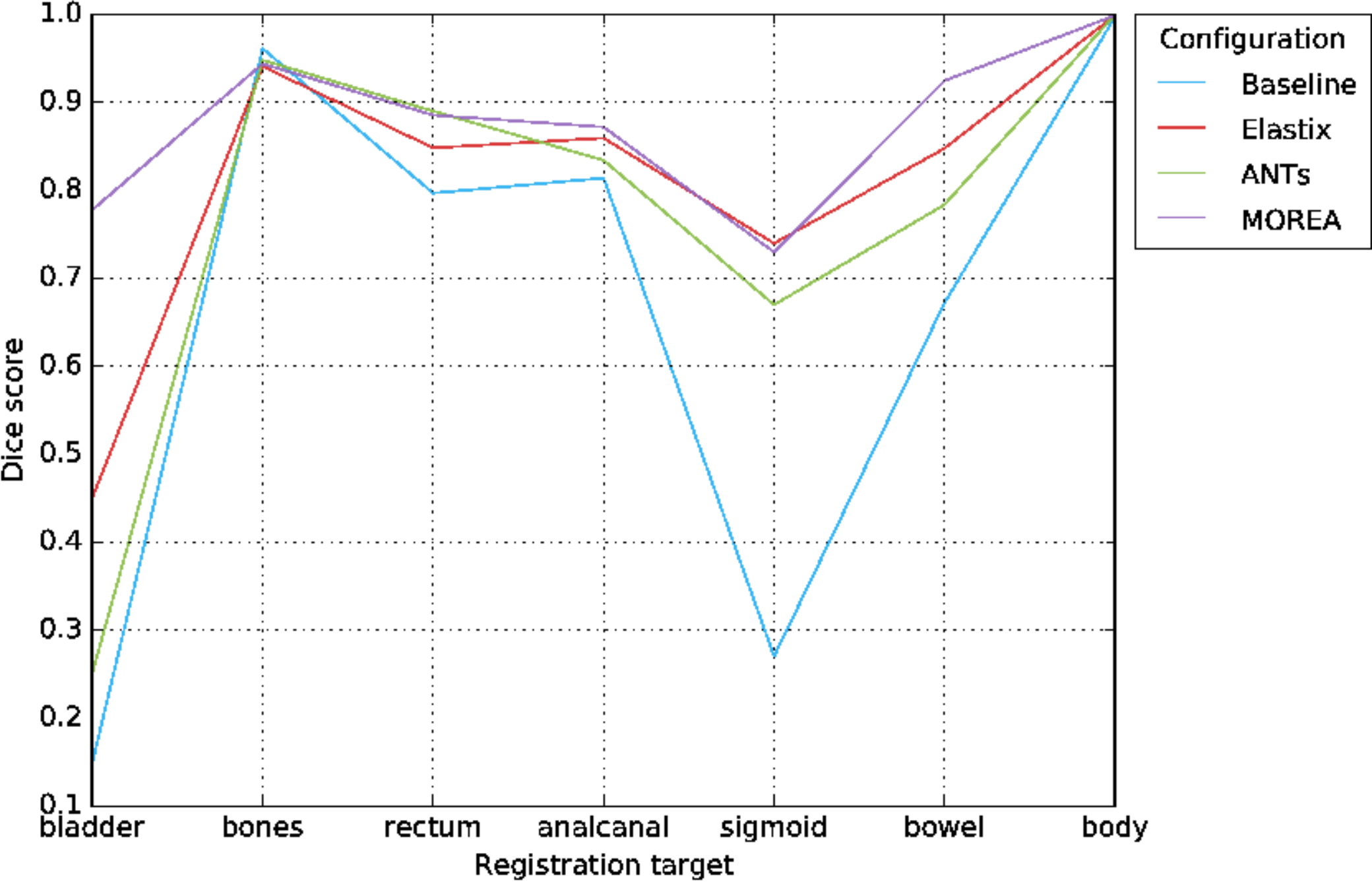}
  \caption{Patient 2.}
  \label{fig:comparison:dice:611}
\end{subfigure}%
\hspace{0.019\linewidth}
\begin{subfigure}{.4\linewidth}
  \centering
  \includegraphics[width=\linewidth]{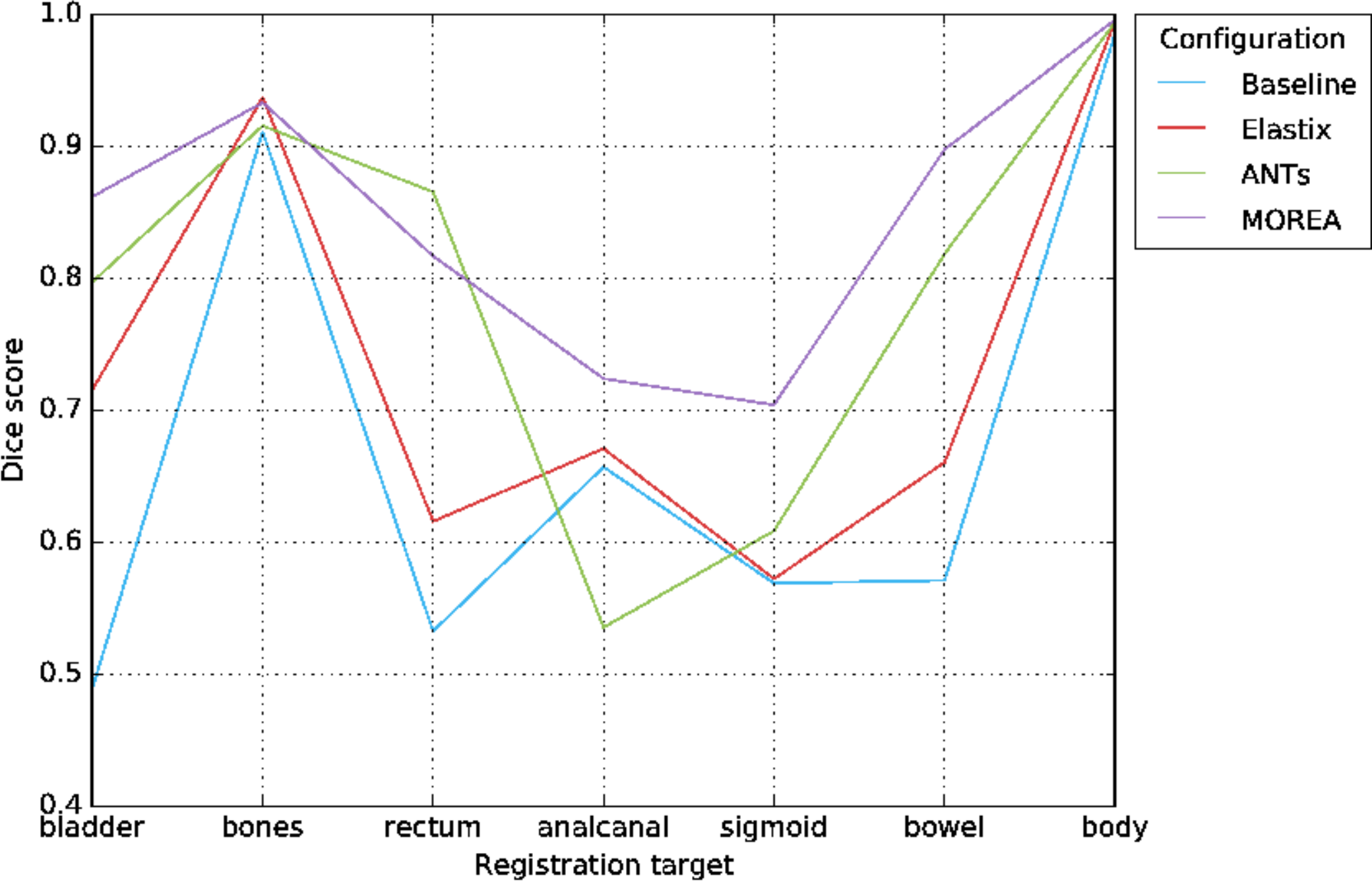}
  \caption{Patient 3.}
  \label{fig:comparison:dice:617}
\end{subfigure}%
\hspace{0.019\linewidth}
\begin{subfigure}{.4\linewidth}
  \centering
  \includegraphics[width=\linewidth]{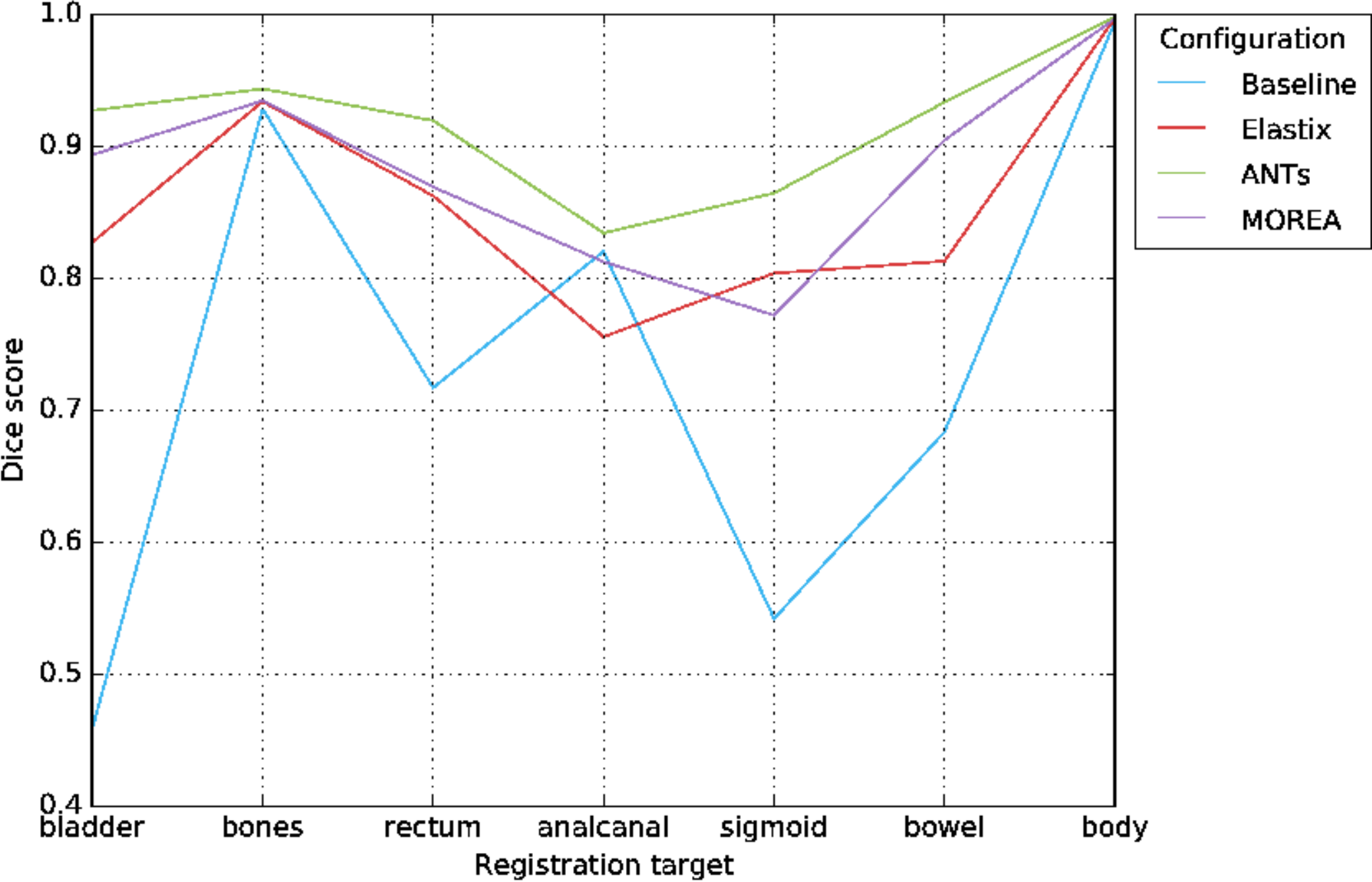}
  \caption{Patient 4.}
  \label{fig:comparison:dice:618}
\end{subfigure}
\vspace{-0.25cm}
\caption{Dice scores for all approaches on all patients. The baseline score after rigid registration is plotted in blue.}
\label{fig:comparison:dice:full}
\vspace{-0.08cm}
\end{figure*}

%% file: sup/figures/tex/results/comparison-hausdorff-full.tex
\begin{figure*}
\centering
\begin{subfigure}{.4\linewidth}
  \centering
  \includegraphics[width=\linewidth]{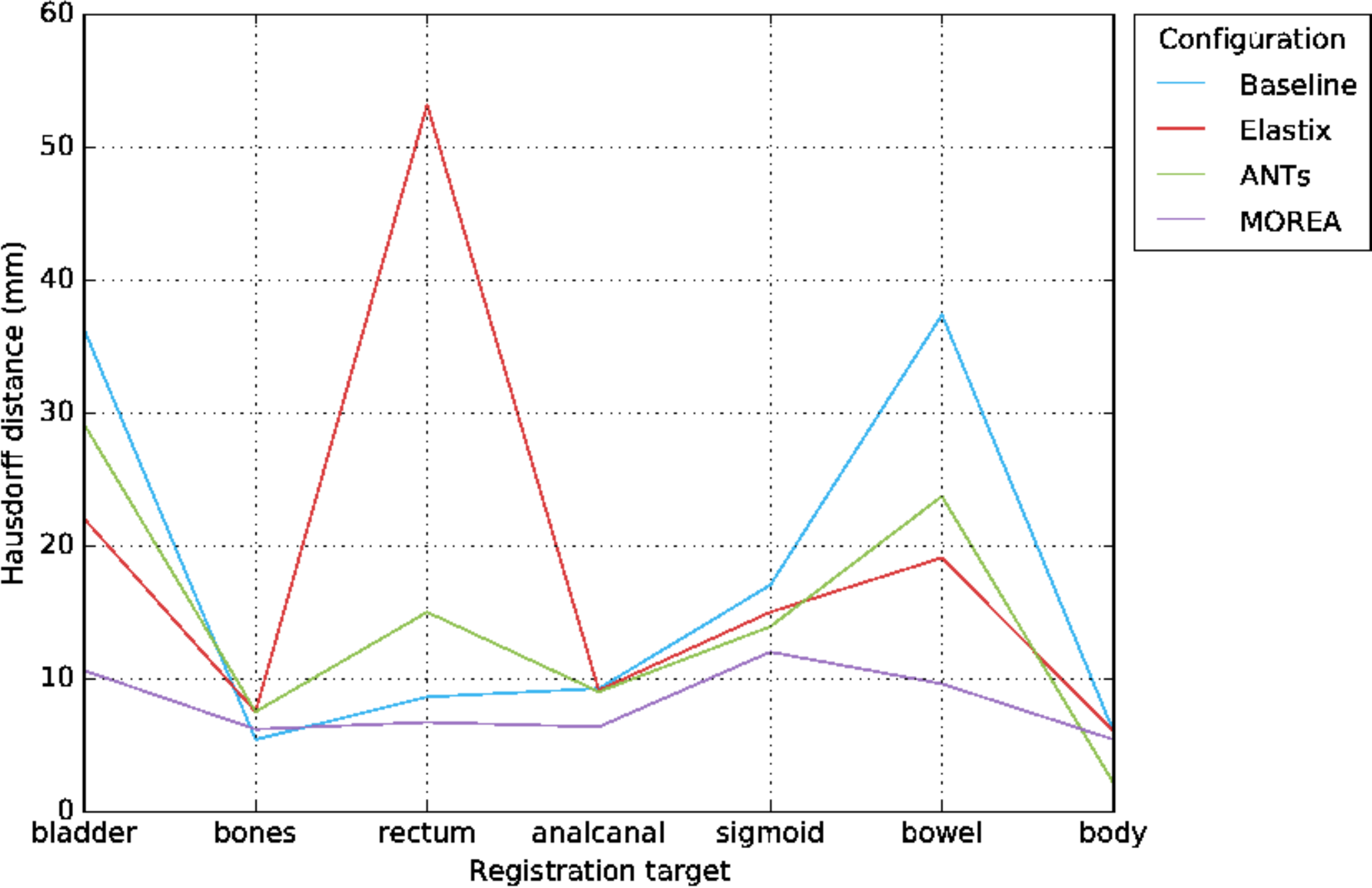}
  \caption{Patient 1.}
  \label{fig:comparison:hausdorff:603}
\end{subfigure}%
\hspace{0.019\linewidth}
\begin{subfigure}{.4\linewidth}
  \centering
  \includegraphics[width=\linewidth]{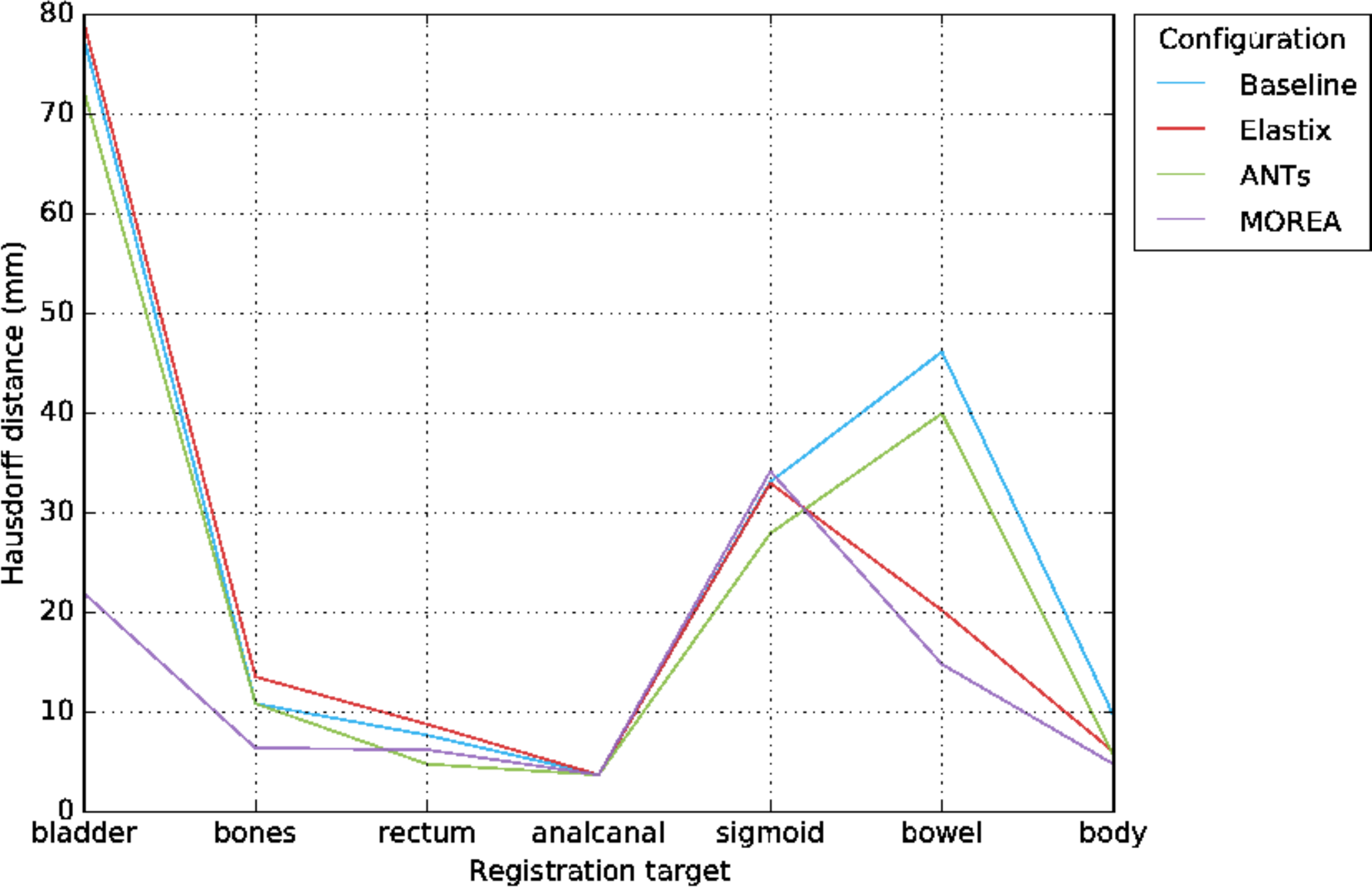}
  \caption{Patient 2.}
  \label{fig:comparison:hausdorff:611}
\end{subfigure}%
\hspace{0.019\linewidth}
\begin{subfigure}{.4\linewidth}
  \centering
  \includegraphics[width=\linewidth]{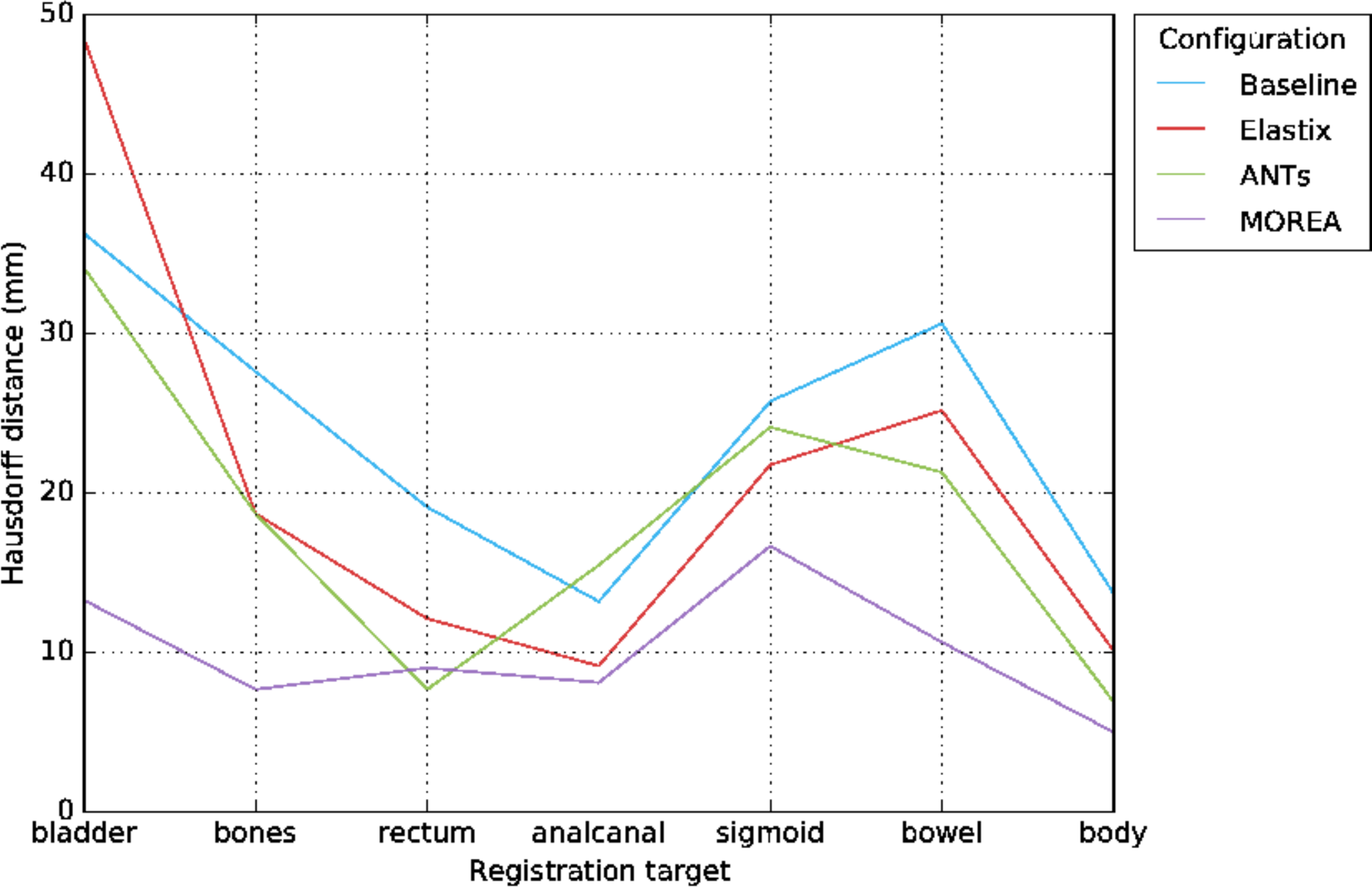}
  \caption{Patient 3.}
  \label{fig:comparison:hausdorff:617}
\end{subfigure}%
\hspace{0.019\linewidth}
\begin{subfigure}{.4\linewidth}
  \centering
  \includegraphics[width=\linewidth]{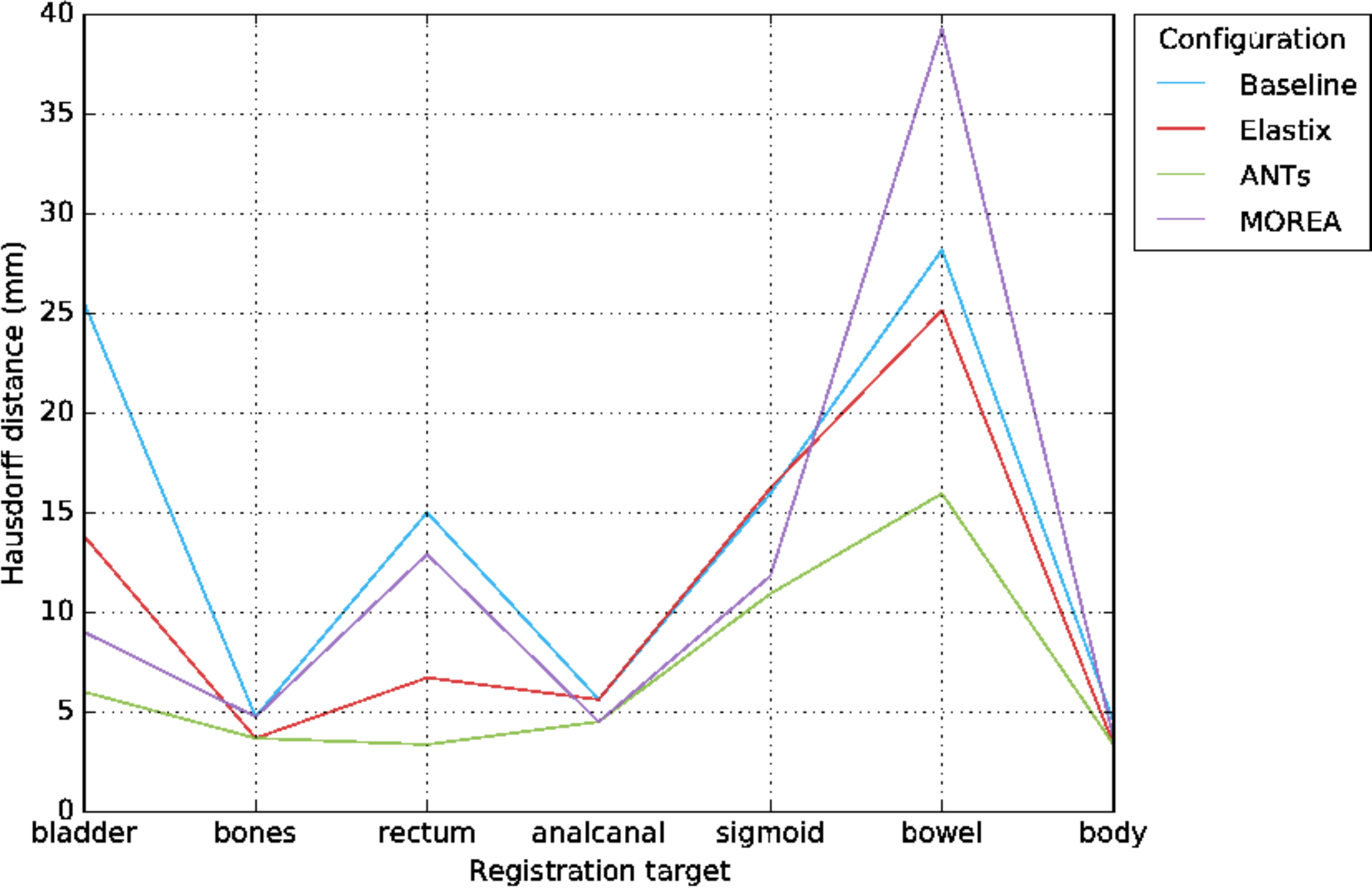}
  \caption{Patient 4.}
  \label{fig:comparison:hausdorff:618}
\end{subfigure}
\vspace{-0.25cm}
\caption{Hausdorff distances for all approaches on all patients. The baseline score after rigid registration is plotted in blue.}
\label{fig:comparison:hausdorff:full}
\end{figure*}

%% file: sup/figures/tex/results/convergence-plots.tex
\begin{figure*}
\centering
\begin{subfigure}{.48\linewidth}
  \centering
  \includegraphics[width=\linewidth]{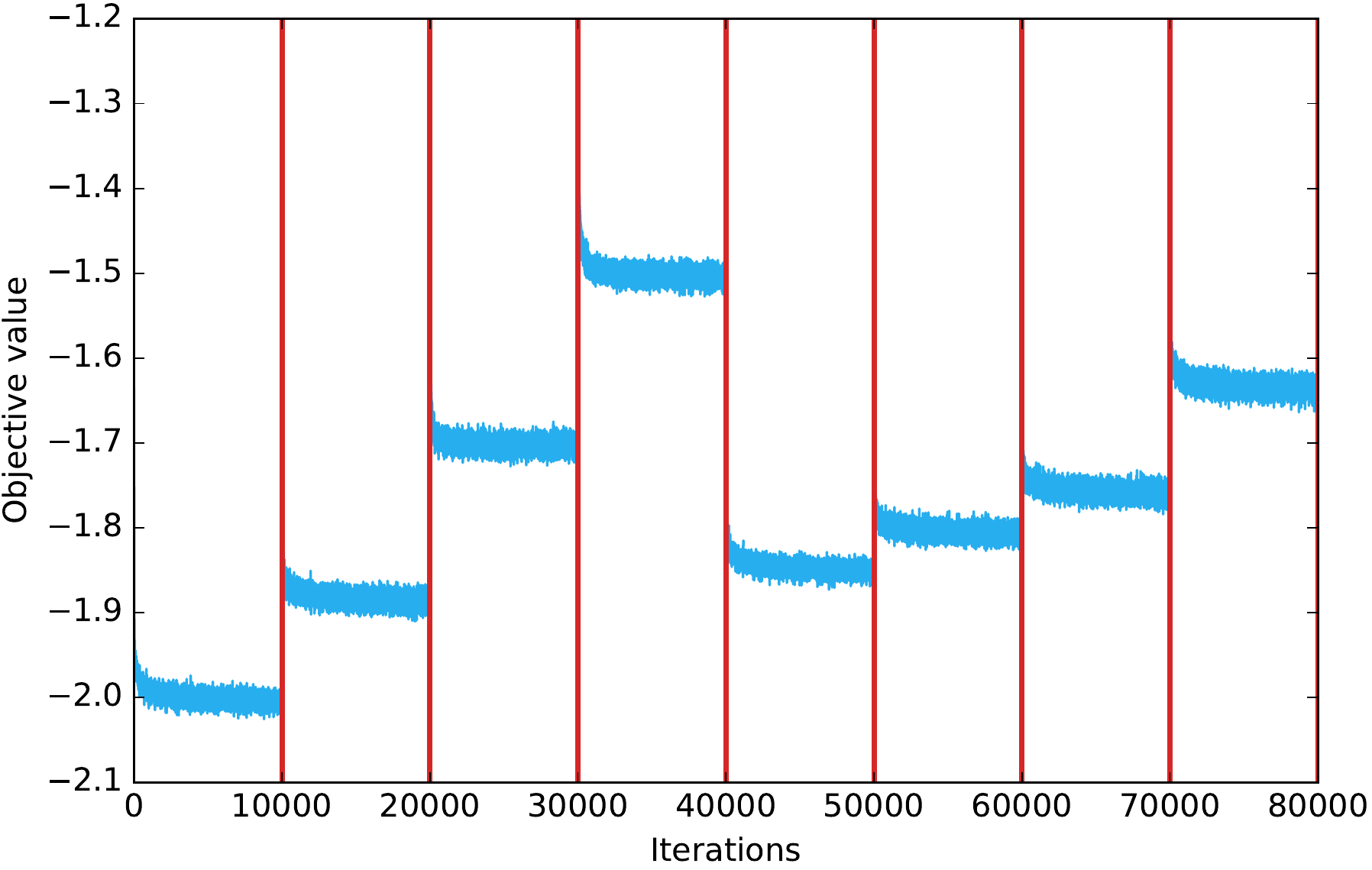}
  \caption{Elastix: objective value at each iteration.}
  \label{fig:convergence:elastix:total}
\end{subfigure}%
\hspace{0.019\linewidth}
\begin{subfigure}{.48\linewidth}
  \centering
  \includegraphics[width=\linewidth]{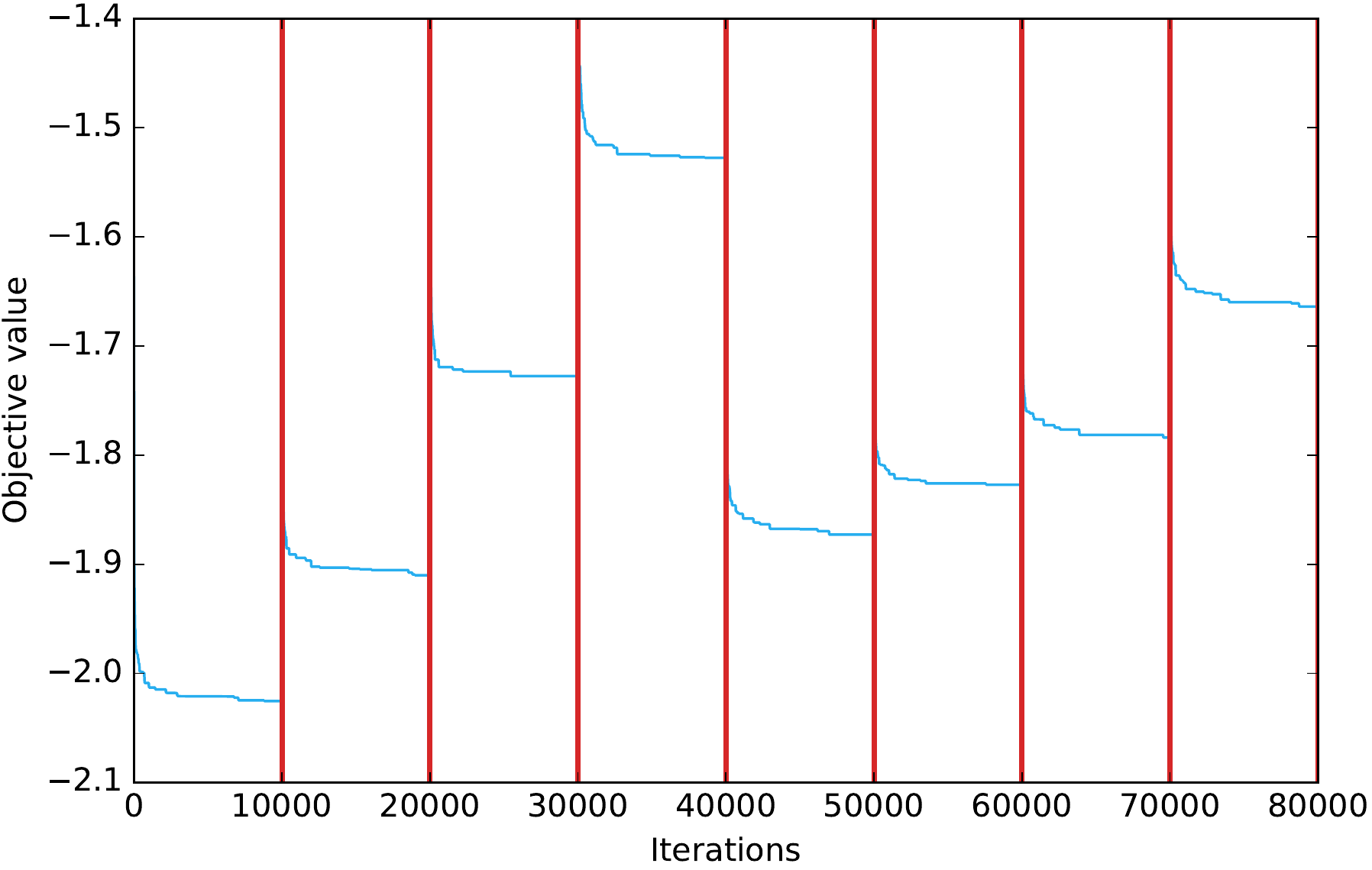}
  \caption{Elastix: best objective value achieved at each point.}
  \label{fig:convergence:elastix:best}
\end{subfigure}
\begin{subfigure}{.48\linewidth}
  \centering
  \includegraphics[width=\linewidth]{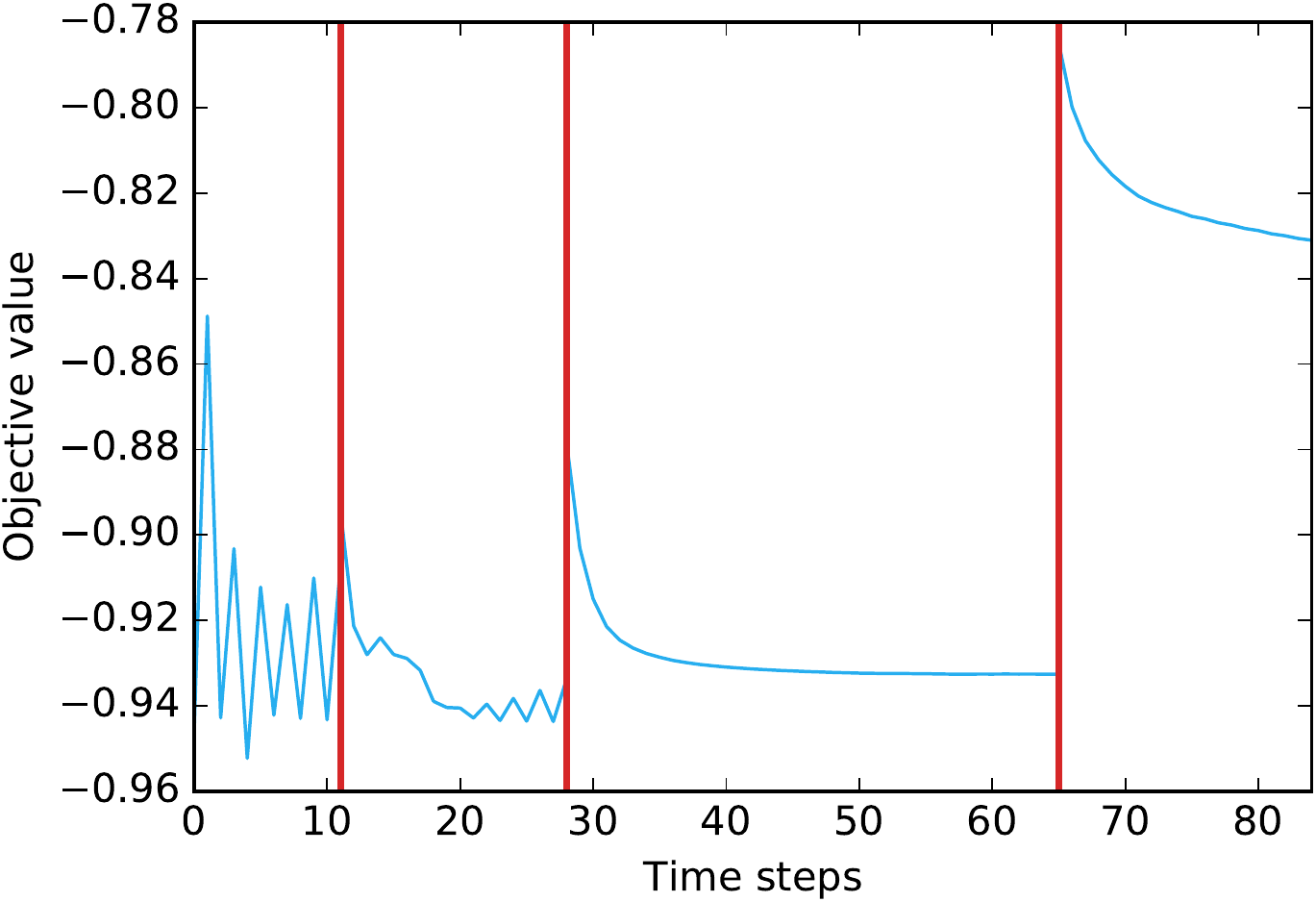}
  \caption{ANTs: objective value at each iteration.}
  \label{fig:convergence:ants:objective}
\end{subfigure}%
\hspace{0.019\linewidth}
\begin{subfigure}{.48\linewidth}
  \centering
  \includegraphics[width=\linewidth]{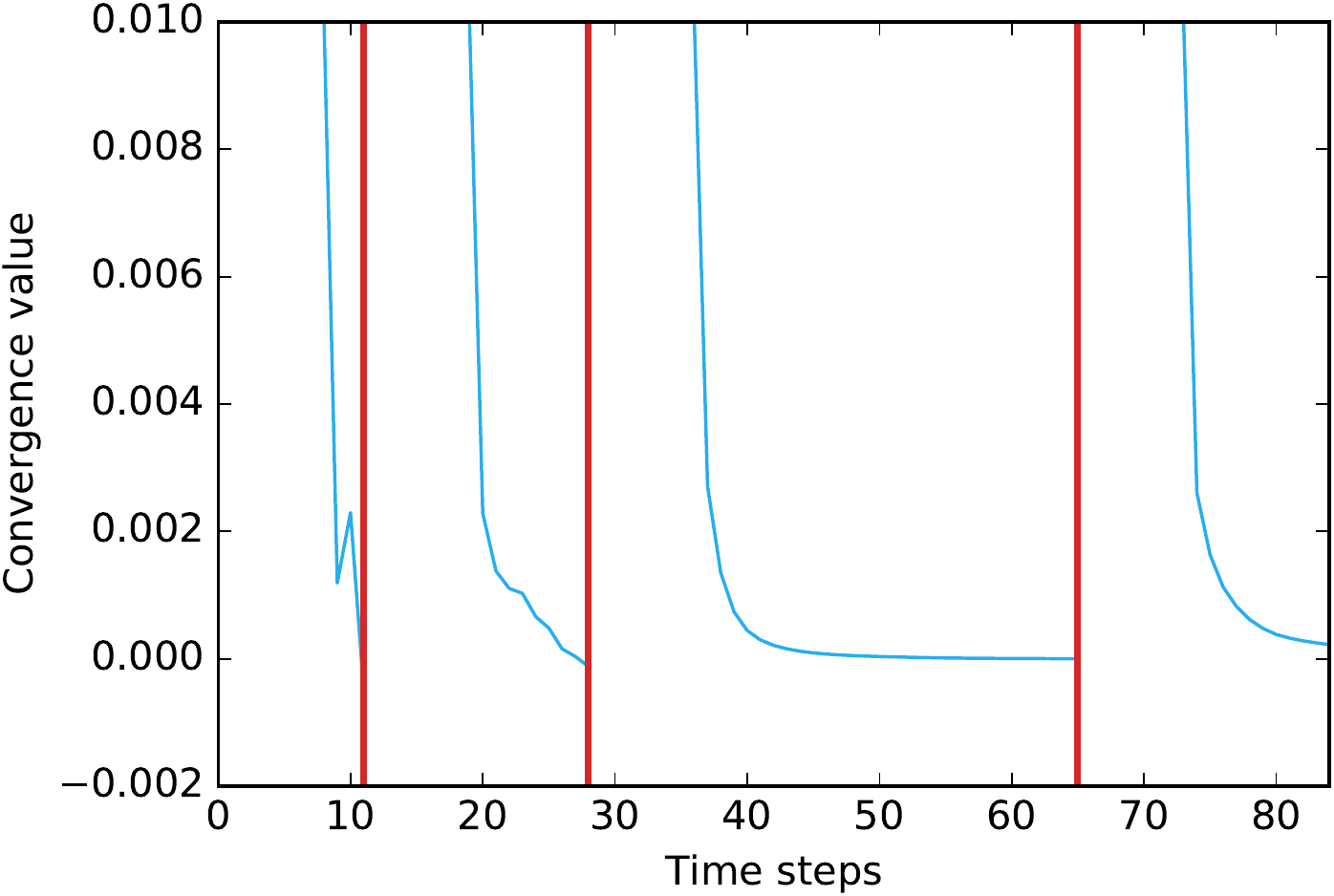}
  \caption{ANTs: convergence measure at each iteration.}
  \label{fig:convergence:ants:conv}
\end{subfigure}
\begin{subfigure}{.48\linewidth}
  \centering
  \includegraphics[width=\linewidth]{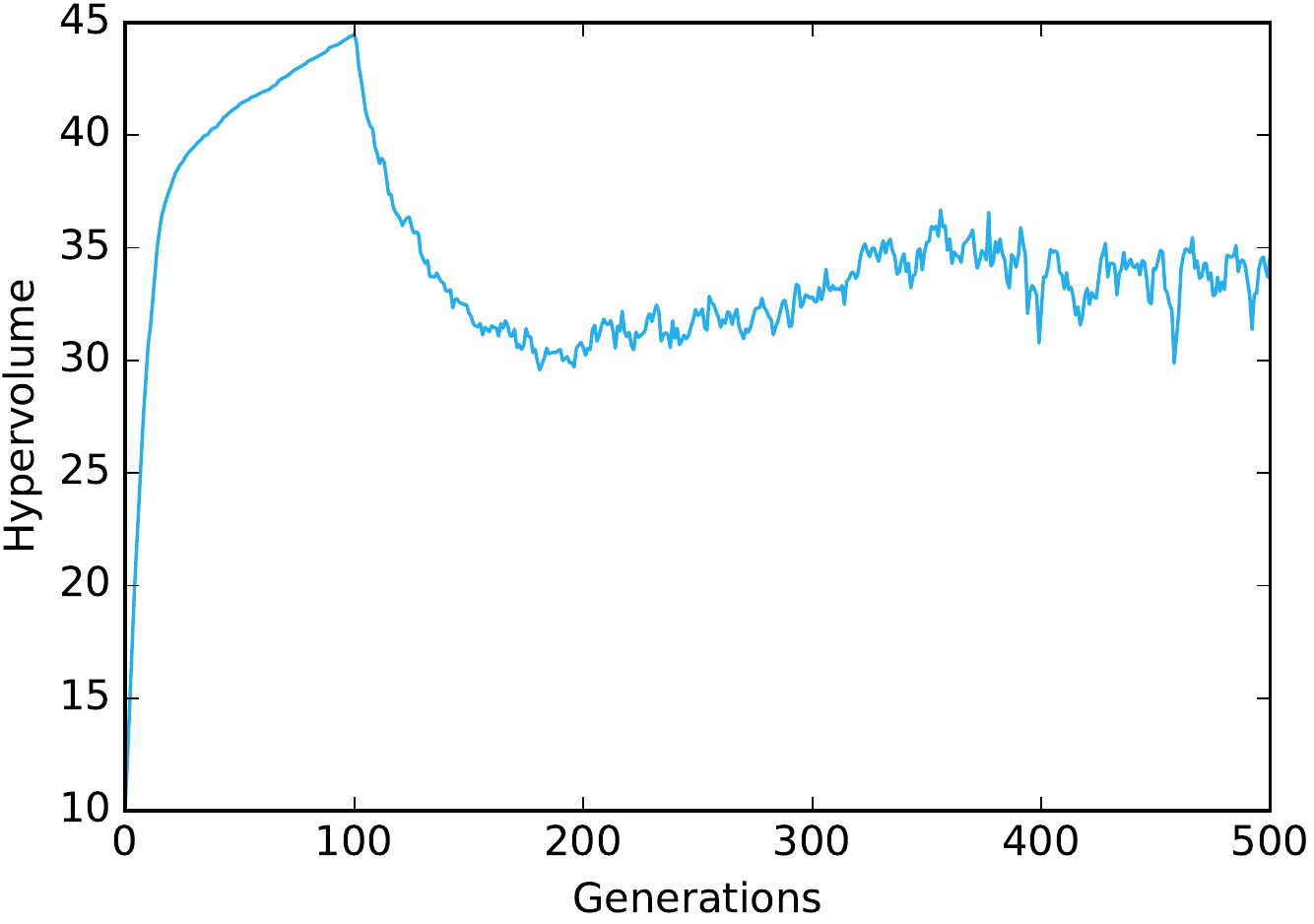}
  \caption{MOREA: hypervolume at each generation.}
  \label{fig:convergence:morea:hv}
\end{subfigure}%
\hspace{0.019\linewidth}
\begin{subfigure}{.48\linewidth}
  \centering
  \includegraphics[width=\linewidth]{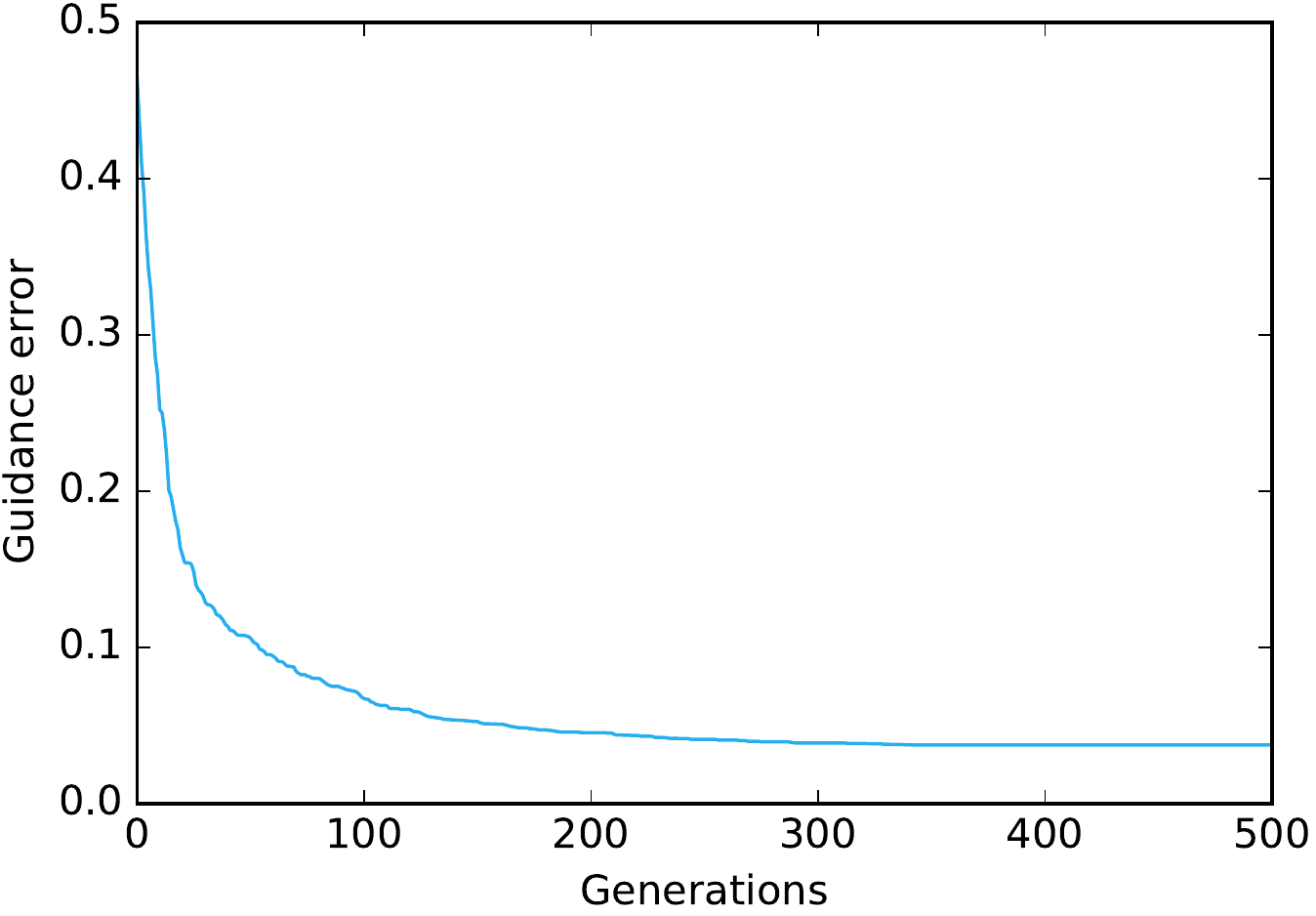}
  \caption{MOREA: best guidance objective value found at each generation.}
  \label{fig:convergence:morea:guidance}
\end{subfigure}
\vspace{-0.25cm}
\caption{Convergence plots for all 3 approaches on one run of Patient~1. Vertical red lines indicate a change of resolution. For ANTs, this leads to 4 optimization segments. For Elastix, we first run a mask registration step (with 4 segments) and then an image registration step (with again 4 segments).}
\label{fig:convergence}
\end{figure*}